\documentclass[acmtog,nonacm]{acmart}
\usepackage{booktabs}
\usepackage{subcaption}
\usepackage[capitalise]{cleveref}
\usepackage{xcolor}

\usepackage[ruled]{algorithm2e}

\SetAlFnt{\small}
\SetAlCapFnt{\small}
\SetAlCapNameFnt{\small}
\SetAlCapHSkip{0pt}

\begin{document}
\title{Progressively Learning Heterogeneous Skills in a Unified Latent Space}

\author{Yue-Yi Zhang}
\email{zhangyy696@mail2.sysu.edu.cn}
\affiliation{
  \institution{Sun Yat-sen University}
  \country{China}
}

\author{Ming Gong}
\email{t2085443@gmail.com}
\affiliation{
  \institution{Sun Yat-sen University}
  \country{China}
}

\author{Linpu He}
\email{linpuhe@163.com}
\affiliation{
  \institution{Sun Yat-sen University}
  \country{China}
}

\author{Wei-Shi Zheng}
\email{wszheng@isee.org}
\affiliation{
  \institution{Sun Yat-sen University}
  \country{China}
}

\author{Zhilin Zhao}
\email{zhaozhlin@mail.sysu.edu.cn}
\affiliation{
  \institution{Sun Yat-sen University}
  \country{China}
}

\begin{abstract}
We propose HetSkills, a novel framework designed to progressively learn heterogeneous skills within a unified latent space for physics-based character control. The core idea is to treat this latent space as a shared executable interface, enabling seamless integration of skills learned from diverse data sources, supervision forms, and tasks. HetSkills begins by learning a tracking skill that establishes a strong foundation in motion control and creates a shared motion decoder, which can be reused across tasks without the need for retraining or separate controllers. To prevent the text-to-motion skill from exploiting shortcut pathways instead of learning language semantics, we introduce motion intuition distillation to ground text-to-motion generation in language semantics and a task-guidance module that dynamically adjusts actions based on high-level language instructions. This enables HetSkills to preserve natural motion while continuously expanding its skill repertoire, making it highly adaptable for long-horizon tasks. Experimental results demonstrate the effectiveness in motion tracking, text-to-motion generation, motion completion, and downstream task adaptation, achieving impressive success rates even under challenging conditions.
\end{abstract}


\begin{teaserfigure}
  \includegraphics[width=1\textwidth]{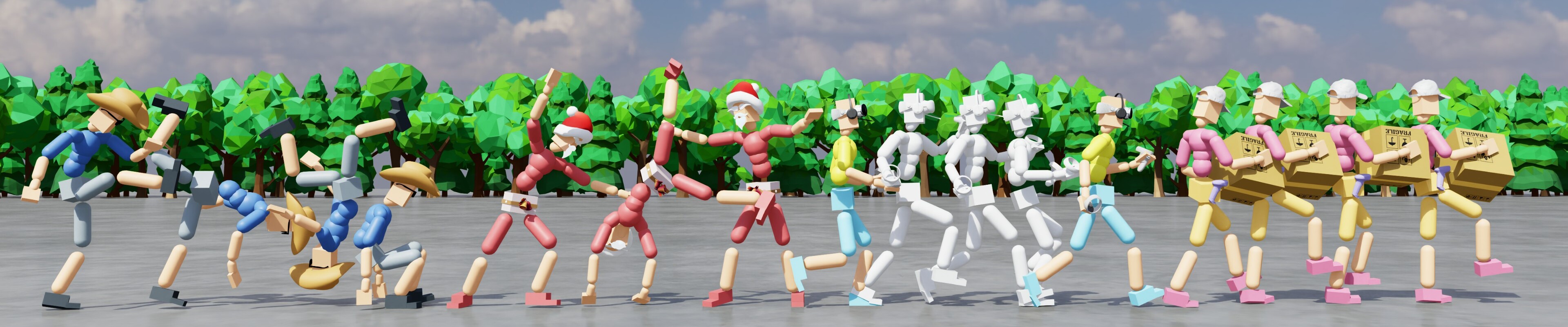}
  \caption{Overview of HetSkills. From left to right, the results illustrate four representative skill categories: motion tracking, text-to-motion generation, motion completion, and downstream task adaptation. Although these skills are learned from heterogeneous data sources, supervision signals, and task objectives, they are all represented in a unified executable latent space and decoded by a shared physics-based motion controller, enabling progressive skill accumulation, reuse, and composition.}
  \label{fig:teaser}
\end{teaserfigure}

\maketitle

\section{Introduction}

Learning physics-based character skills that are natural, composable, and reusable is a fundamental goal in animation, gaming, and robotics. Early approaches~\citep{peng2022ase,tessler2023calm,zhu2023neural} typically learn task-specific controllers by imitating reference motions from small, narrowly curated motion datasets. While effective for individual tasks, such methods offer limited reusability and compositionality. More recent efforts substantially expand the coverage of motion priors and behavior models by leveraging large-scale data~\citep{wu2025uniphys}. However, despite this progress, physics-based character control still lacks a unified and executable skill representation that can support diverse behaviors within a single framework. As a result, it remains difficult to accumulate skills from heterogeneous data sources, compose them flexibly, and transfer them efficiently to new downstream tasks while preserving natural human-like motion.

Large-scale motion priors and language-conditioned control~\citep{tessler2024maskedmimic,juravsky2024superpadl} suggest that language can provide a unified and flexible interface for accessing rich human motion knowledge. This creates the possibility of reusing previously acquired motion capabilities without collecting new demonstrations or training a dedicated controller for every new behavior. Existing methods~\citep{luo2023universal,mu2025smp} partially move in this direction by leveraging pretrained motion priors to guide downstream control. Yet these approaches typically apply pretrained motion models as external guidance or regularization, rather than as a shared skill representation in which newly acquired capabilities can be progressively integrated and directly reused. Consequently, they remain limited when tasks involve heterogeneous objectives, diverse interaction patterns, or behaviors that go beyond the support of the original motion data.

To address the lack of a unified and executable skill representation for progressively accumulating and reusing heterogeneous skills in physics-based character control, we propose HetSkills, a physics-based character control framework that progressively learns heterogeneous skills within a unified latent space. Specifically, heterogeneous skills refer to capabilities acquired from different data sources, supervision forms, and training stages, including motion tracking, text-to-motion generation, motion completion, and downstream task adaptation. Instead of training a separate controller for each capability, HetSkills continuously accumulates these skills into a shared latent representation that serves as a common executable interface for skill acquisition, composition, and reuse. This unified design enables the controller to grow with newly introduced abilities while maintaining a coherent motion prior and a consistent control space.

More specifically, HetSkills is built around a unified latent space that serves as a shared interface for acquiring, composing, and reusing heterogeneous skills. To improve compositional control and generalization, we adopt a part-wise character decomposition~\citep{bae2023pmp,bae2025plt}, which allows different body parts to receive specialized yet coordinated control signals within a common latent architecture. The latent space is anchored by an end-to-end tracking skill that is preserved and directly reused throughout subsequent learning stages, thereby avoiding an additional distillation step and its associated performance degradation. To support robust text-to-motion generation, we further introduce Motion Intuition Distillation (MID), which encourages the model to ground its predictions in language semantics rather than shortcut pathways. For downstream tasks, HetSkills dispenses with reference motions and task-specific demonstrations. It performs lightweight latent-space adaptation, with language specifying high-level behavioral objectives for different body parts within a coordinated whole-body control framework.

Experiments demonstrate the effectiveness of HetSkills in three key aspects. First, HetSkills supports stable accumulation and reuse of heterogeneous skills within a shared latent space, enabling reliable compositional control across behaviors learned from different sources and supervision forms. Second, it achieves robust text-to-motion generation under both standard and challenging initializations, reaching a success rate of $96.9\%$ under standard initialization and $81.7\%$ from a neutral pose. Third, HetSkills transfers effectively to novel downstream tasks through lightweight latent-space adaptation, without requiring task-specific demonstrations or reference motions. These results indicate that a unified latent space can serve as an effective substrate for progressively expanding physics-based character capabilities while preserving natural motion quality.

\section{Related Work}

Physics-based character control is a long-standing focus of research in animation, robotics, and gaming, aiming to generate realistic, reusable, and composable character behaviors \cite{yin2007simbicon,wang2009optimizing,liu2017learning}. Over the years, methods have addressed different aspects of motion learning, including imitation, motion tracking, language-conditioned control, and downstream task adaptation. However, existing methods often struggle with integrating heterogeneous skills into a single, unified framework, requiring either retraining or relying on fragmented modules for new tasks. We summarize the key lines of related work below, situating them with respect to our goal of progressively integrating heterogeneous skills within a unified framework.

\paragraph{Physics-based Motion Imitation and Skill Learning.}
Physics-based character control traditionally relies on motion capture data and reinforcement learning to produce physically plausible behaviors. Early work~\citep{peng2018deepmimic} establishes the foundational paradigm of training tracking policies against reference motion clips, enabling robust imitation of a wide range of physically plausible skills. To reduce reliance on hand-crafted tracking rewards, adversarial approaches~\citep{peng2021amp, peng2022ase, tessler2023calm, zhang2025physics, won2020scalable, dou2023c, hassan2023synthesizing} are introduced to encourage stylistic realism without explicit motion matching, with some also learning reusable latent skill spaces from unstructured motion data for high-level task reuse. In parallel, VAE-based methods~\citep{ling2020character, yao2022controlvae, zhu2023neural, merel2018neural} provide complementary benefits through probabilistic latent modeling, learning structured and diverse skill representations that support behavioral diversity and downstream task reuse. Diffusion-based methods~\citep{mu2025smp, truong2024pdp, huang2025diffuse, serifi2024robot} further broaden the generative toolkit by leveraging pretrained motion diffusion models as either reusable behavioral priors or direct policy parameterizations for physics-based control. Despite these advances, both families of methods are typically designed around a fixed training corpus with homogeneous supervision, making it challenging to incrementally expand the skill repertoire as new data sources or conditioning modalities become available.

\paragraph{Large-scale Motion Priors and Language-conditioned Control.}
To improve coverage and generalization beyond small task-specific datasets, a line of work~\citep{luo2023perpetual, luo2023universal} scales motion tracking to large corpora and distills the acquired motor skills into a universal physics-based latent space, enabling diverse downstream tasks to reuse a shared motor representation. Building on this, methods such as~\citep{tessler2024maskedmimic, wu2025uniphys} unify motion tracking with richer conditioning signals such as language and kinematic constraints within a single model, significantly broadening the interface through which users can direct character behavior. Similarly, works like~\citep{juravsky2022padl, juravsky2024superpadl, ren2023insactor} train physics-based controllers directly conditioned on language commands, scaling to thousands of diverse skills. However, distillation-based approaches may introduce some capability loss relative to the original tracking experts, and adapting these models to complex downstream tasks can remain challenging, as the breadth of the learned skill space makes it difficult to reliably activate task-relevant behaviors without additional guidance.

\begin{figure*}[t]
    \centering
    \includegraphics[width=\textwidth]{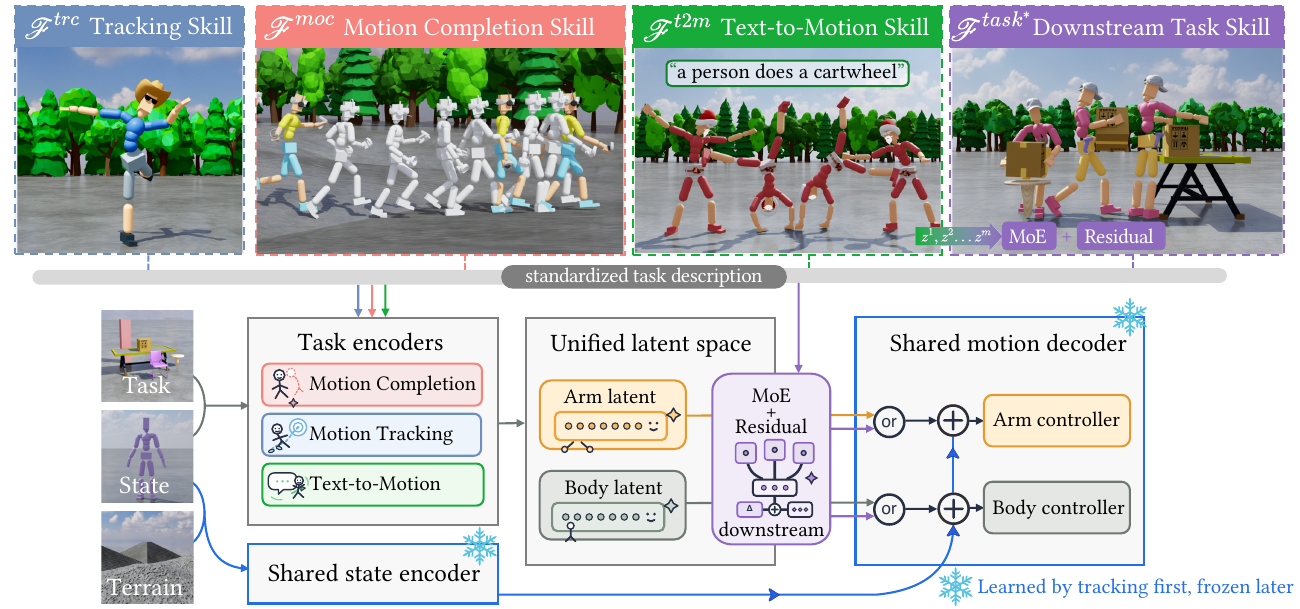}
    \caption{HetSkills progressively integrates four representative heterogeneous skills, including motion tracking $\mathscr{F}^{trc}$, text-to-motion $\mathscr{F}^{t2m}$, motion completion $\mathscr{F}^{moc}$, and downstream task adaptation $\mathscr{F}^{task^*}$, within a unified latent space. Each skill maps its heterogeneous goal signals to part-wise arm and body latents via its own task encoder. After motion tracking learns the shared state encoder and motion decoder, these modules are frozen and reused by all subsequent skills. The downstream task adaptation adopts a language-guided compositional module to blend frozen motion priors with task-conditioned residual corrections, enabling lightweight adaptation without task-specific demonstrations or reference motions.}
    \label{fig:overview}
\end{figure*}

\paragraph{Part-Wise Motion Learning.}
Part-wise decomposition is widely explored in kinematics-based motion generation to improve motion controllability, diversity, and compositionality~\citep{jang2022motion, wan2024tlcontrol}. In physics-based character control, related works further leverage part-wise structure to compose partial motion priors, decouple imitation objectives, or facilitate part-wise planning, enabling the synthesis of more diverse and interaction-rich behaviors~\citep{bae2023pmp, xu2023composite, khoshsiyar2024partwisempc}. Another line introduces part-wise latent priors or modular skill representations within hierarchical control pipelines, enabling more structured composition and improved task adaptability~\citep{bae2025plt, huang2025modskill}. While these works collectively demonstrate the benefits of part-wise decomposition for compositional control and generalization, they typically instantiate part-wise structure within a specific motion-prior, imitation, planning, or modular-control framework. This makes their skill spaces effective for composing behaviors under a fixed interface, but less suited for progressively accommodating heterogeneous skill types, supervision signals, and conditioning modalities within a unified representation.

\paragraph{Downstream Task Adaptation with Motion Priors.}
A natural paradigm for downstream task learning is to freeze a pretrained motion prior and train a high-level policy that operates in the learned latent space~\citep{peng2022ase, tessler2023calm, luo2023universal, zhu2023neural, hassan2023synthesizing}, allowing task policies to inherit motion naturalness without retraining the low-level controller. To better handle complex or contact-rich scenarios, token-based adaptation methods~\citep{pan2025tokenhsi, vainshtein2025task} introduce task-specific tokens as a lightweight interface between a frozen pretrained policy and new task objectives. Another direction repurposes pretrained generative models~\citep{mu2025smp, tevet2024closd} as reusable behavioral priors that provide motion naturalness constraints or planning guidance during task optimization. While these approaches demonstrate the value of structured motion priors for downstream learning, the underlying skill representation is generally fixed after pretraining, and the semantic interface it exposes is limited, making it difficult to flexibly compose and steer behavioral distributions toward diverse downstream objectives without additional supervision or data.

\section{Problem Formulation}

We formulate physics-based character control as a goal-conditioned Markov decision process \cite{liu2022goal}, where a policy acts according to the current state $s_t$ and a goal signal $g_t$. At each timestep $t$, the character executes an action $a_t$, transitions to the next state $s_{t+1} \sim p(\cdot \mid s_t, a_t)$, and receives a scalar reward $r_t$. The objective is to maximize the expected discounted return
\begin{equation}
    J(\pi) = \mathbb{E}\!\left[\sum_{t=0}^{T} \gamma^t r_t\right],
\end{equation}
where $\gamma \in [0,1)$ is the discount factor~\cite{sutton1998reinforcement}. In our setting, actions are target joint rotations that are converted to torques through proportional-derivative (PD) controllers~\cite{tan2011stable} in the physics simulator.

The key challenge is that the goal signal takes heterogeneous forms across skills. Depending on the task, $g_t$ may correspond to future reference poses for motion tracking, language descriptions for text-to-motion generation, sparse temporal constraints for motion completion, or task-specific observations and language conditions for downstream adaptation. These goals differ in modality, supervision, and data distribution, making it difficult to consolidate them under a single controller.

Our objective is therefore to learn a unified latent space that serves as a shared control interface across heterogeneous skills. Instead of training a separate low-level controller for each task, each skill predicts a latent $z_t$ from its own goal specification, and a shared decoder maps $z_t$ and $s_t$ to the final physical action. Under this formulation, the central problem is to progressively construct a latent control space that can incorporate new skills from different training stages while preserving previously learned behaviors and supporting efficient transfer to downstream tasks.

\section{HetSkills: Progressive Heterogeneous Skill Learning}

The proposed HetSkills is a physics-based character control framework that progressively learns heterogeneous skills within a unified latent space. As shown in ~\cref{fig:overview}, the key idea is to treat this latent space as a shared executable interface for skill acquisition, composition, and reuse. Accordingly, skills learned from different data sources, supervision forms, and training stages can be incorporated into a common control representation. To support scalable reuse and downstream transfer, HetSkills further combines this unified latent space with a standardized task interface, allowing newly introduced skills to be integrated without retraining separate low-level controllers.

Specifically, HetSkills begins by learning a tracking skill $\mathscr{F}^{trc}$, which simultaneously constructs the unified latent space and produces the shared motion decoder reused across all subsequent stages. To improve compositional control and generalization, $\mathscr{F}^{trc}$ adopts a part-wise character decomposition that factorizes the humanoid into coordinated body partitions, each governed by a dedicated latent command within a common control architecture. After training, the state encoder and part-wise decoders are frozen and directly reused, avoiding an additional distillation step and its associated performance degradation.

The remaining skills are built on top of this fixed control substrate, each introducing a distinct form of goal conditioning over the shared latent space. $\mathscr{F}^{t2m}$ grounds natural language descriptions into latent motion commands through Motion Intuition Distillation (MID), improving semantic robustness under challenging initializations. $\mathscr{F}^{moc}$ handles sparse or partial observations under a unified sparse-goal formulation, covering VR-driven tracking, motion in-betweening, and human-scene interaction. $\mathscr{F}^{tsk*}$ reuses the frozen text-conditioned prior for downstream tasks, composing part-wise language instructions through a learned routing mechanism without requiring task-specific demonstrations or retraining. Across all stages, skills are formulated under a standardized task description, which provides a skill-agnostic interface for consistent skill integration and flexible sequential composition in long-horizon tasks.

\section{$\mathscr{F}^{trc}$: Tracking Skill} \label{sec:tracking}

We begin by learning a tracking skill $\mathscr{F}^{trc}$ to construct the unified latent space that underlies all subsequent skills. The key idea is to learn a general tracking controller with a compact and reusable latent space that remains expressive for diverse later skills. Therefore, subsequent skills can operate in the same control space without relearning low-level dynamics. To support compositionality, we adopt a part-wise architecture that factorizes the character into coordinated body partitions, while preserving a shared global context.

\begin{figure}[t]
    \centering
    \includegraphics[width=\linewidth,trim=0 0 0 0,clip]{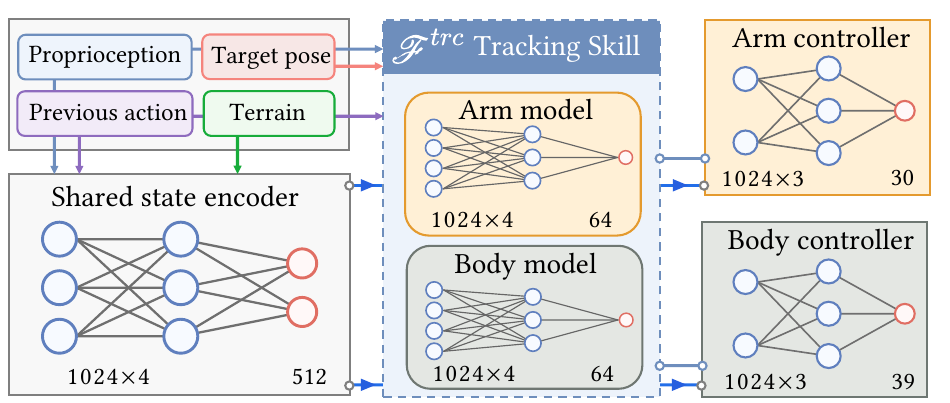}
    \caption{Architecture of the tracking skill $\mathscr{F}^{trc}$. The tracking skill takes proprioception, the previous action, and future target poses as input, then predicts separate latent commands for the arm and body branches. A shared state encoder extracts the current physical context, while part-wise controllers decode the arm and body latents into joint-level target actions. After tracking training, the shared state encoder and part-wise controllers are frozen as the common motion decoder for all later skills, while $\mathscr{F}^{trc}$ itself is retained as one heterogeneous skill.}
    \label{fig:trc_arch}
\end{figure}

\cref{fig:trc_arch} provides an overview of the tracking architecture. The tracker consists of a shared state encoder $\mathcal{E}_s$, two part-wise future encoders $\mathscr{F}^{trc}_a$ and $\mathscr{F}^{trc}_b$, and two part-wise low-level controllers $\mathcal{D}_a$ and $\mathcal{D}_b$. The future encoders predict compact deterministic latent commands $z_a^t$ and $z_b^t$ for the arm and body partitions, which are decoded into joint-level actions by the controllers conditioned on the shared state feature. After training, $\mathcal{E}_s$, $\mathcal{D}_a$, and $\mathcal{D}_b$ are frozen and reused as a shared motion controller across all subsequent stages, avoiding an additional distillation step and its associated performance degradation. The following subsections detail the model representation, part-wise architecture, and training objective in turn.

\subsection{Model Representation}
\begin{figure}[t]
    \centering
    \includegraphics[width=\linewidth,trim=0 0 0 0,clip]{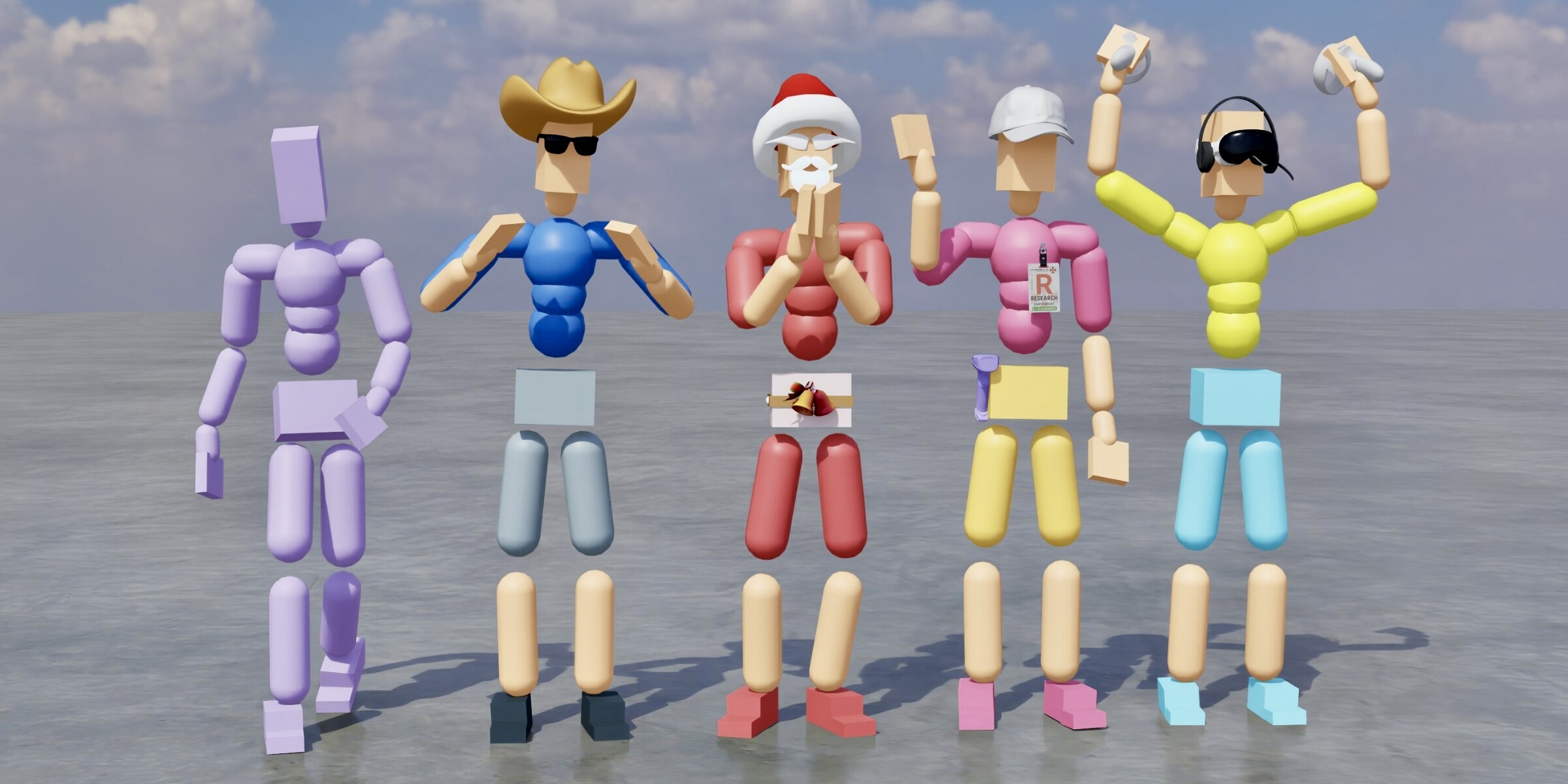}
    \caption{Visualization models of different HetSkills stages. The leftmost character is the physically simulated humanoid used for control, while the other four styled characters visualize the learned skill families, including a cowboy for motion tracking $\mathscr{F}^{trc}$, a Santa for text-to-motion $\mathscr{F}^{t2m}$, a gamer for motion completion $\mathscr{F}^{moc}$, and a courier for downstream task adaptation $\mathscr{F}^{task^*}$. These visual appearances distinguish skills in demonstrations. All skills share the same underlying physical character and latent controller.}
    \label{smpl}
\end{figure}

As shown in ~\cref{smpl}, we use a physically simulated humanoid based on the neutral SMPL~\cite{loper2015smpl} body model with $69$ degrees of freedom, following prior physics-based character control works~\cite{luo2022universal,luo2023perpetual,tessler2024maskedmimic}. All rotations are represented using the continuous 6D parameterization~\cite{zhou2019continuity}. For notational convenience, we apply $\delta_r(\cdot)$ to denote a quantity expressed relative to the current root frame, and $\delta_g(\cdot)$ to denote the difference between a target quantity and its current counterpart.

\paragraph{Proprioception.} The policy observes the current body configuration as
\begin{equation}
    s^t = \bigl(\, \delta_r(\theta^t),\; \dot{\theta}^t,\; \delta_r(v^t),\; h^t_\text{root} \,\bigr),
\end{equation}
where $\theta^t$ and $\dot{\theta}^t$ denote the joint rotations and angular velocities, $v^t$ denotes the linear velocities, and $h^t_\text{root}$ is the root height above the ground.

\paragraph{Goal observation.} The goal input consists of the next $K$ target poses from the reference motion, $g^t = [\hat{f}^{t+1}, \ldots, \hat{f}^{t+K}]$. Each target pose $\hat{f}$ is represented as
\begin{equation}
    \hat{f} = \bigl(\, \delta_r(\hat{p}),\; \delta_g(p),\; \delta_r(\hat{\theta}),\; \delta_g(\theta),\; \delta_g(v),\; \delta_g(\dot{\theta}) \,\bigr),
\end{equation}
where the first two terms denote the target body positions and position errors in the current root frame, the next two denote the target rotations and rotation errors, and the last two denote the linear and angular velocity differences between the current and target poses.

\paragraph{Action Space.} The tracking skill takes the current proprioception $s^t$, future reference goal $g^t$ and the previous action $a^{t-1}$ as input, and predicts part-wise latent actions in the corresponding latent space:
\begin{equation}
    z_p^t = \mathscr{F}^{trc}_p(s^t, g^t, a^{t-1}), \qquad p \in \{a,b\},
    \label{eq:trc_latent}
\end{equation}
where $z_a^t$ and $z_b^t$ denote the latent actions for the arm and body partitions, respectively. These latent actions are then decoded into joint-level target actions through the corresponding state feature:
\begin{equation}
    c^t = \mathcal{E}_s(s^t), \quad
    a^t = \bigl[\mathcal{D}_a(c^t, z_a^t),\; \mathcal{D}_b(c^t, z_b^t)\bigr],
    \label{eq:trc_action}
\end{equation}
where $\mathcal{E}_s$ is the shared state encoder, and $\mathcal{D}_a$ and $\mathcal{D}_b$ are the part-wise low-level controllers for the arm and body partitions. The final action $a^t$ consists of target joint rotations for all actuated degrees of freedom, which are converted to joint torques via PD controllers~\cite{tan2011stable}. PD controllers are widely used in physics-based character animation~\cite{xu2025parc,yu2025skillmimic,peng2018deepmimic} because they provide a stable and intuitive interface between learned policies and physical simulation, abstracting away low-level torque computation while remaining responsive to perturbations.

\subsection{Part-wise Architecture}
Part-wise architecture is designed to improve compositionality and latent controllability while preserving coordinated whole-body behavior. The core idea is to factorize the humanoid into a small number of semantically meaningful body partitions, and to let each partition predict its own latent action within a shared control framework. This design encourages specialization for different motion patterns, while maintaining global consistency through a common state representation. In addition, we construct the latent space to be deterministic and compact. Therefore, it can serve as a stable substrate for subsequent skill learning and downstream composition.

\paragraph{Part-wise Decomposition.} We decompose the humanoid into two kinematic partitions: an \emph{arm} part that groups both hands and arms, and a \emph{body} part that contains all remaining joints. This decomposition is motivated by the observation that motion datasets often include fine-grained upper-limb behaviors, such as gesturing, punching, and object interaction, whose high-frequency and low-inertia dynamics differ substantially from those of locomotion and balance. A dedicated arm branch therefore allows the model to specialize in these behaviors without entangling them with lower-body control. Compared with finer-grained decompositions~\citep{huang2025modskill,bae2025plt}, the two-part split provides a favorable trade-off between expressiveness and simplicity. It achieves strong tracking quality across diverse motion categories while avoiding redundant parameters and excessive latent channels that would complicate downstream skill composition. We validate this design choice in~\cref{result:trc_and_moc}.

\paragraph{Network Structure.} With this factorization, the tracking policy consists of five modules, including a shared state encoder $\mathcal{E}_s$, two part-wise future encoders $\mathscr{F}_{a}^{trc}$ and $\mathscr{F}_{b}^{trc}$, and two part-wise low-level controllers $\mathcal{D}_a$ and $\mathcal{D}_b$. At each timestep, $\mathcal{E}_s$ maps the current observation to a shared state feature $c^t$ that captures the full-body context. In parallel, $\mathscr{F}_{a}^{trc}$ and $\mathscr{F}_{b}^{trc}$ each take a two-frame window of future reference motion and predict compact latent commands $z_a^t$ and $z_b^t$ for the arm and body partitions, respectively. We adopt two future frames because this improves tracking accuracy and provides stronger supervision when the tracker later serves as an expert. At inference time, a single reference frame is sufficient by duplicating it to form the two-frame input. The shared feature $c^t$ is concatenated with each part-wise latent and passed to the corresponding controller to produce joint-level actions. Because both controllers condition on the same shared state feature, the two branches retain access to a consistent global body context and can coordinate without explicit cross-branch communication. At the same time, the separate latent commands and decoders allow each branch to specialize in its own body partition. Therefore, the latent action captures the intended motion while the shared feature anchors it to the current physical state.

\begin{figure}[t]
    \centering
    \includegraphics[width=\linewidth,trim=0 0 0 0,clip]{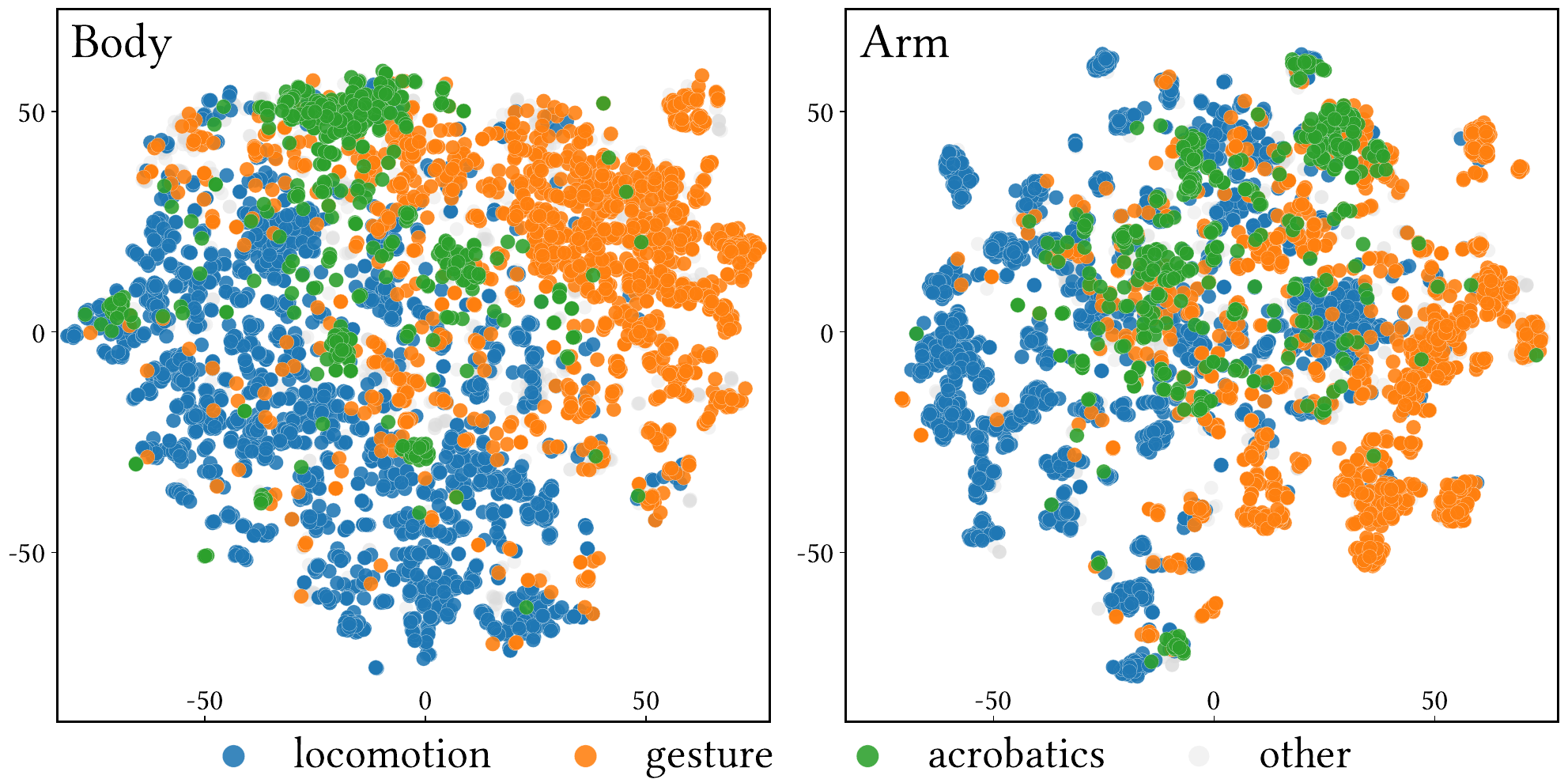}
    \caption{Learned body and arm latent spaces. t-SNE projections show that motions from different categories, including locomotion, gesture, acrobatics, and other behaviors, form distinguishable distributions, even though no explicit category labels are used during latent learning. This indicates that the deterministic and compact latent representation captures semantically meaningful motion structure, which supports later skill reuse and composition.}
    \label{fig:tsne}
\end{figure}

\paragraph{Deterministic Latent Space.} Existing prior latent-variable policies ~\citep{liu2022motor,ling2020character,tessler2024maskedmimic} typically model the latent space as a stochastic distribution, while our future encoders produce deterministic latent representations \cite{holden2017phase,zhang2018mode}. In stochastic formulations, exploration noise is injected directly into the latent space \cite{plappert2017parameter}, which encourages the encoder to increase latent magnitudes in order to preserve discriminability among different skills. This can degrade tracking precision, destabilize the latent scale, and introduce an additional sensitive KL-divergence coefficient \cite{kingma2013auto,burgess2018understanding}. We instead keep the latent space deterministic and inject exploration noise only at the final action space, where it promotes exploration without distorting the learned latent geometry. To further regularize the latent representation, we apply two complementary objectives: a latent magnitude penalty that encourages compact representations, and an AR(1) temporal smoothness penalty~\citep{merel2018neural} that discourages abrupt latent changes between consecutive timesteps:
\begin{equation}
    \mathcal{L}_{mr} = \sum_{p \in \{a,b\}} \mathbb{E}\left[\|z_{p}^t\|_2^2\right], \quad
    \mathcal{L}_{ar} = \sum_{p \in \{a,b\}} \mathbb{E}\left[\|z_{p}^t-\phi z_{p}^{t-1}\|_2\right].
    \label{eq:tracker_reg}
\end{equation}
Together, the deterministic design and these regularizers yield a compact and well-structured latent space. As illustrated by the t-SNE visualization in ~\cref{fig:tsne}, motions from different categories, including locomotion, gesture, acrobatics, and other behaviors, exhibit distinct distributional tendencies in the learned latent space, suggesting that the representation captures semantically meaningful structure without explicit category supervision. This structure is also beneficial for downstream skill composition, since later modules only need to predict a latent point per part rather than match an entire latent distribution.

\subsection{Training Objective}

The tracking skill is trained to imitate reference motions while simultaneously shaping the latent space to be compact, smooth, and reusable. To this end, we optimize the controller with reinforcement learning using motion-tracking rewards, and augment the policy objective with latent-space regularization. We further adopt sampling and termination strategies that improve training efficiency on large and diverse motion datasets.

We train the tracking policy with proximal policy optimization (PPO) \cite{schulman2017proximal} to imitate reference motions. Each reward term takes the form $r(x, k) = \exp(-k\|x\|)$. The full tracking reward is defined as
\begin{equation}
    r^t = \sum_{j \in \mathcal{J}} w^{j}\, r\!\left(\delta_g(x^{t,j}),\, k^{j}\right)  + w^{ct} r^{t,ct} + w^{sm} r^{t,sm} + w^{eg} r^{t,eg},
    \label{eq:reward}
\end{equation}
where $\mathcal{J} = \{gp, gr, jv, jav, rh\}$ corresponds to global joint positions, global joint rotations, joint velocities, joint angular velocities, and root height, respectively. Here, $r^{t,ct}$ denotes the contact reward, which encourages correct foot contact behavior. The final two terms correspond to an action smoothness penalty and an energy penalty, which jointly encourage smoother motions. Detailed reward weights and coefficients are provided in the supplementary material.

The overall training objective combines the PPO loss with the latent regularization terms introduced in ~\cref{eq:tracker_reg}:
\begin{equation}
    \mathcal{L}_{trc} = \mathcal{L}_{\mathrm{ppo}} + \lambda_{mr}\mathcal{L}_{mr} + \lambda_{ar}\mathcal{L}_{ar}.
    \label{eq:tracker_loss}
\end{equation}
This objective improves tracking fidelity and encourages the learned latent actions to remain compact and temporally coherent, which is important for later skill reuse and downstream composition.

To improve training efficiency on diverse motion datasets, we additionally adopt early termination~\citep{peng2018deepmimic} and prioritized motion sampling~\citep{luo2023universal}. An episode is terminated when any joint position deviates from the reference beyond a predefined threshold, preventing training from being dominated by undesirable states. Motions with higher failure rates are sampled more frequently, allowing training to focus on challenging and underrepresented behaviors. After training, we freeze $\mathcal{E}_s$, $\mathcal{D}_a$, and $\mathcal{D}_b$ and reuse them as the shared decoder in subsequent stages.

\section{$\mathscr{F}^{t2m}$: Text-to-Motion Skill}
\label{sec:t2m}

\begin{figure}[t]
    \centering
    \includegraphics[width=\linewidth,trim=0 0 0 0,clip]{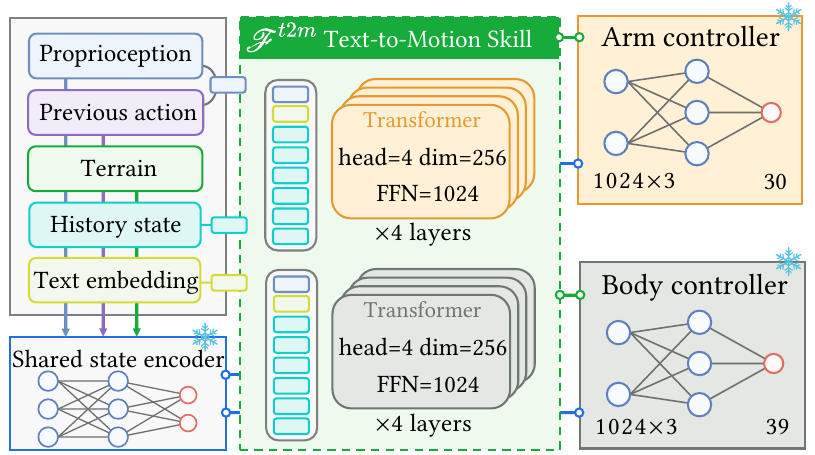}
    \caption{Architecture of the text-to-motion skill $\mathscr{F}^{t2m}$. The frozen TMR text encoder provides a semantic language embedding, while the current proprioception-action input and recent history states are projected into tokens. Two part-wise Transformer encoders predict arm and body latents in the shared latent space, which are then decoded by the frozen state encoder and part-wise controllers into physical actions. This design isolates language grounding from low-level control and allows text-conditioned motion generation to reuse the tracking-learned controller.}
    \label{fig:t2m_arch}
\end{figure}

We learn a text-to-motion skill $\mathscr{F}^{t2m}$ that maps natural language descriptions to part-wise latent actions within the unified latent space, with the low-level controller frozen from the tracking skill. The goal is to ground natural language descriptions into reusable latent actions while preserving the motion quality and controllability provided by the tracking decoder. A major challenge is that text-conditioned models can easily exploit shortcut pathways, such as privileged future-motion cues or overly regular training initialization, instead of learning meaningful language semantics. Our design therefore emphasizes semantic grounding under imperfect context, so that the learned skill remains robust under changed initial states, mismatched histories, and other challenging inference conditions.

The architecture of the text-to-motion skill is illustrated in \cref{fig:t2m_arch}. It consists of two part-wise text encoders $\mathscr{F}^{t2m}_a$ and $\mathscr{F}^{t2m}_b$, which predict latent actions for the arm and body partitions, respectively, while reusing the frozen state encoder $\mathcal{E}_s$ and part-wise controllers $\mathcal{D}_a$ and $\mathcal{D}_b$ from the tracking stage. The following subsections detail MID, the model architecture, and the training objective in turn.

\subsection{Motion Intuition Distillation}
\label{sec:mid}
The key idea of Motion Intuition Distillation(MID) is to remove those training shortcuts that would otherwise allow the model to bypass semantic understanding. Instead of predicting motion under near-perfect future guidance or always starting from a matched initial state, the model must infer the intended behavior from language, the current state, and imperfect historical context. This encourages the text-conditioned policy to acquire a more robust motion intuition that generalizes beyond the training distribution.

\paragraph{Shortcut Issue.} A major challenge in learning text-to-motion skills within a unified latent space is that the model can easily rely on shortcut pathways instead of learning meaningful language semantics. When future motion information is indirectly accessible, whether through residual branches, future-motion conditions, or other forms of privileged information, the model tends to shift the main modeling burden to these easier pathways, since predicting the next latent from nearby future frames requires far less abstraction than grounding language descriptions into motion. As a result, the residual structure may acquire overly strong compensatory behavior that effectively bypasses the text encoder, while the component responsible for semantic understanding remains insufficiently trained. In addition, when training episodes always start from the ground-truth initial frame together with its matching motion history, the model can overfit to local state-transition patterns and achieve high in-distribution accuracy without learning to handle changed initial states, mismatched histories, or motion transitions.

\paragraph{Formulation.} To address this issue, we propose MID, which removes dependence on future motion and standard initialization. The text-to-motion encoders $\mathscr{F}^{t2m}_a$ and $\mathscr{F}^{t2m}_b$ receive only the current state, past history, and the text embedding. As a result, the model infers the intended motion from the language condition, the current state, and imperfect historical context. We first adopt Reference State Initialization (RSI)~\citep{peng2018deepmimic}, where, with probability $p_{\mathrm{rsi}}=0.7$, the initial state of a training episode is sampled uniformly from any frame along the reference clip rather than taken from the first frame of the target motion. We further introduce Randomized Memory Initialization (RMI). With probability $p_{\mathrm{rmi}}=0.2$, we uniformly sample another motion sequence from the motion library, use its terminal frame as the initial state, and fill the history buffer with a real history segment from the end of that randomly sampled sequence. Under this training scheme, the model cannot always rely on an initial frame and history prefix that strictly match the target motion. Instead, it learns to infer the subsequent behavior from language semantics and the current observation. We validate the necessity of each component in~\cref{result:Text2Motion}.

\subsection{Model Architecture}
The text-to-motion architecture reuses the latent space and shared decoder learned by the tracking skill, and only learns the mapping from text-conditioned observations to latent actions. This design keeps the low-level control substrate fixed and shifts the learning burden to semantic grounding in latent space. As a result, the model can leverage a strong motion prior learned from large unstructured motion data while adapting it to language supervision with comparatively limited text-annotated data. Concretely, at each step the skill receives the proprioception $s^t$, the previous action $a^{t-1}$, an observation history $H^t$, and a text embedding $e$, and maps them to part-wise latent actions:
\begin{equation}
    z_p^t = \mathscr{F}^{t2m}_p(s^t, a^{t-1}, H^t, e), \qquad p \in \{a,b\},
    \label{eq:t2m_latent}
\end{equation}
where $z_a^t$ and $z_b^t$ denote the latent actions predicted for the arm and body partitions, respectively. These latent actions are then decoded into the final joint-level action through the frozen shared decoder:
\begin{equation}
    a^t = \bigl[\mathcal{D}_a(\mathcal{E}_s(s^t), z_a^t),\; \mathcal{D}_b(\mathcal{E}_s(s^t), z_b^t)\bigr].
    \label{eq:t2m_action}
\end{equation}

\paragraph{Input Representation.}
The proprioception $s^t$ follows the same definition as in the tracking skill. The text embedding $e$ is obtained from a frozen Text-to-Motion Retrieval (TMR) text encoder~\cite{petrovich2023tmr}, which is trained through contrastive learning to align language and motion in a shared embedding space. This provides a semantically structured representation that facilitates grounding text descriptions into latent motion commands. The observation history $H^t$ consists of six frames uniformly subsampled from the past two seconds of simulation, providing sufficient temporal context for the model to infer the current motion phase and dynamics.

\paragraph{Skill Encoders.} Two part-wise Transformer~\citep{vaswani2017attention} encoders $\mathscr{F}^{t2m}_a$ and $\mathscr{F}^{t2m}_b$ take $s^t$, $a^{t-1}$, $e$, and $H^t$ as input and predict latent actions $z_a^t$ and $z_b^t$ in the arm and body latent spaces, respectively. The two encoders share the same architecture but maintain separate parameters, enabling each branch to specialize in the dynamics of its corresponding body partition. The predicted latents are then decoded by the frozen state encoder $\mathcal{E}_s$ and part-wise controllers $\mathcal{D}_a$ and $\mathcal{D}_b$ to produce joint-level actions.

\paragraph{Shared Decoder Reuse.} Freezing the decoder provides two practical benefits. First, it removes the need to jointly learn low-level control during text-to-motion training, reducing the optimization problem to latent prediction alone. Second, it allows the text-to-motion skill to build on top of a controller already trained on a large and diverse motion dataset. Therefore, even a comparatively small text-annotated dataset can suffice to learn semantically grounded motion skills on top of a general-purpose motor foundation.

\subsection{Training Objective}
We train the text-to-motion encoders using the frozen tracking skill $\mathscr{F}^{trc}$ as the expert teacher \cite{ross2011reduction}. During training, the text-to-motion skill autonomously interacts with the environment, and at each time step, $\mathscr{F}^{trc}$ observes the ground-truth future reference frames to provide the expert action $\hat{a}^t$ for the current state. Rather than supervising in the latent space, we compare the final joint-level actions $a^t$ and $\hat{a}^t$ produced after decoding through the shared controllers. Therefore, the loss naturally accounts for the nonlinear mapping from latent to action space. The overall training objective is
\begin{equation}
    \mathcal{L}_{t2m} = \mathbb{E}\!\left[\|a^t - \hat{a}^t\|_2\right]
    + \lambda_{mr}\mathcal{L}_{mr} + \lambda_{\mathrm{ar}}\mathcal{L}_{\mathrm{ar}}.
\end{equation}
The two regularization terms follow the same form as in~\cref{eq:tracker_reg}, penalizing latent magnitude and encouraging temporal smoothness. These regularizers help maintain a compact and well-structured latent distribution. Since our downstream module predicts residual latent adjustments on top of the text-to-motion output, a well-regularized base distribution makes such residual learning more stable and effective.

\section{$\mathscr{F}^{moc}$: Motion Completion Skill}

We introduce $\mathscr{F}^{moc}$ as a motion completion skill that extends the latent space to tasks requiring the recovery of physically plausible full-body motion from sparse or partial observations~\citep{harvey2020robust,cohan2024flexible,qin2022motion,oreshkin2023motion,hwang2025motion}. These tasks differ in the form of their conditioning signals. However, they all require the controller to infer coherent whole-body behavior from incomplete information. Some involve spatially sparse observations, as in VR-driven body tracking where only a small set of end-effector trajectories is available, while others involve temporally sparse observations, as in motion in-betweening where only scattered keyframe poses are provided. The key idea of this stage is to express these diverse signals uniformly as partial goal specifications and handle them within the same latent space.

The motion completion skill therefore serves as a unified control interface for sparse-observation tasks. Instead of introducing a separate controller for each conditioning type, we learn a single skill family that maps the current proprioception together with sparse target observations to latent actions compatible with the shared decoder. In this work, we instantiate this framework in three representative settings, including VR tracking, motion in-betweening, and human-scene interaction. These cases cover both spatially sparse and temporally sparse conditioning, and together illustrate that the shared latent space can support a broader family of goal specifications beyond motion tracking and language-guided generation.

\subsection{Unified Formulation}
Let $\mathcal{O}^t = \{(t_k,\tilde{g}_k^t)\}_{k=1}^{K}$ denote the sparse target observations, where each $t_k$ is a relative time offset and $\tilde{g}_k^t$ is the corresponding partial goal feature. The motion completion skill takes the current proprioception $s^t$, the previous action $a^{t-1}$, and the sparse observation set $\mathcal{O}^t$ as input, and predicts part-wise latent actions:
\begin{equation}
    z_p^t = \mathscr{F}^{moc}_p(s^t, a^{t-1}, \mathcal{O}^t), \qquad p \in \{a,b\},
    \label{eq:moc_latent}
\end{equation}
where $z_a^t$ and $z_b^t$ denote the latent actions for the arm and body partitions, respectively. The resulting latent actions are decoded into the final joint-level target action through the shared decoder:
\begin{equation}
    c^t = \mathcal{E}_s(s^t), \qquad
    a^t = \bigl[\mathcal{D}_a(c^t, z_a^t),\; \mathcal{D}_b(c^t, z_b^t)\bigr].
    \label{eq:moc_action}
\end{equation}

\paragraph{VR Tracking.} In VR tracking, the conditioning signal consists of the full kinematic state of three end-effector bodies, namely the head and both hands, from the next reference frame. This corresponds to a spatially sparse observation setting, where only a small subset of body parts is directly specified and the controller infers the remaining full-body motion. We train the skill with a behavior cloning objective,
\begin{equation}
    \mathcal{L}_{\mathrm{BC}} = \mathbb{E}\!\left[\|a^t - \hat{a}^t\|_2\right],
\end{equation}
using the stage-1 tracker as the expert teacher. We additionally apply the latent magnitude regularization term $\mathcal{L}_{mr}$ from ~\cref{eq:tracker_reg}. The AR(1) smoothness term $\mathcal{L}_{ar}$ is omitted in this case, since strong immediate fidelity to sparse spatial targets is more important than long-horizon temporal smoothing.

\paragraph{Motion In-betweening.} In motion in-betweening \cite{starke2023motion,kim2022conditional,kaufmann2020convolutional}, the conditioning signal consists of a future full-body pose together with its time offset $\tau^t$. The target frame is uniformly sampled from a future horizon of $5$ to $30$ frames, and once the character reaches it, a new target frame is resampled from the same range. This corresponds to a temporally sparse observation setting, where the controller synthesizes plausible intermediate motion that connects scattered target poses. Training follows the same behavior cloning objective $\mathcal{L}_{\mathrm{BC}}$, while both $\mathcal{L}_{mr}$ and $\mathcal{L}_{ar}$ are applied to encourage compact latent representations and temporally coherent transitions across keyframes.

\begin{figure}[t]
    \centering
    \includegraphics[width=\linewidth,trim=0 0 0 0,clip]{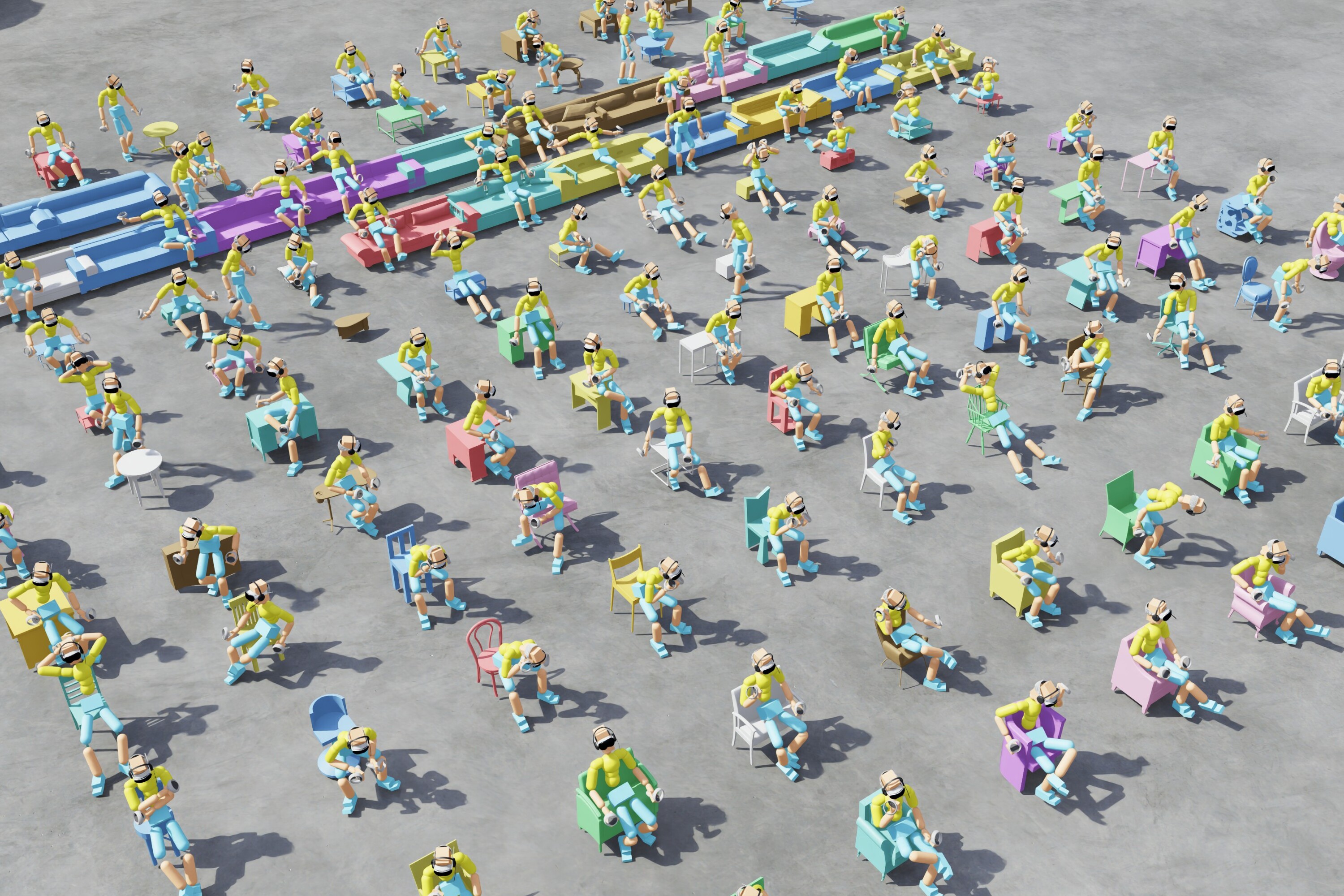}
    \caption{Examples of human-scene interaction generated by HetSkills. The characters interact with diverse everyday objects such as chairs, sofas, and tables, producing sitting, reclining, and other object-conditioned motions within the same shared control framework.}
    \label{fig:samp}
\end{figure}

\paragraph{Human-scene Interaction.} Human-scene interaction \cite{starke2019neural,hassan2021populating} can be cast under the same formulation as motion in-betweening \cite{hwang2025scenemi}. The conditioning signal consists of a future target interaction state together with its time offset, sampled from the same rolling future horizon. We train this skill on the SAMP dataset~\citep{hassan2021stochastic}, which contains motions of characters interacting with everyday objects such as chairs and sofas, shown in \cref{fig:samp}. SAMP lies outside the distribution of the tracker training data. However, the generalization capacity of the shared latent space allows the same control framework to be reused without modification. For this setting, we adopt reinforcement learning instead of behavior cloning, as direct optimization against a tracking-style reward yields stronger performance under distribution shift. A key advantage of the unified latent space is that each motion completion skill can be trained independently on its own motion distribution while remaining fully compatible with the same shared control interface.

\section{$\mathscr{F}^{tsk^*}$: Language-Guided Downstream Adaptation}

We introduce $\mathscr{F}^{tsk^*}$ as a language-guided downstream adaptation skill that reuses the language-conditioned motion distribution induced by $\mathscr{F}^{t2m}$ for new task objectives. Rather than retraining the motion prior or the low-level controllers, $\mathscr{F}^{tsk^*}$ learns lightweight task-specific guidance on top of the shared latent space. Therefore, downstream behaviors remain natural and human-like while adapting to novel tasks. To accommodate heterogeneous downstream objectives, this stage combines a standardized task interface with compositional guidance based on part-wise language priors and task-conditioned residual correction.

Concretely, $\mathscr{F}^{tsk^*}$ consists of three components: a standardized task description that provides a unified interface for representing and organizing tasks, a compositional task guidance module that combines multiple part-wise language priors under the current task context, and a lightweight adaptation objective that optimizes only the downstream guidance modules while keeping the pretrained motion prior fixed. The following subsections detail these three components in turn.

\subsection{Standardized Task Description}
\label{sec:task_description}

To support heterogeneous skill learning and composition, HetSkills represents all skills through a unified task interface. Specifically, both previously learned skills and downstream task skills are formulated using a standardized task description that is independent of their specific training procedures. This abstraction provides a consistent way to specify, organize, compose, and extend skills, and further serves as the basis for long-horizon task execution in the shared latent space.

\paragraph{Task Unit.} We abstract the execution of each skill as a standardized task unit
\begin{equation}
    \mathcal{T}_i = \big(\mathrm{Initialization}_i,\ \mathrm{Condition}_i,\ \mathrm{Terminate}_i\big),
\end{equation}
where each component plays a distinct role. \textit{Initialization} specifies the initialization protocol for the skill, including any required state resets, memory initialization, or environment configuration that must be established before execution begins. \textit{Condition} encodes the task-specific guidance that governs the behavior during execution. Depending on the skill type, this may take the form of a language instruction, a target goal state, a reference trajectory, or another modality that parameterizes the desired behavior. \textit{Terminate} defines the criterion under which the skill is considered complete, such as a fixed execution horizon, a goal-reaching condition, or a learned termination signal.

\paragraph{Task composition.} This interface is deliberately agnostic to skill type. A language-conditioned motion generation skill, a human-scene interaction skill, and a goal-conditioned locomotion skill can all be expressed under the same $\mathcal{T}_i$ abstraction, differing only in how each component is instantiated. Since all skills share this interface, they can be executed independently or organized sequentially as
\begin{equation}
    \mathrm{Seq}(\mathcal{T}_1, \mathcal{T}_2, \ldots, \mathcal{T}_N),
\end{equation}
which provides a unified mechanism for sequential scheduling and long-horizon skill composition.

\subsection{Compositional Task Guidance}
A single language condition is often insufficient to represent the full complexity of a downstream task, especially when the desired behavior involves blending multiple motion styles or switching between behaviors over time. Our solution is to combine multiple part-wise language priors through a learned routing mechanism and then refine the resulting latent action with a task-conditioned residual. This produces a flexible adaptation module that remains grounded in the pretrained motion prior while retaining task-specific expressiveness.

\paragraph{Instruction Set.} For each body part, we consider a set of $M$ language conditions
\begin{equation}
    \mathcal{I}_{p} = \{e_{p}^{m}\}_{m=1}^{M}, \qquad p \in \{a,b\},
\end{equation}
where each instruction $e_{p}^{m}$ provides a distinct semantic description of the target behavior. Conditioned on these part-wise language inputs, the corresponding frozen language priors produce multiple latent candidates:
\begin{equation}
    z_{p}^{t,m} = \mathscr{F}_{p}^{t2m}(s^t, a^{t-1}, H^t, e_{p}^{m}),
    \qquad p \in \{a,b\}.
\label{eq:prior}
\end{equation}

\paragraph{Latent Routing.} Since the most appropriate motion style depends on the current task context, we learn a gating function $\mathcal{G}_{p}$ for each body part. It takes the current proprioceptive state and a task-specific observation $o^t_{\mathrm{task}}$ as input and outputs normalized routing weights over the $M$ instruction branches:
\begin{equation}
    \boldsymbol{\alpha}_{p}^{t} = \mathrm{softmax}\ \!\big(\mathcal{G}_{p}(s^t, a^{t-1}, o^t_{\mathrm{task}})\big), \qquad p \in \{a,b\}.
\label{eq:gating}
\end{equation}
Here, $\boldsymbol{\alpha}_{p}^{t} = \{\alpha_{p}^{t,m}\}_{m=1}^{M}$ denotes the routing weights, where $\alpha_{p}^{t,m}$ is the weight assigned to the $m$-th branch for part $p$.

\paragraph{Residual Refinement.} While a weighted combination of language-prior latents approximates the target motion style, the resulting latent remains loosely coupled to task-specific objectives, as the language priors encode general motion distributions without direct awareness of task constraints. We therefore introduce a task-conditioned residual module $\mathcal{R}_p$ for each body part and define the final part-wise latent as
\begin{equation}
    \hat{z}_p^t = \sum_{m=1}^{M} \alpha_p^{t,m} z_p^{t,m} + \mathcal{R}_p(s^t, a^{t-1}, o^t_{\mathrm{task}}), \qquad p \in \{a,b\}.
\label{eq:aggregation}
\end{equation}
The resulting latents $\hat{z}_a^t$ and $\hat{z}_b^t$ are then decoded by the frozen part-wise controllers using the same procedure as in the pretrained model.

\begin{algorithm}[t]
\SetAlgoLined
\KwIn{Frozen priors $\mathscr{F}_{p}^{t2m}$, frozen decoder $\mathcal{D}$;
      instruction sets $\mathcal{I}_p = \{e_p^m\}_{m=1}^{M}$, $p\in\{a,b\}$;
      gating networks $\mathcal{G}_p$, residual networks $\mathcal{R}_p$, critic $V_\phi$;
      fixed regularization coefficients $\lambda_{\mathrm{lmp}}, \lambda_{\mathrm{smooth}}, z_{\mathrm{bound}}$}
\KwOut{Trained MoE gating networks $\mathcal{G}_p$, part-wise residual networks $\mathcal{R}_p$, and value network $V_\phi$}
\For{\emph{each training iteration}}{
    \For{\emph{each environment} $i$ \emph{(parallel)}}{
        Observe $s^t$, $a^{t-1}$, $H^t$, $o^t_{\mathrm{task}}$\;
        \For{\emph{each language condition} $e_p^m \in \mathcal{I}_p$}{
            Compute $z_p^{t,m}$ via Eq.~\eqref{eq:prior}\;
        }
        Compute gating weights $\boldsymbol{\alpha}_p^t$ via Eq.~\eqref{eq:gating}\;
        Compute aggregated latent $\hat{z}_p^t$ via Eq.~\eqref{eq:aggregation}\;
        $\hat{z}^t \leftarrow [\hat{z}_a^t;\, \hat{z}_b^t]$\;
        Sample $z^t \sim \mathcal{N}(\hat{z}^t, \mathrm{diag}(\sigma^2))$, store $\log\pi_\theta(z^t)$\;
        $a^t \leftarrow \mathcal{D}(s^t, z^t)$, collect reward $r^t$, store transition\;
    }
    Compute advantages $\hat{A}^t$ and returns $\hat{R}^t$ via GAE\;
    \For{\emph{each mini-batch from rollout buffer}}{
        Compute $\mathcal{L}_{\mathrm{tsk}}$ via Eq.~\eqref{eq:update}\;
        Update $\mathcal{G}_p, \mathcal{R}_p, V_\phi$ via $\nabla\mathcal{L}_{\mathrm{tsk}}$\;
    }
}
\Return $\mathscr{F}^{tsk^*} = \{\mathcal{I}_p, \mathcal{G}_p, \mathcal{R}_p\}_{p\in\{a,b\}}$\;
\caption{Language-Guided Downstream Adaptation}
\label{alg:lgda}
\end{algorithm}

\subsection{Task Adaptation}
\begin{figure*}[t]
    \centering
    \begin{subfigure}[t]{\textwidth}
        \includegraphics[width=0.0990\linewidth]{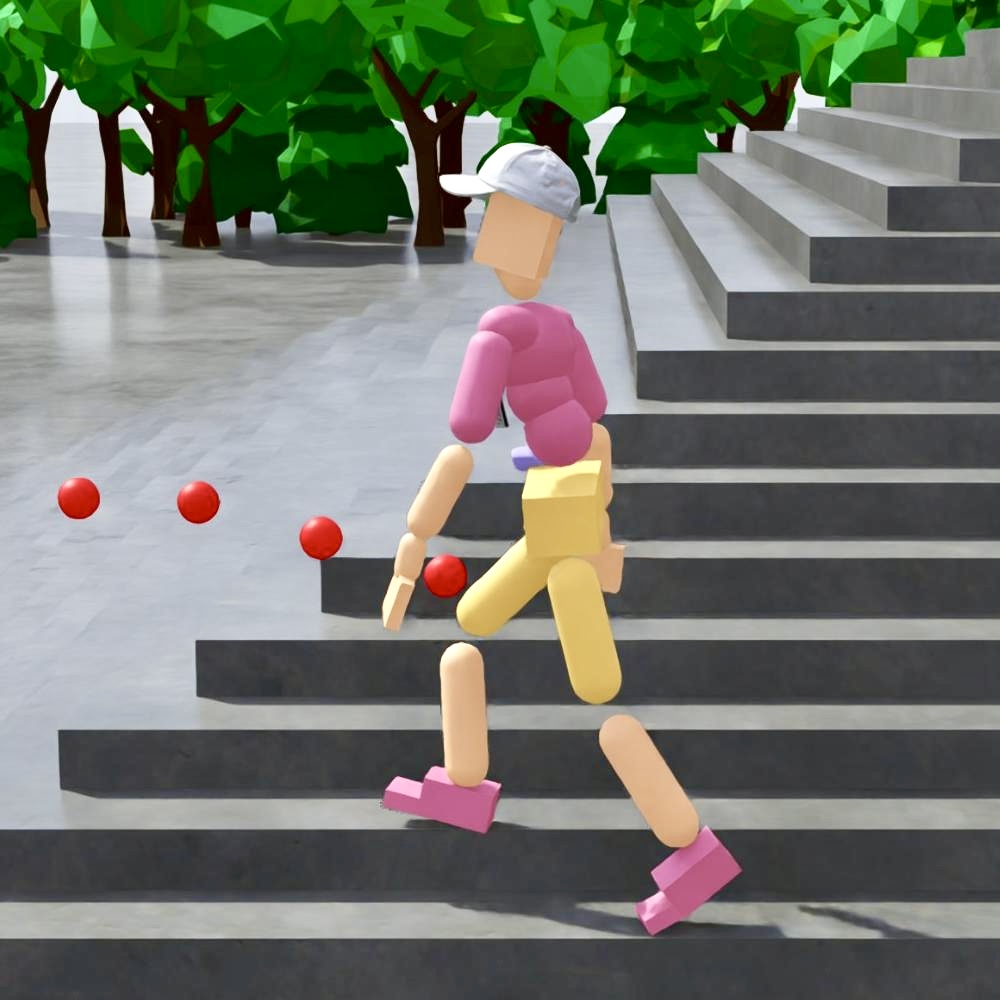}\hfill
        \includegraphics[width=0.0990\linewidth]{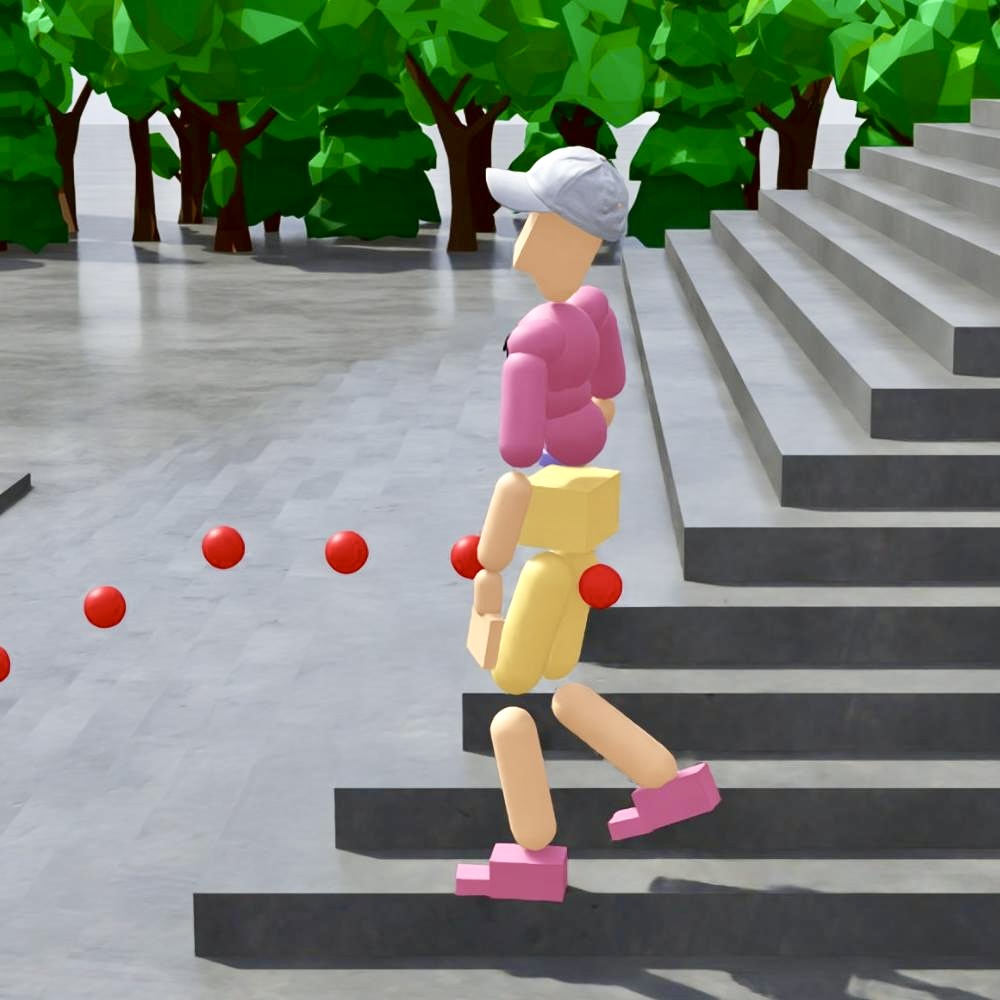}\hfill
        \includegraphics[width=0.0990\linewidth]{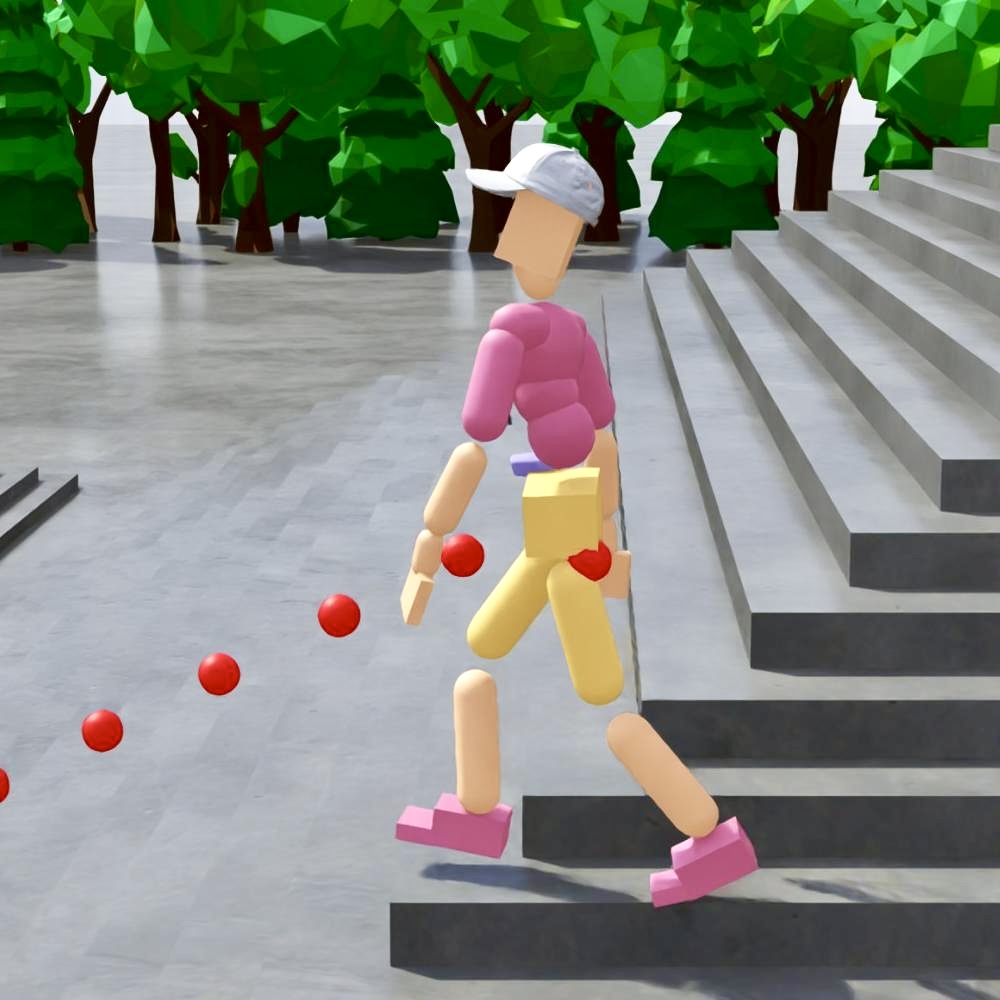}\hfill
        \includegraphics[width=0.0990\linewidth]{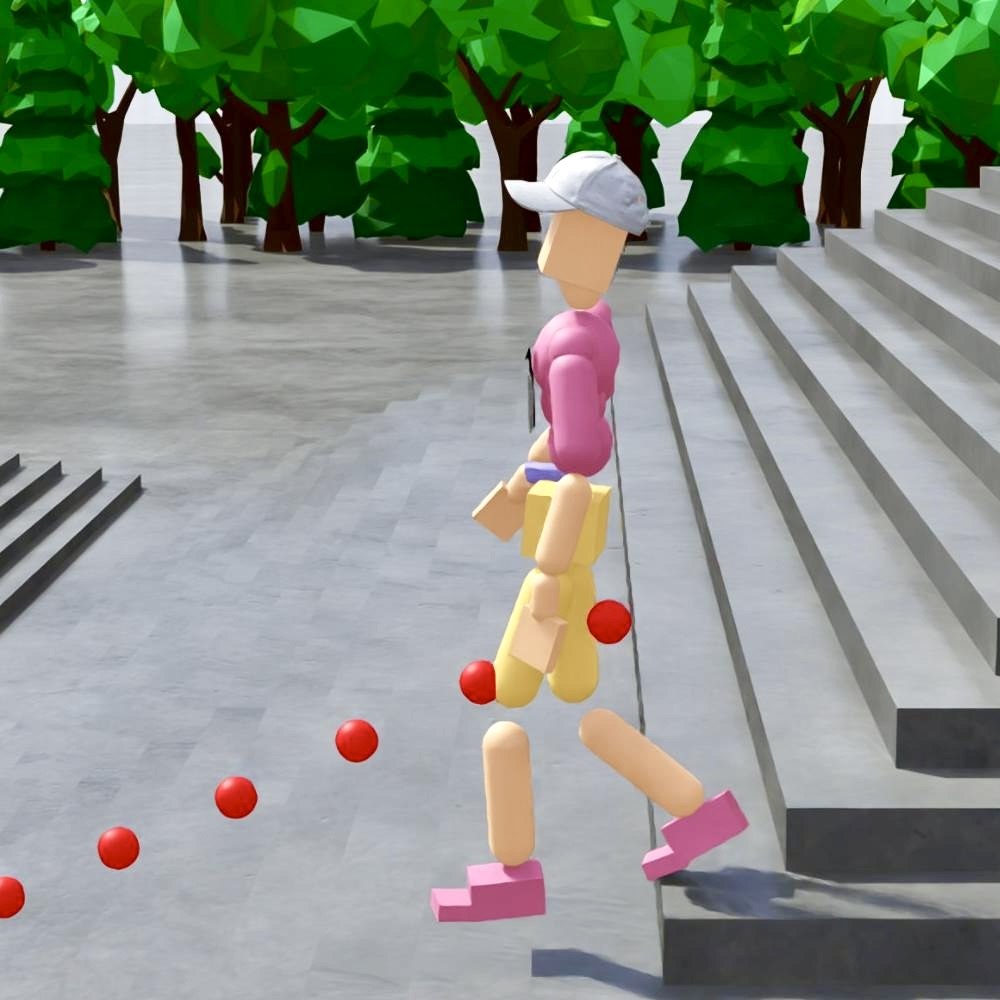}\hfill
        \includegraphics[width=0.0990\linewidth]{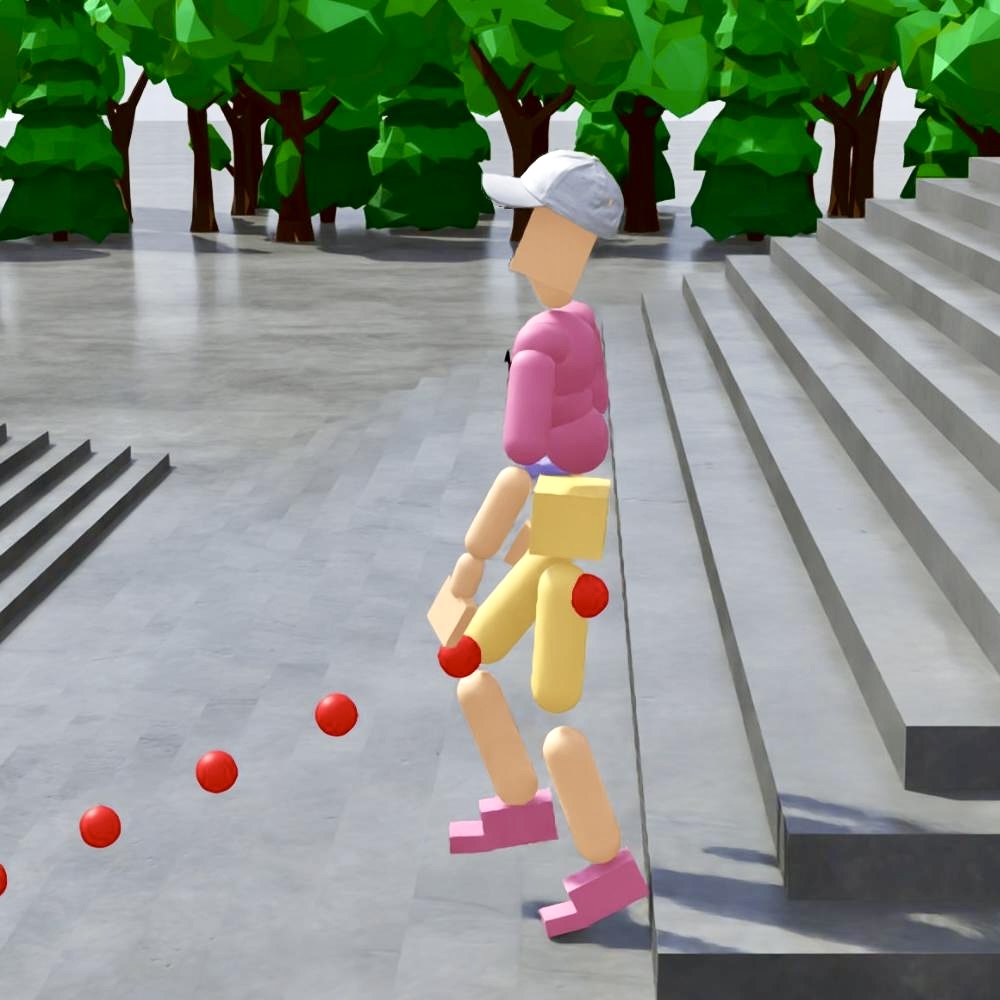}\hfill
        \includegraphics[width=0.0990\linewidth]{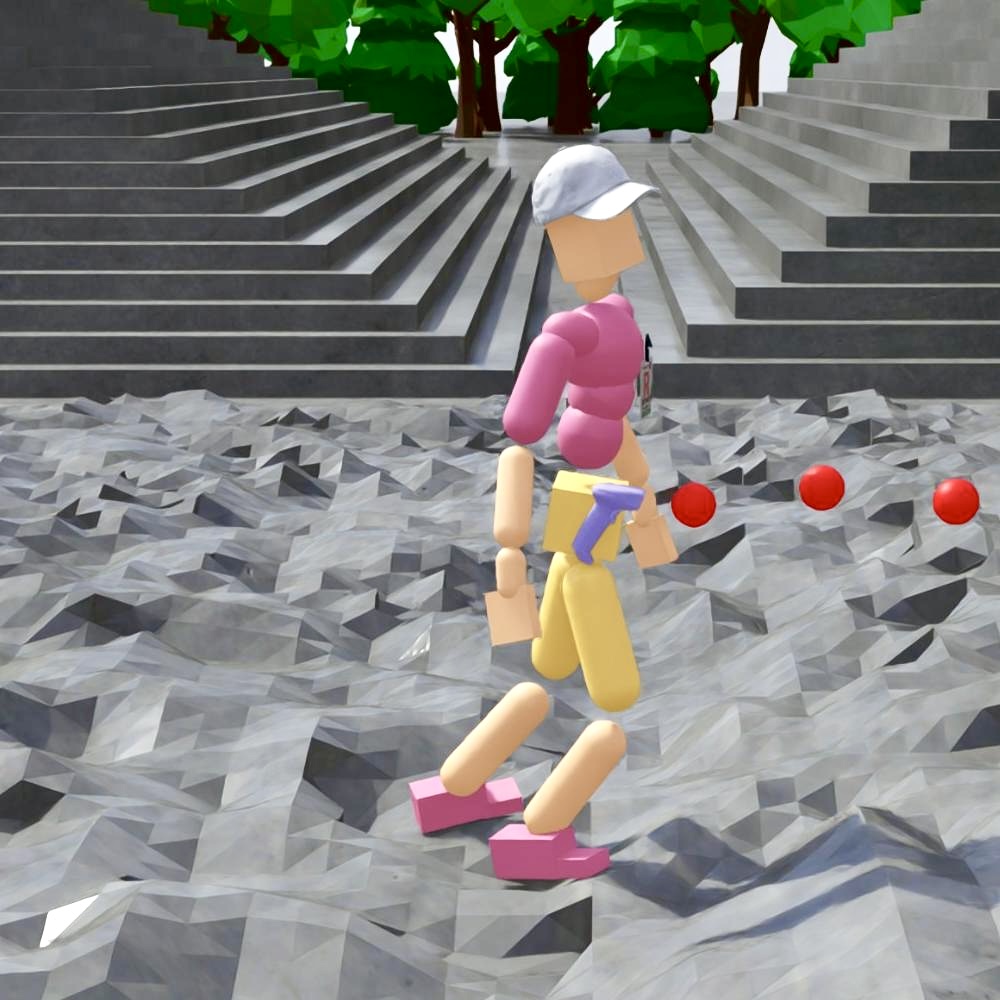}\hfill
        \includegraphics[width=0.0990\linewidth]{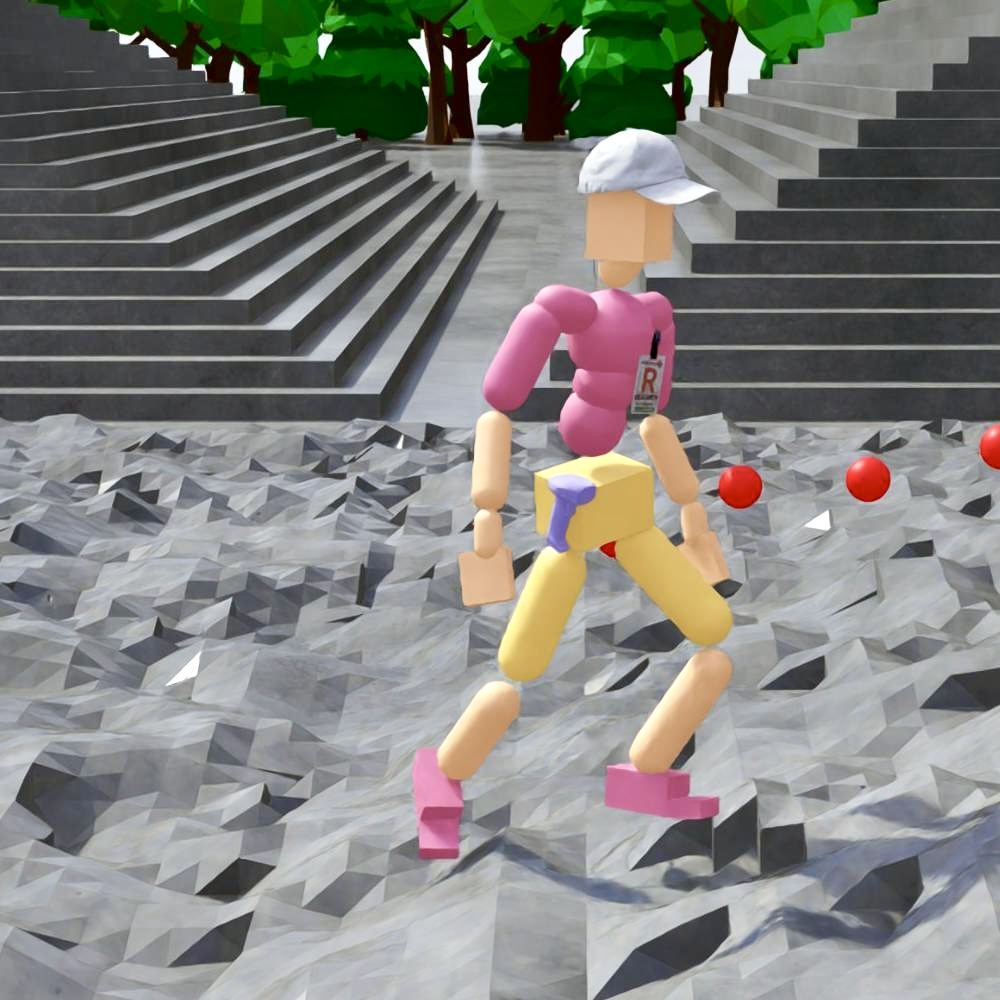}\hfill
        \includegraphics[width=0.0990\linewidth]{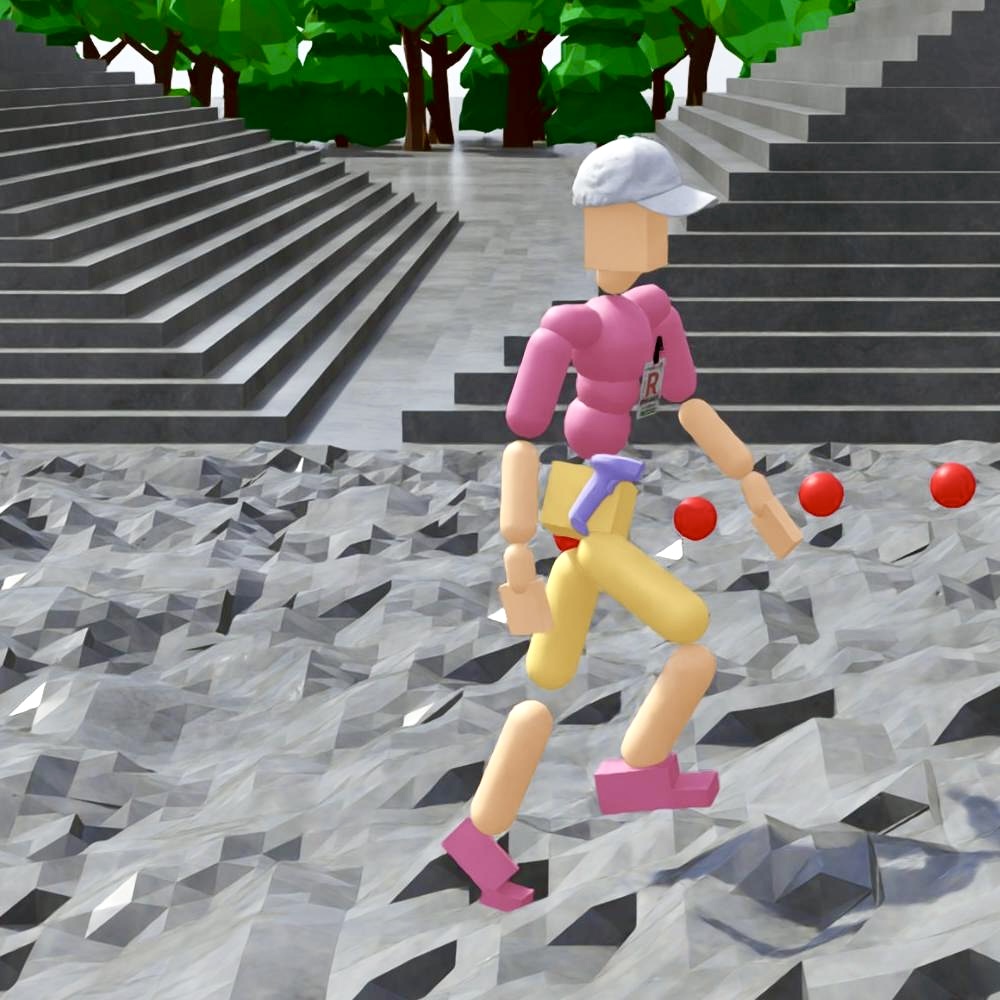}\hfill
        \includegraphics[width=0.0990\linewidth]{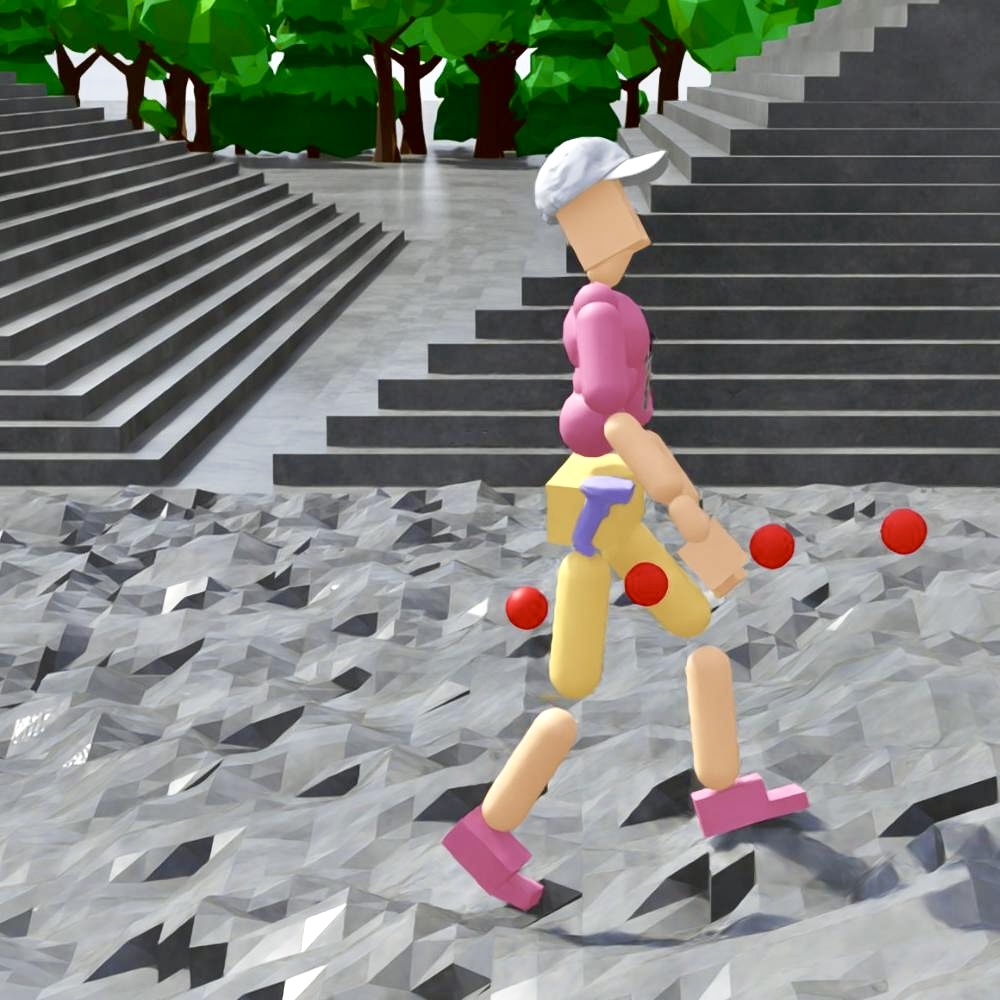}\hfill
        \includegraphics[width=0.0990\linewidth]{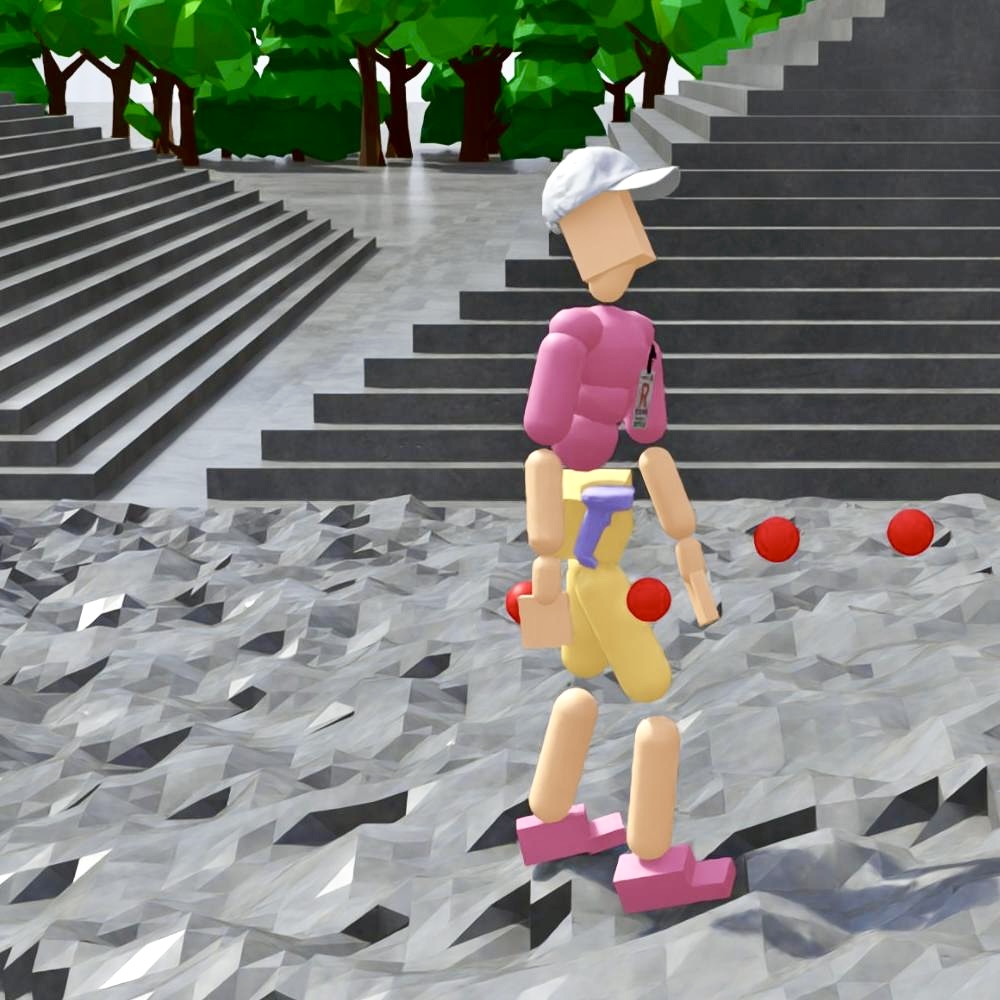}
        \caption{Path Follow}
        \label{fig:task_pathfollow}
    \end{subfigure}\\[0.5em]
    \begin{subfigure}[t]{0.497\textwidth}
        \includegraphics[width=0.1980\linewidth]{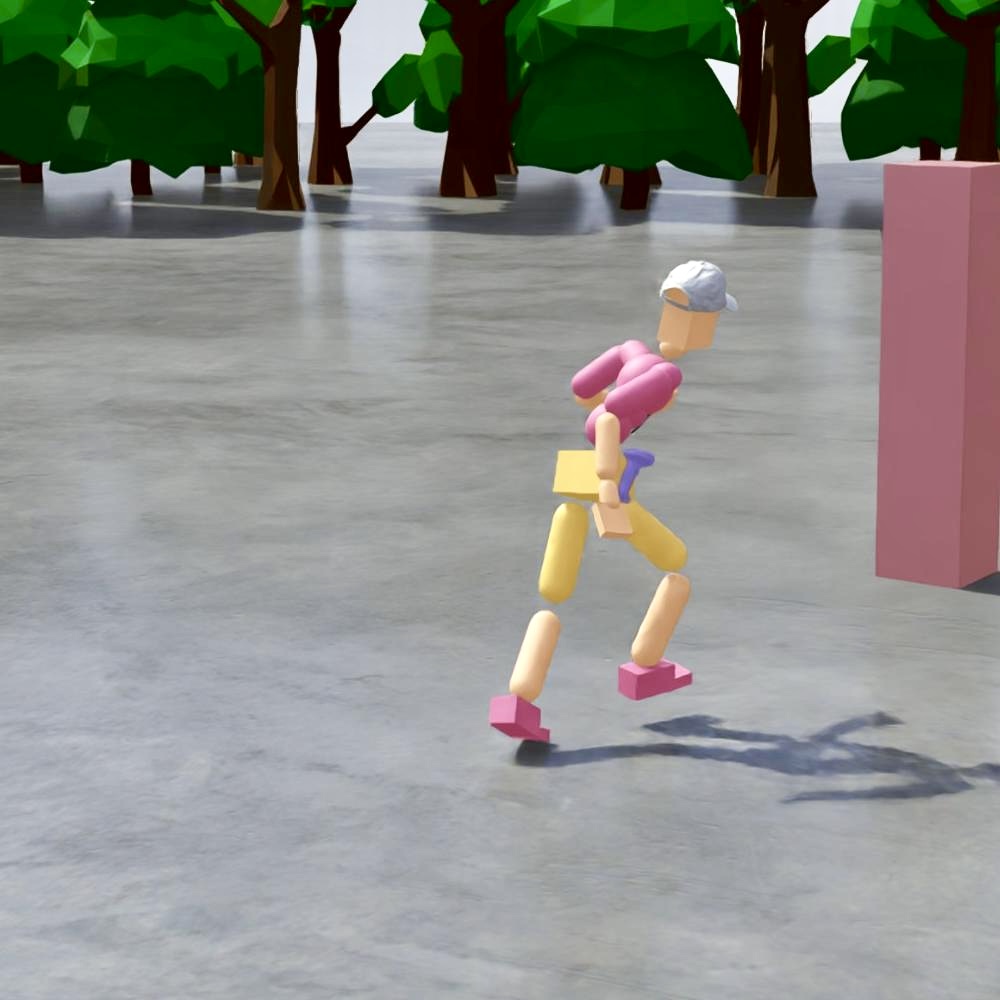}\hfill
        \includegraphics[width=0.1980\linewidth]{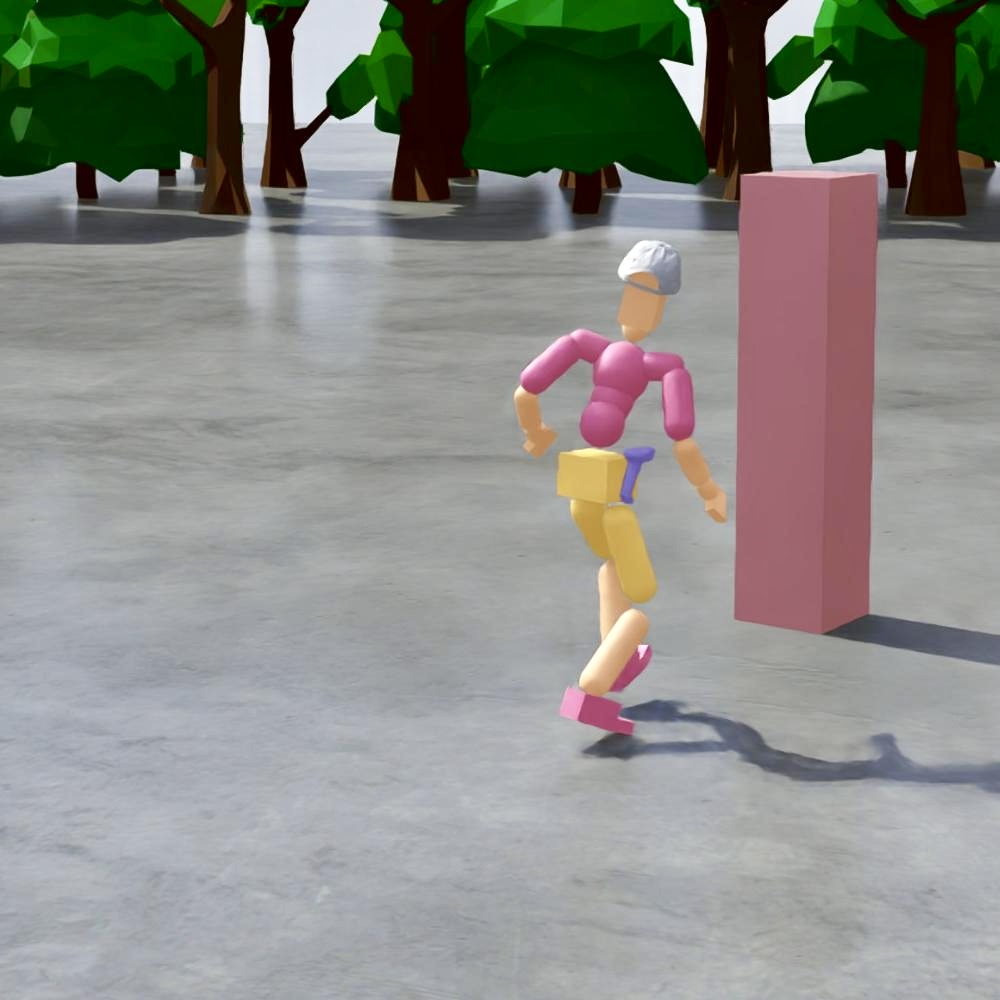}\hfill
        \includegraphics[width=0.1980\linewidth]{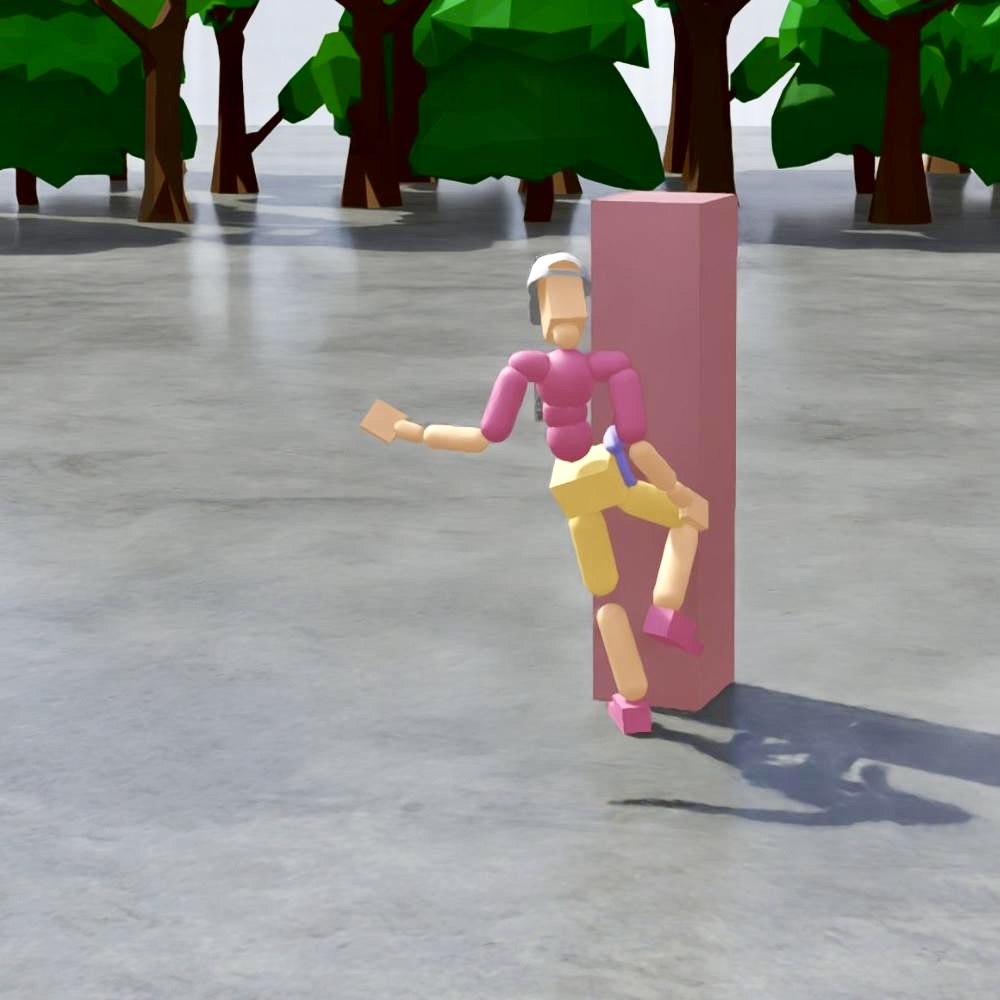}\hfill
        \includegraphics[width=0.1980\linewidth]{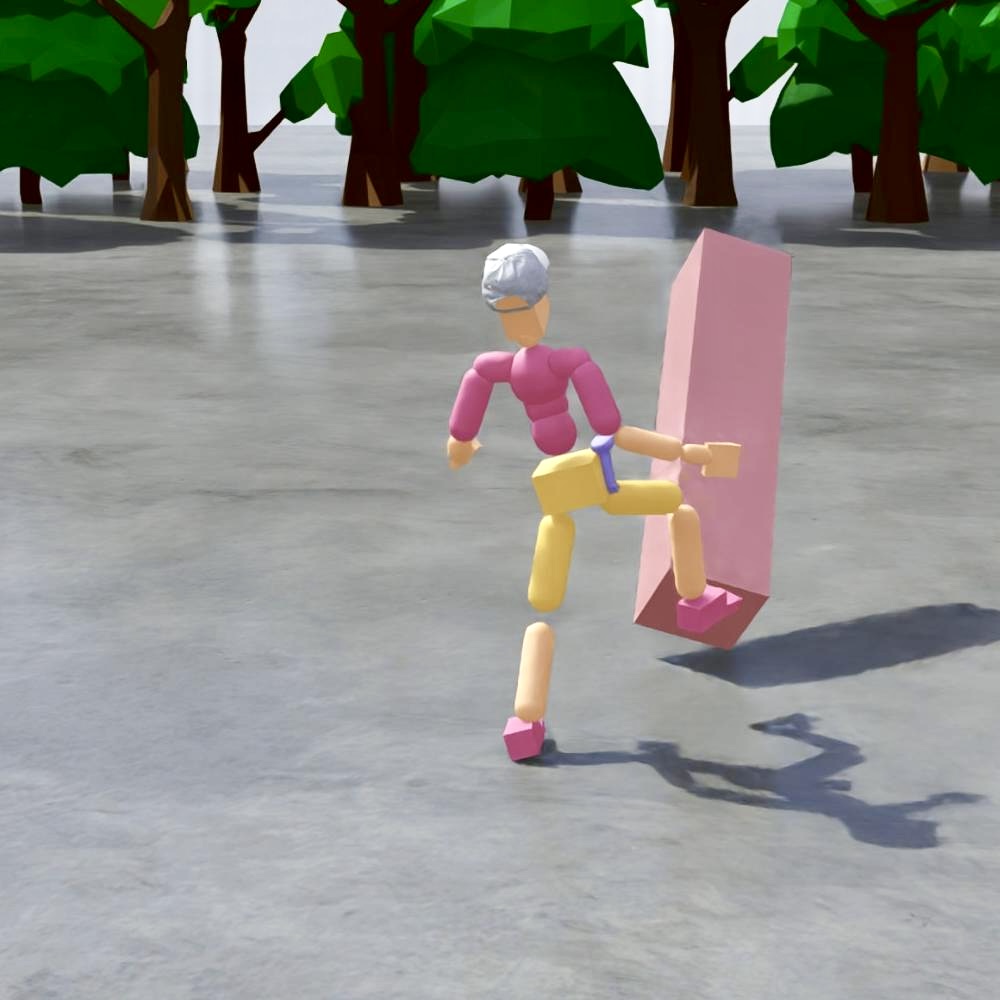}\hfill
        \includegraphics[width=0.1980\linewidth]{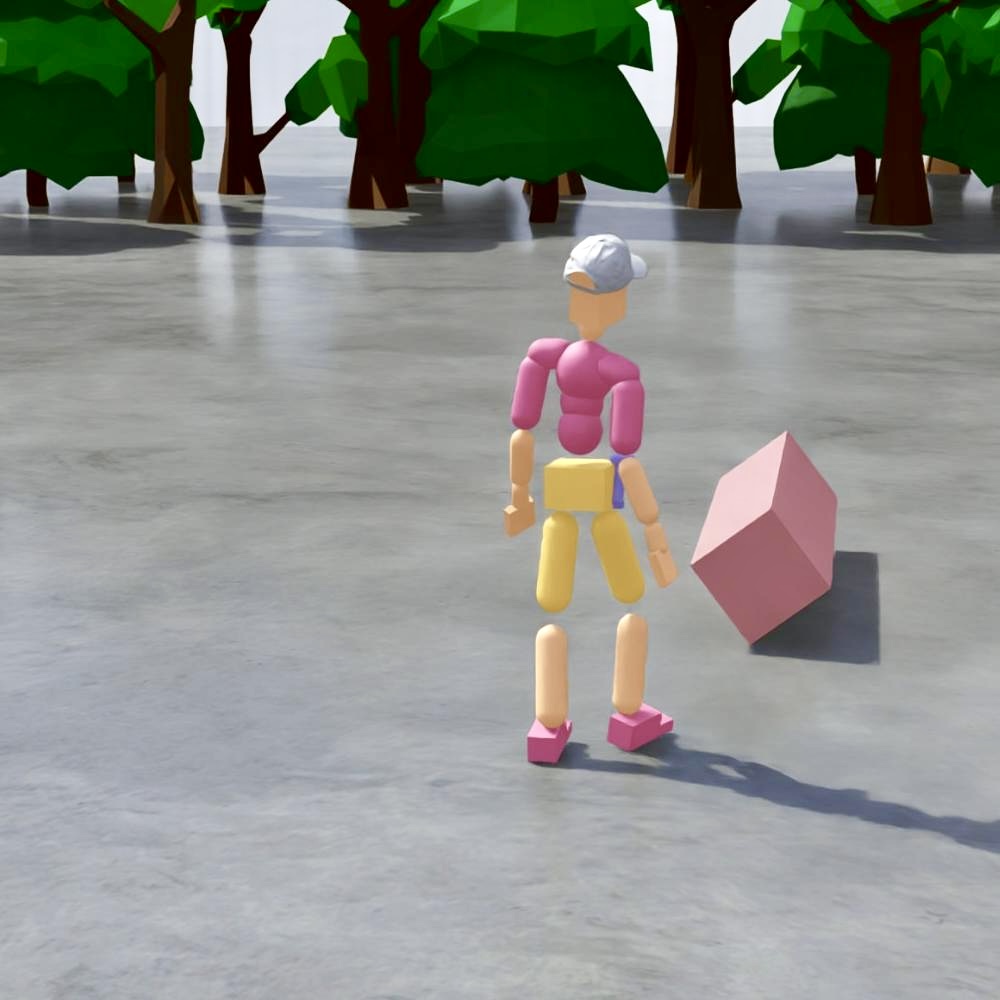}
        \caption{Strike Kick}
        \label{fig:task_kick}
    \end{subfigure}\hfill
    \begin{subfigure}[t]{0.497\textwidth}
        \includegraphics[width=0.1980\linewidth]{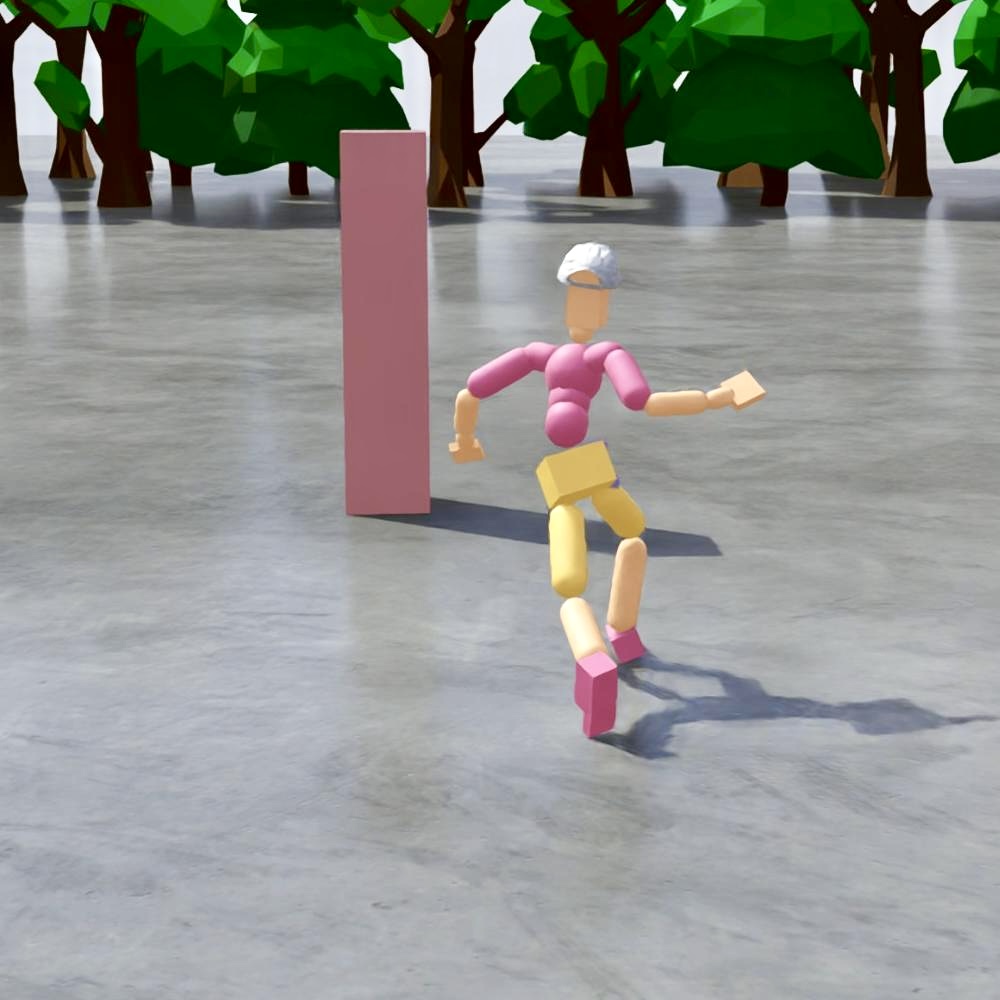}\hfill
        \includegraphics[width=0.1980\linewidth]{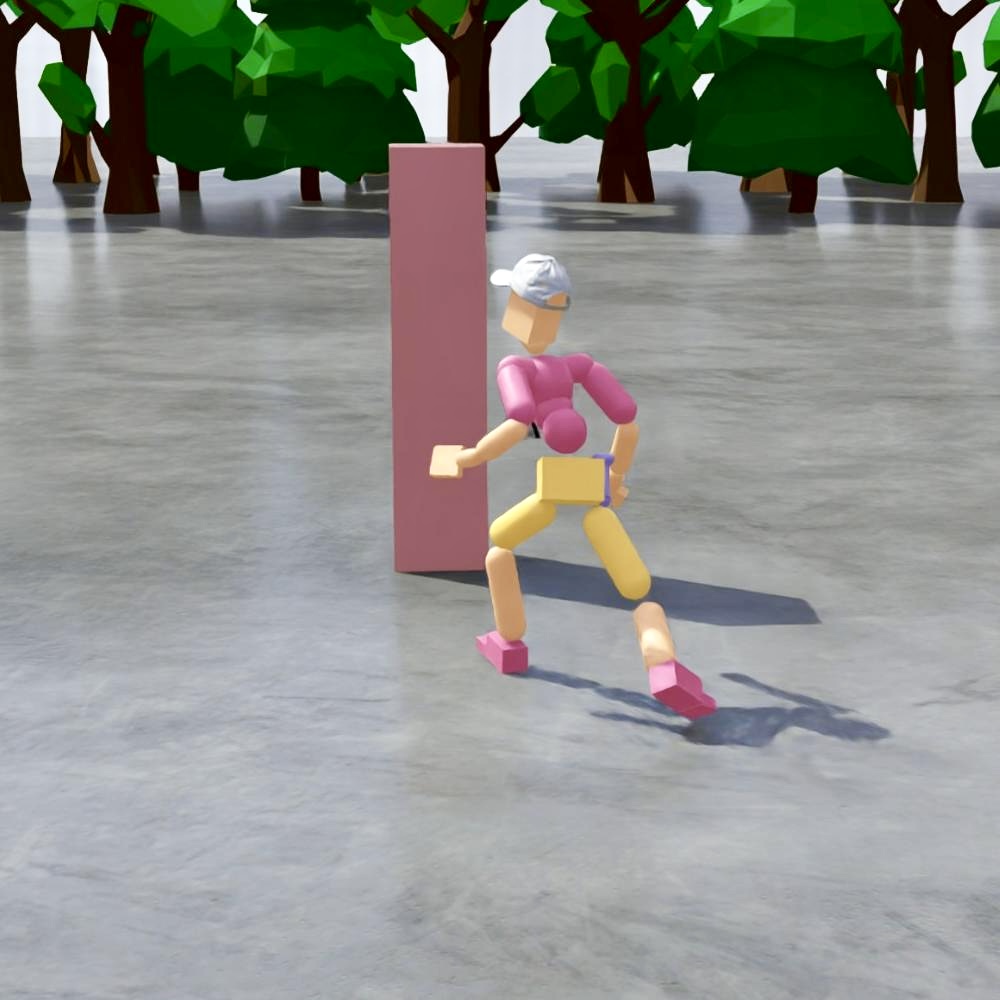}\hfill
        \includegraphics[width=0.1980\linewidth]{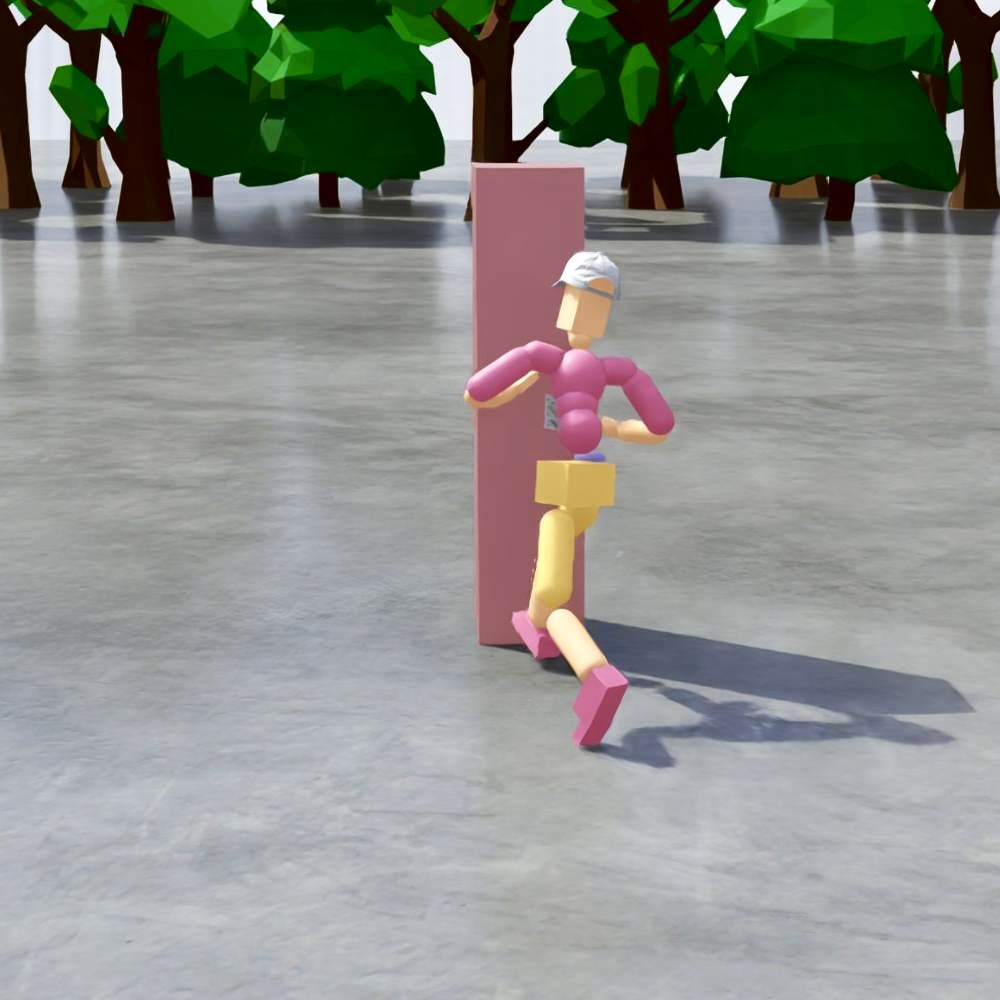}\hfill
        \includegraphics[width=0.1980\linewidth]{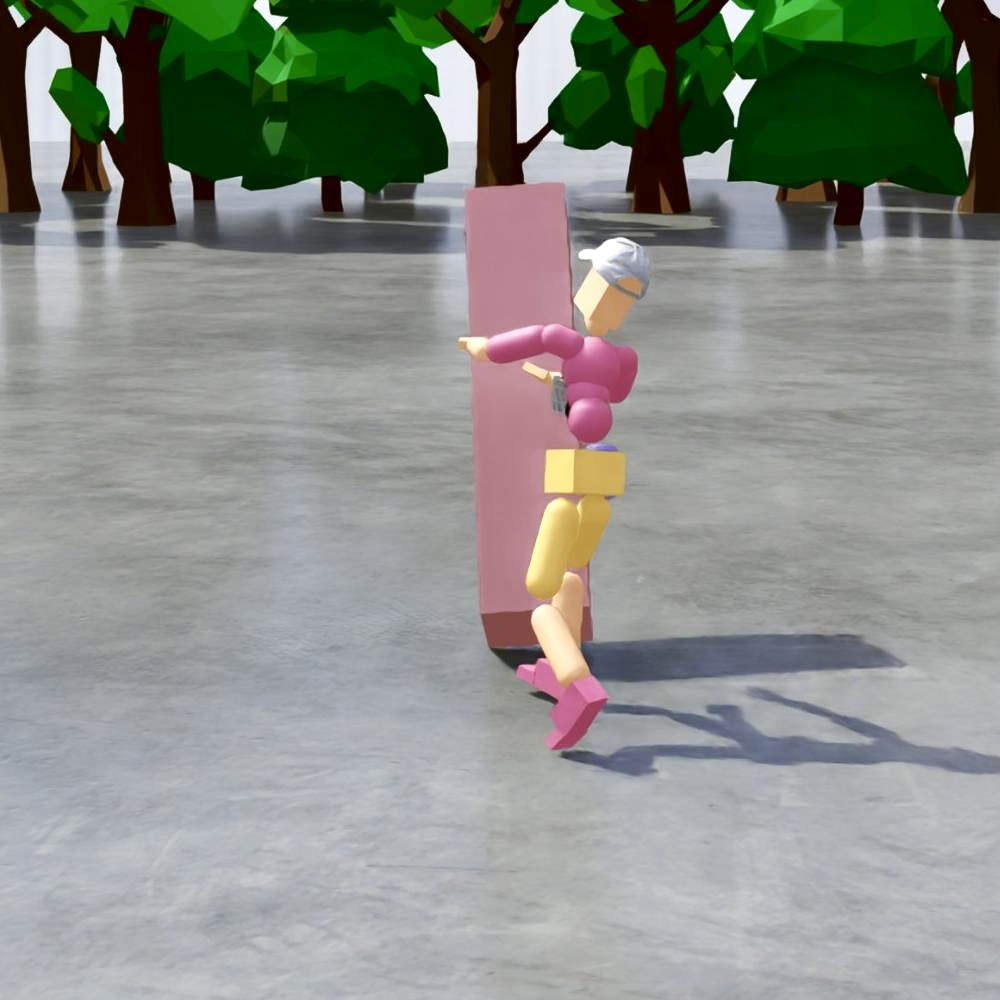}\hfill
        \includegraphics[width=0.1980\linewidth]{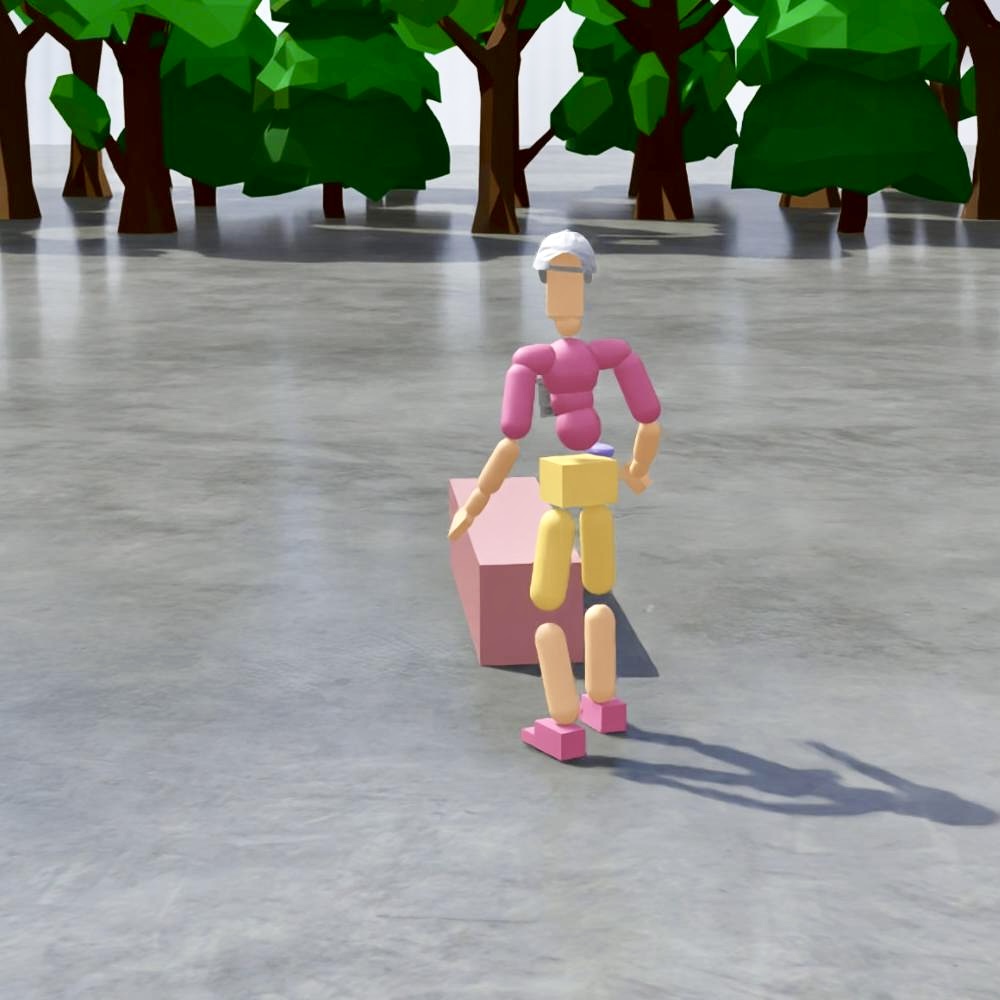}
        \caption{Strike Push}
        \label{fig:task_push}
    \end{subfigure}\\[0.5em]
    \begin{subfigure}[t]{\textwidth}
        \includegraphics[width=0.0990\linewidth]{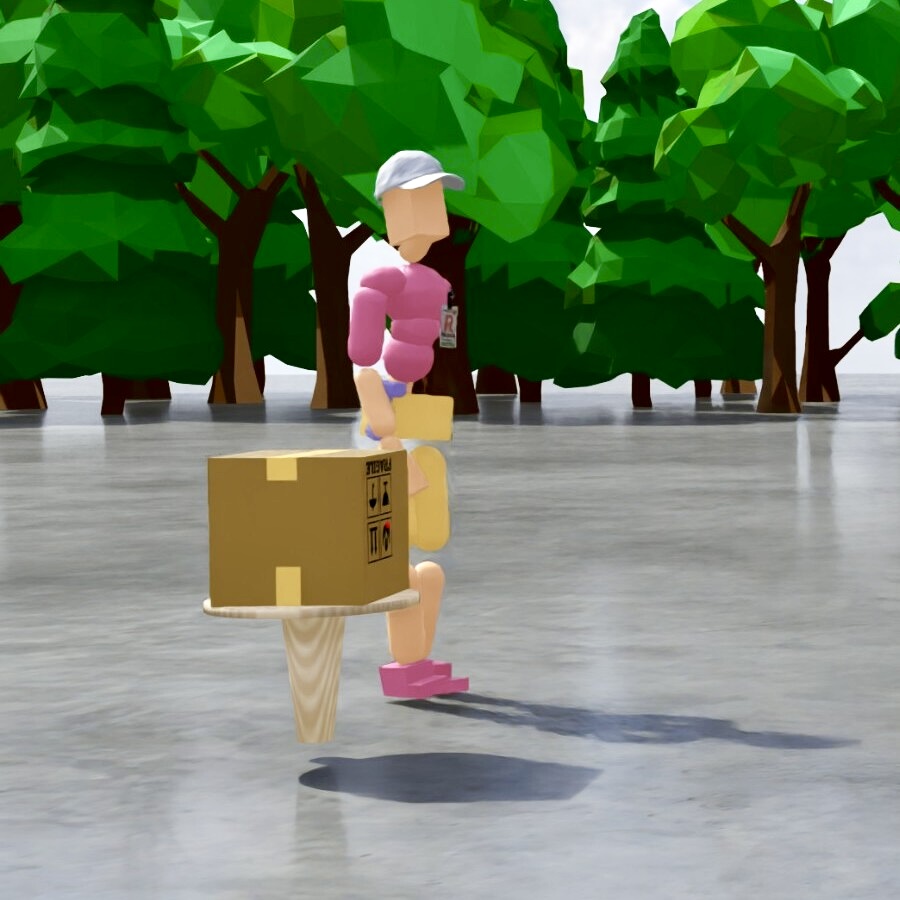}\hfill
        \includegraphics[width=0.0990\linewidth]{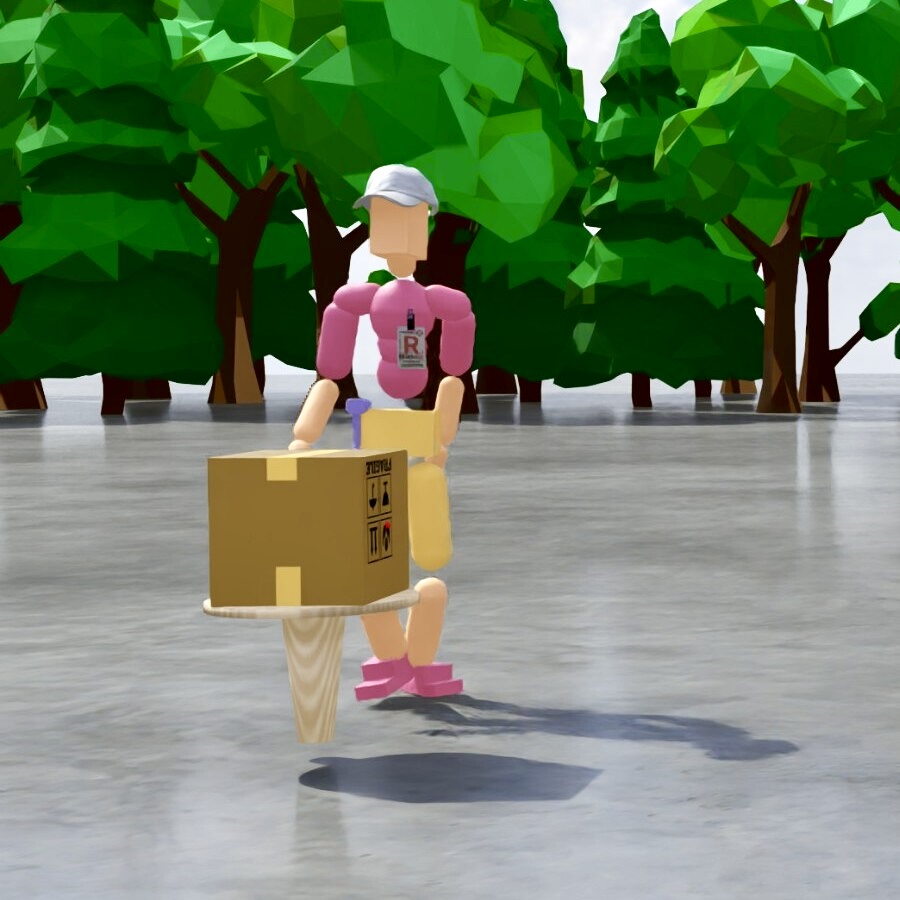}\hfill
        \includegraphics[width=0.0990\linewidth]{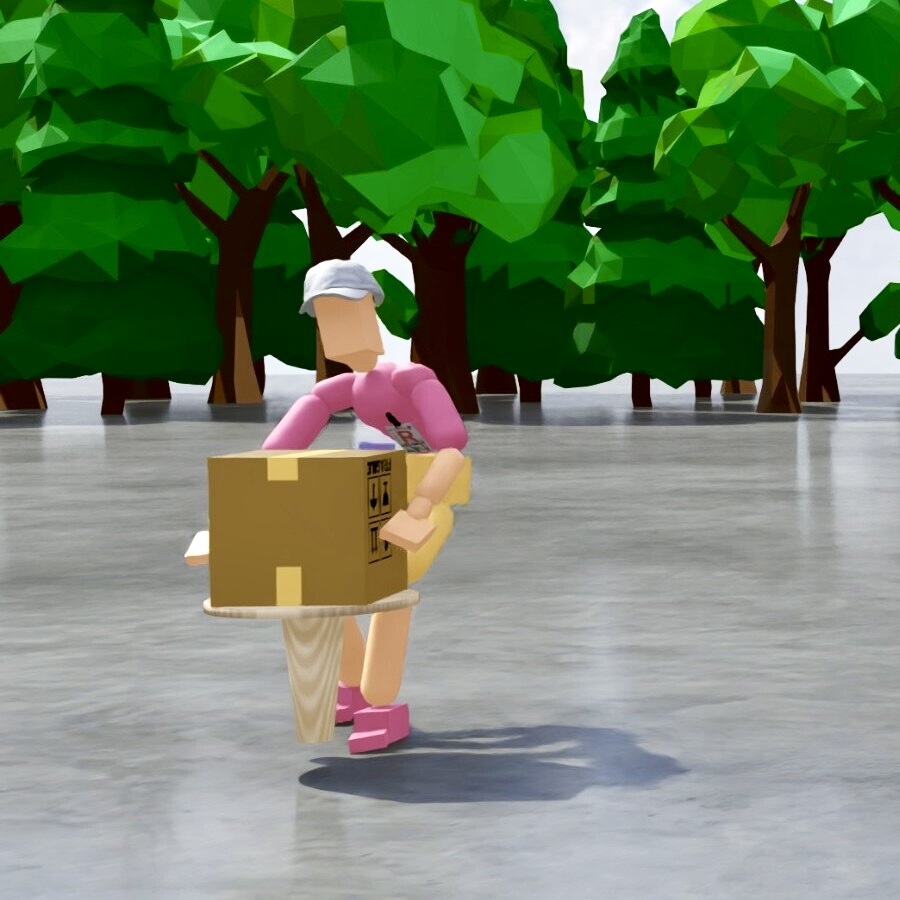}\hfill
        \includegraphics[width=0.0990\linewidth]{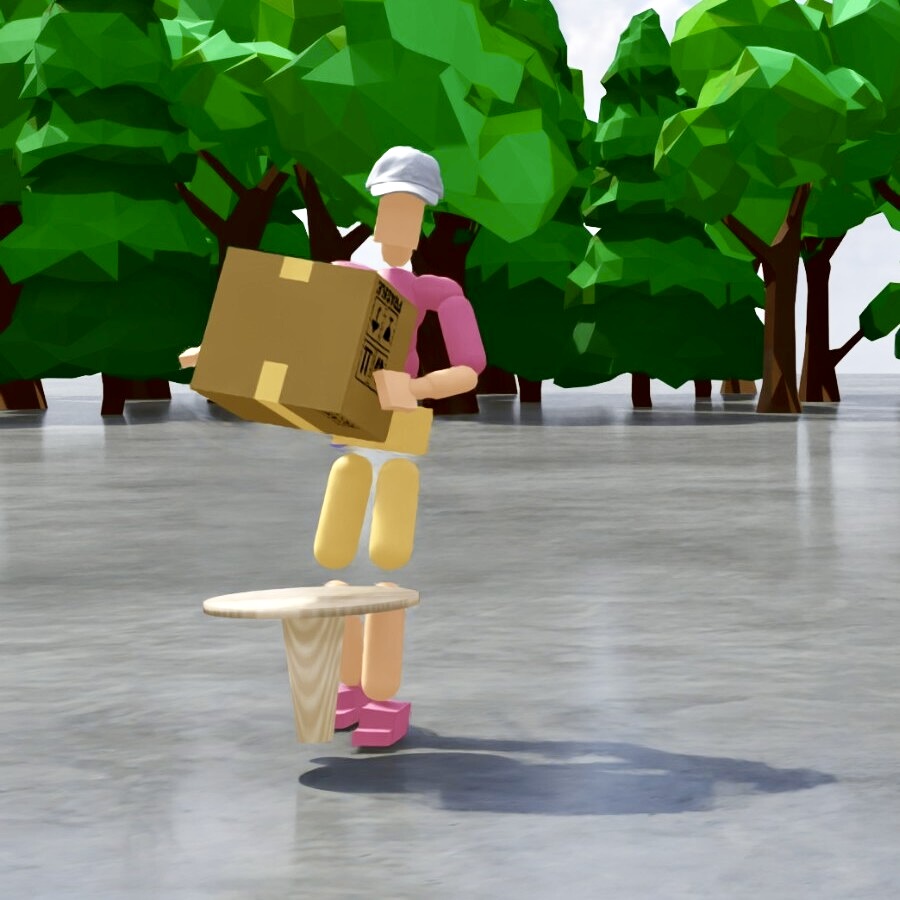}\hfill
        \includegraphics[width=0.0990\linewidth]{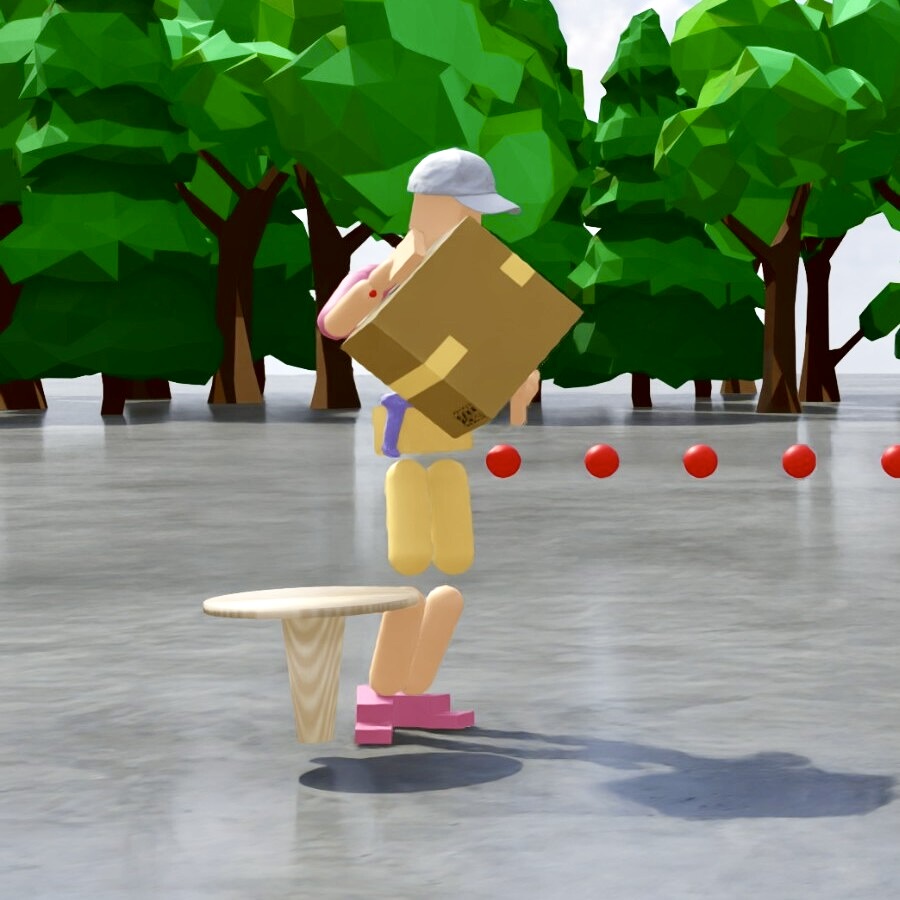}\hfill
        \includegraphics[width=0.0990\linewidth]{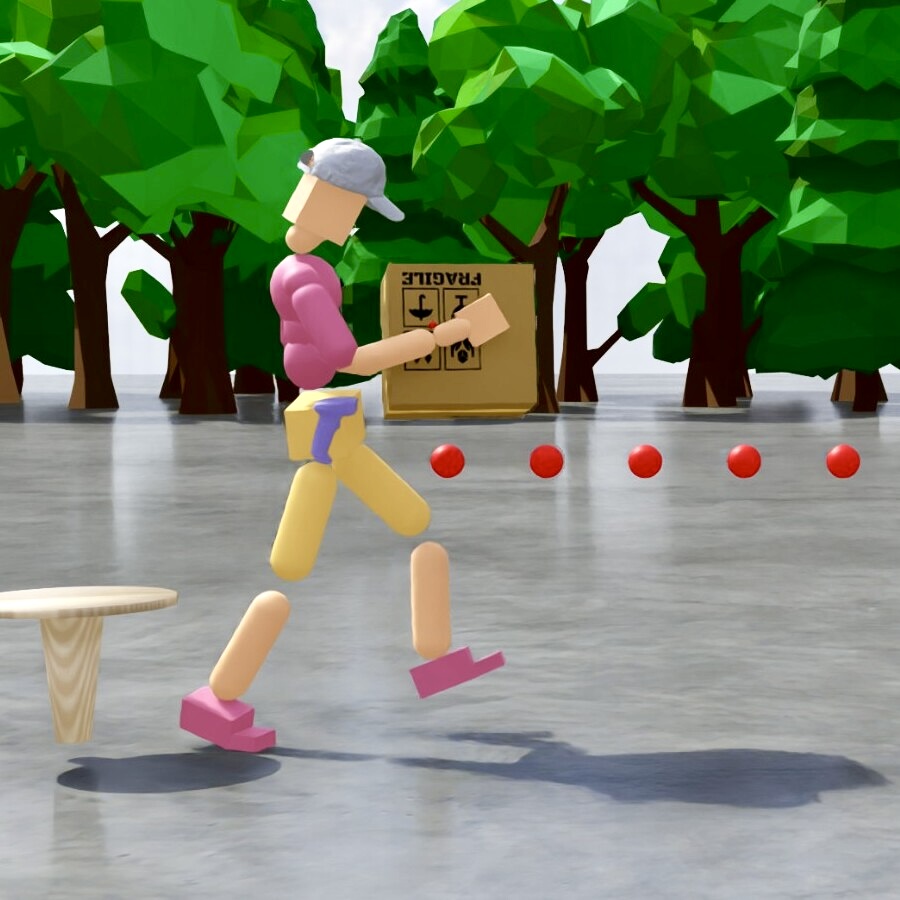}\hfill
        \includegraphics[width=0.0990\linewidth]{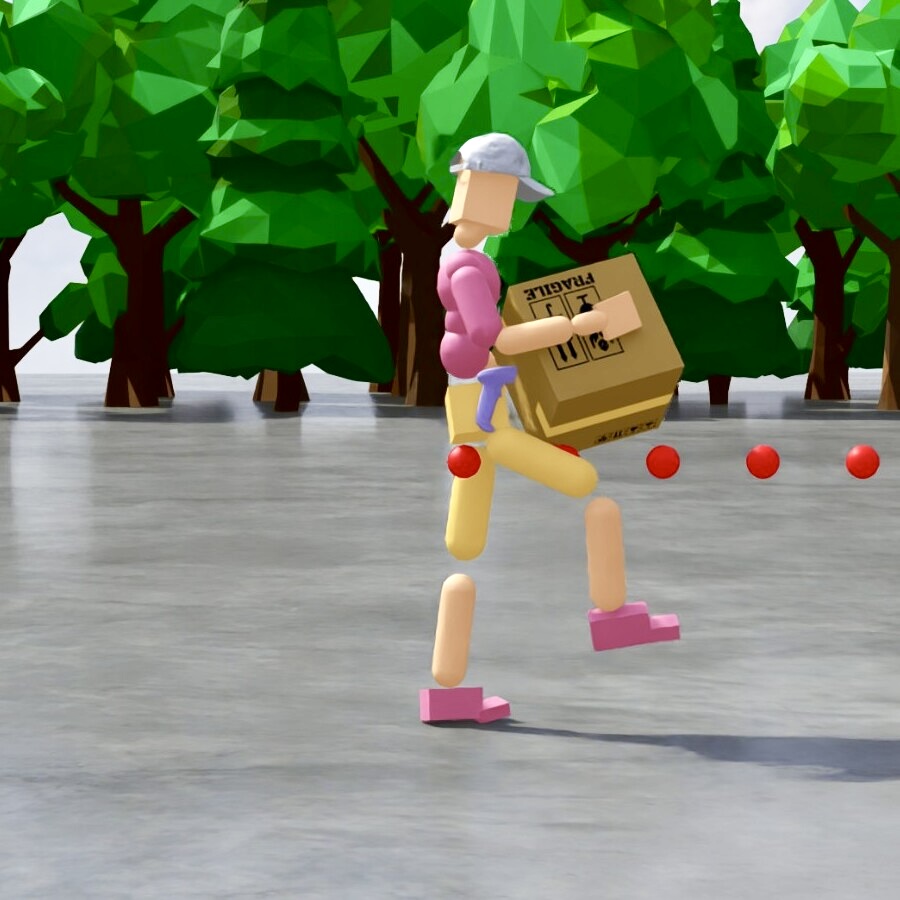}\hfill
        \includegraphics[width=0.0990\linewidth]{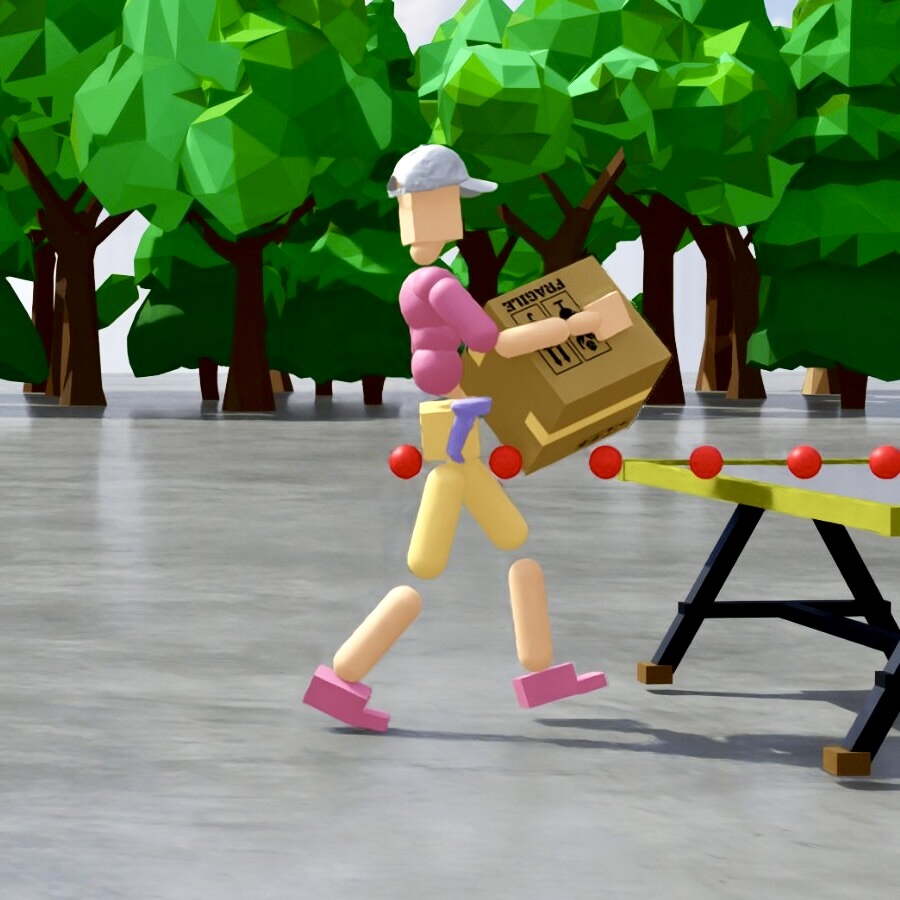}\hfill
        \includegraphics[width=0.0990\linewidth]{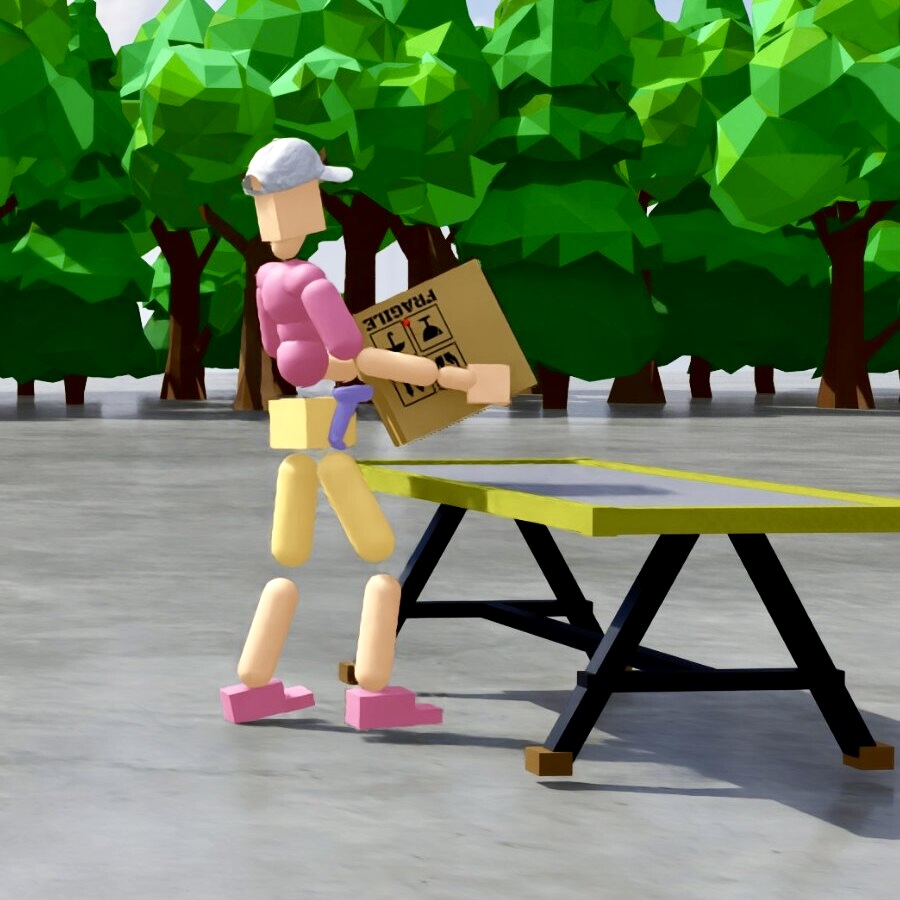}\hfill
        \includegraphics[width=0.0990\linewidth]{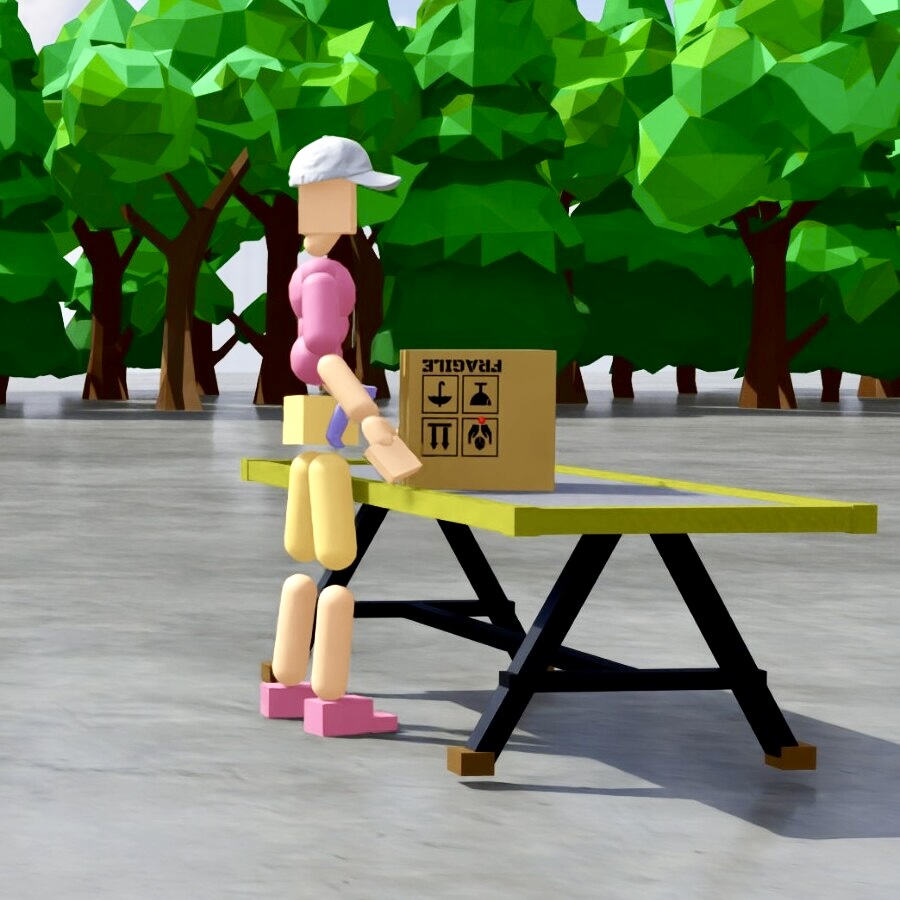}
        \caption{Pick-and-Place}
        \label{fig:task_pap}
    \end{subfigure}

    \caption{Downstream task examples solved by HetSkills. The same frozen latent controller is adapted to diverse downstream tasks: (a) path follow, (b) strike with a kick, (c) strike with a push, and (d) pick-and-place. No task-specific motion demonstrations are used. Each task is learned through reward-driven lightweight adaptation that composes language-conditioned latent priors with task-conditioned residual corrections.}
    \label{fig:task_qualitative}
\end{figure*}

Downstream adaptation optimizes only the lightweight guidance modules while keeping all pretrained priors and decoders fixed. This design preserves the motion naturalness encoded in the shared latent space and restricts task learning to the level of latent composition and correction. In practice, however, optimizing only task rewards can still drive the latent actions away from the motion-prior distribution. We therefore regularize the adapted latents to maintain stable and natural behavior. \cref{fig:task_qualitative} shows representative downstream task examples under this adaptation setting.

\paragraph{Trainable modules.} We train the compositional task guidance modules $\mathcal{G}_p$ and $\mathcal{R}_p$ using PPO~\citep{schulman2017proximal}, while keeping all pretrained priors and the decoder frozen. The full procedure is summarized in Algorithm~\ref{alg:lgda}. After training, a new downstream skill is characterized by its instruction set and learned guidance module:
\begin{equation}
    \mathscr{F}^{tsk^*} = \{\mathcal{I}_{p},\, \mathcal{G}_p,\, \mathcal{R}_p\}, \qquad p \in \{a,b\}.
\end{equation}

\paragraph{Latent regularization.} Empirically, optimizing only the task reward tends to drive the latent action away from the motion-prior distribution, resulting in motion jitter and degraded naturalness. To mitigate this effect, we introduce two regularization terms. The latent magnitude penalty $\mathcal{L}_{\mathrm{lmp}}$ is activated only when $\hat{z}$ exceeds a predefined threshold $z_{\mathrm{bound}}$:
\begin{equation}
    \mathcal{L}_{\mathrm{lmp}} = \mathbb{E}\left[\left(\max\left(|\hat{z}| - z_{\mathrm{bound}}, 0\right)\right)^2\right].
\end{equation}
The latent smoothness penalty $\mathcal{L}_{\mathrm{smooth}}$ encourages temporal consistency by penalizing large differences between latent actions at consecutive timesteps:
\begin{equation}
    \mathcal{L}_{\mathrm{smooth}} = \mathbb{E}\left[\|\hat{z}^t - \hat{z}^{t-1}\|^2\right].
\end{equation}

\paragraph{Optimization objective.} The full downstream training objective combines the PPO loss with the two regularization terms:
\begin{equation}
    \mathcal{L}_{\mathrm{tsk}} = \mathcal{L}_{\mathrm{ppo}} + \lambda_{\mathrm{lmp}} \mathcal{L}_{\mathrm{lmp}} + \lambda_{\mathrm{smooth}} \mathcal{L}_{\mathrm{smooth}}.
\label{eq:update}
\end{equation}
Together, these terms keep the adapted latent actions within a stable region of the shared control space and improve the naturalness and robustness of the resulting motions.

\section{Experimental Setup}
All experiments are conducted in Isaac Lab~\cite{mittal2025isaac} using ProtoMotions \cite{ProtoMotions}, with physics simulation running at 120~Hz and control policy execution at 30~Hz. Our framework is trained in four progressive stages using two consumer-grade NVIDIA RTX 5090 GPUs. Detailed hyperparameters and implementation specifics are provided in the appendix.

\subsection{Datasets}

Our progressive training paradigm naturally supports heterogeneous data sources across different stages, eliminating the need to unify all data under a single annotation format. This approach allows each stage to leverage the data best suited to its task objectives. 

Specifically, for motion tracking, we use AMASS~\cite{mahmood2019amass} as the base motion dataset and follow the filtering pipeline of PHC~\cite{luo2023perpetual} to clean the data. This pipeline removes clips that exhibit non-physical artifacts, such as limb penetration, body floating, and interactions with unmodeled objects, yielding a high-quality set of training sequences. This filtered set is also reused for VR tracking and motion in-betweening, where sparse conditioning signals are constructed directly from the same clips. For VR tracking, only the kinematic states of the head and both hands are retained as spatially sparse end-effector observations. For motion in-betweening, a future full-body pose is sampled from a rolling horizon of $5$ to $30$ frames ahead, providing temporally sparse keyframe targets. For text-to-motion, we use the HumanML3D~\cite{guo2022generating} dataset to introduce natural language annotations for motion clips. Following SuperPADL~\cite{juravsky2024superpadl}, we discard clips shorter than $2$ seconds or longer than $9$ seconds, as such clips tend to be dominated by idle poses, redundant pauses, or compound actions, all of which degrade training stability and supervision quality. For human-object interaction, we incorporate the SAMP~\cite{hassan2021stochastic} dataset, which provides high-quality motion capture data covering typical furniture interactions, such as sitting and lying down. The dataset records both body motion and object spatial information simultaneously, which is essential for training models that can understand and predict interaction-based behaviors.

\subsection{Tasks}
\label{sec:tasks}
To evaluate the downstream task adaptation module, we design three tasks that do not require task-specific demonstration data. For all tasks, we apply an energy penalty $\lambda \cdot \mathcal{P}$ ($\lambda = 10^{-5}$) to suppress unnecessary high-power motions. In the path follow task, this penalty is applied only to the leg joints. Full reward specifications and hyperparameters are provided in the appendix. Specifically:

\paragraph{Path Follow} The agent tracks an online-generated random path through complex terrain. At each timestep, the agent observes the next $10$ waypoints in its local coordinate frame, and the reward is defined as:
\begin{equation}
    r = \exp\left(-\|p_{\text{target}}^{xy} - p_{\text{root}}^{xy}\|^2\right) - \lambda \cdot \mathcal{P}_{\text{leg}}
\end{equation}
where the first term encourages the root to stay close to the current path target in the horizontal plane.
\paragraph{Strike} The agent approaches a randomly placed target and knocks it over using designated body parts, either hands for pushing or leg segments for kicking. The reward is defined as
\begin{equation}
    r = 0.6\, r_{\text{rot}} + 0.2\, r_{\text{vel}} + 0.1\, r_{\text{prog}} + 0.1\, r_{\text{toward}} - \lambda \cdot \mathcal{P}
\end{equation}
where $r_{\text{rot}}$ rewards target tilt, $r_{\text{vel}}$ encourages approaching the target at a desired speed, $r_{\text{prog}}$ rewards reducing the distance to the target, and $r_{\text{toward}}$ rewards facing the target.
\paragraph{Pick-and-Place} The task consists of three stages, including pick-up, carry-to, and put-down. In the pick-up stage, the agent must approach a box randomly placed on a source platform and lift it to a target height using both hands. The reward is:
\begin{equation}
    r = 0.4\, r_{\text{lift}} + 0.3\, r_{\text{grasp}} + 0.2\, r_{\text{force}} + 0.1\, r_{\text{face}} - \lambda \cdot \mathcal{P}
\end{equation}
where $r_{\text{lift}}$ rewards lifting the box, $r_{\text{grasp}}$ rewards approaching the grasp points, $r_{\text{force}}$ rewards effective bilateral contact, and $r_{\text{face}}$ rewards facing the box. The carry-to and put-down stages require the agent to walk along a path while maintaining the grasp, and to place the box onto a target platform, respectively. We use pick-up as the baseline comparison task, as it is the most challenging of the three stages.

\subsection{Evaluation Dimensions}

To tackle a wide variety of tasks, including precise motion tracking, generating realistic motions from high-level language inputs, and adapting to new tasks without requiring retraining, we evaluate the performance of HetSkills across three key dimensions, with an additional long-horizon skill composition demonstration. Specifically:

\paragraph{Tracking and Motion Completion.}
We evaluate the motion tracking skill $\mathscr{F}^{trc}$ on both the training and test splits of AMASS, and report the success rate and MPJPE (Mean Per Joint Position Error, in mm)~\citep{luo2023universal,tessler2024maskedmimic}. We also conduct ablation studies to justify our part-wise decomposition design. For the motion completion skill $\mathscr{F}^{moc}$, we evaluate three instantiations that cover both spatially and temporally sparse conditioning. In VR-driven body tracking, only the head and both hands are provided as visible goal constraints, while all other body segments are masked. In motion in-betweening, the success rate is reported on the training and test splits of AMASS under the same MPJPE failure threshold. For human-object interaction, we provide qualitative examples to demonstrate the ability of the framework to handle complex scenarios.

\paragraph{Text-to-Motion.} We evaluate the text-to-motion skill $\mathscr{F}^{t2m}$ from two complementary aspects. First, we evaluate pose-level robustness under two initialization protocols: starting from the ground-truth first frame of each target clip and starting from a neutral pose. A rollout is considered a failure if the MPJPE exceeds a predefined threshold at any frame, and we report the success rates under both protocols. Second, we evaluate semantic alignment using the HumanML3D retrieval protocol \cite{guo2022generating} with a pretrained TMR model \cite{petrovich2023tmr}. In this evaluation, the generated motion is used to retrieve its corresponding language description from candidate texts in the shared text-motion embedding space. We report R-Precision (R@N) and MedR, where higher R@N and lower MedR indicate better semantic alignment between the generated motion and the input language. This query-based evaluation complements MPJPE by measuring whether the generated motion is recognizable as the intended textual action.

\paragraph{Downstream Task Adaptation.} We train and evaluate $\mathscr{F}^{tsk^*}$ on the three tasks described in~\cref{sec:tasks}, comparing our results against relevant baselines. The language instructions used to condition each task are provided in Appendix~\ref{app:instructions}.

\section{Experimental Results}
HetSkills progressively builds a unified latent space that supports motion tracking, text-to-motion generation, motion completion, and downstream task adaptation within a single shared architecture. In this section, we present the experimental results to demonstrate the effectiveness of each stage in the HetSkills pipeline. We also highlight the utility of our part-wise decomposition design, which enhances both performance and interpretability, and show how the unified latent space generalizes across heterogeneous skills and tasks. Specifically, we evaluate HetSkills on motion tracking, text-to-motion generation, and downstream task adaptation, and analyze how these capabilities contribute to the overall flexibility and robustness of the model. Building on these skills, we further provide a long-horizon composition example to show that heterogeneous abilities can be organized through the standardized task interface and executed within the same shared model.

\begin{figure}[t]
    \centering
    \begin{subfigure}[t]{\columnwidth}
        \includegraphics[width=0.1970\linewidth]{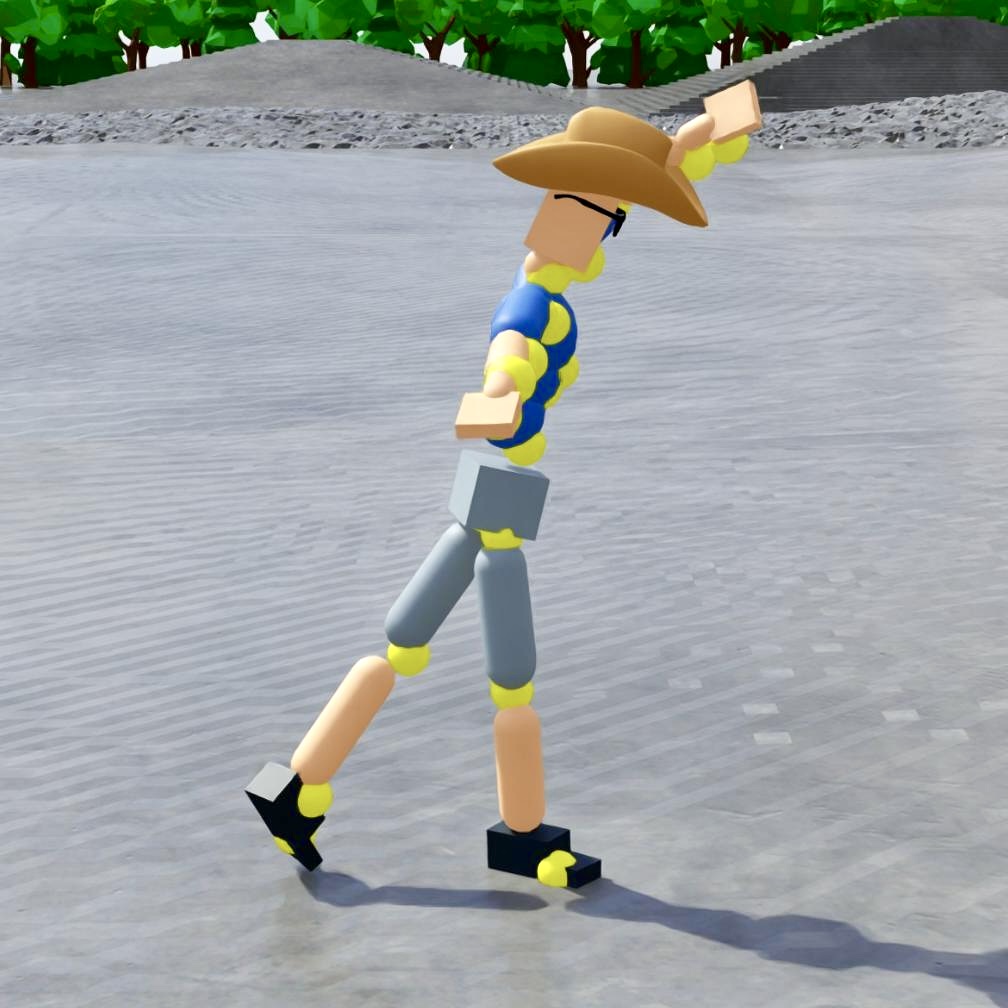}\hfill
        \includegraphics[width=0.1970\linewidth]{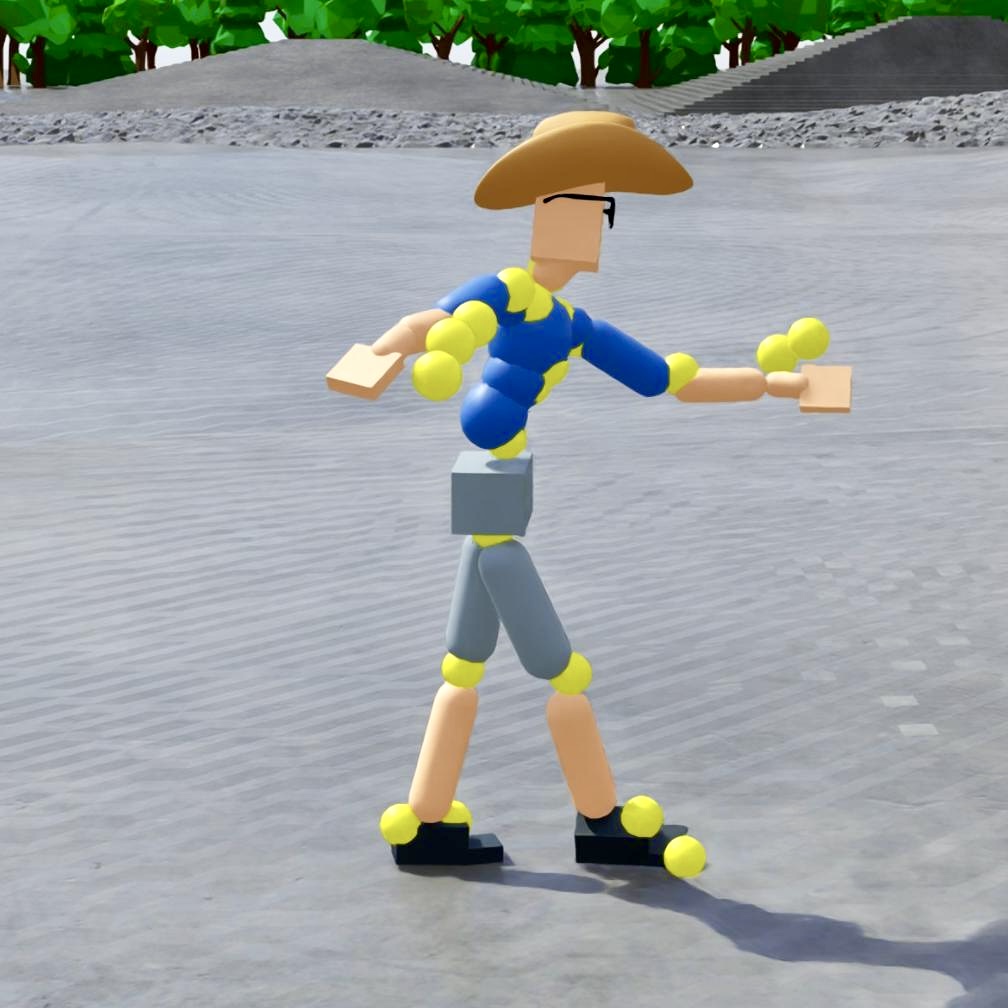}\hfill
        \includegraphics[width=0.1970\linewidth]{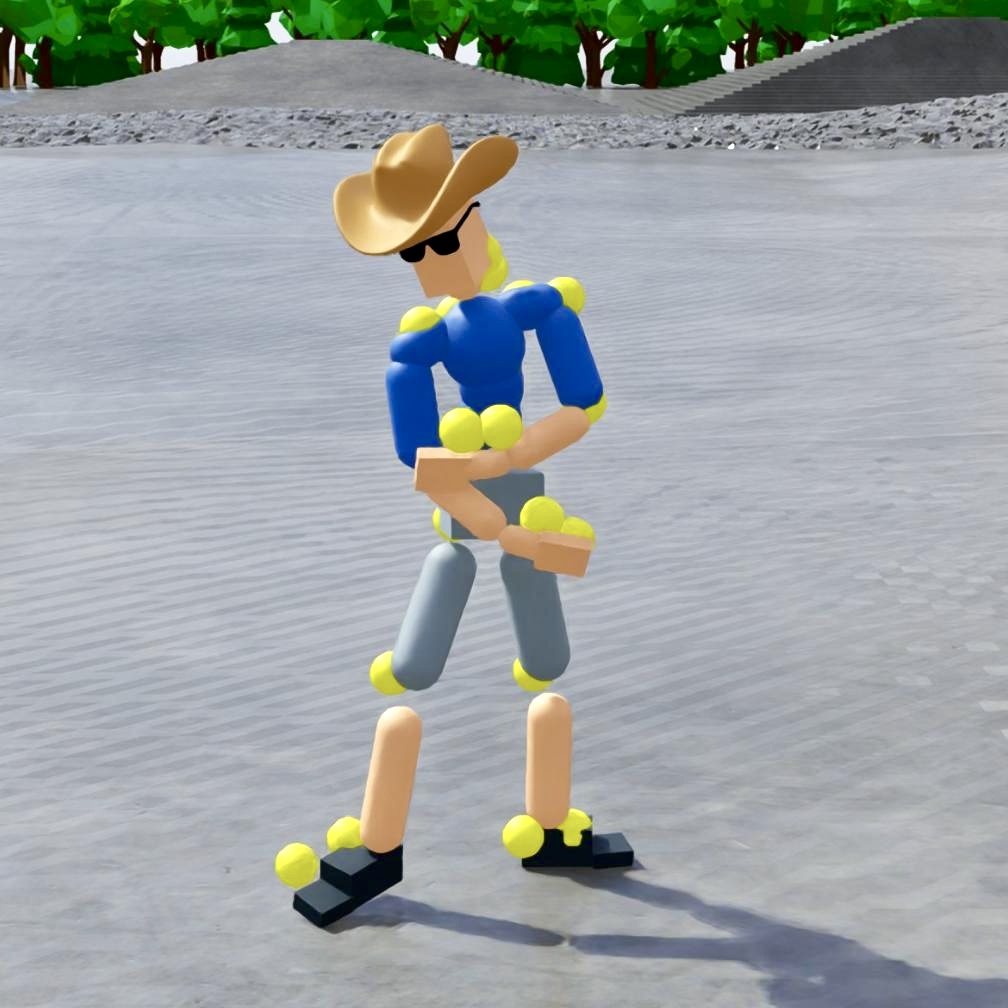}\hfill
        \includegraphics[width=0.1970\linewidth]{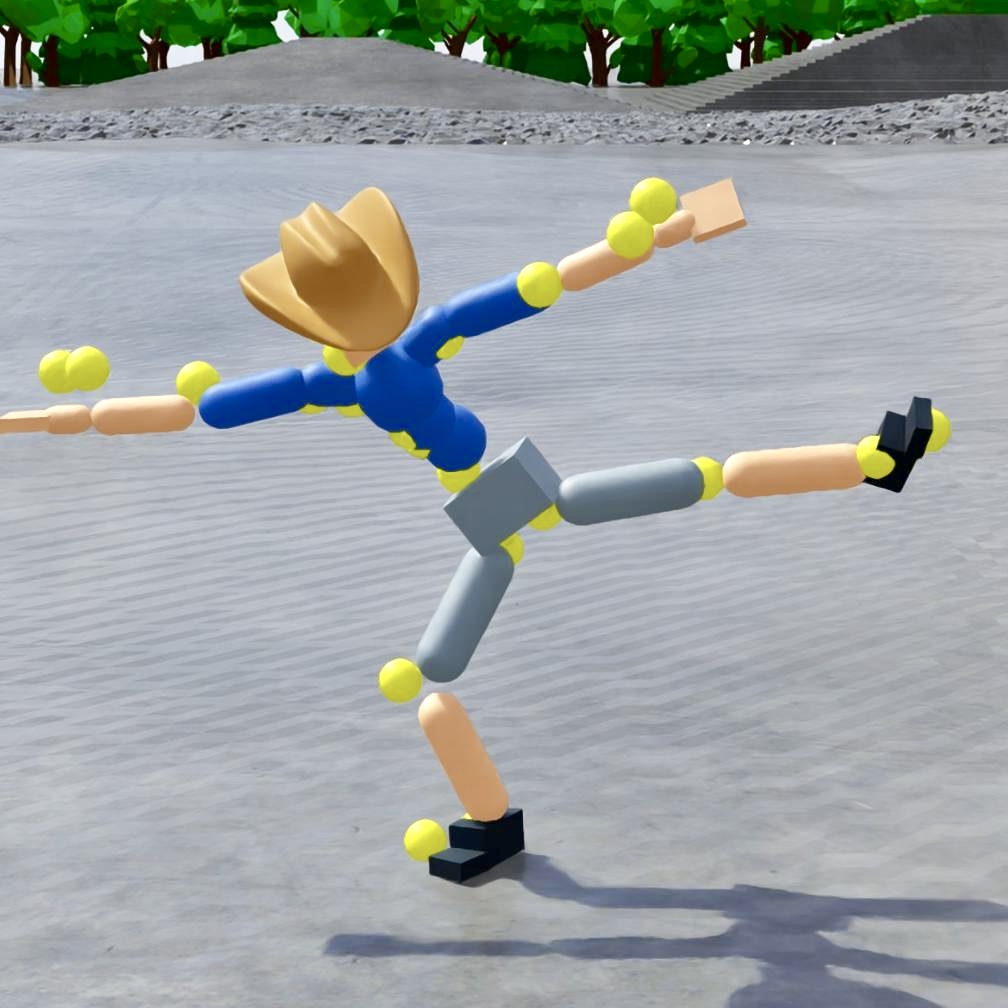}\hfill
        \includegraphics[width=0.1970\linewidth]{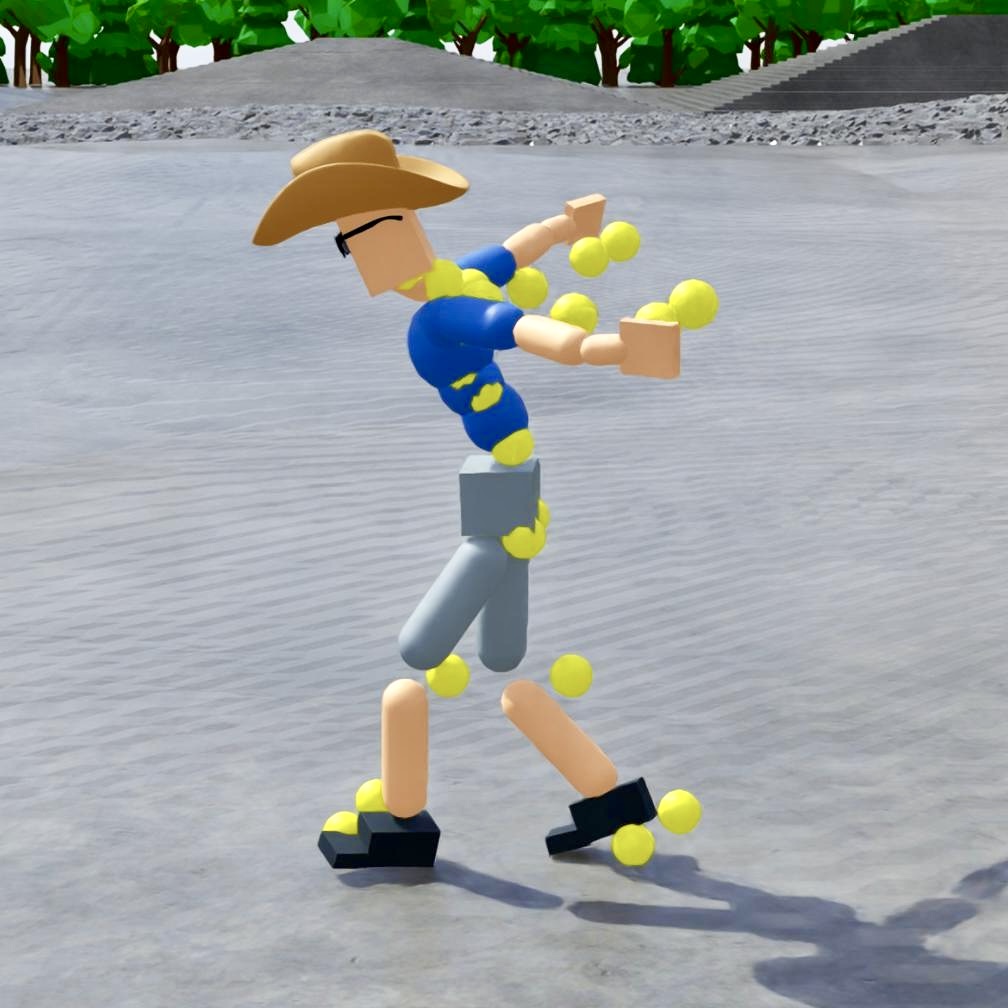}
        \label{fig:trc_ballet}
    \end{subfigure}\\[0.1em]
    \begin{subfigure}[t]{\columnwidth}
        \includegraphics[width=0.1970\linewidth]{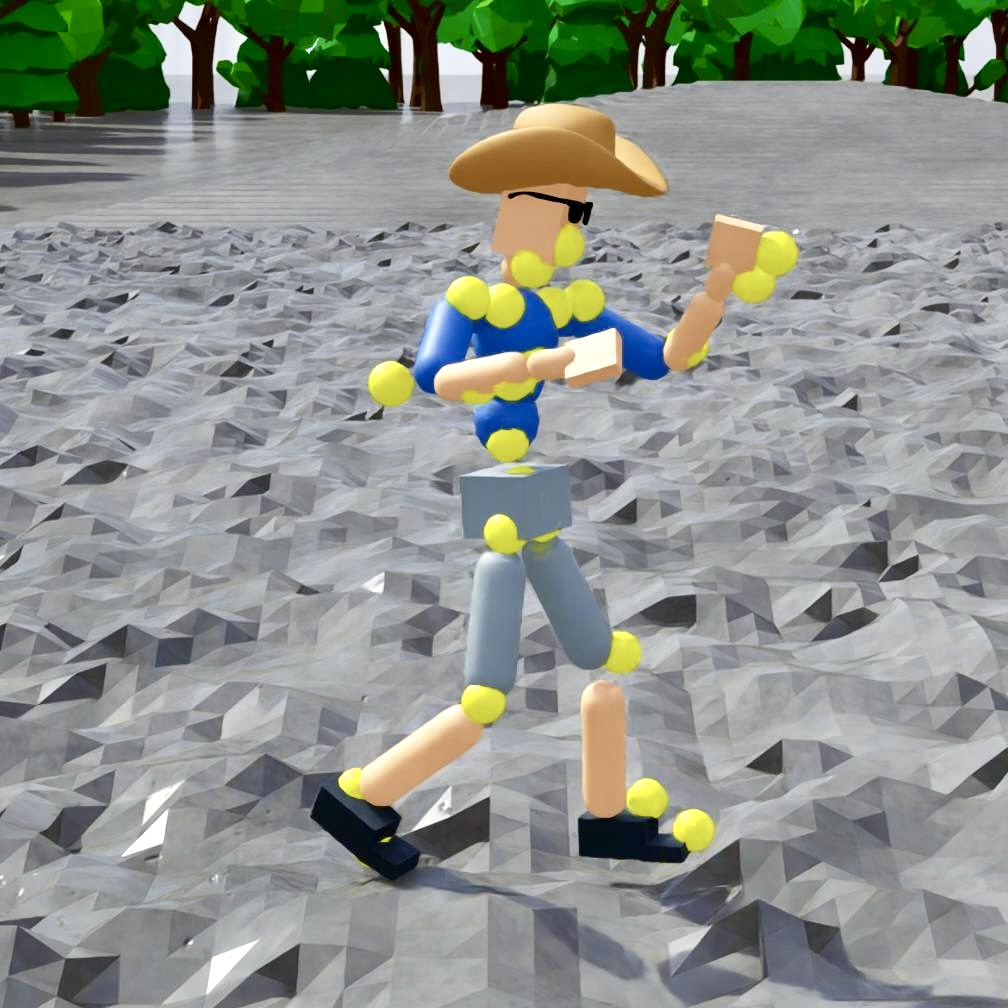}\hfill
        \includegraphics[width=0.1970\linewidth]{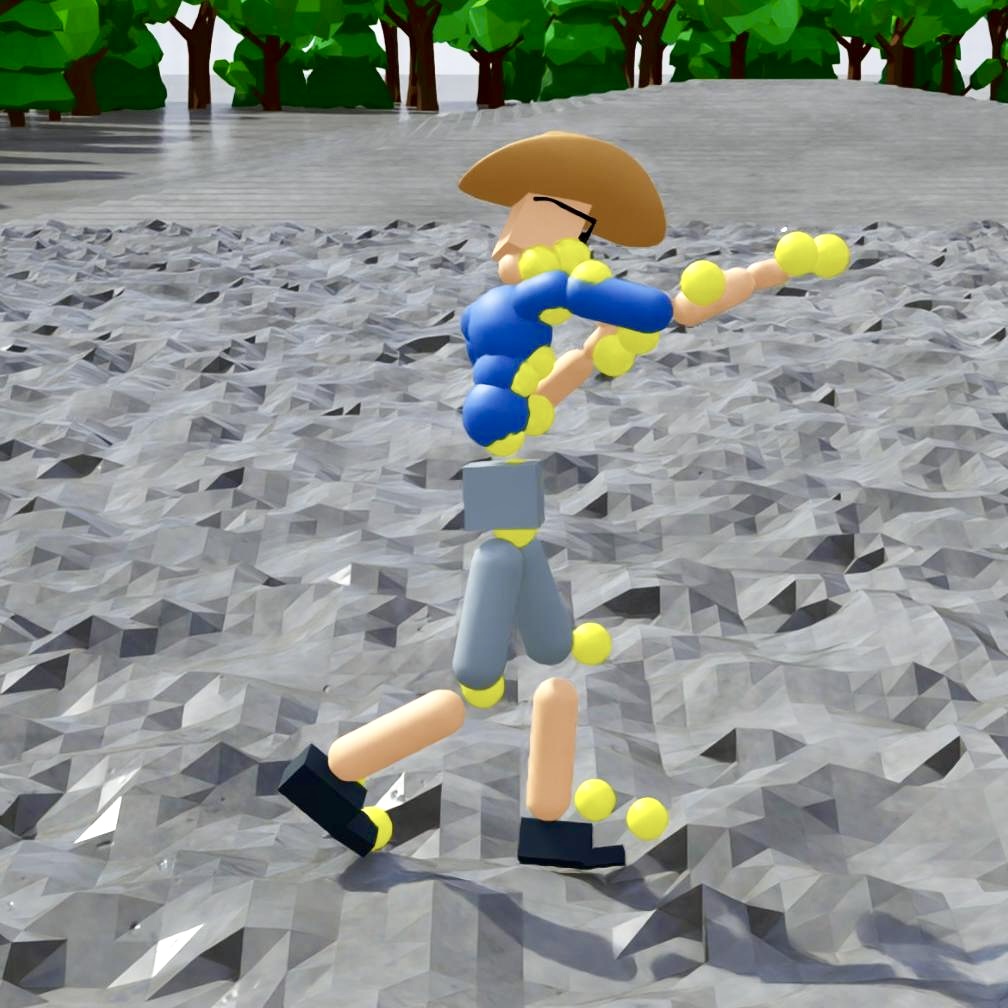}\hfill
        \includegraphics[width=0.1970\linewidth]{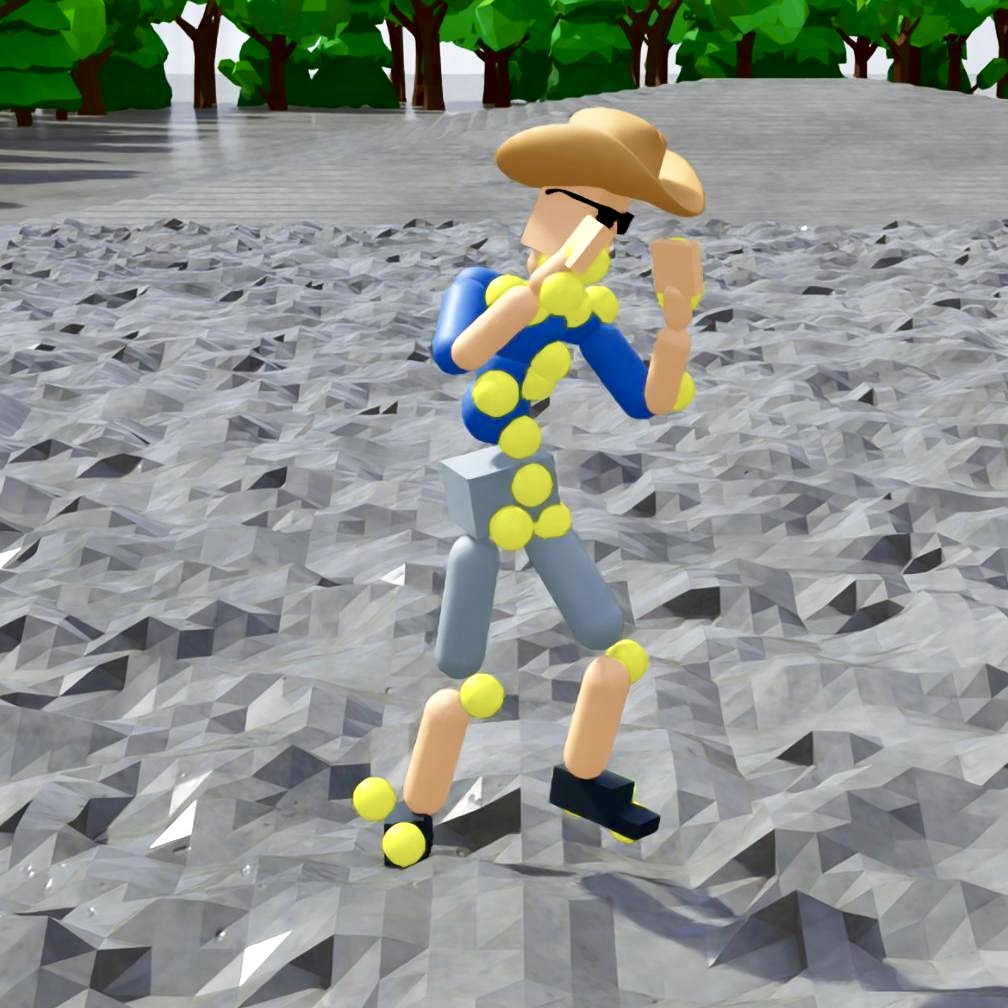}\hfill
        \includegraphics[width=0.1970\linewidth]{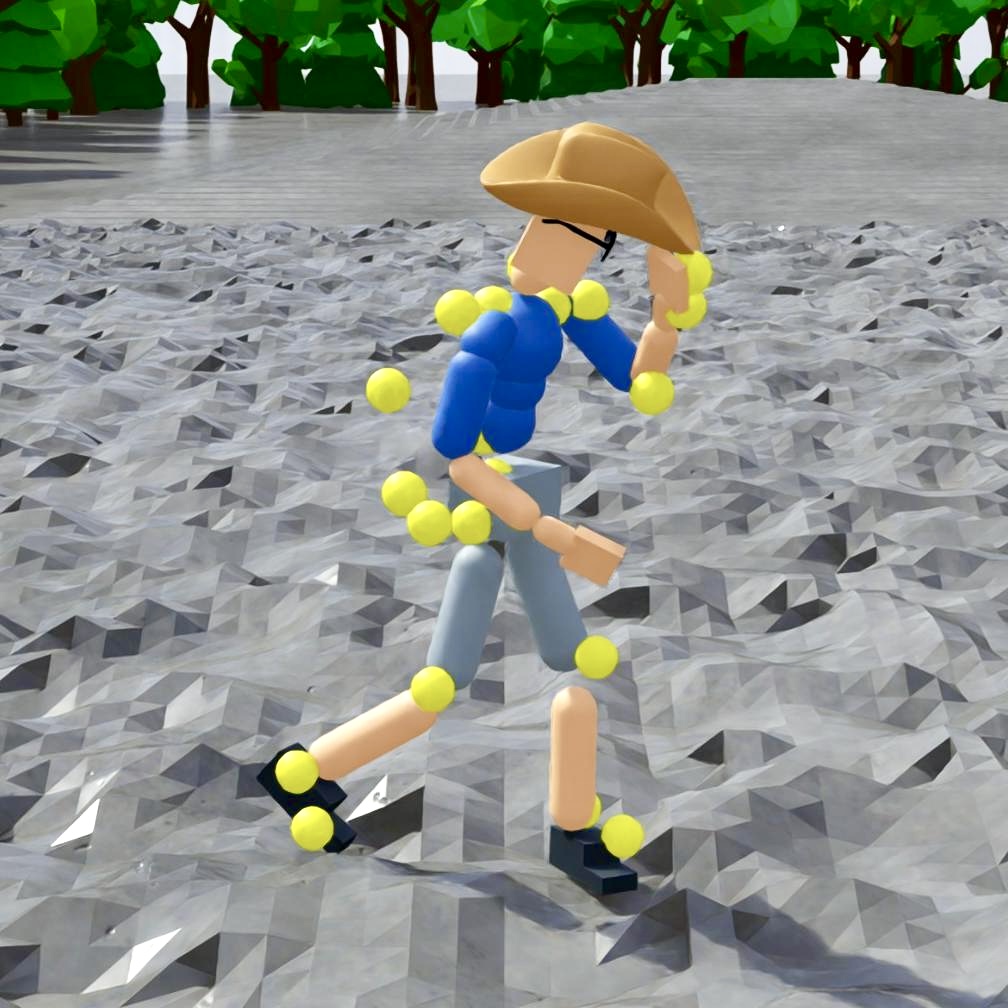}\hfill
        \includegraphics[width=0.1970\linewidth]{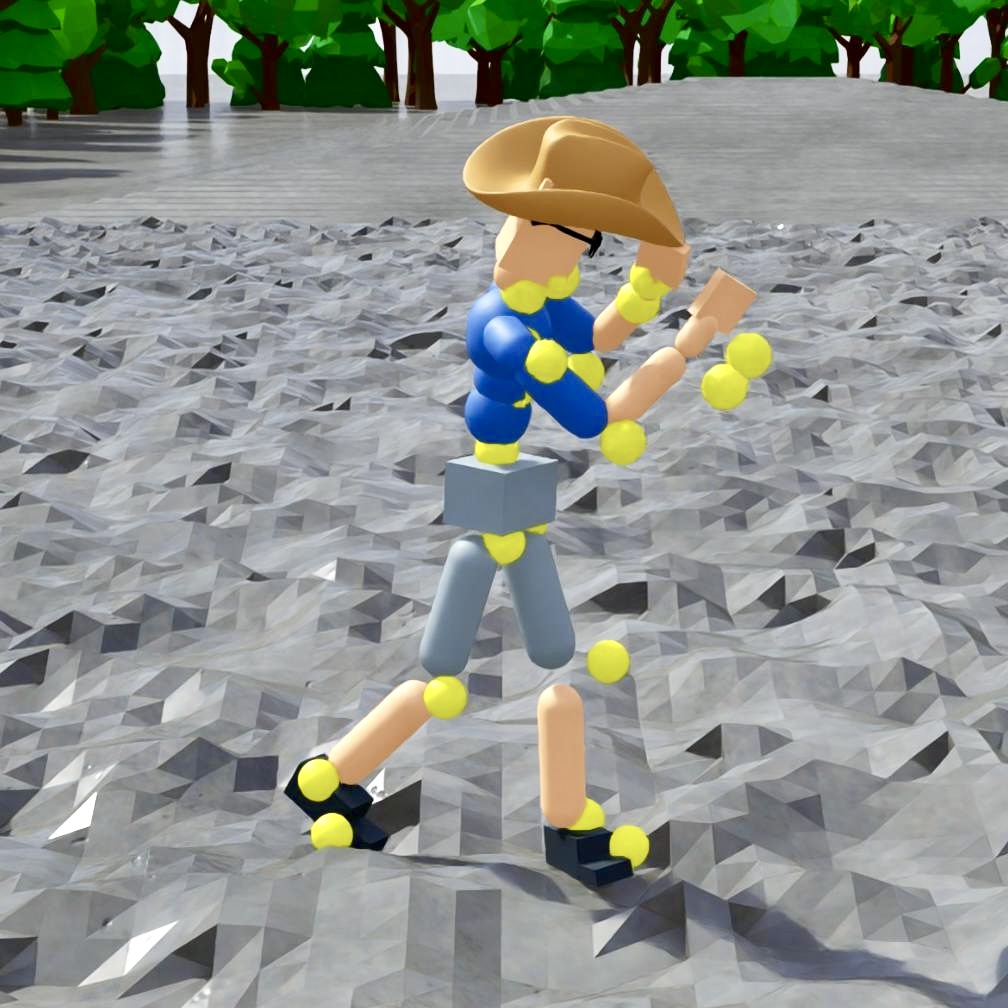}
        \label{fig:trc_box}
    \end{subfigure}\\[0.1em]
    \begin{subfigure}[t]{\columnwidth}
        \includegraphics[width=0.1970\linewidth]{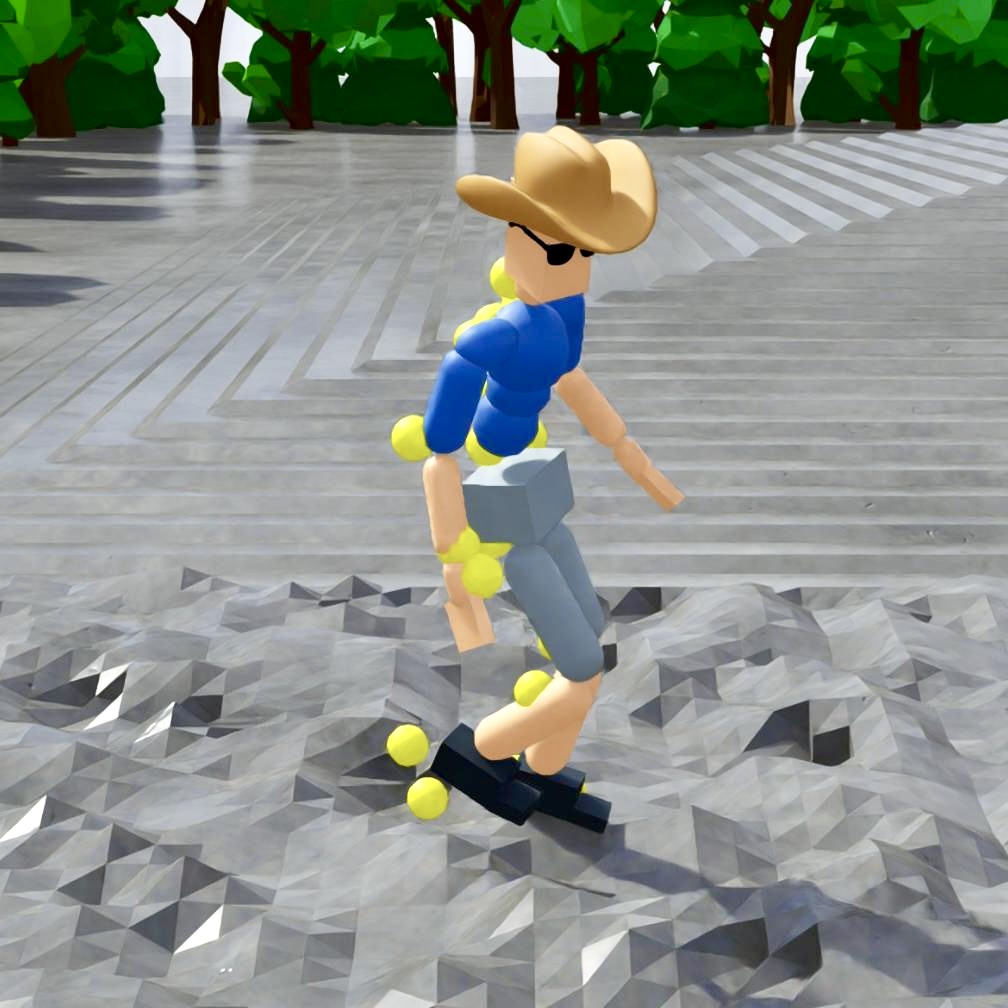}\hfill
        \includegraphics[width=0.1970\linewidth]{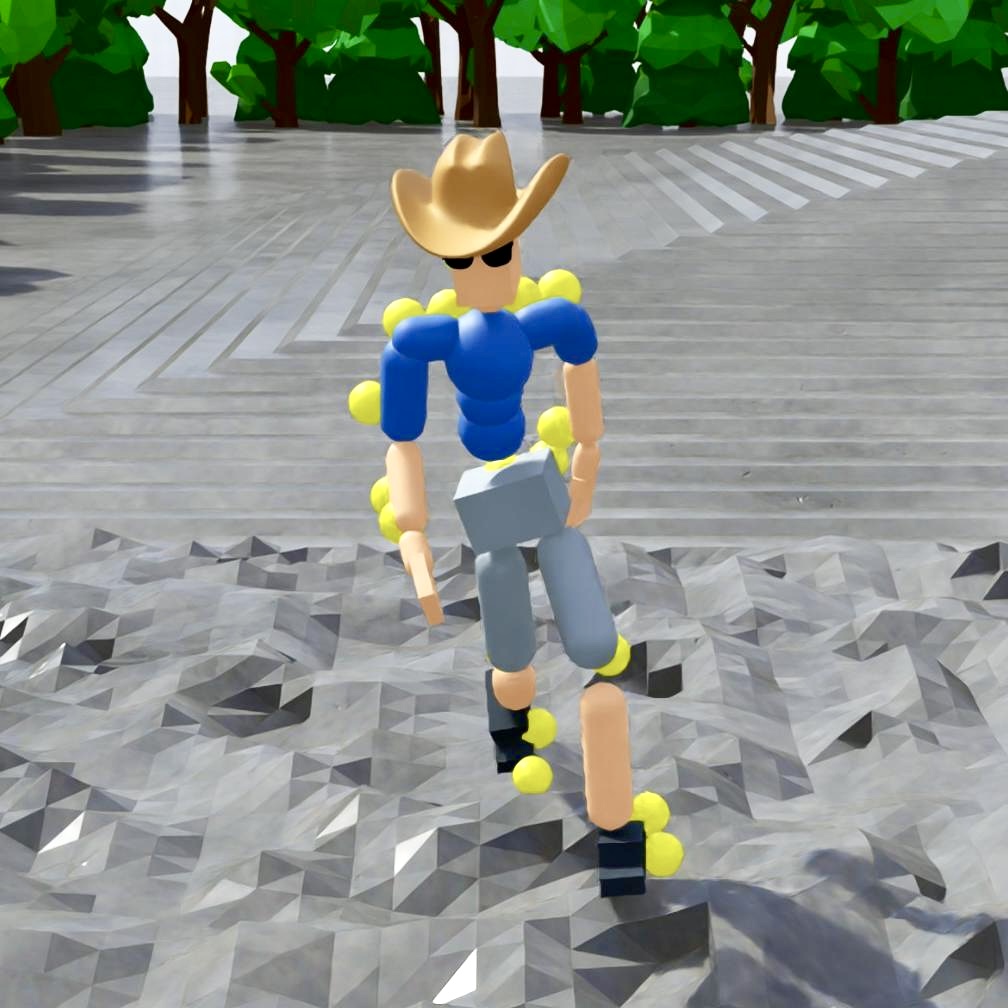}\hfill
        \includegraphics[width=0.1970\linewidth]{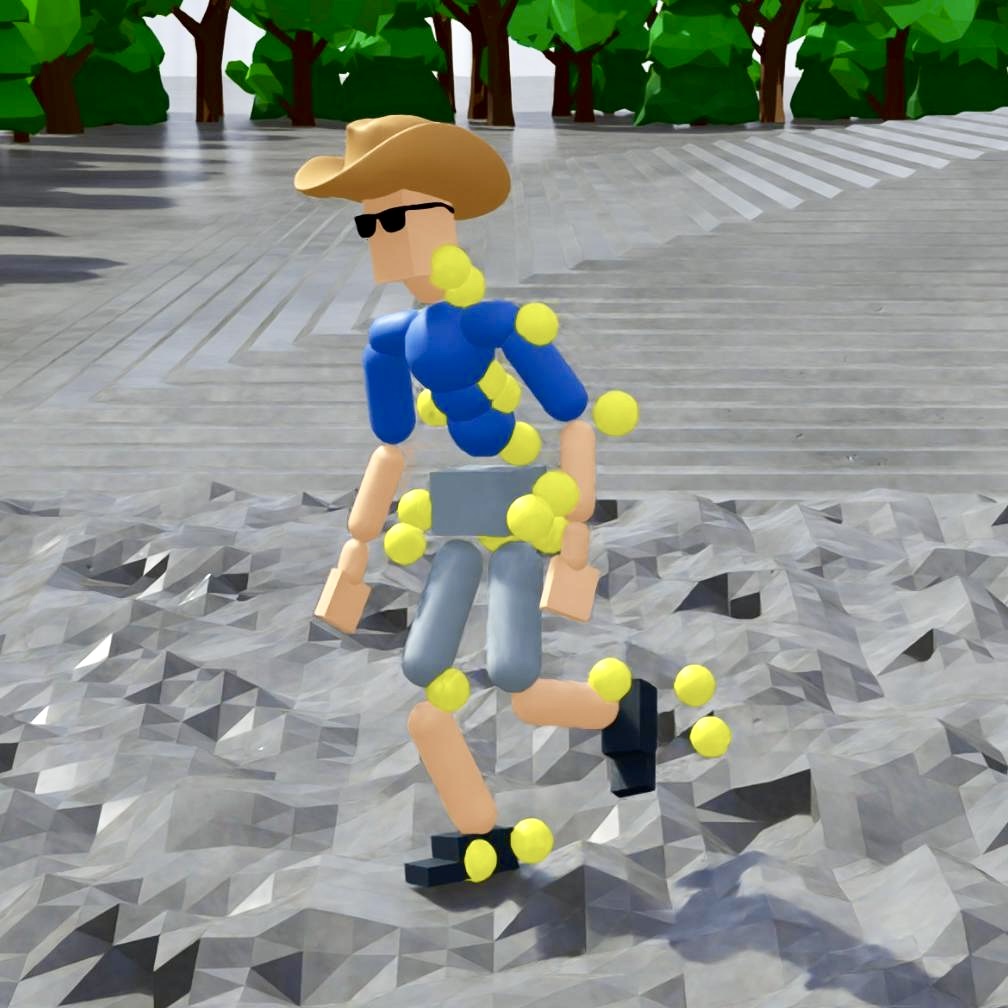}\hfill
        \includegraphics[width=0.1970\linewidth]{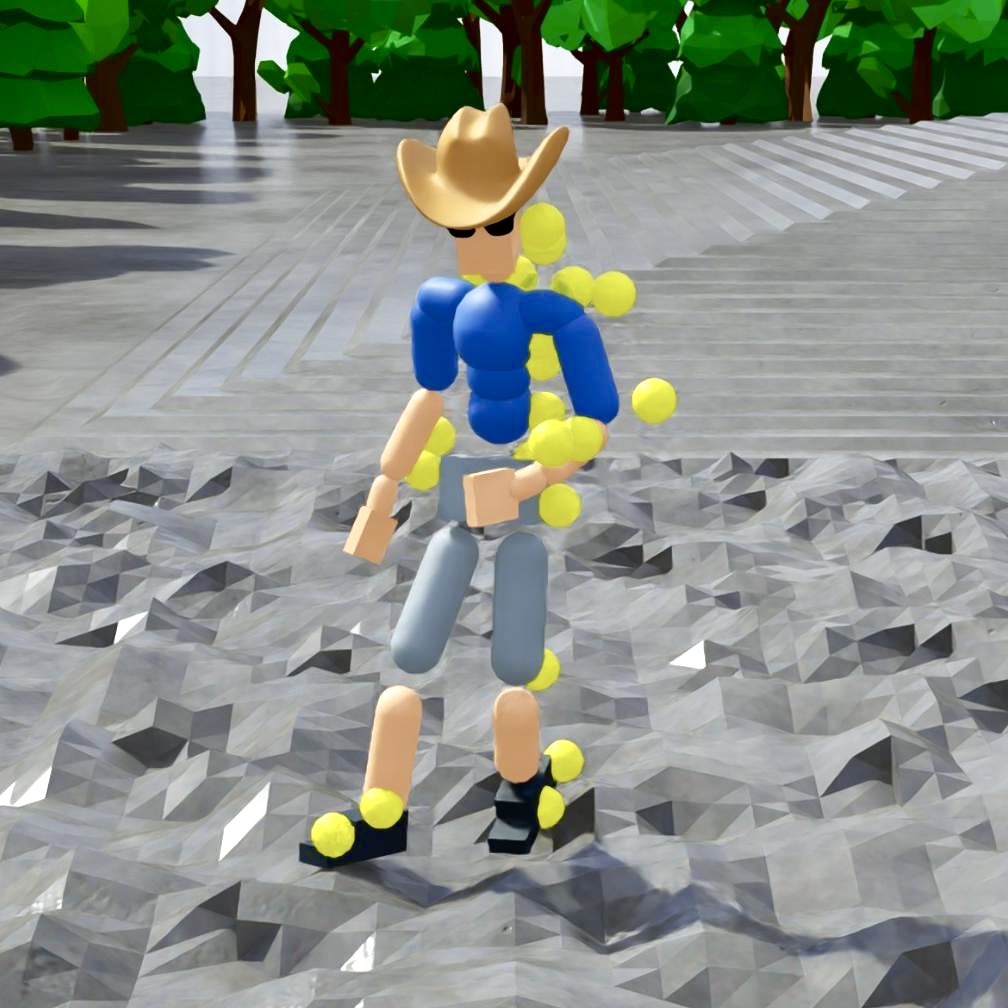}\hfill
        \includegraphics[width=0.1970\linewidth]{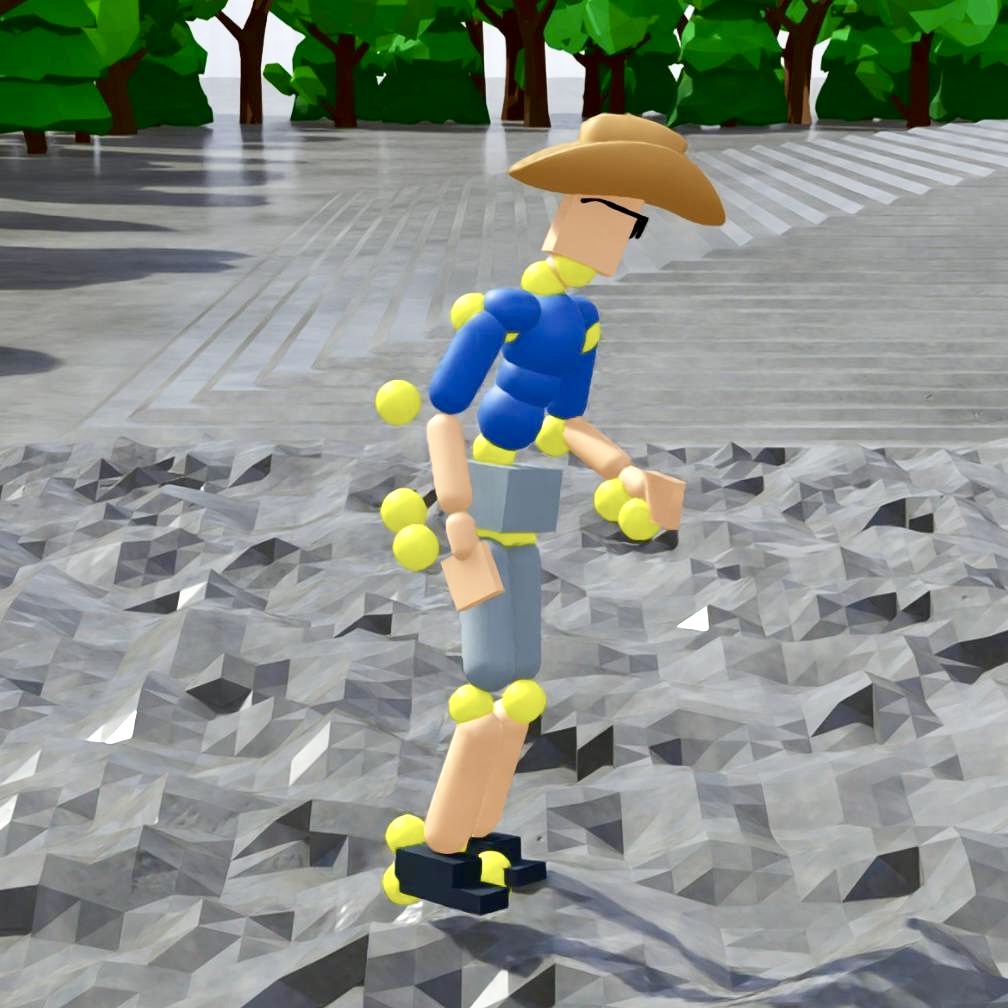}
        \label{fig:trc_run}
    \end{subfigure}
    \caption{Qualitative results of motion tracking. Each row shows temporally ordered rollout frames generated by the tracking skill $\mathscr{F}^{trc}$. The examples demonstrate that the shared decoder can reproduce diverse reference motions with stable balance, plausible contacts, and coordinated full-body control.}
    \label{fig:trc_qualitative}
\end{figure}

\begin{table}[t]
\centering
\caption{Motion-tracking performance on AMASS. Success rate reports the percentage of rollouts that remain below the 0.5 m MPJPE failure threshold on flat terrain. The upper block compares HetSkills with MaskedMimic and PULSE, including one-frame and two-frame inference variants of HetSkills. The lower block ablates the number of latent body partitions, showing the trade-off between tracking accuracy, part-wise interpretability, and computational efficiency.}
\label{tab:tracking_results}
\setlength{\tabcolsep}{6pt}
\renewcommand{\arraystretch}{1.1}
\begin{tabular}{l|cc|cc}
\hline\hline
& \multicolumn{2}{c|}{Train} & \multicolumn{2}{c}{Test} \\
\cline{2-5}
Method & Success & MPJPE & Success & MPJPE \\
\hline
MaskedMimic       & 99.4\% & 32.9 & 99.2\% & 35.1 \\
PULSE             & 99.8\% & 39.2 & 97.1\% & 54.1 \\
HetSkills-1step   & 99.9\% & 29.2 & 99.3\% & 43.6 \\
HetSkills-2step & 99.9\% & 28.2 & 100\% & 39.4 \\
\hline\hline
HetSkills-1part   & 99.9\% & 31.8 & 100\% & 42.3 \\
HetSkills-5part   & 100\% & 26.0 & 100\% & 35.5 \\
\hline\hline
\end{tabular}
\end{table}

\subsection{Tracking and Motion Completion}
\label{result:trc_and_moc}

We evaluate HetSkills on motion tracking and motion completion. These tasks assess the ability of the model to generate precise motions and handle sparse conditioning. First, we present motion tracking results and compare HetSkills with baselines. Next, we evaluate motion completion using partial input data, highlighting the ability to handle varying partial observations.

\begin{figure*}[t]
    \centering
    \begin{subfigure}[t]{\textwidth}
        \centering
        \includegraphics[width=0.2200\linewidth]{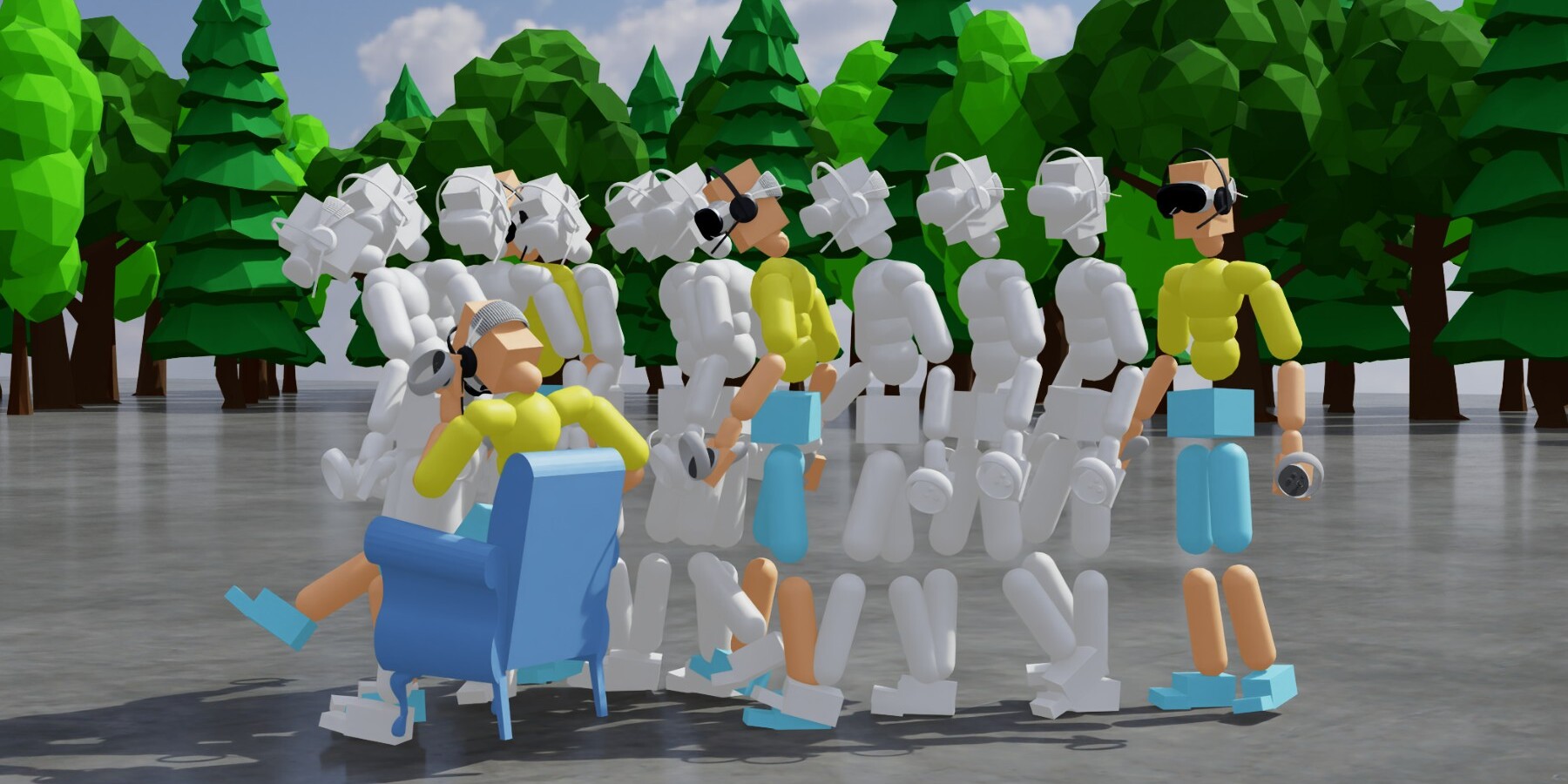}\hfill
        \includegraphics[width=0.1100\linewidth]{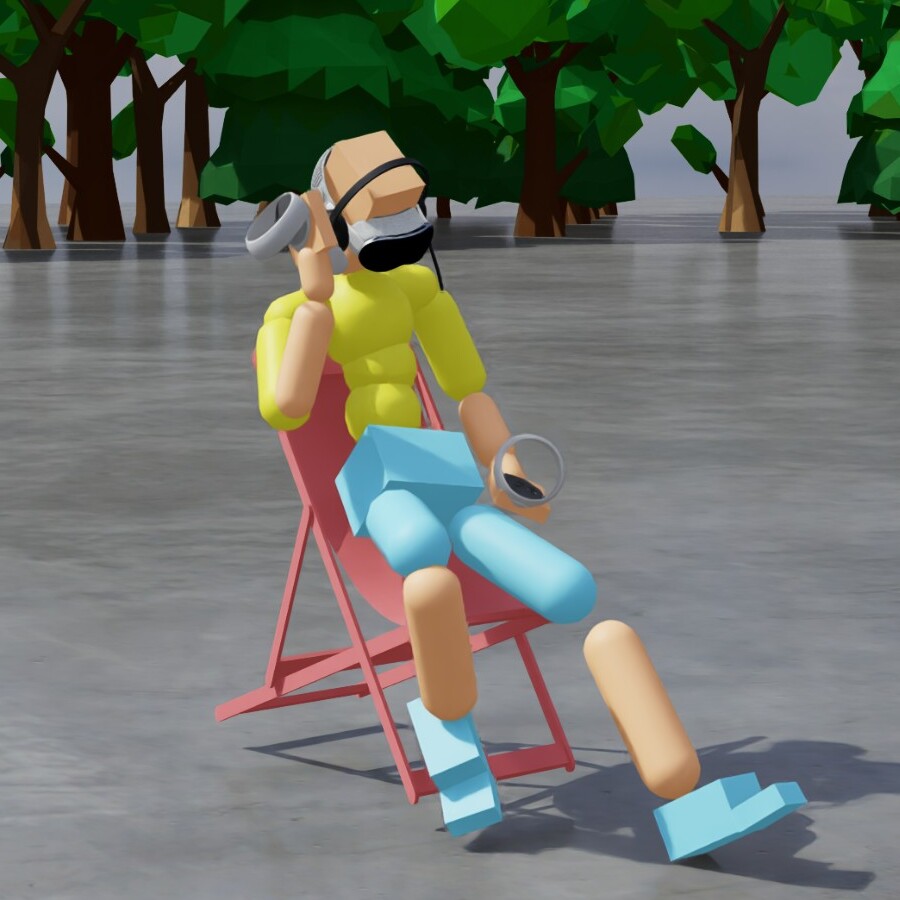}\hfill
        \includegraphics[width=0.1100\linewidth]{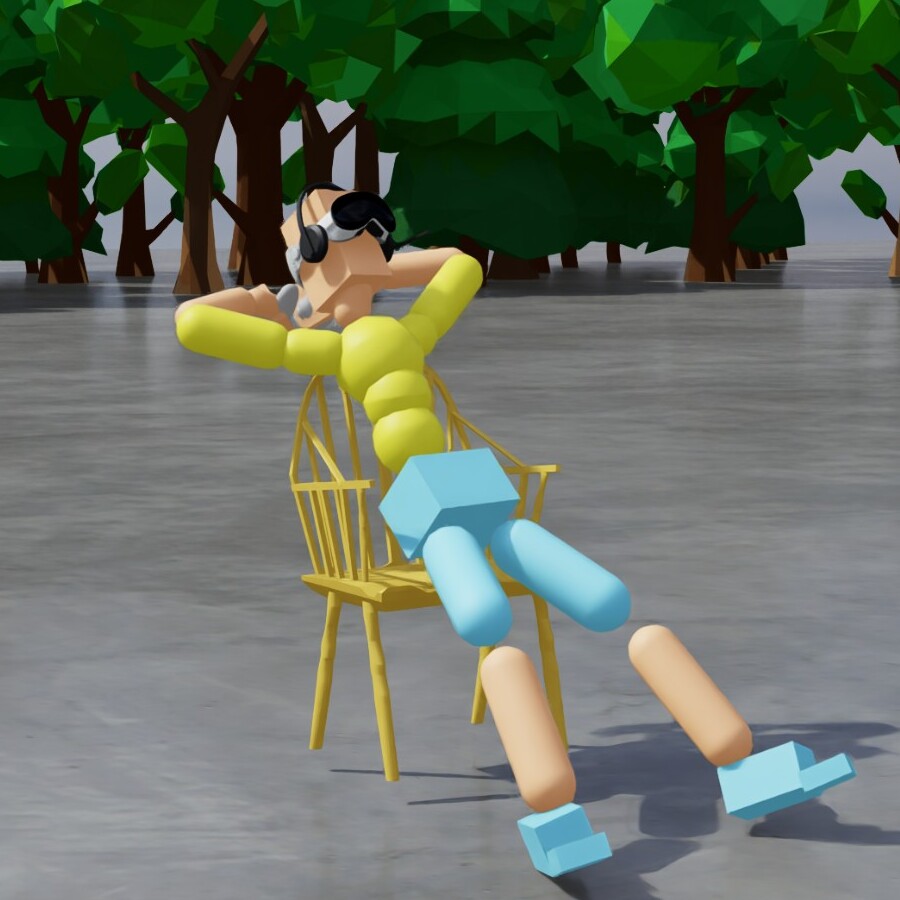}\hfill
        \includegraphics[width=0.1100\linewidth]{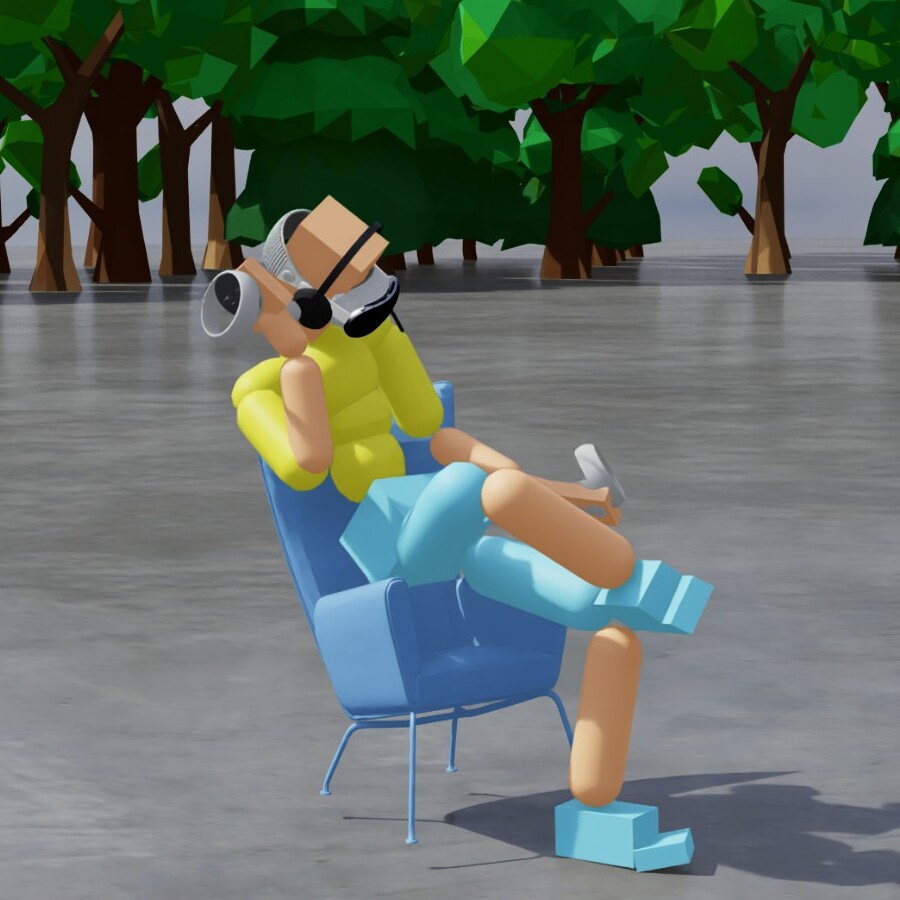}\hfill
        \includegraphics[width=0.1100\linewidth]{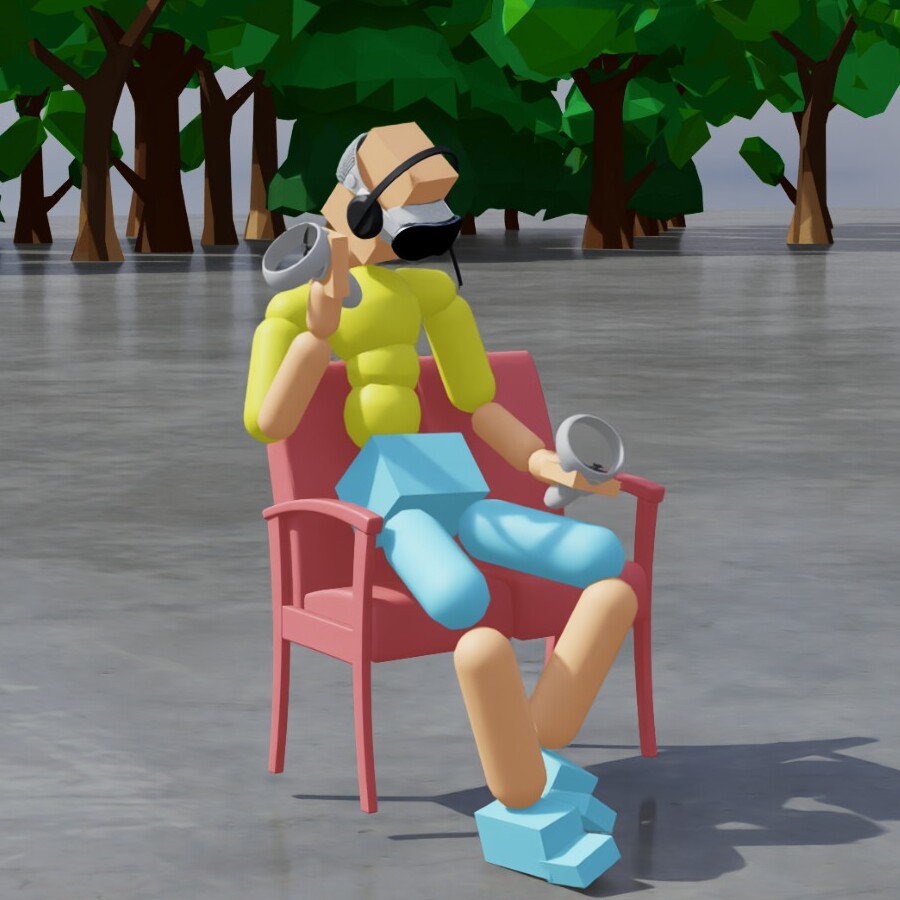}\hfill
        \includegraphics[width=0.1100\linewidth]{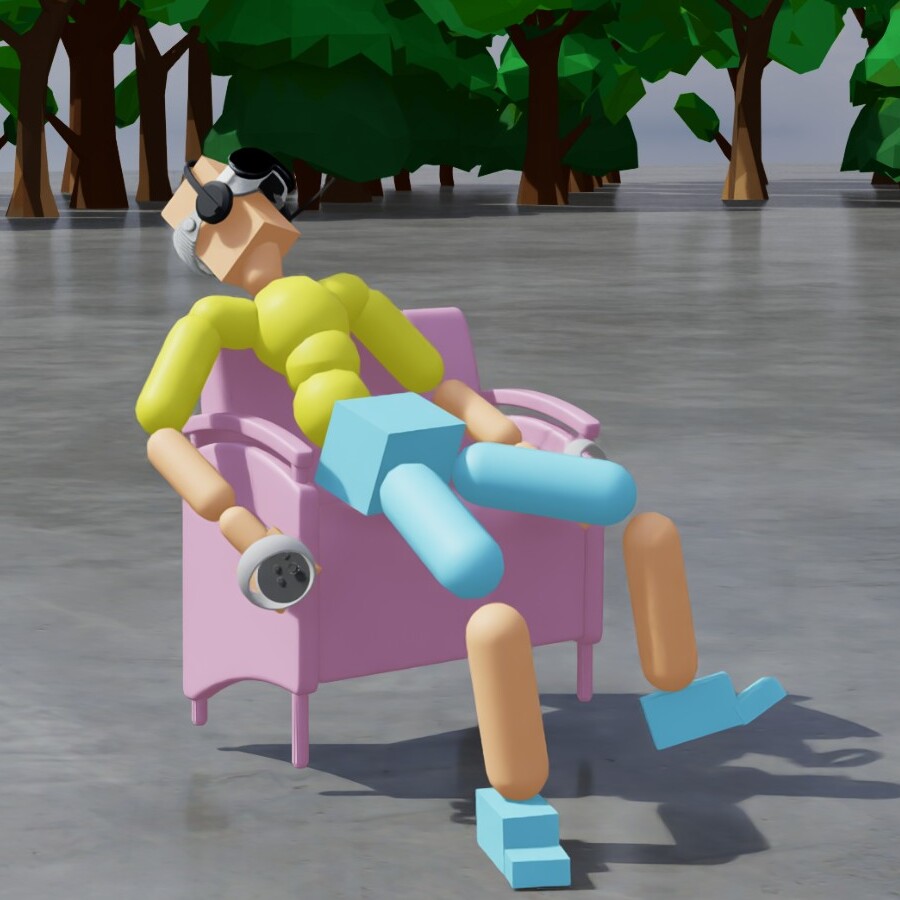}\hfill
        \includegraphics[width=0.1100\linewidth]{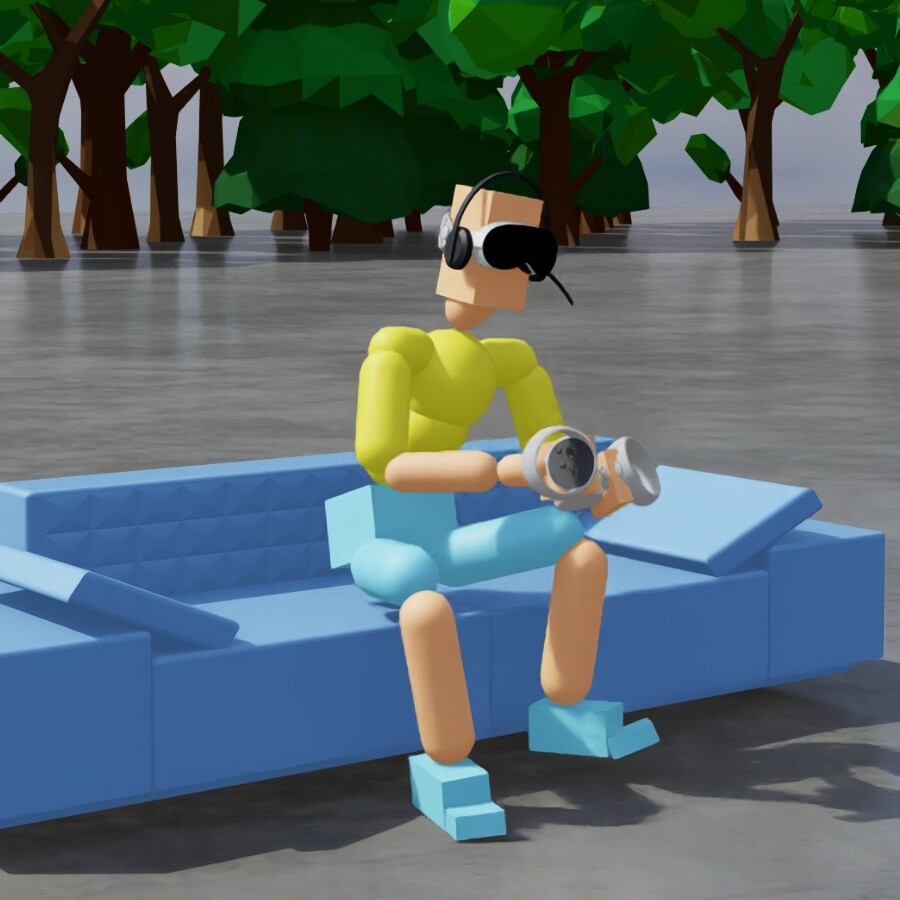}\hfill
        \includegraphics[width=0.1100\linewidth]{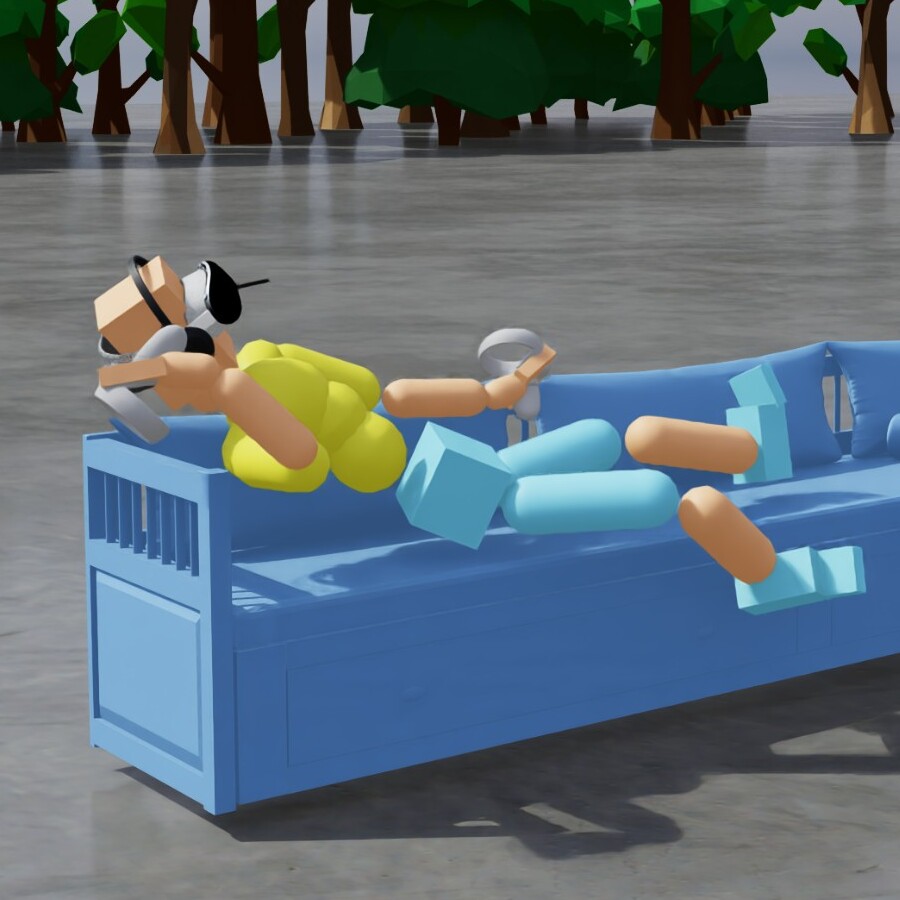}
        \caption{\textbf{Human-Scene Interaction:} the character naturally interacts with diverse everyday objects in a variety of poses.}
    \end{subfigure}
    \begin{subfigure}[t]{\textwidth}
        \centering
        \includegraphics[width=0.3054\linewidth]{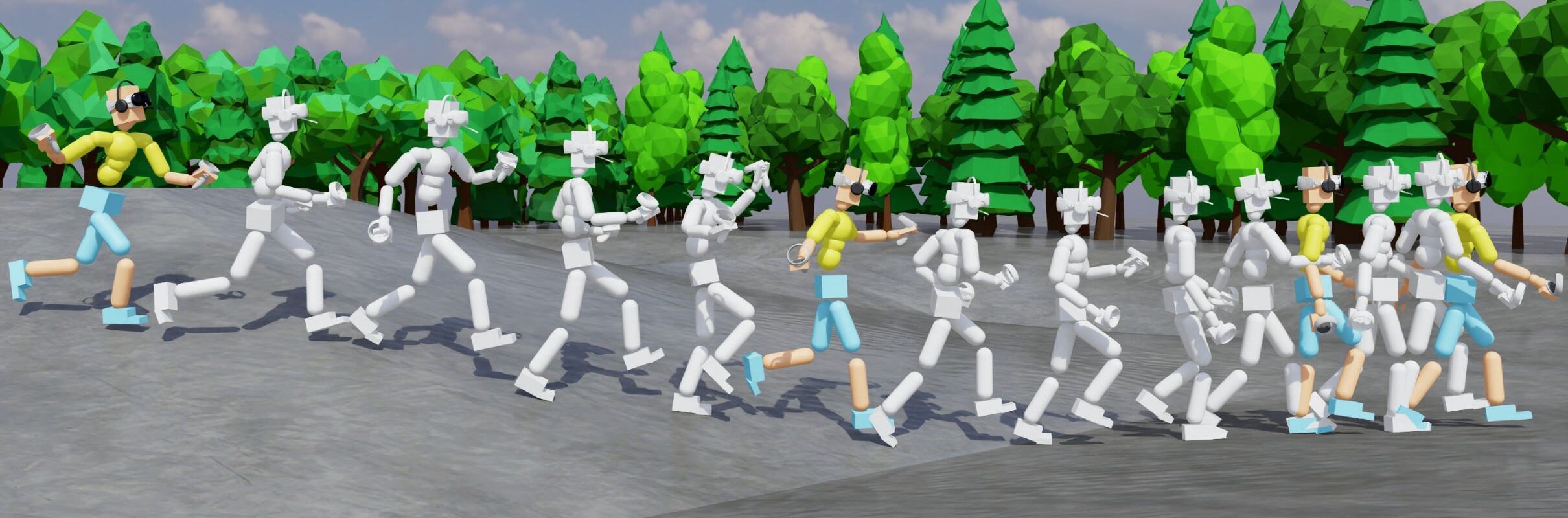}\hfill
        \includegraphics[width=0.2776\linewidth]{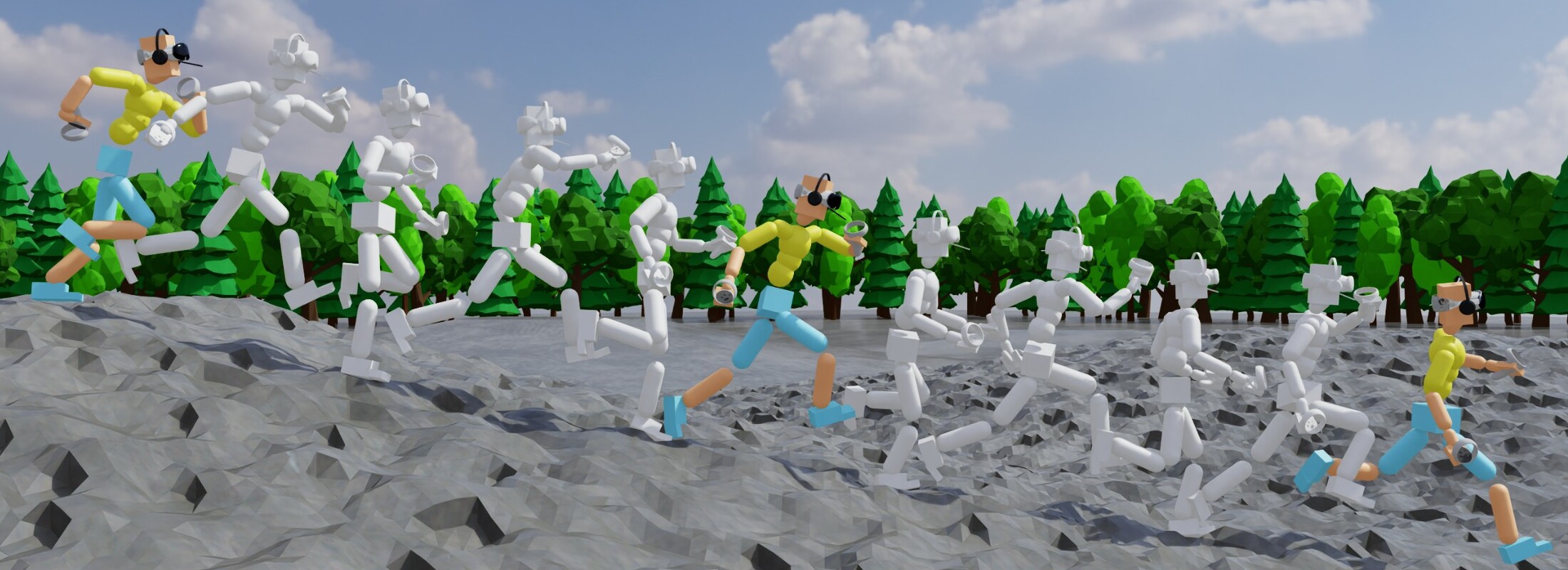}\hfill
        \includegraphics[width=0.2322\linewidth]{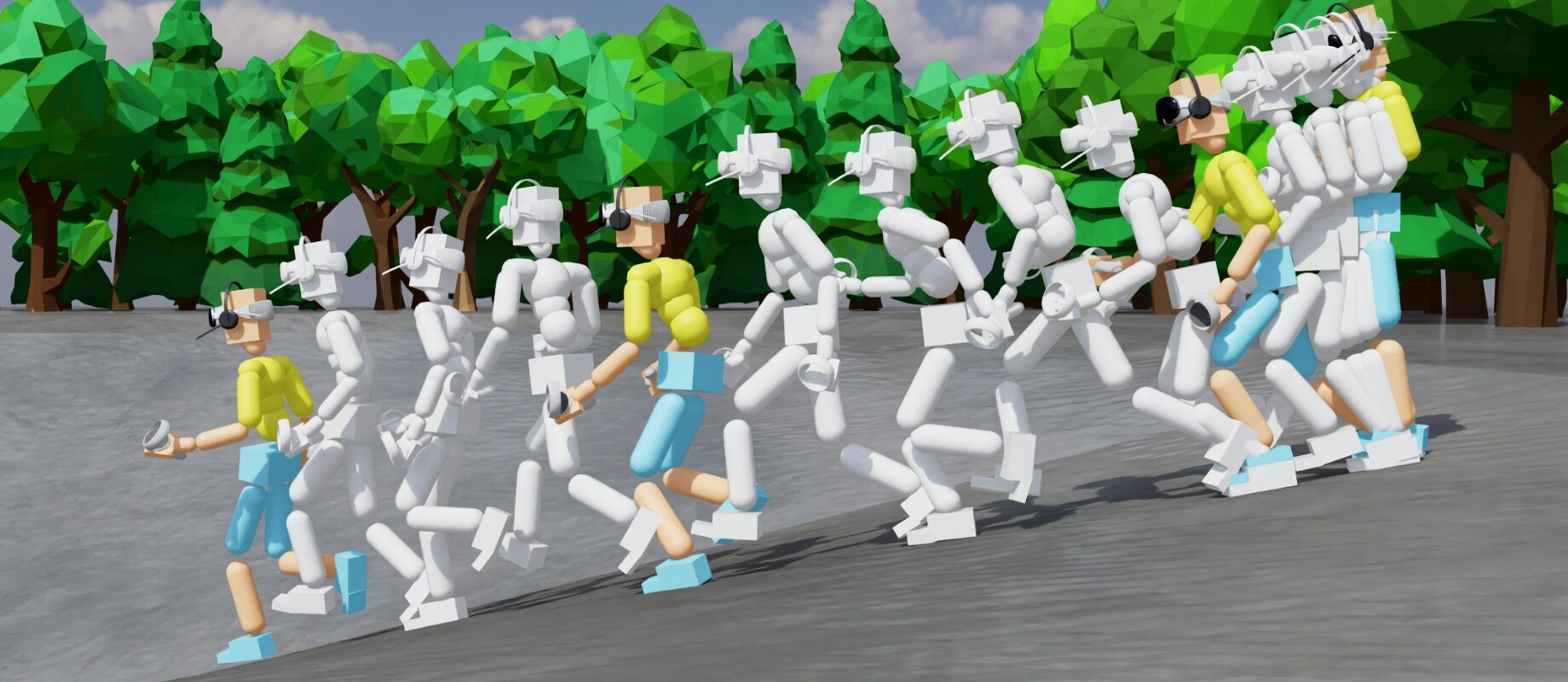}\hfill
        \includegraphics[width=0.1742\linewidth]{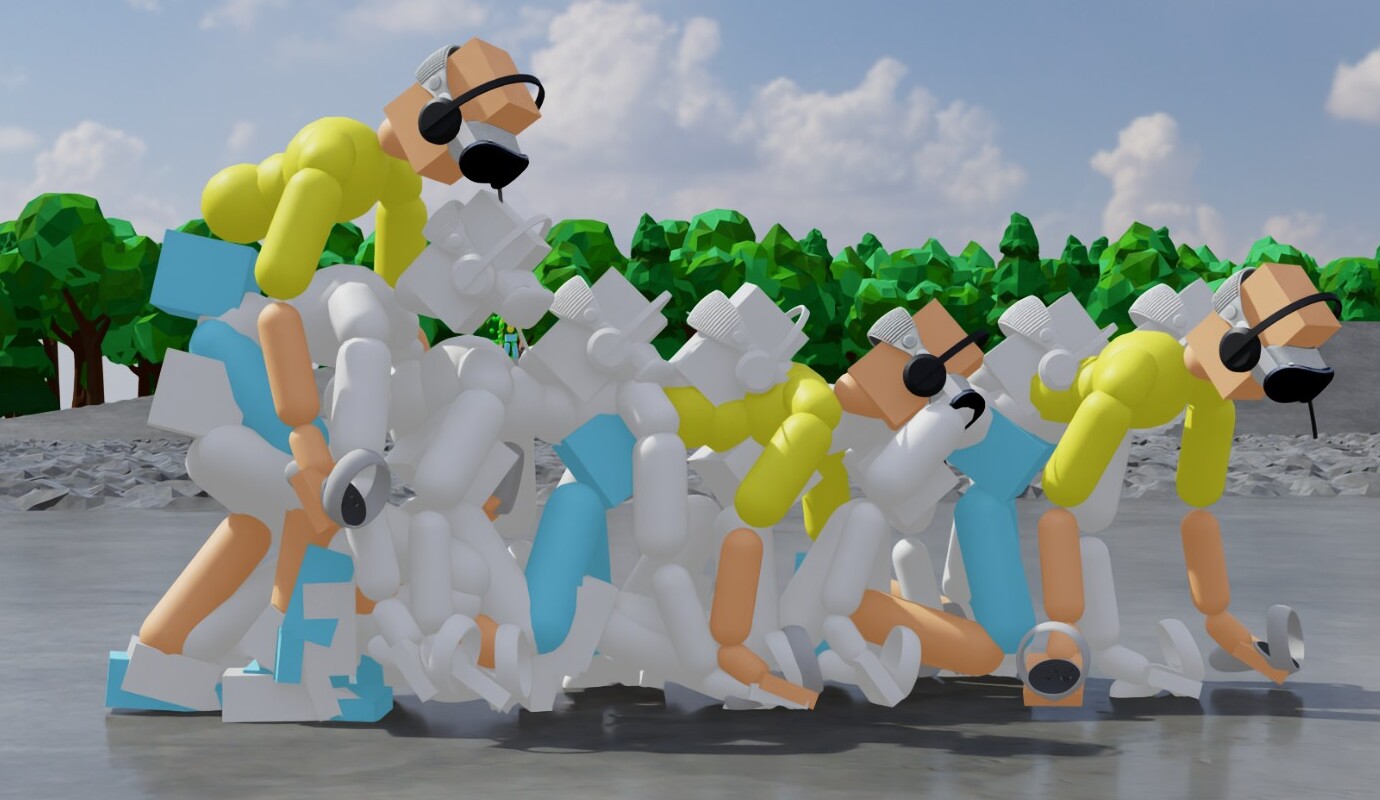}
        \caption{\textbf{Motion In-Betweening:} the character synthesizes physically plausible intermediate frames (white) between sparse keyframe poses (colored).}
    \end{subfigure}
    \begin{subfigure}[t]{\textwidth}
        \begin{subfigure}[t]{0.4995\textwidth}
            \centering
            \includegraphics[width=0.1975\linewidth]{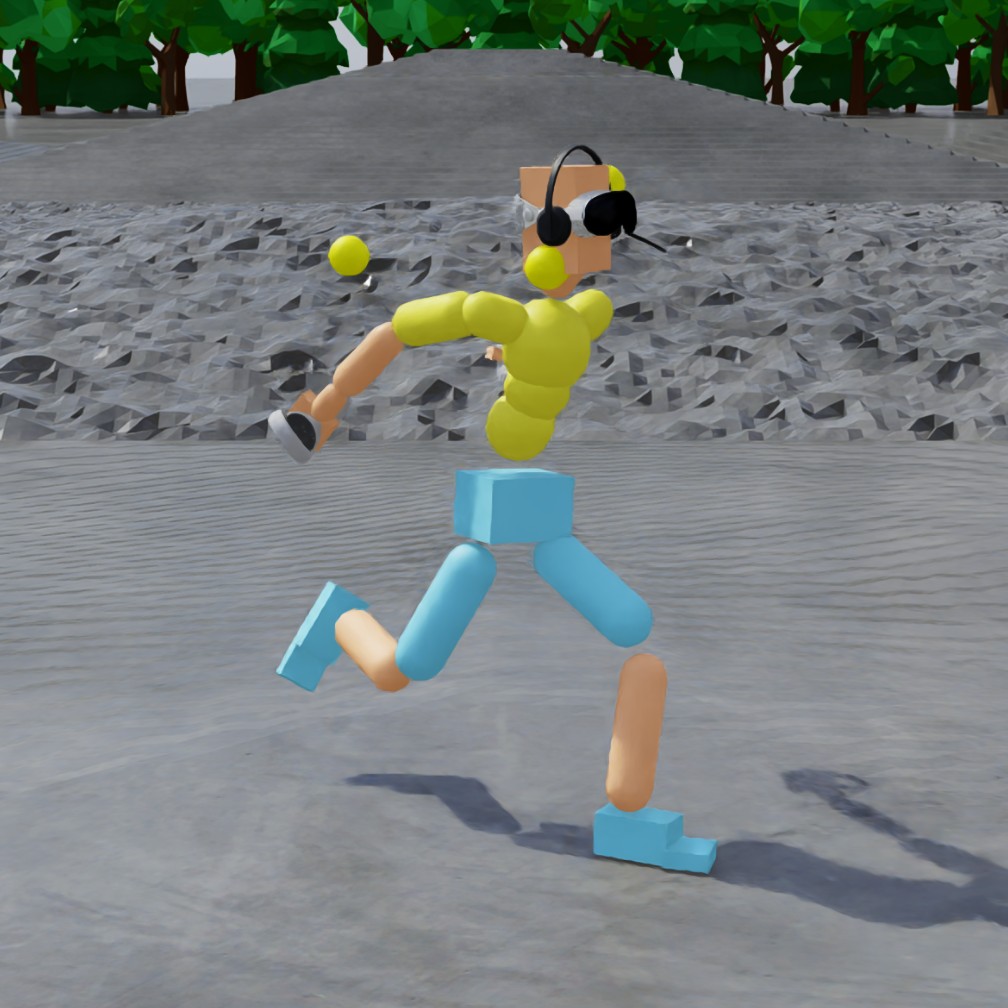}\hspace{0.5pt}%
            \includegraphics[width=0.1975\linewidth]{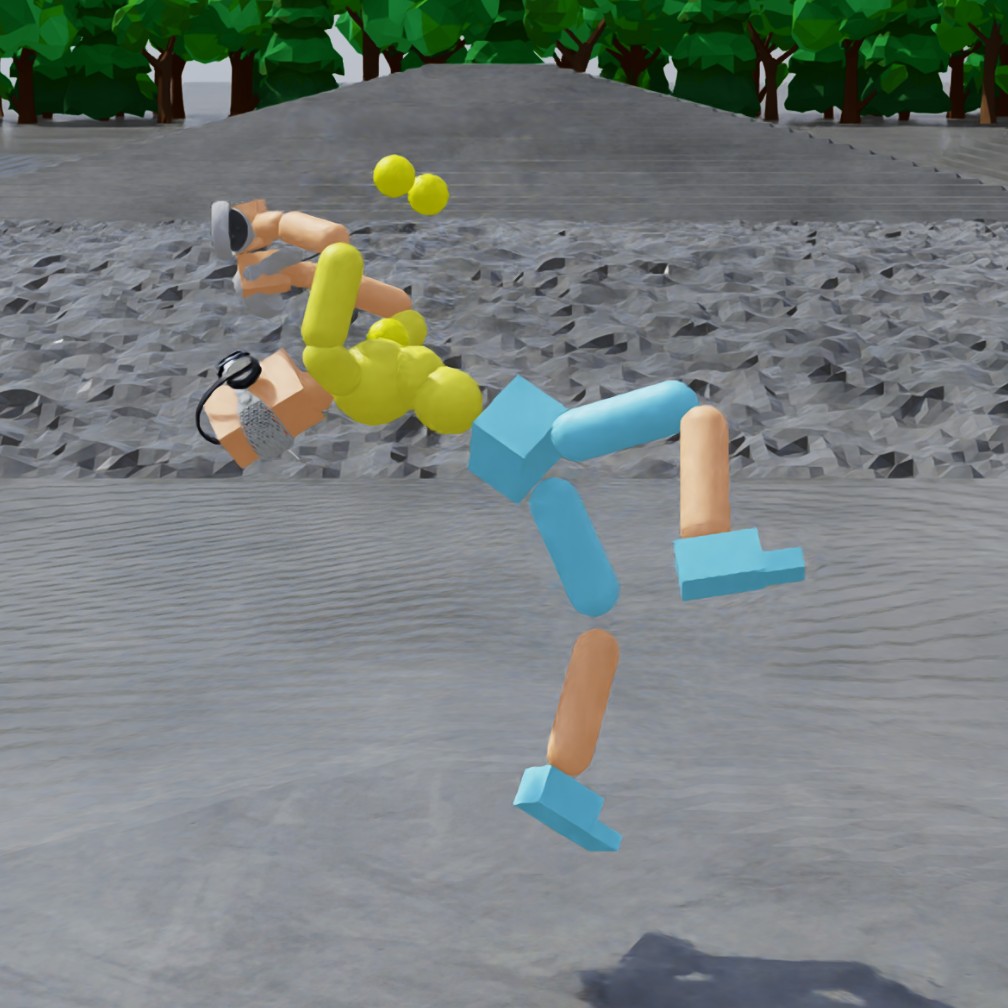}\hspace{0.5pt}%
            \includegraphics[width=0.1975\linewidth]{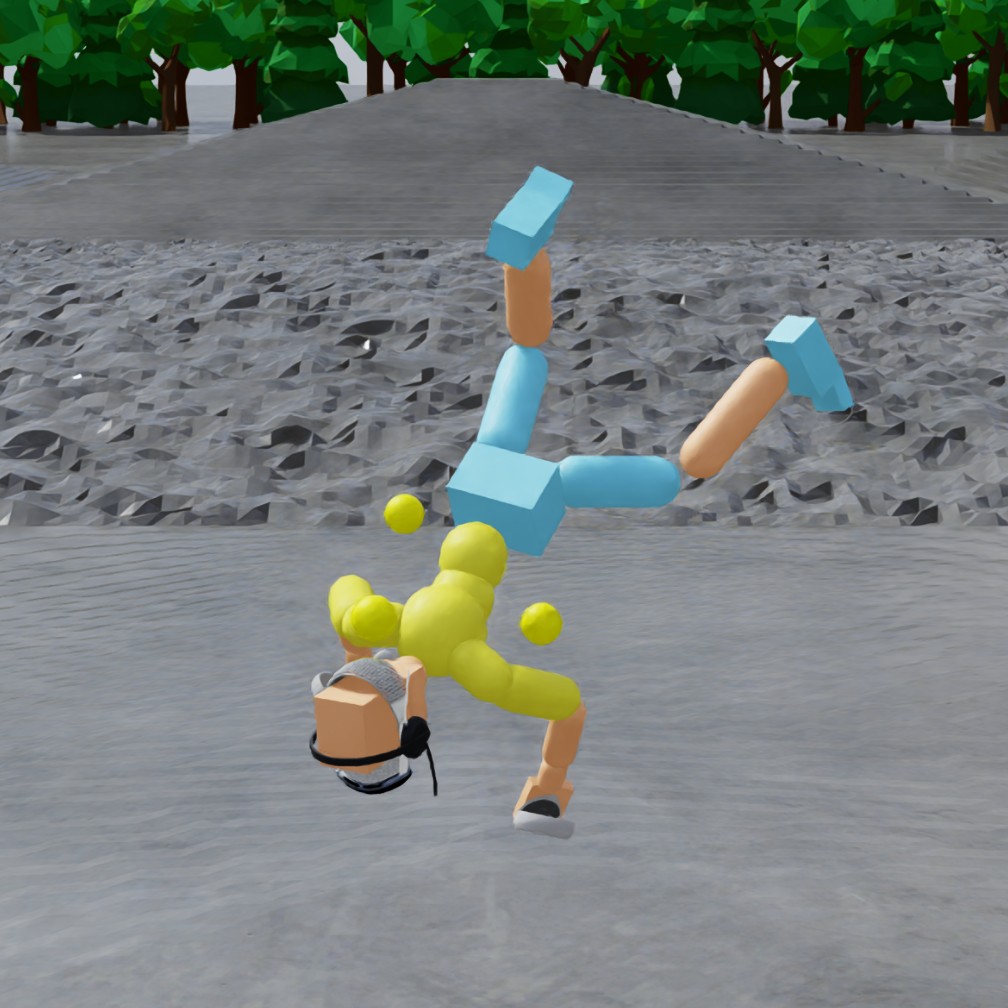}\hspace{0.5pt}%
            \includegraphics[width=0.1975\linewidth]{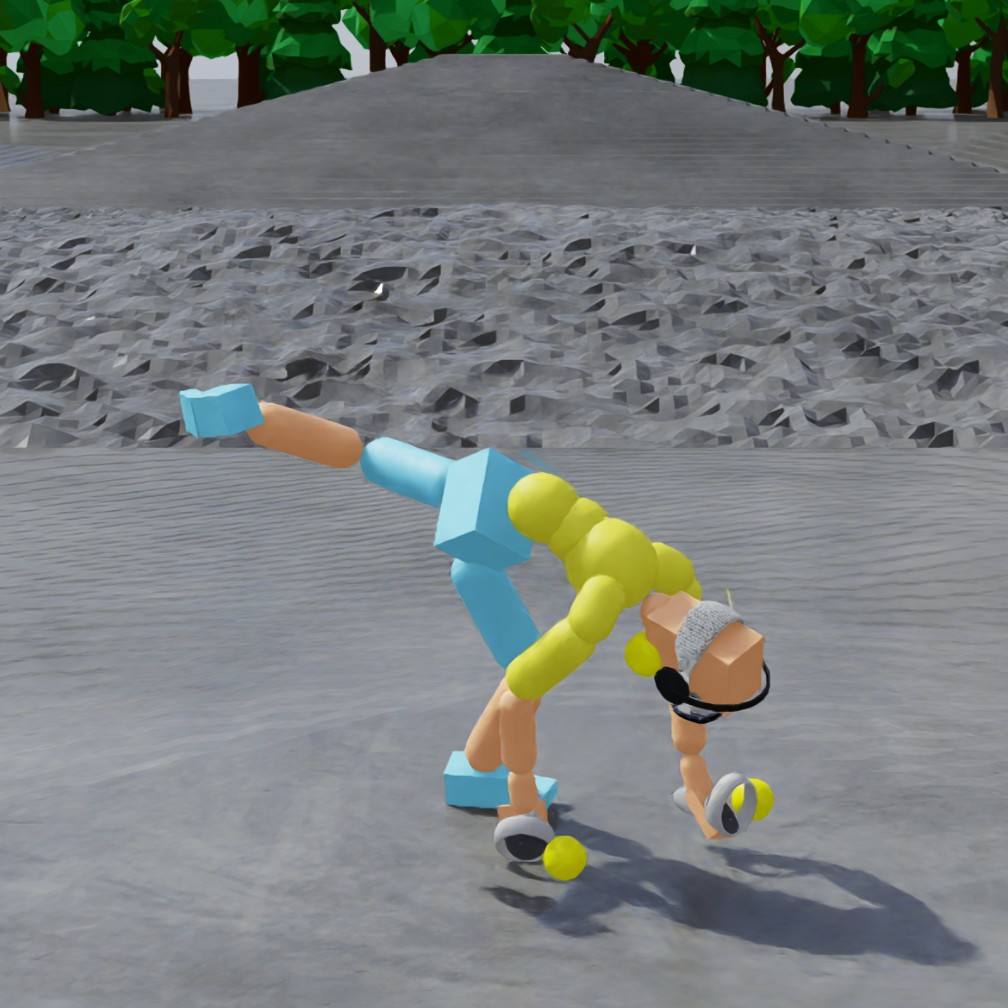}\hspace{0.5pt}%
            \includegraphics[width=0.1975\linewidth]{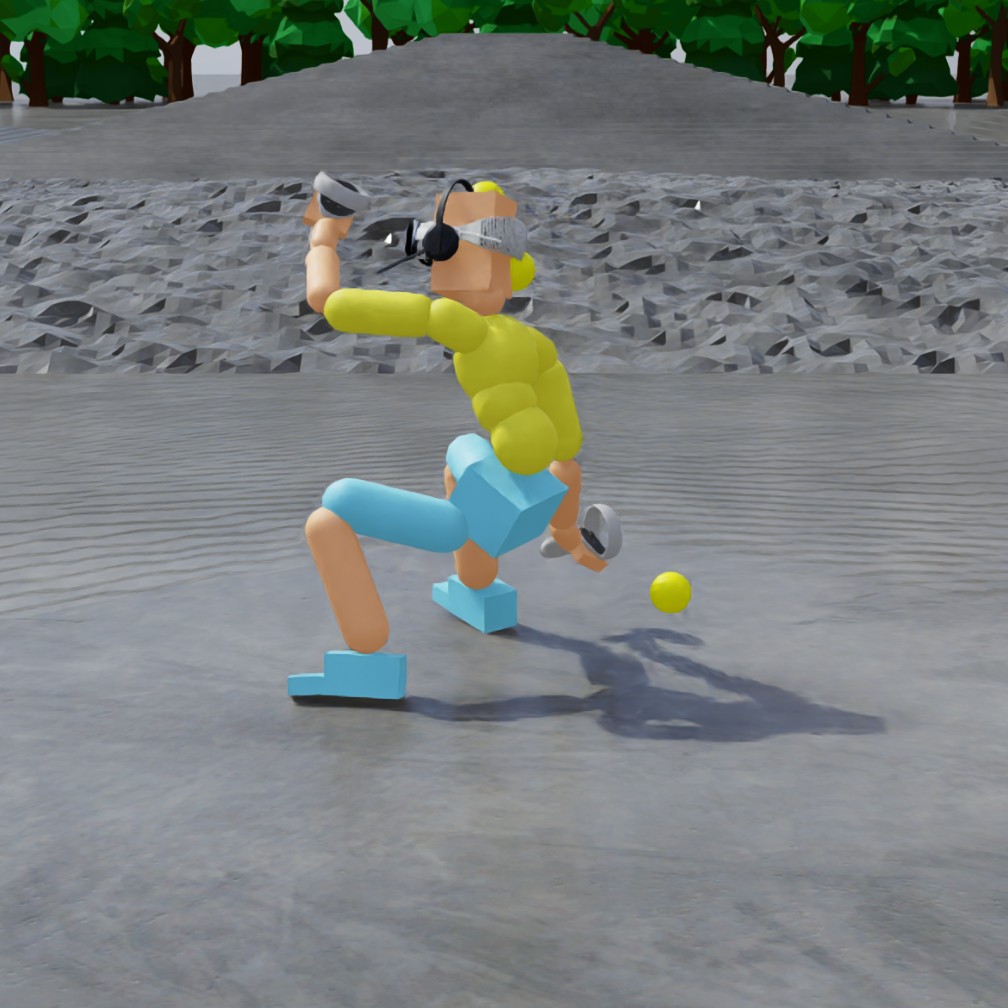}
        \end{subfigure}\hfill
        \begin{subfigure}[t]{0.4995\textwidth}
            \centering
            \includegraphics[width=0.1975\linewidth]{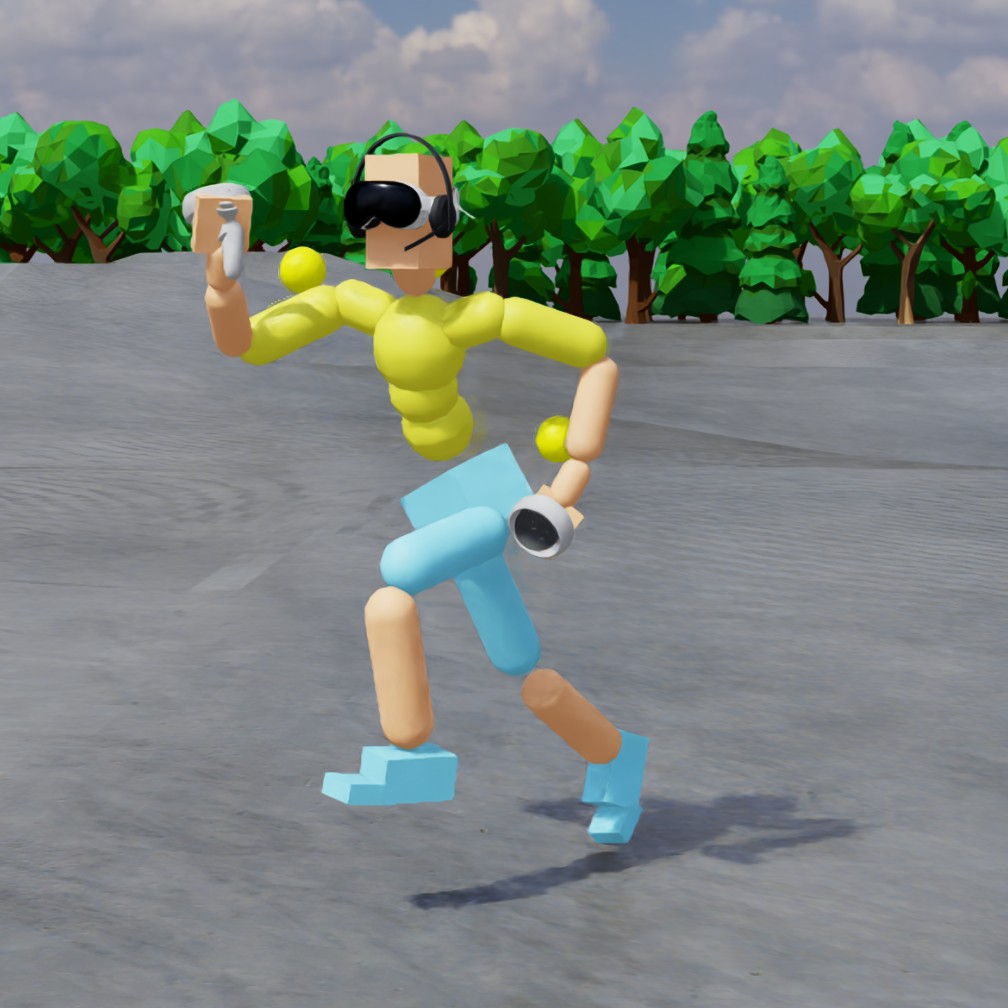}\hspace{0.5pt}%
            \includegraphics[width=0.1975\linewidth]{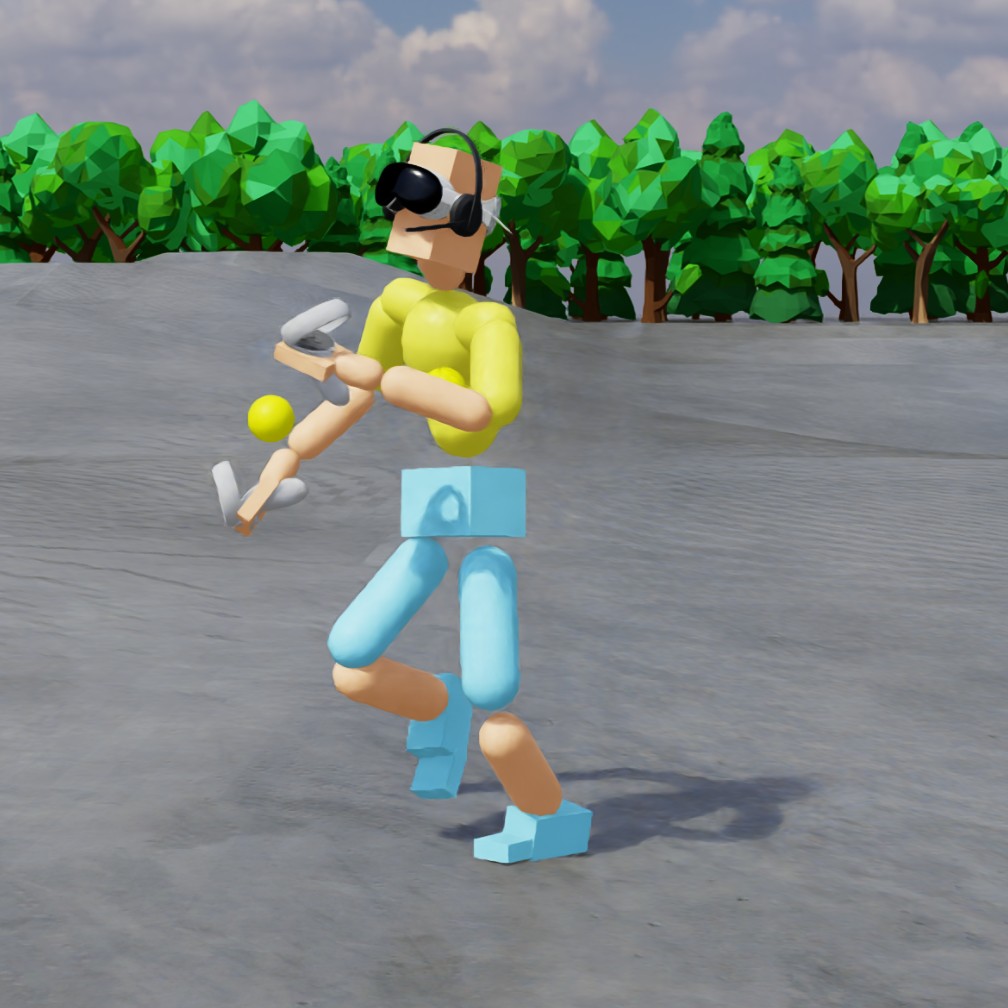}\hspace{0.5pt}%
            \includegraphics[width=0.1975\linewidth]{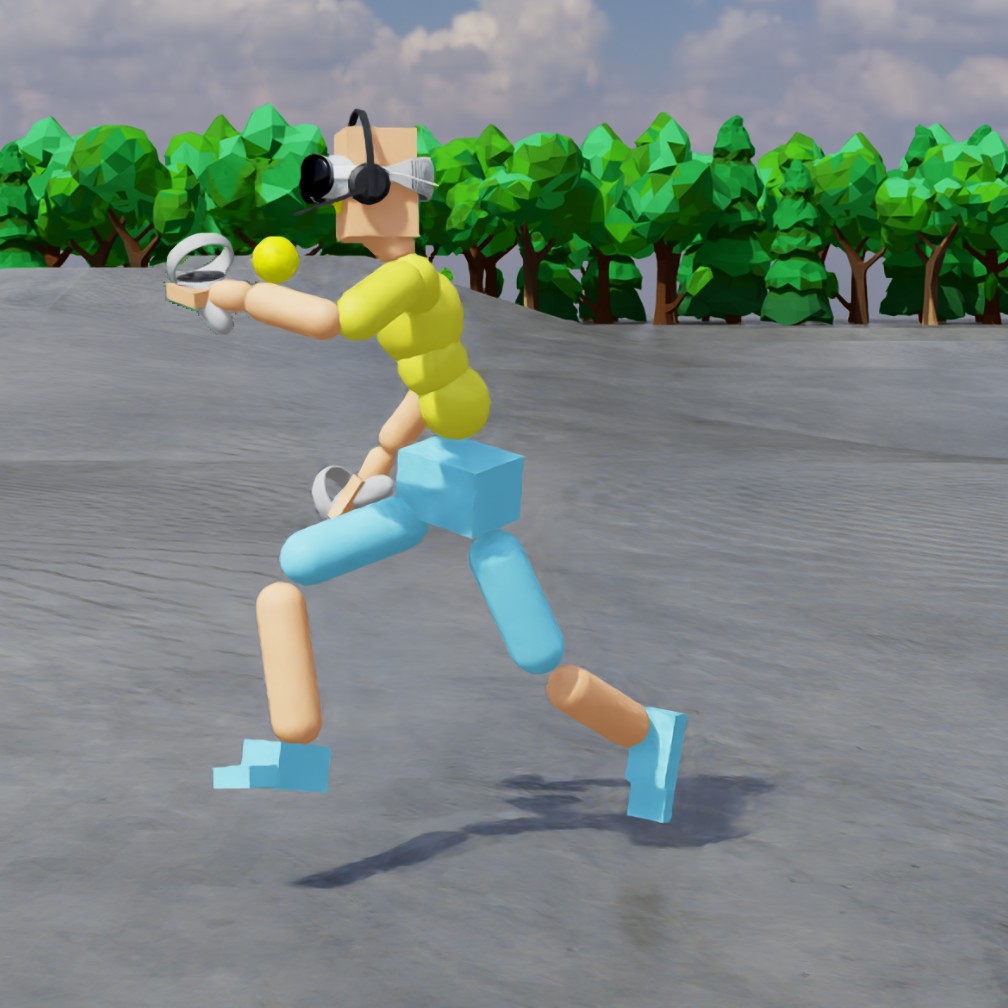}\hspace{0.5pt}%
            \includegraphics[width=0.1975\linewidth]{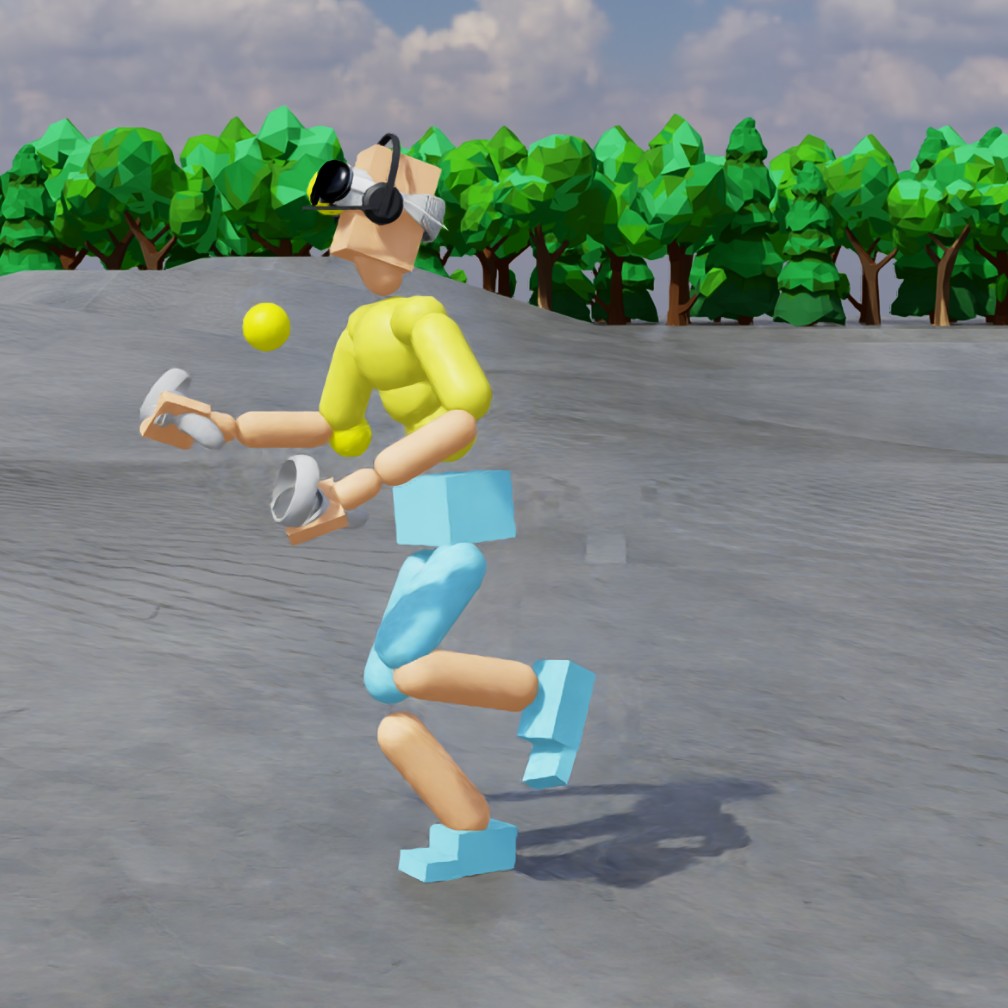}\hspace{0.5pt}%
            \includegraphics[width=0.1975\linewidth]{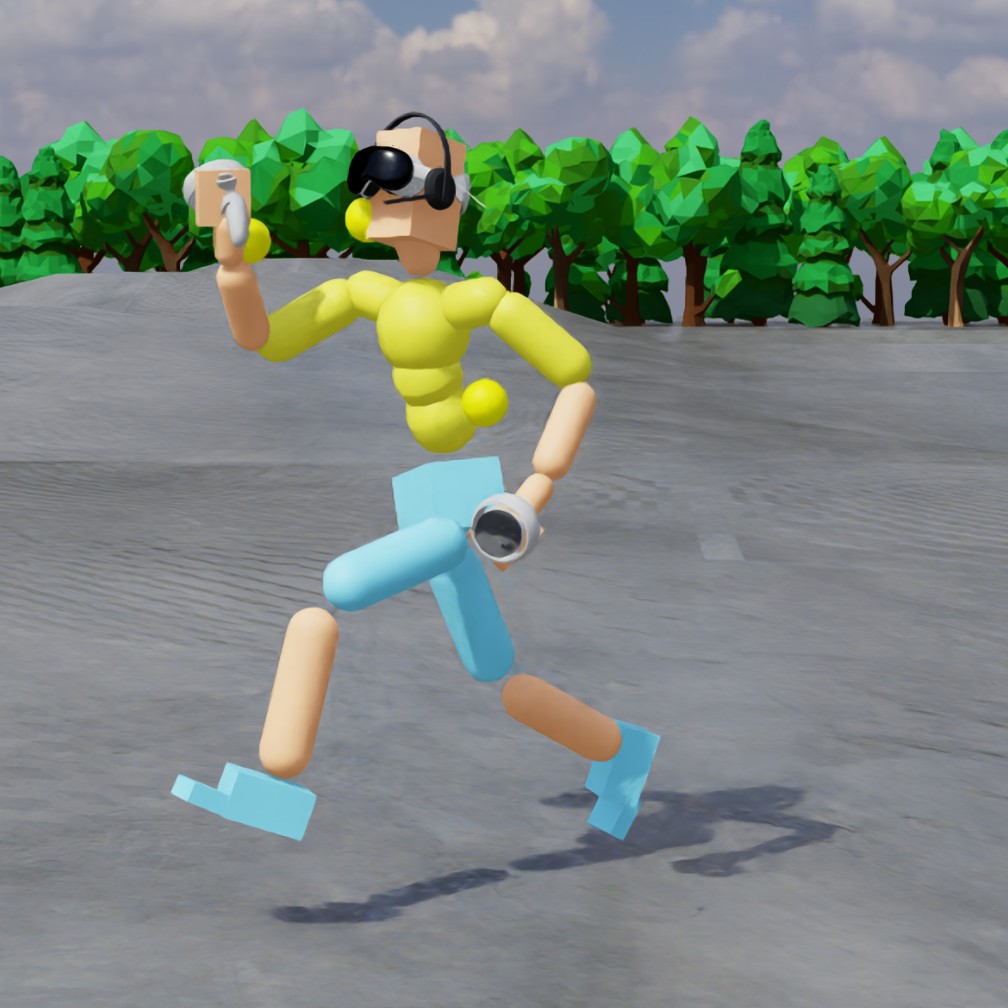}
        \end{subfigure}
        \caption{\textbf{VR Tracking:} physically plausible full-body motion completed from only head and hand constraints.}
    \end{subfigure}
    \caption{HetSkills supports diverse motion completion tasks including (a) human-scene interaction, (b) motion in-betweening, and (c) VR tracking, recovering coherent full-body motion from spatially or temporally sparse observations within a single shared controller.}
    \label{fig:vr_qualitative}
\end{figure*}

 \paragraph{Motion Tracking.}~\cref{tab:tracking_results} reports motion tracking results on the training and test splits of AMASS~\cite{mahmood2019amass}. We evaluate each method using success rate and MPJPE (Mean Per Joint Position Error, in mm). Following prior work, a rollout is considered a failure if the mean per-joint position error exceeds 0.5 m at any frame~\citep{luo2023universal,tessler2024maskedmimic}. MPJPE measures the average Euclidean distance between the predicted and reference joint positions over all joints and frames. Additional tracking rollouts are shown in~\cref{fig:trc_qualitative}.

The upper block of~\cref{tab:tracking_results} reports results for our two-part decomposition. Since $\mathscr{F}^{trc}$ is trained with a two-frame future window, we report two inference variants: HetSkills-2step uses two frames as during training, while HetSkills-1step duplicates a single observed frame to fill the two-frame input. HetSkills-1step achieves the highest success rate among all methods under the fair inference setting. Its MPJPE is higher than MaskedMimic only on the test set, which is expected: the distribution shift introduced by removing one future frame at inference, combined with the prioritization of success on challenging motions, leads to a slight MPJPE increase. The small gap between the two variants nonetheless confirms that duplicating a single frame suffices in practice.

The lower block of~\cref{tab:tracking_results} investigates the effect of part granularity. Increasing the number of parts from one to five consistently reduces the MPJPE. Specifically, when increasing from the one-part configuration (HetSkills-1part) to the five-part configuration (HetSkills-5part), the MPJPE on the test set decreases from 42.3 mm to 35.5 mm, showing a clear improvement in tracking accuracy. This reduction in MPJPE confirms that part-wise decomposition improves tracking precision. However, finer decompositions reduce the semantic interpretability of each part latent and increase computational overhead. Based on these trade-offs, we adopt the two-part configuration (arm part and body part) for all subsequent experiments, as it strikes a good balance between accuracy and efficiency.

\paragraph{Motion Completion.} The VR-driven body tracking task, where only the head and both hands are observable as goal constraints, is shown in~\cref{tab:vr}. A rollout is considered a failure if the mean tracking error of the head and both hands exceeds 0.5 m at any frame. HetSkills achieves success rates of $99.9\%$ and $97.8\%$ on the training and test splits, respectively, demonstrating that the unified latent space has the capacity to support this sparse-conditioning skill along with other heterogeneous skills. HetSkills significantly outperforms other large-scale motion priors in terms of success rates, suggesting that the structured latent space provides a more robust foundation for generalizing to new conditioning forms. The higher MPJPE on the test set is expected, as the lower body is underdetermined with only three upper-body endpoints provided as constraints, allowing many plausible configurations and positional errors that do not necessarily reflect the motion quality.

For motion in-betweening, HetSkills achieves success rates of $99.9\%$ on the training split and $100\%$ on the test split of AMASS, with both tasks using a 0.5 m full-body MPJPE failure threshold. However, success rate alone is not a comprehensive measure, as motion in-betweening allows for a wide range of plausible intermediate trajectories. A rollout that deviates from the reference is not necessarily incorrect. For human-scene interaction, HetSkills achieves a success rate of $96.1\%$ on the SAMP dataset, demonstrating the ability of the unified latent space to generalize to out-of-distribution interaction patterns without modifying the shared decoder. This result shows that each skill can be trained independently on its respective motion distribution while remaining fully compatible with the common control space. These results indicate that the unified latent space can accommodate heterogeneous skills from various conditioning modalities and motion distributions within a single shared framework. As quantitative metrics alone do not fully capture motion naturalness and plausibility, we encourage viewers to refer to the supplementary video for a qualitative demonstration of the capabilities of the system.

\begin{table}[t]
\centering
\caption{VR-driven body tracking from sparse end-effector observations. Only the head and both hands are provided as goal constraints, while the remaining full-body motion must be inferred by the controller. Success rate measures whether the visible end-effectors remain within the 0.5 m tracking-error threshold, and MPJPE reports full-body reconstruction error in millimeters on the AMASS train and test splits.}
\label{tab:vr}
\setlength{\tabcolsep}{6pt}
\renewcommand{\arraystretch}{1.1}
\begin{tabular}{l|cc|cc}
\hline\hline
& \multicolumn{2}{c|}{Train} & \multicolumn{2}{c}{Test} \\
\cline{2-5}
Method & Success & MPJPE & Success & MPJPE \\
\hline
HetSkills    & 99.9\% & 39.9  & 97.8\% & 82.3  \\
PULSE        & 99.5\% & 57.8  & 93.4\% & 88.6  \\
MaskedMimic  & 98.6\% & 50.0  & 98.1\% & 58.1  \\
ASE          & 79.8\% & 103.0 & 37.6\% & 120.5 \\
\hline\hline
\end{tabular}
\end{table}

\subsection{Text-to-Motion.}
\label{result:Text2Motion}

\begin{figure*}[t]
    \centering
    \begin{subfigure}[t]{0.33333\textwidth}
        \centering
        \includegraphics[width=0.2425\linewidth]{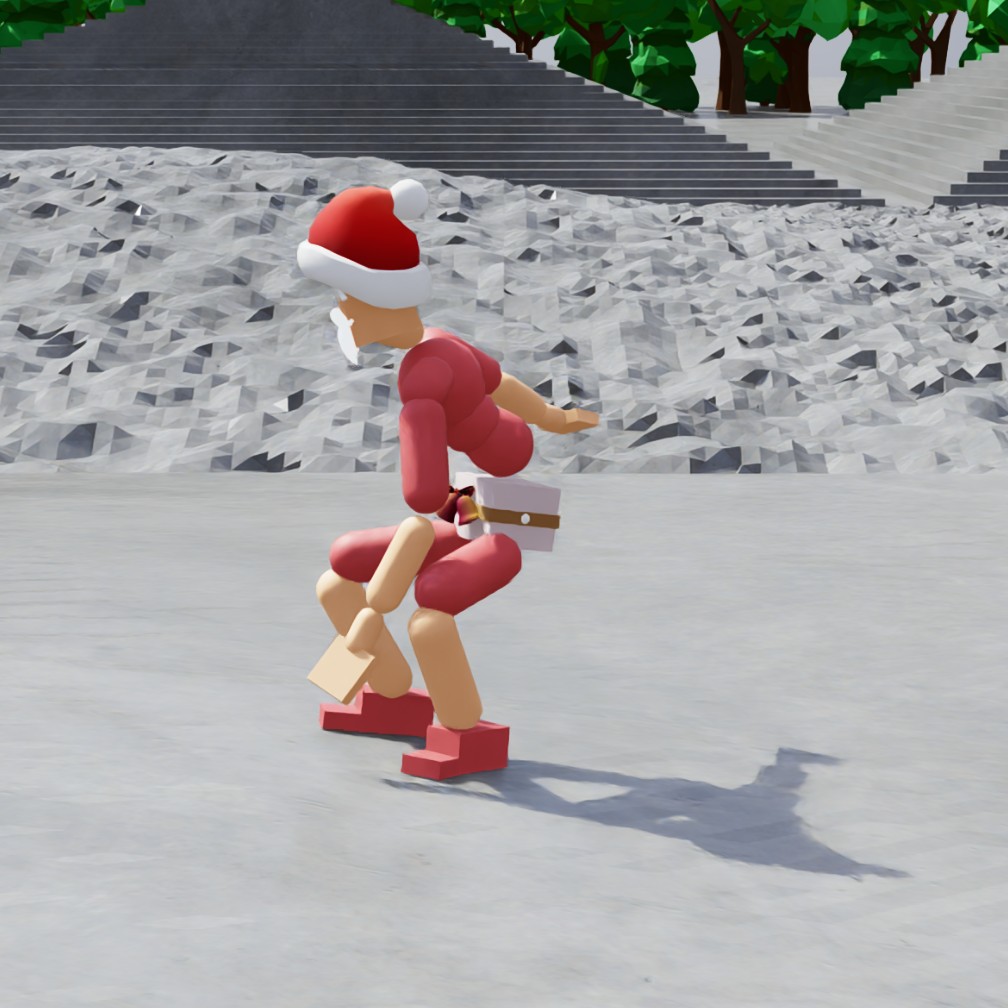}\hspace{0.5pt}%
        \includegraphics[width=0.2425\linewidth]{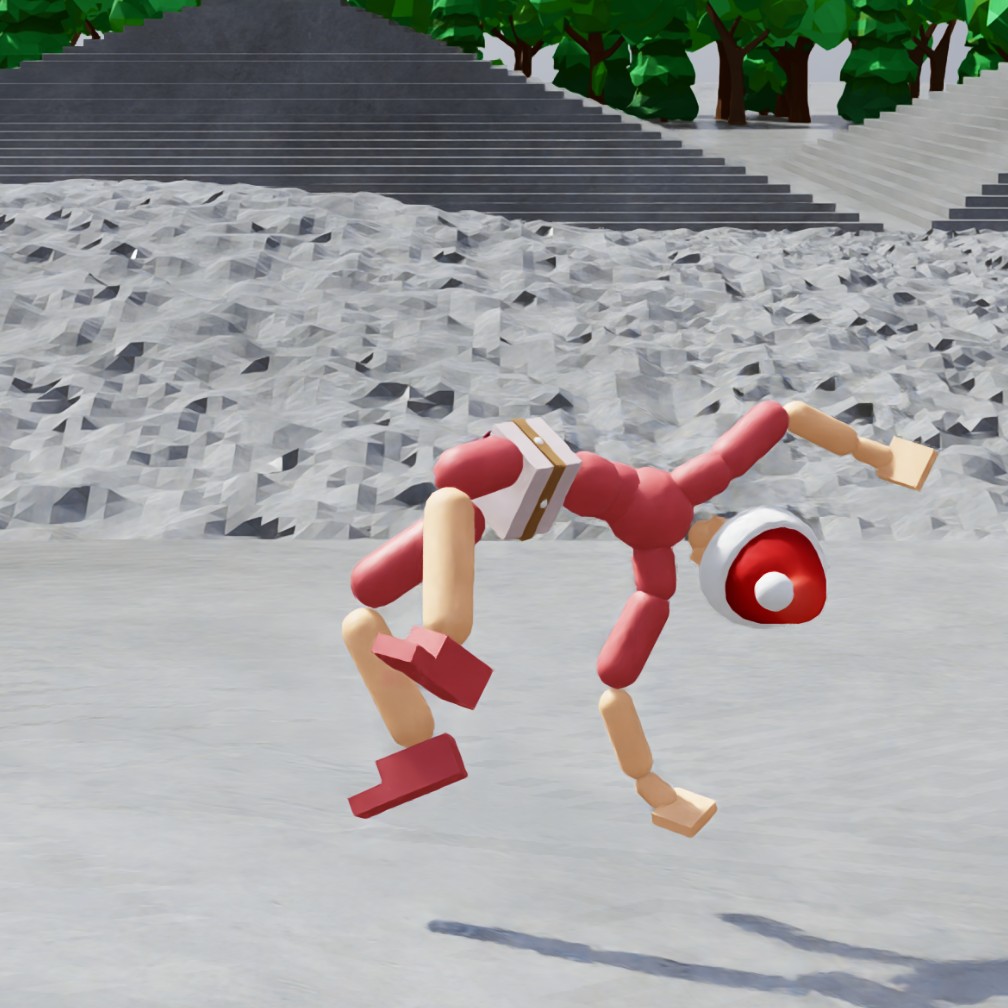}\hspace{0.5pt}%
        \includegraphics[width=0.2425\linewidth]{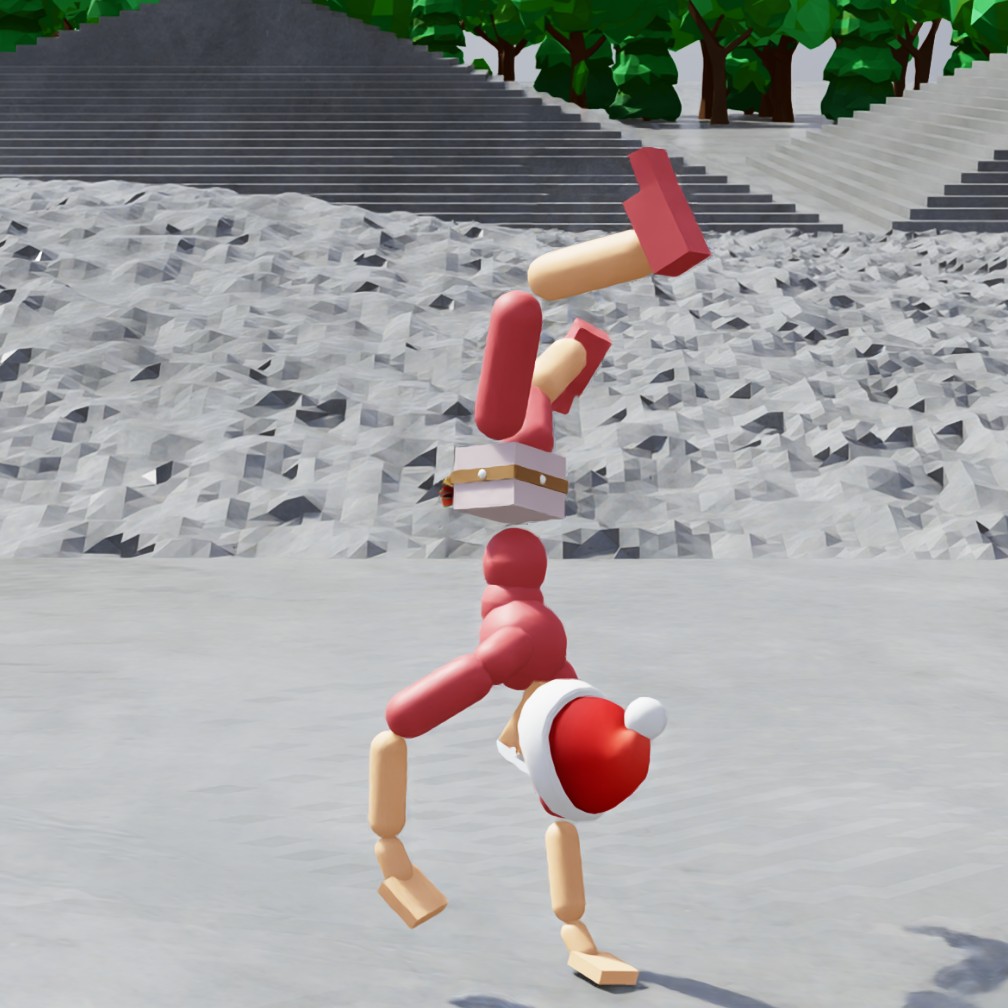}\hspace{0.5pt}%
        \includegraphics[width=0.2425\linewidth]{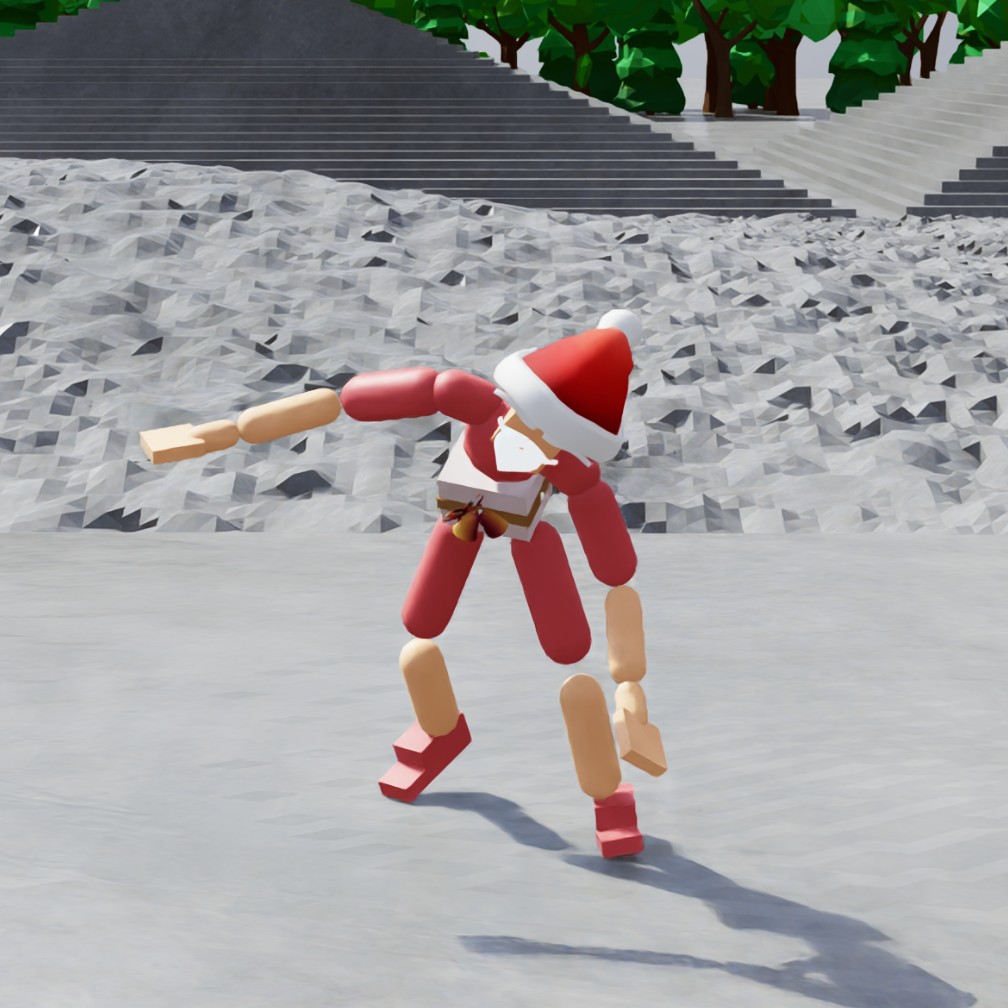}
        \caption{A person does a backflip.}
        \label{fig:t2m_backflip}
    \end{subfigure}\hfill
    \begin{subfigure}[t]{0.33333\textwidth}
        \centering
        \includegraphics[width=0.2425\linewidth]{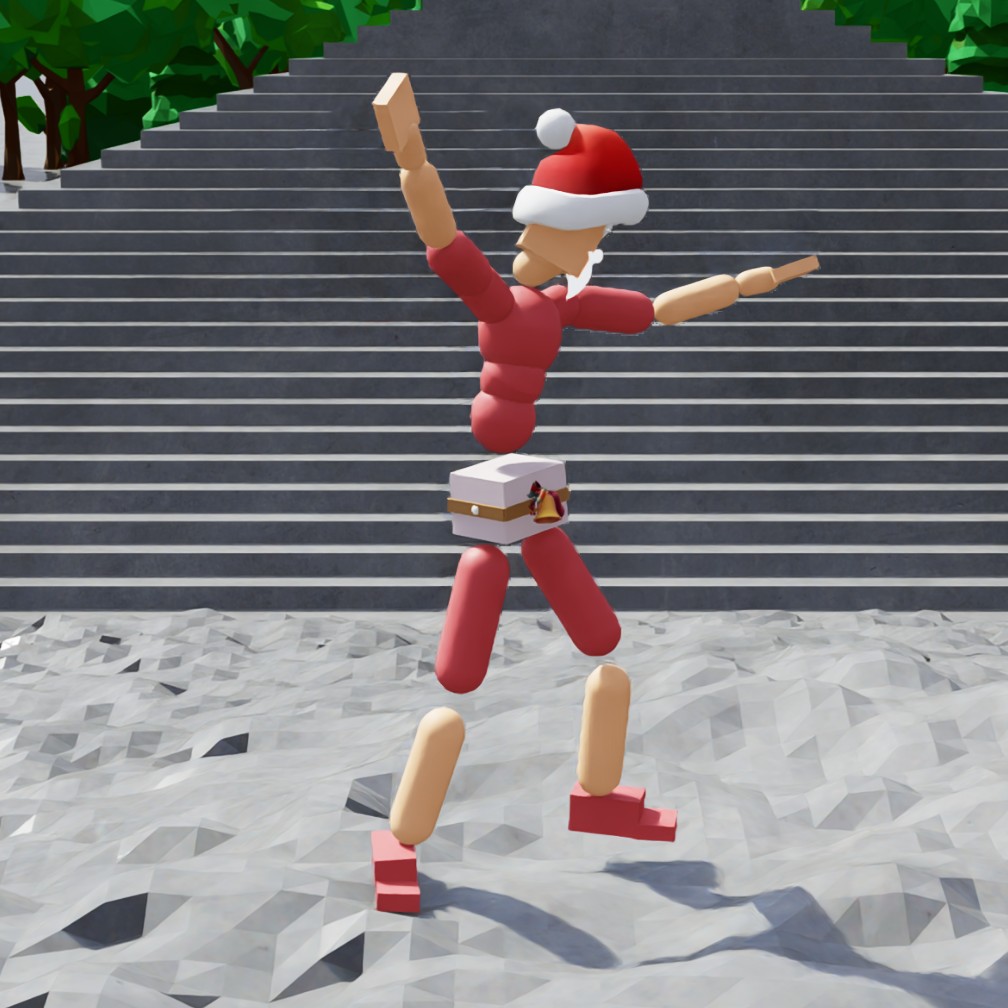}\hspace{0.5pt}%
        \includegraphics[width=0.2425\linewidth]{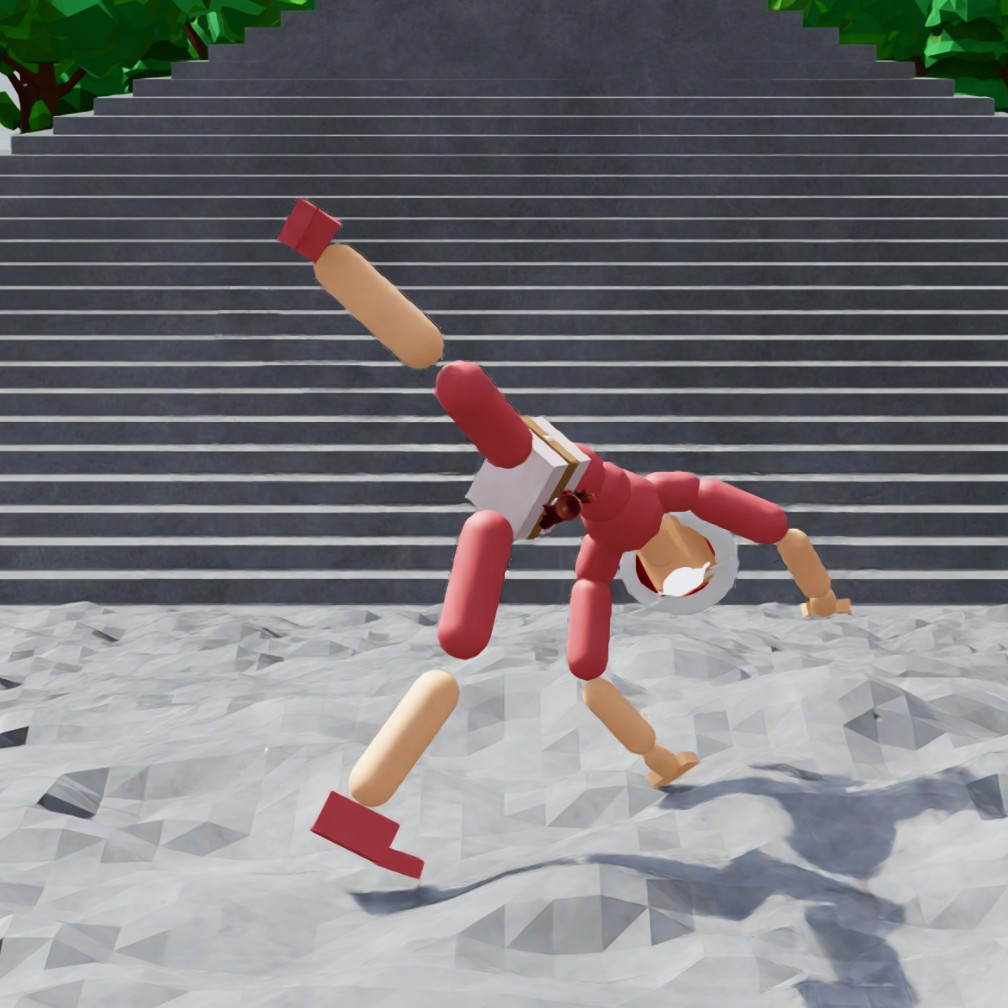}\hspace{0.5pt}%
        \includegraphics[width=0.2425\linewidth]{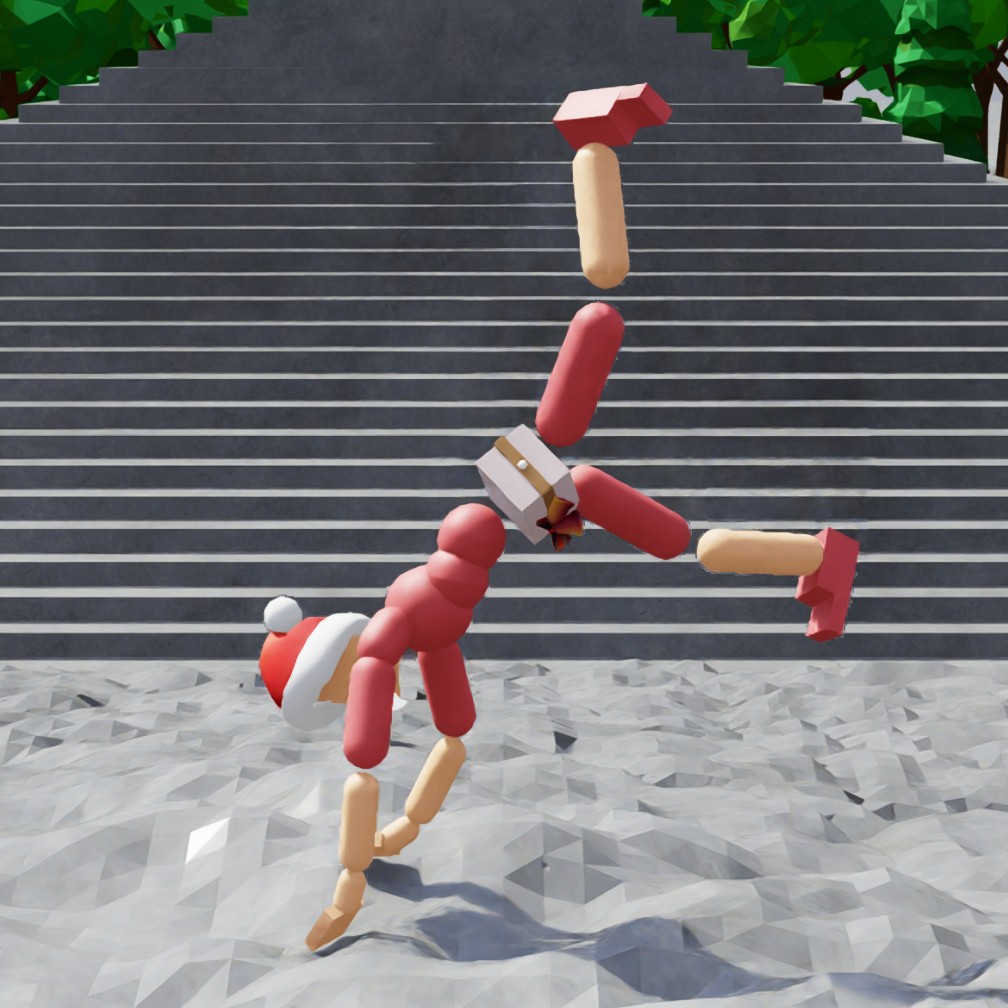}\hspace{0.5pt}%
        \includegraphics[width=0.2425\linewidth]{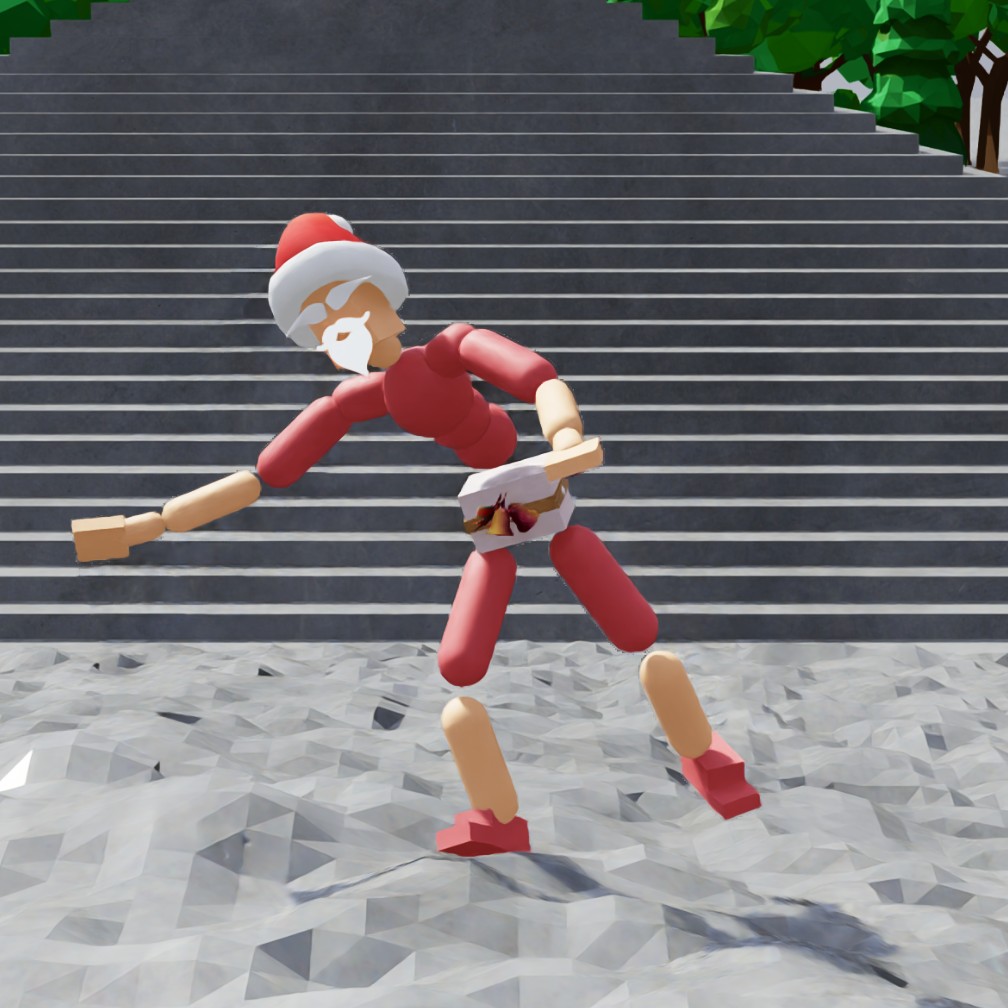}
        \caption{A person does a cartwheel.}
        \label{fig:t2m_cartwheel}
    \end{subfigure}\hfill
    \begin{subfigure}[t]{0.33333\textwidth}
        \centering
        \includegraphics[width=0.2425\linewidth]{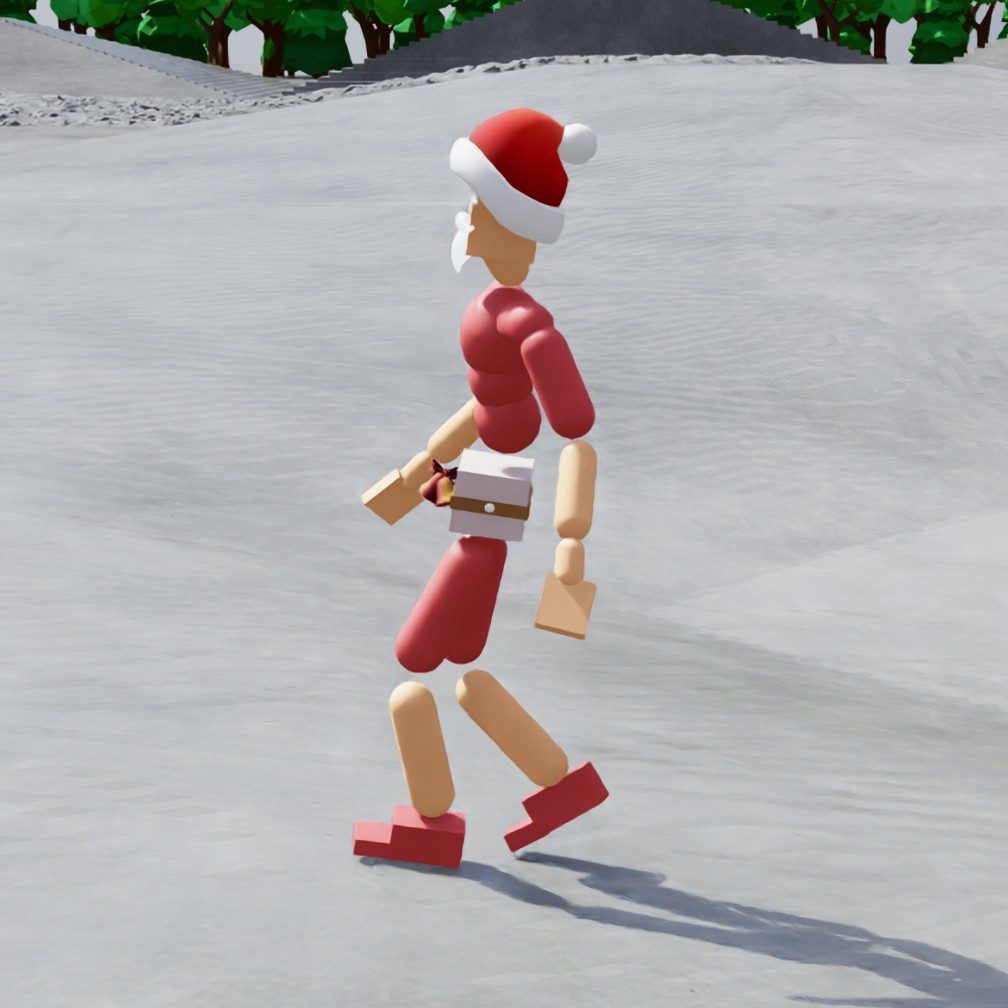}\hspace{0.5pt}%
        \includegraphics[width=0.2425\linewidth]{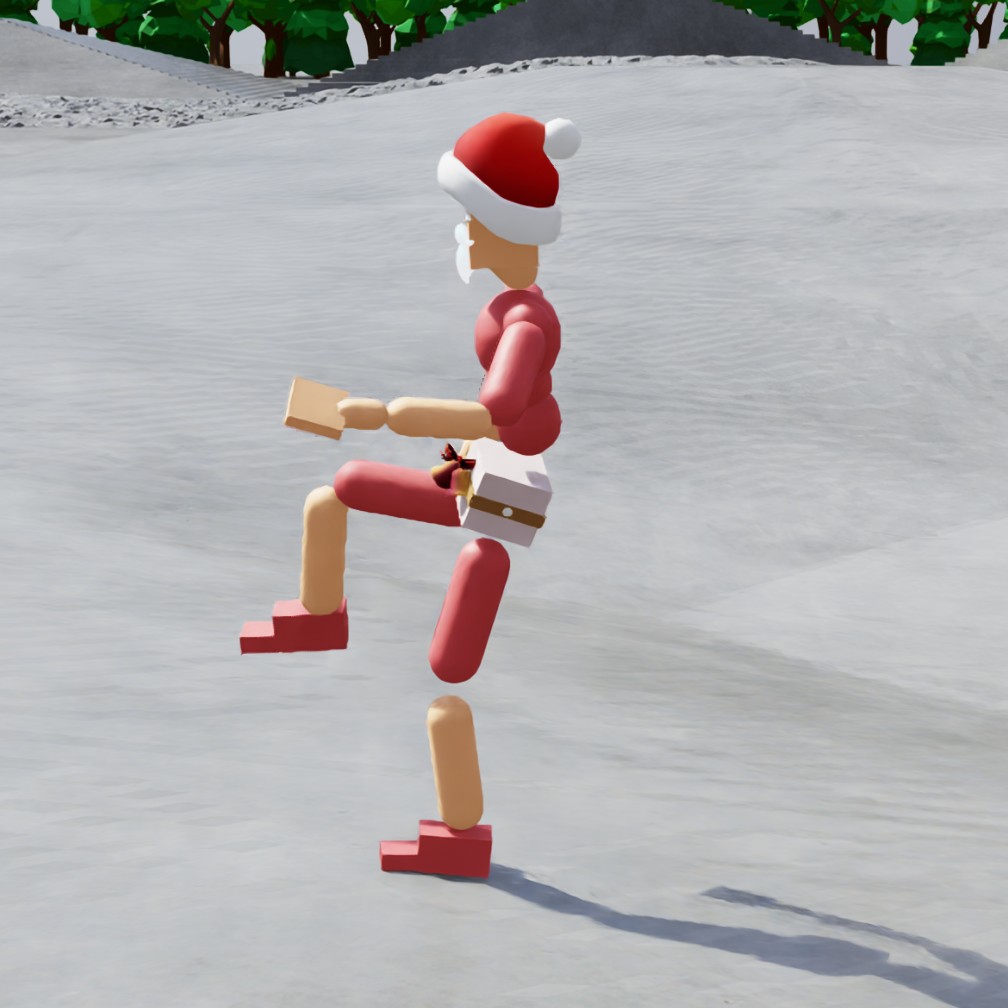}\hspace{0.5pt}%
        \includegraphics[width=0.2425\linewidth]{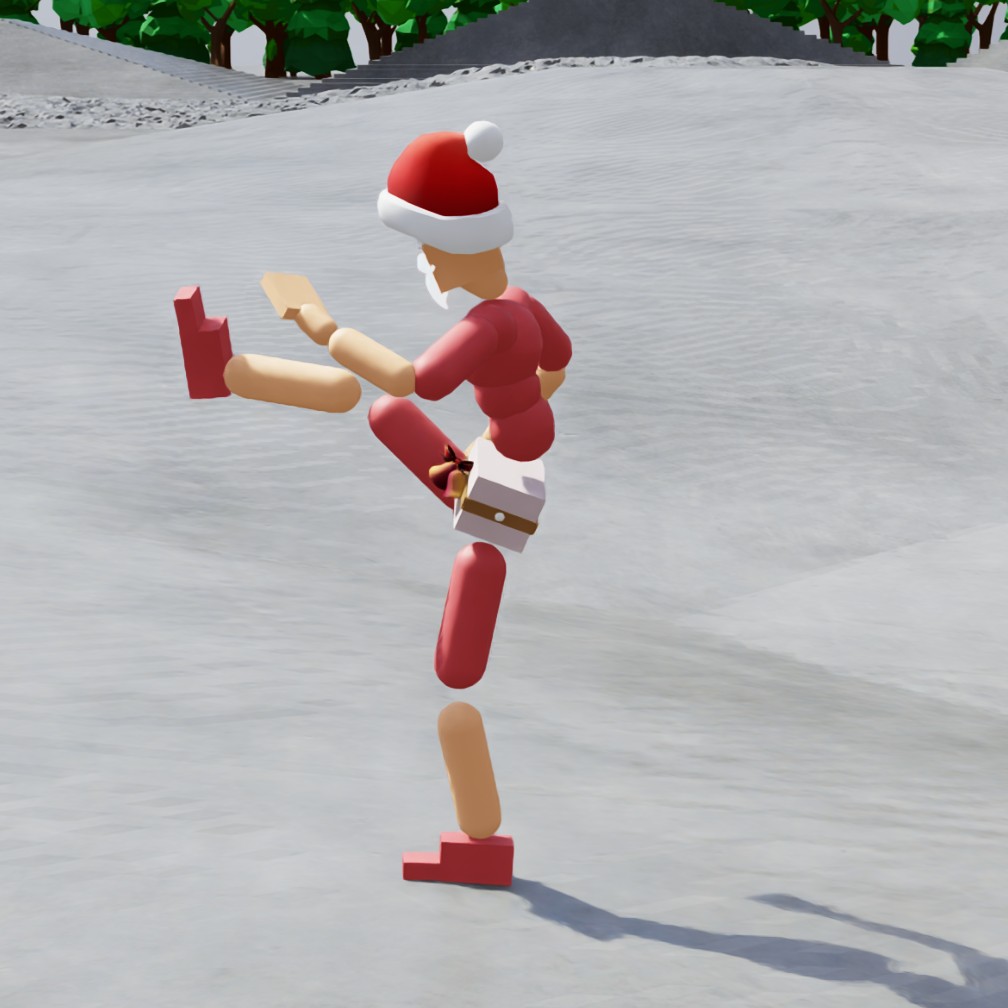}\hspace{0.5pt}%
        \includegraphics[width=0.2425\linewidth]{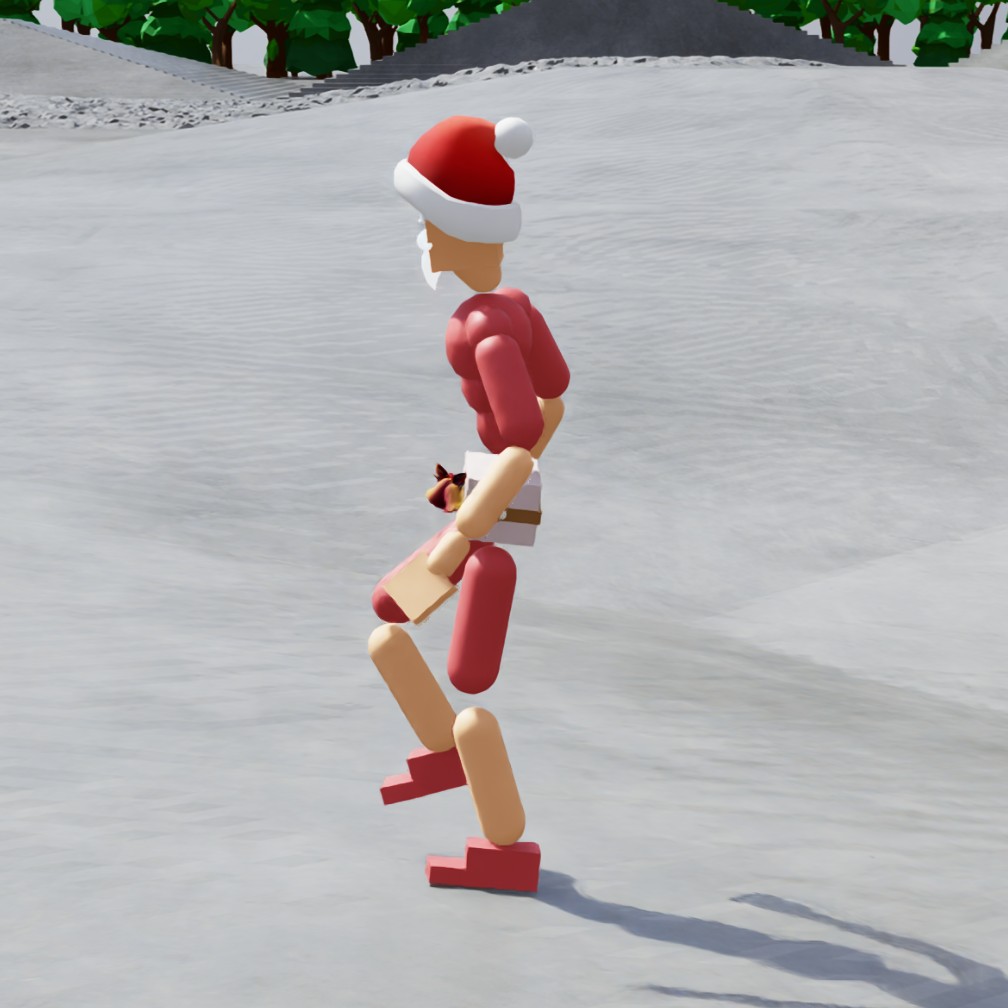}
        \caption{A person kicks forward.}
        \label{fig:t2m_kick}
    \end{subfigure}\\[0.5em]
    \begin{subfigure}[t]{0.33333\textwidth}
        \centering
        \includegraphics[width=0.2425\linewidth]{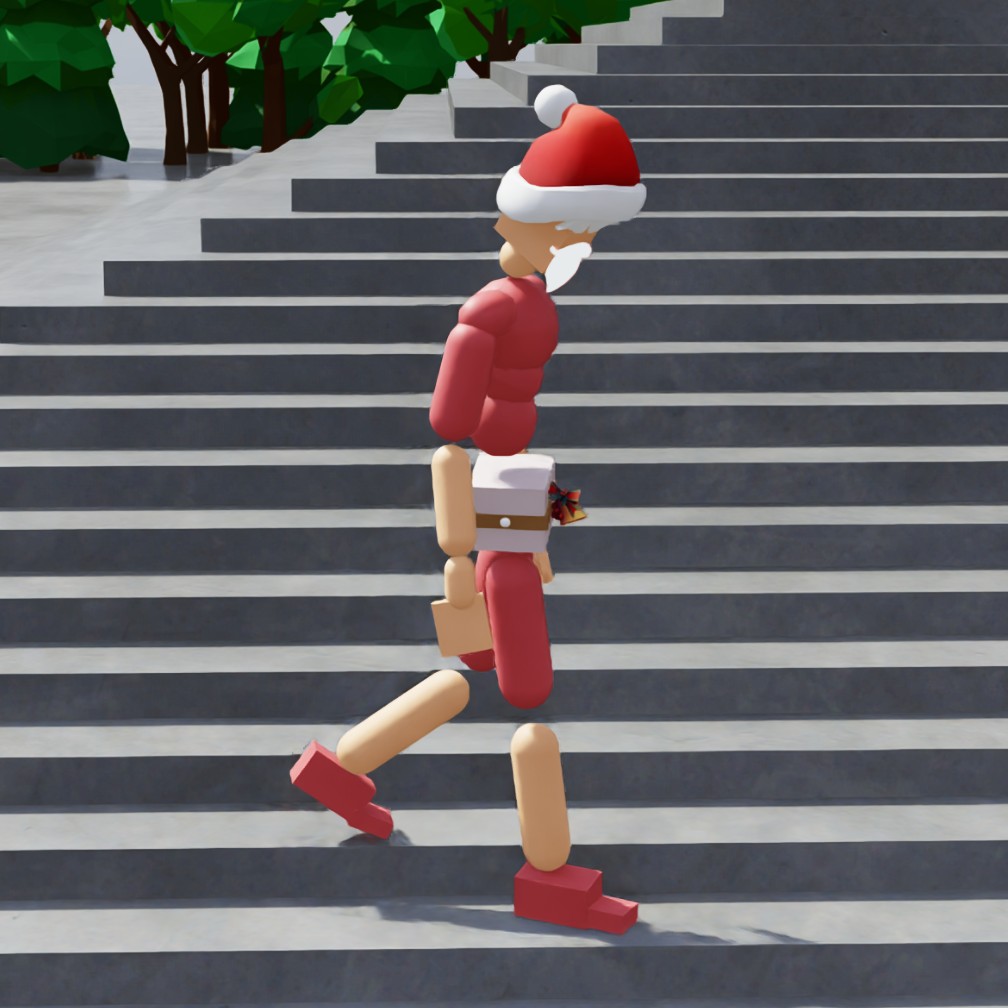}\hspace{0.5pt}%
        \includegraphics[width=0.2425\linewidth]{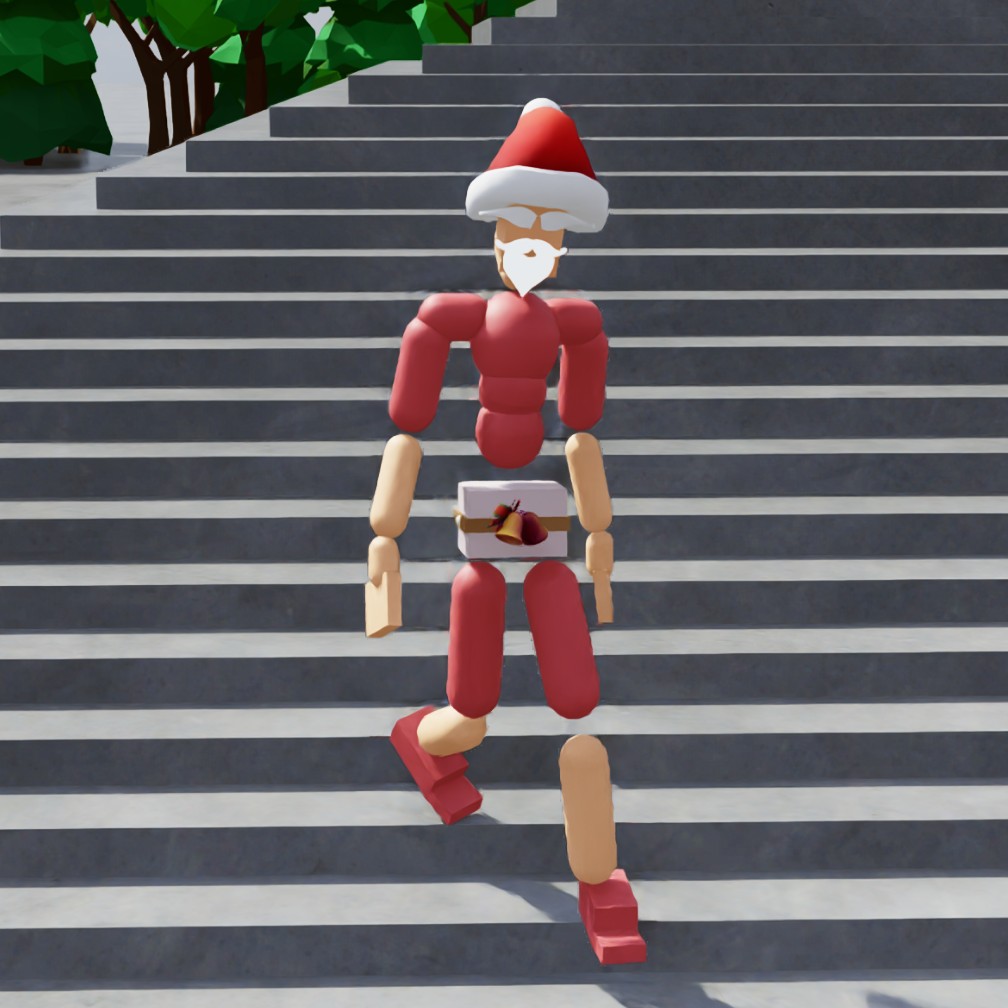}\hspace{0.5pt}%
        \includegraphics[width=0.2425\linewidth]{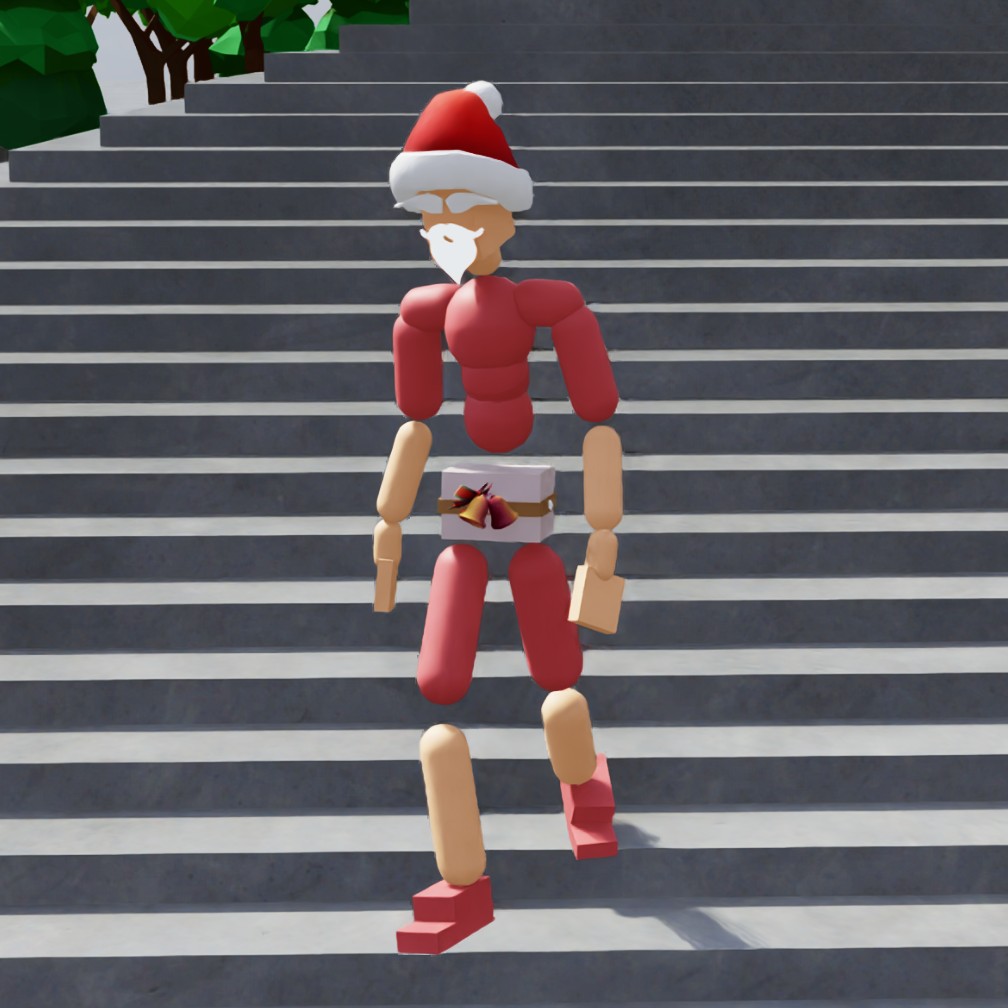}\hspace{0.5pt}%
        \includegraphics[width=0.2425\linewidth]{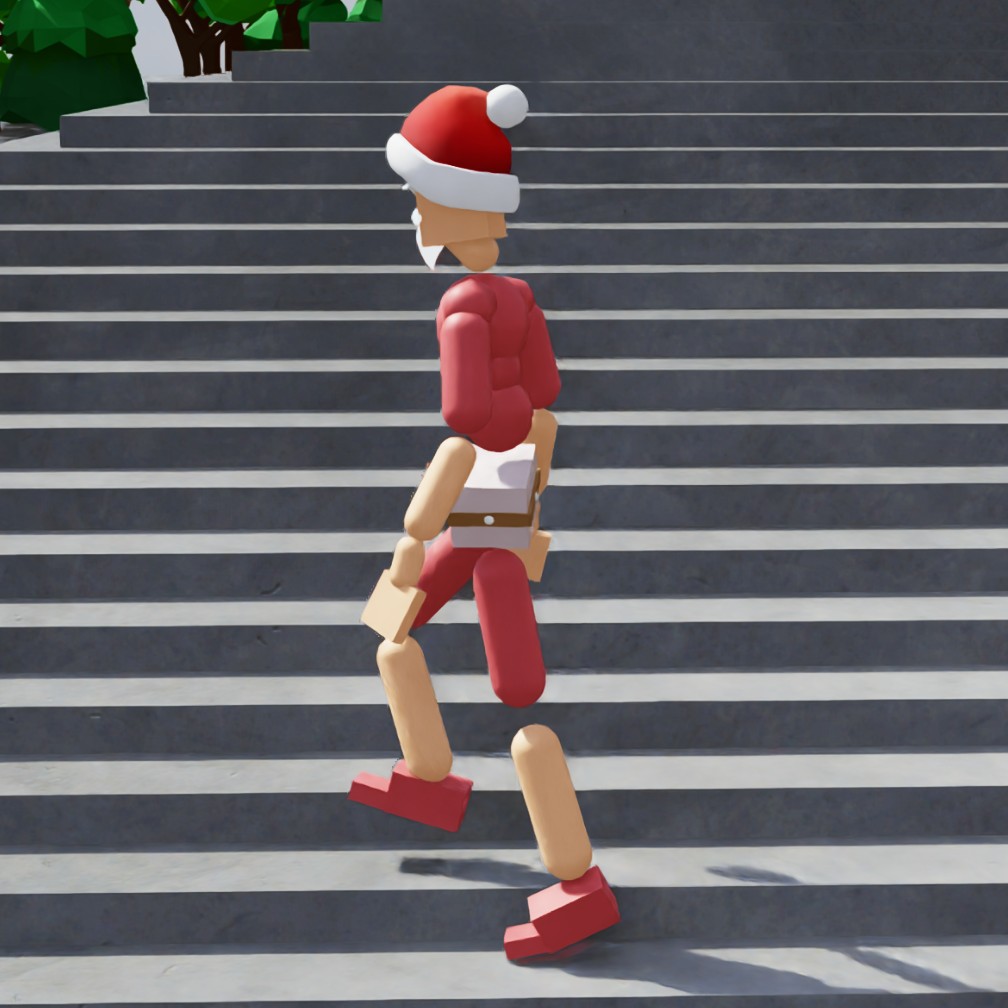}
        \caption{A person walks in a circle.}
        \label{fig:t2m_circle}
    \end{subfigure}\hfill
    \begin{subfigure}[t]{0.33333\textwidth}
        \centering
        \includegraphics[width=0.2425\linewidth]{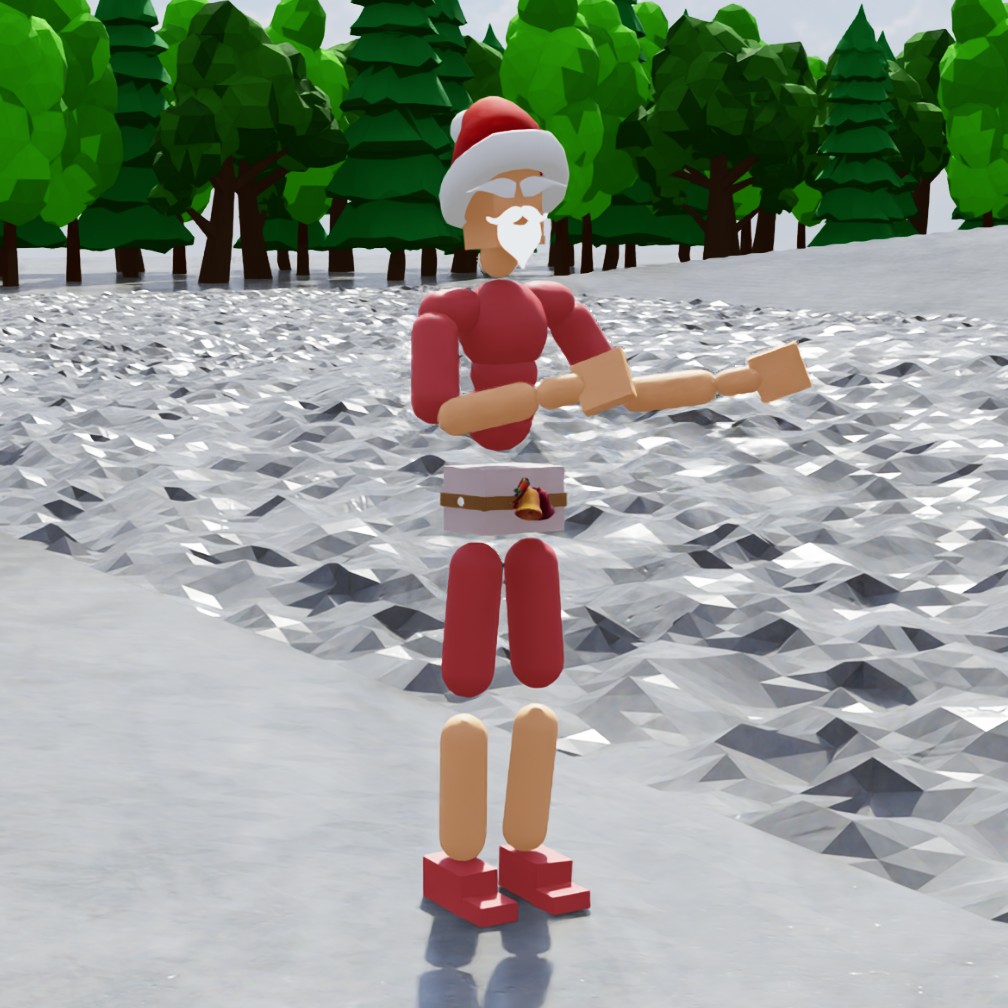}\hspace{0.5pt}%
        \includegraphics[width=0.2425\linewidth]{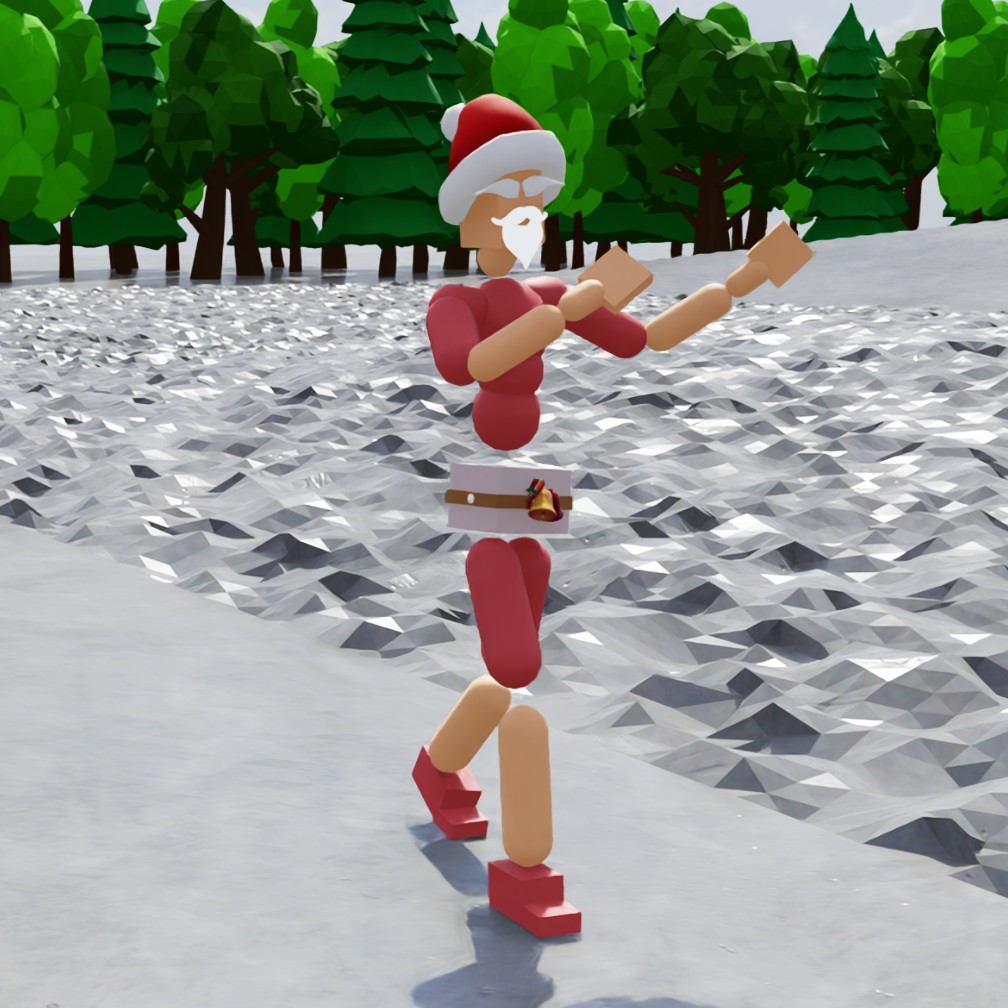}\hspace{0.5pt}%
        \includegraphics[width=0.2425\linewidth]{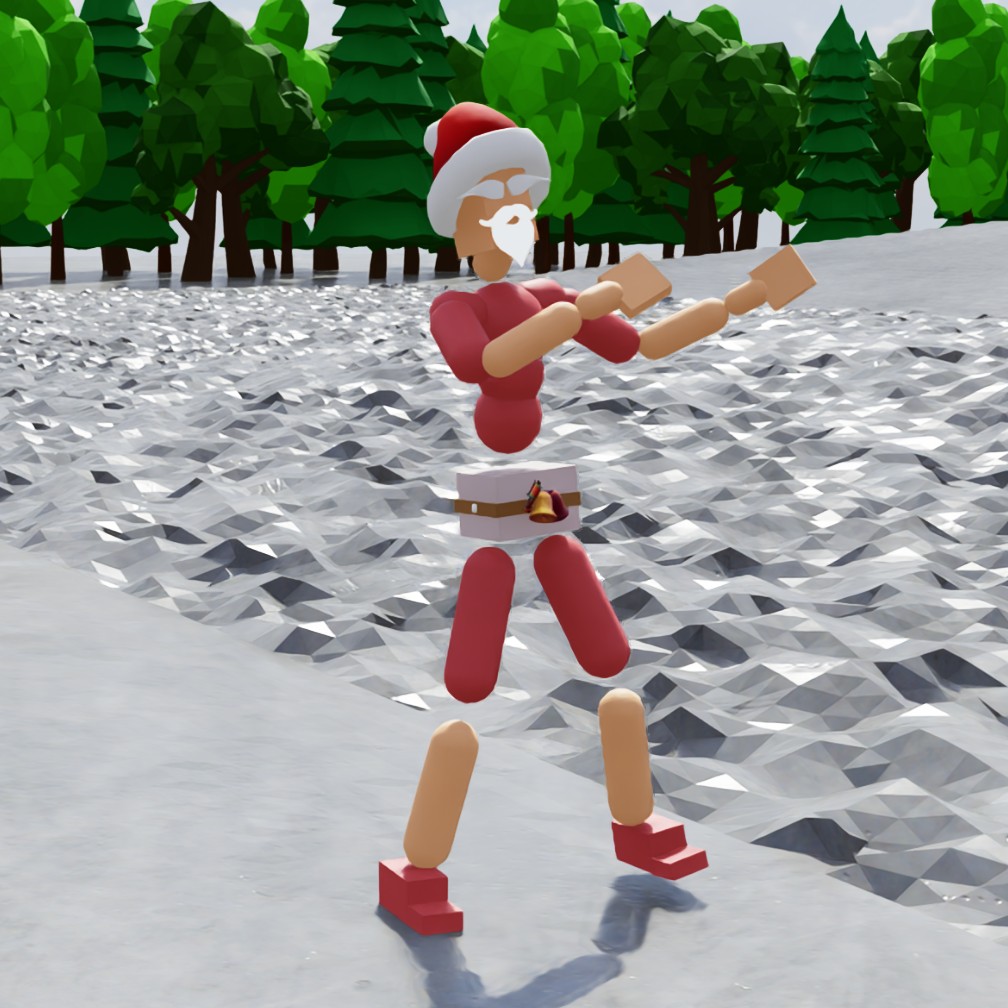}\hspace{0.5pt}%
        \includegraphics[width=0.2425\linewidth]{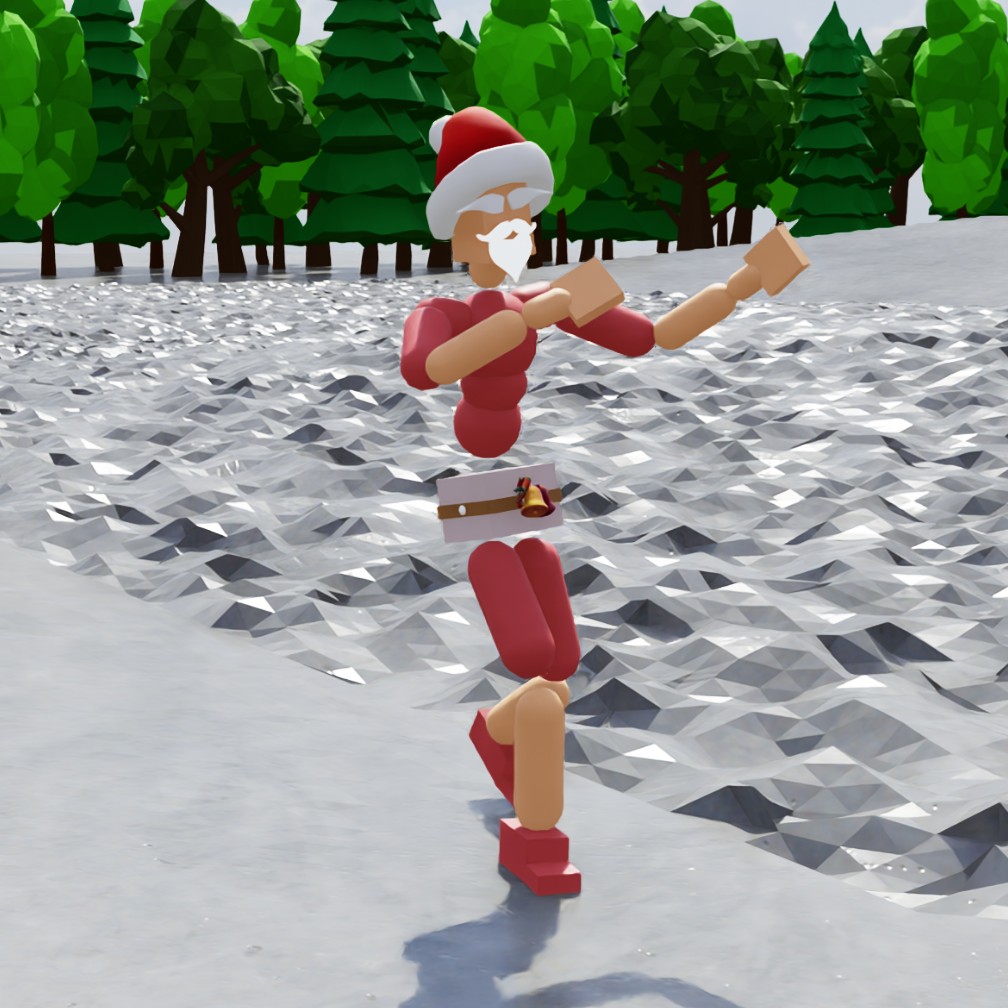}
        \caption{A person dances the waltz.}
        \label{fig:t2m_waltz}
    \end{subfigure}\hfill
    \begin{subfigure}[t]{0.33\textwidth}
        \centering
        \includegraphics[width=0.2425\linewidth]{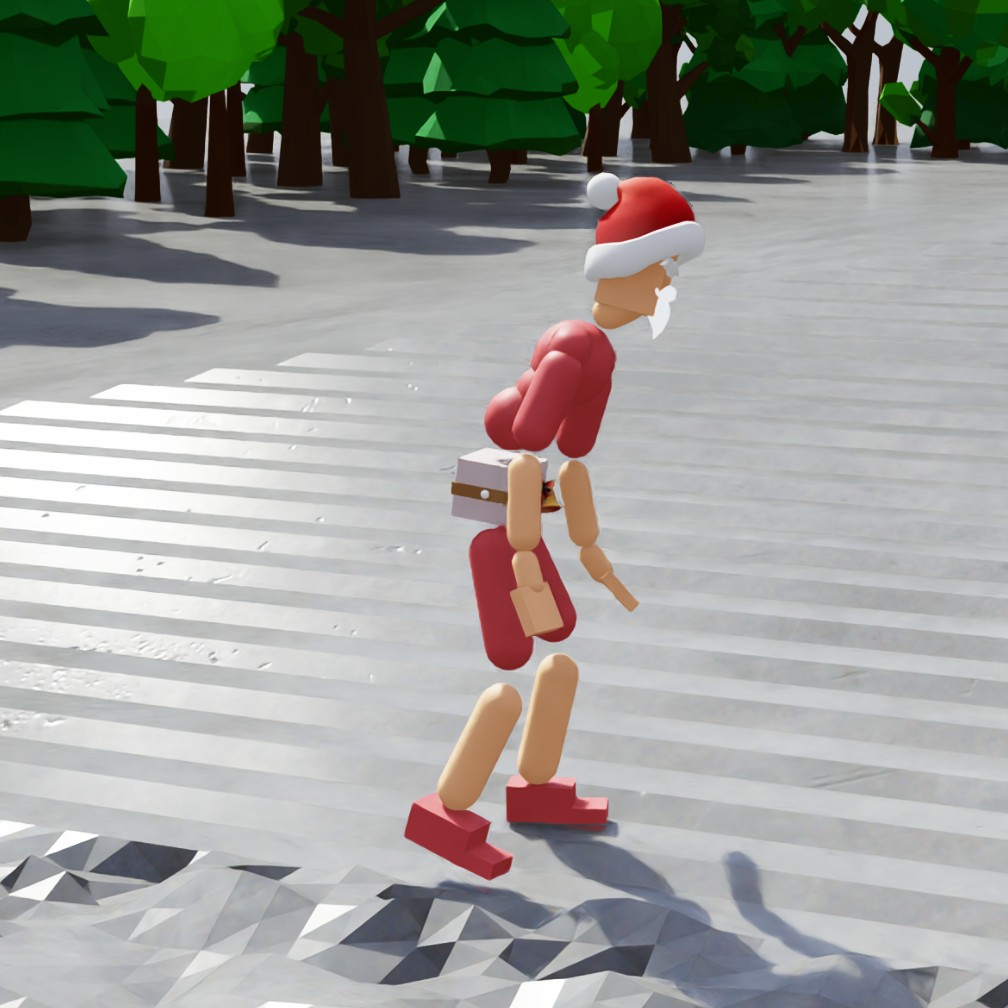}\hspace{0.5pt}%
        \includegraphics[width=0.2425\linewidth]{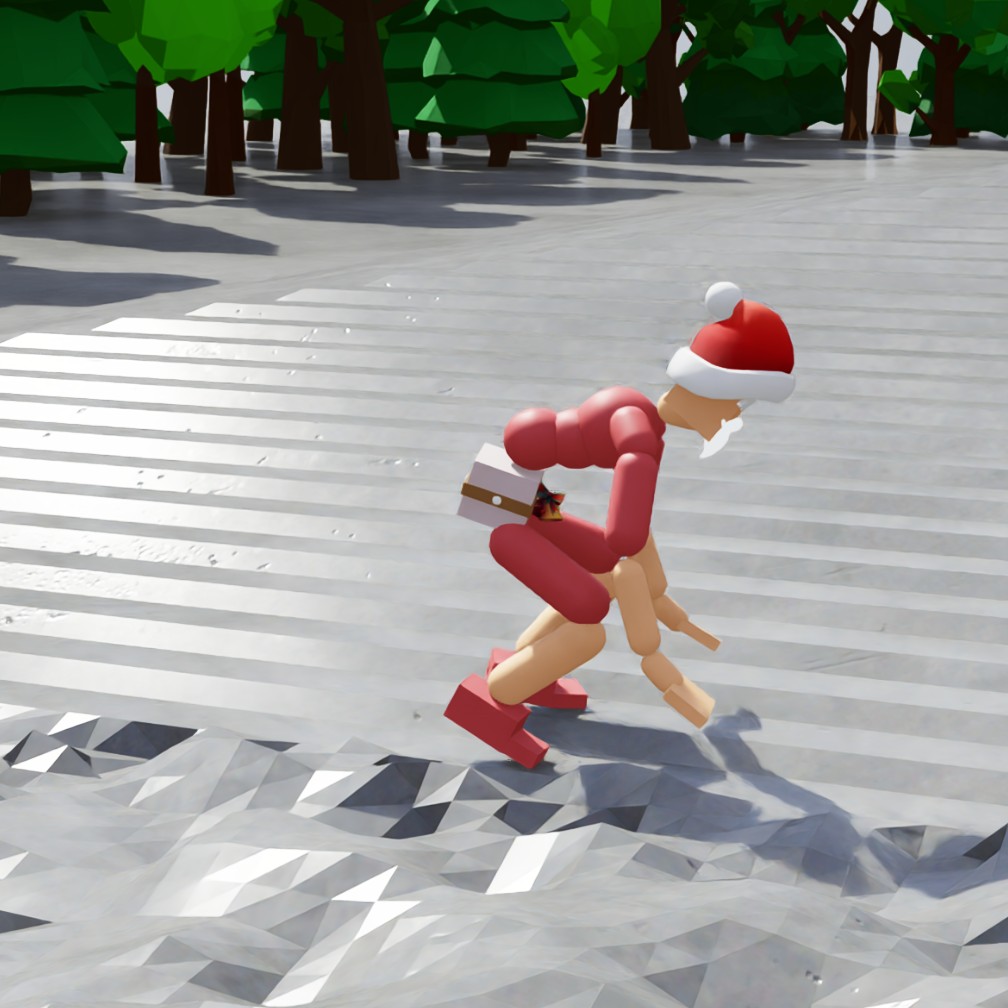}\hspace{0.5pt}%
        \includegraphics[width=0.2425\linewidth]{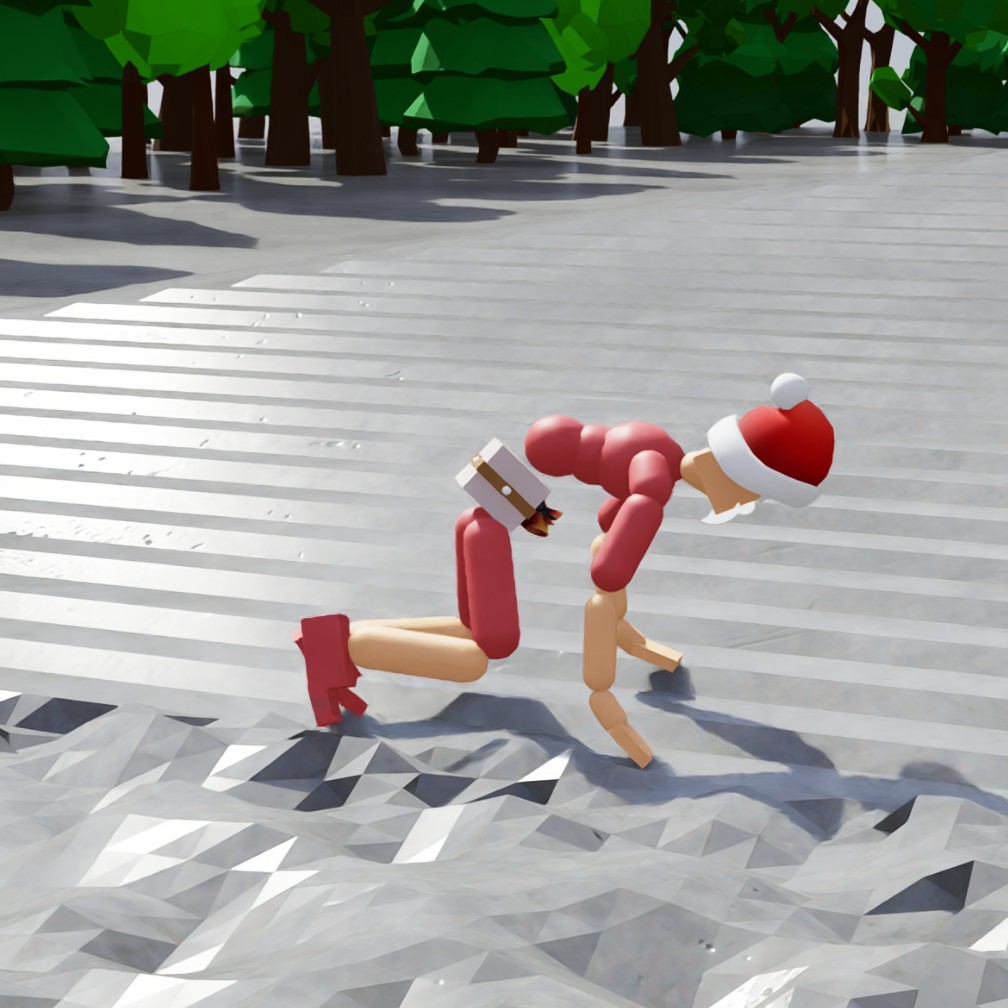}\hspace{0.5pt}%
        \includegraphics[width=0.2425\linewidth]{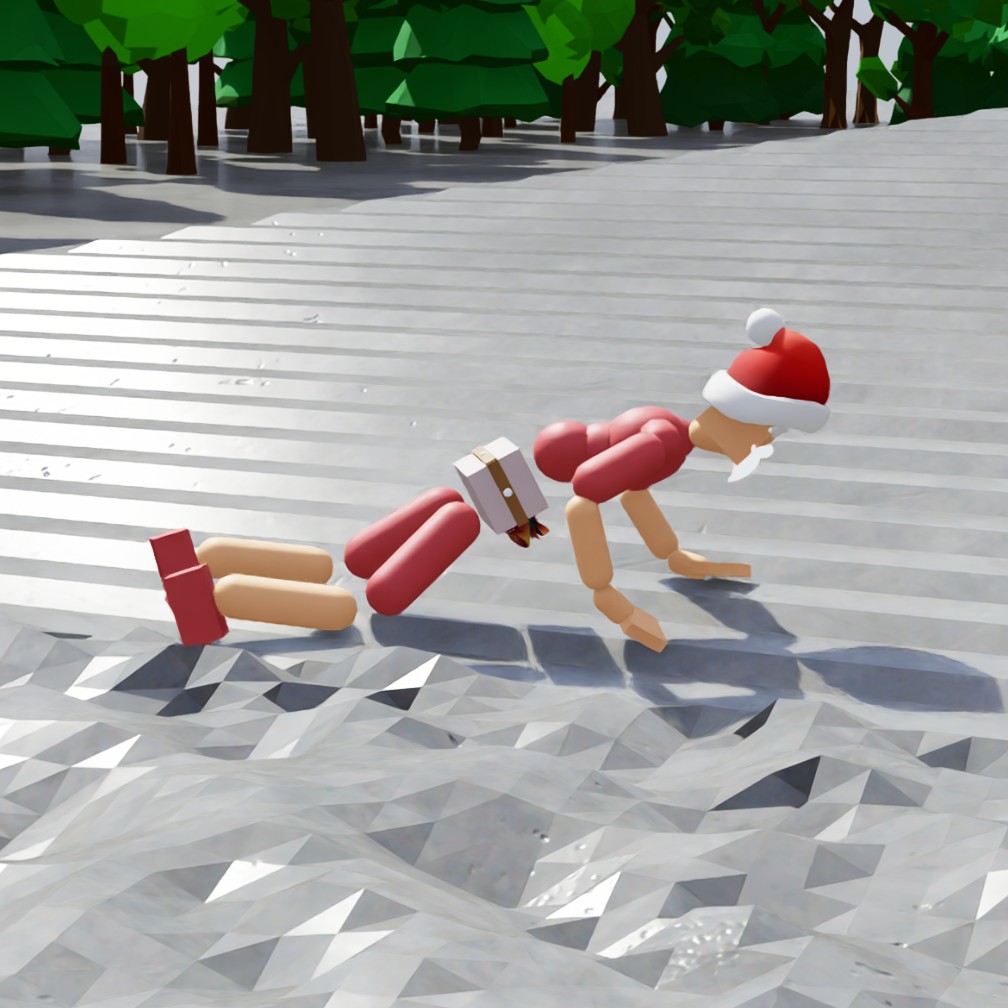}
        \caption{A person crouches and then lies down.}
        \label{fig:t2m_lie}
    \end{subfigure}\\[0.5em]
    \begin{subfigure}[t]{0.33333\textwidth}
        \centering
        \includegraphics[width=0.2425\linewidth]{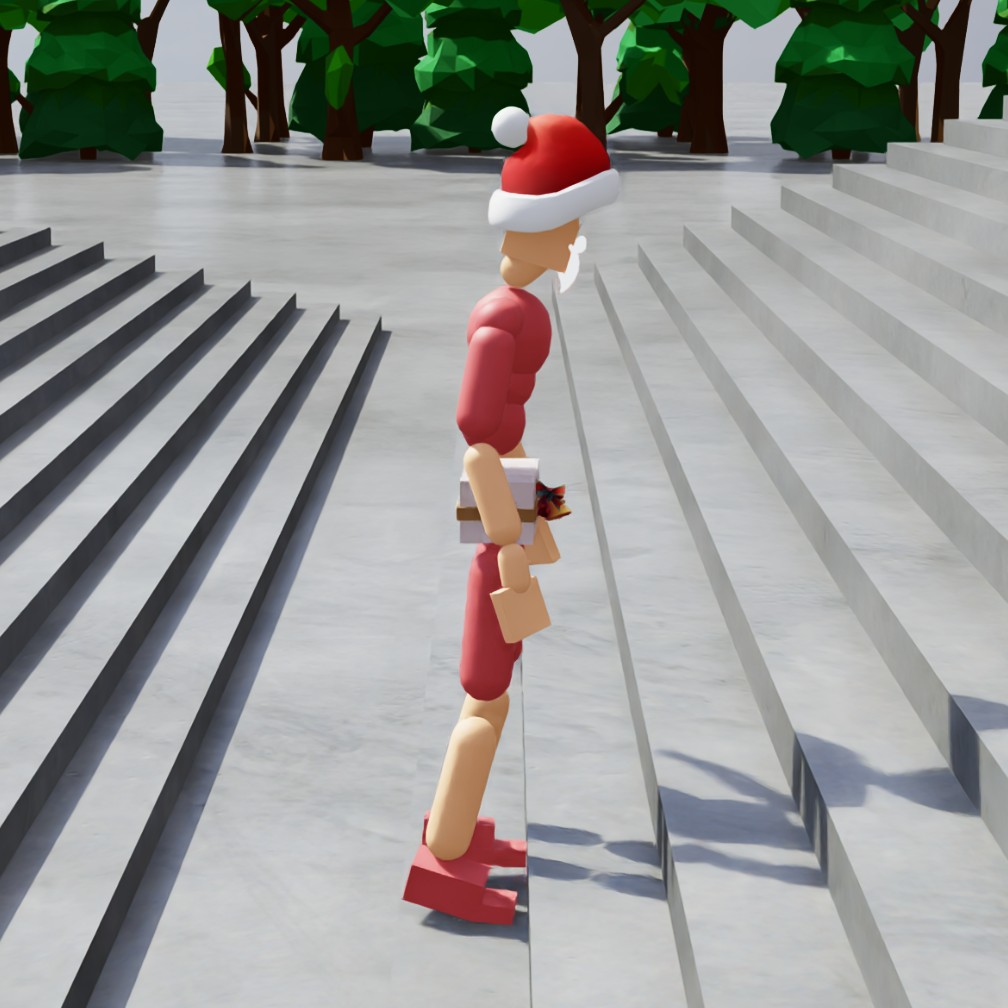}\hspace{0.5pt}%
        \includegraphics[width=0.2425\linewidth]{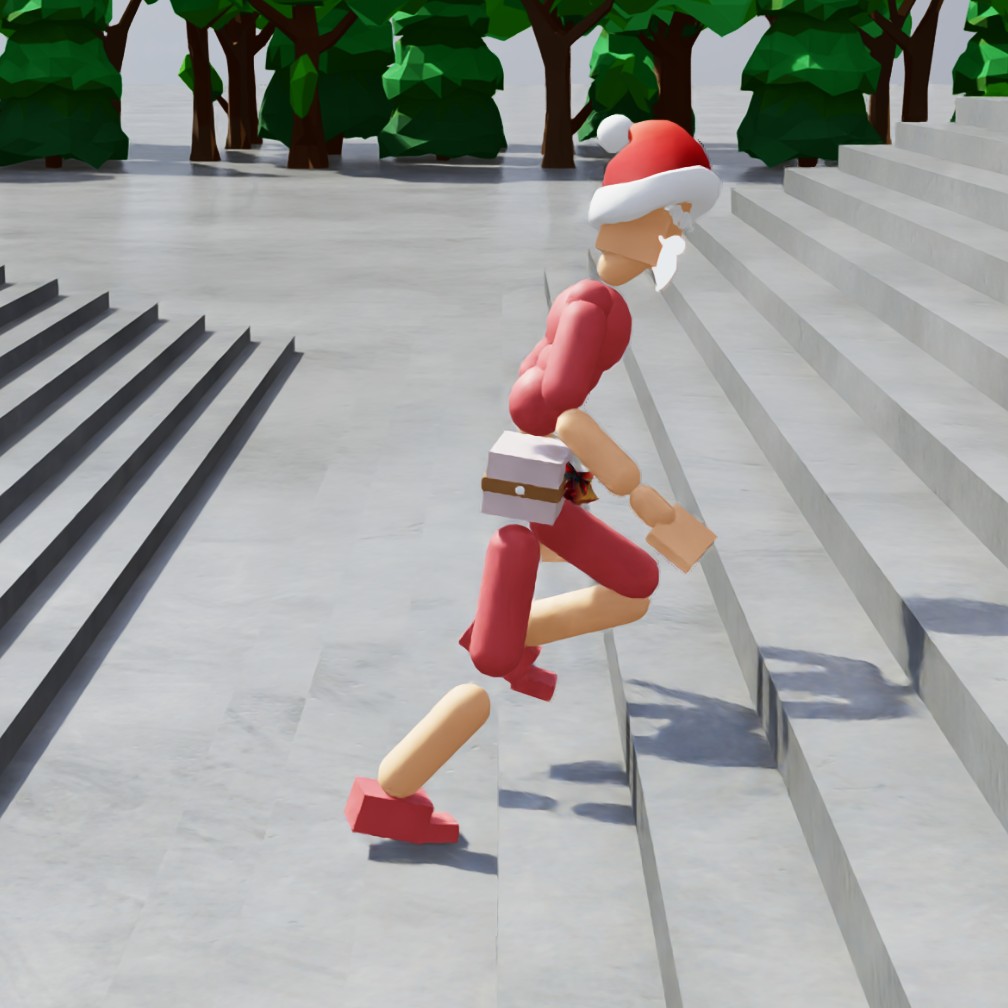}\hspace{0.5pt}%
        \includegraphics[width=0.2425\linewidth]{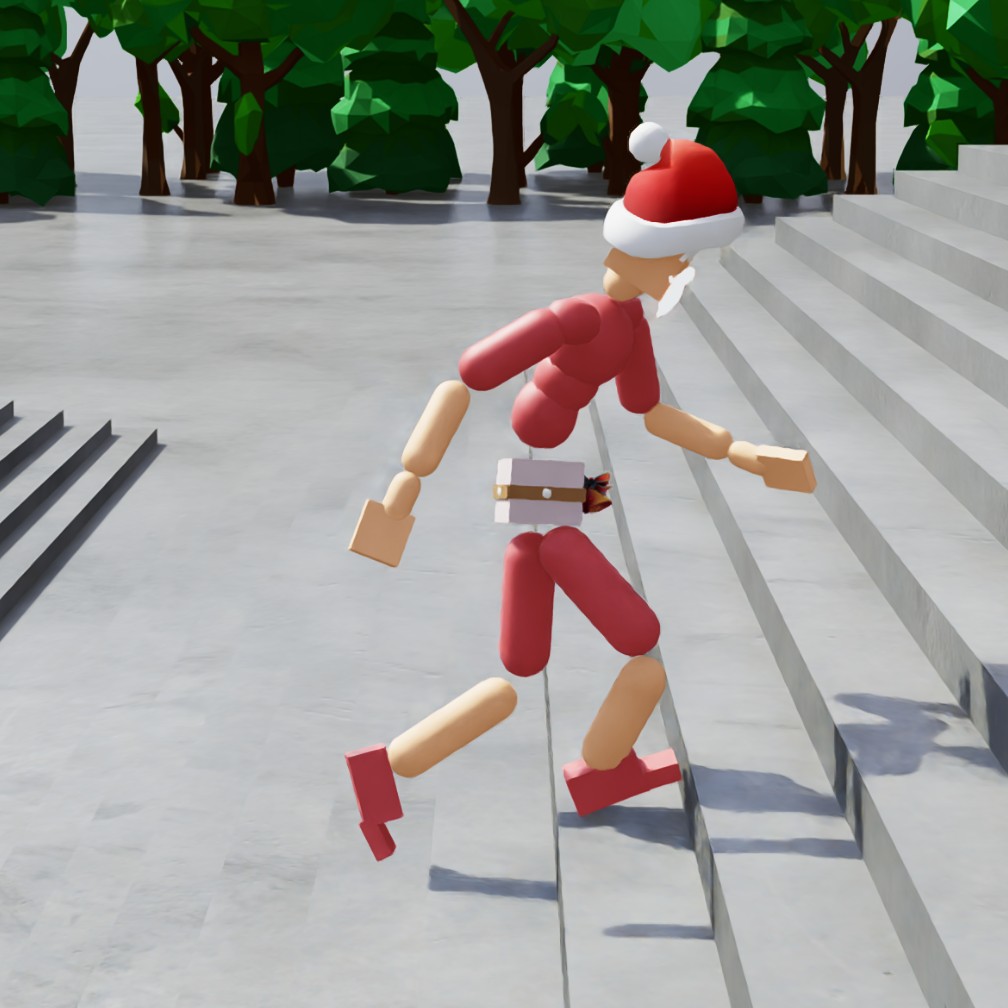}\hspace{0.5pt}%
        \includegraphics[width=0.2425\linewidth]{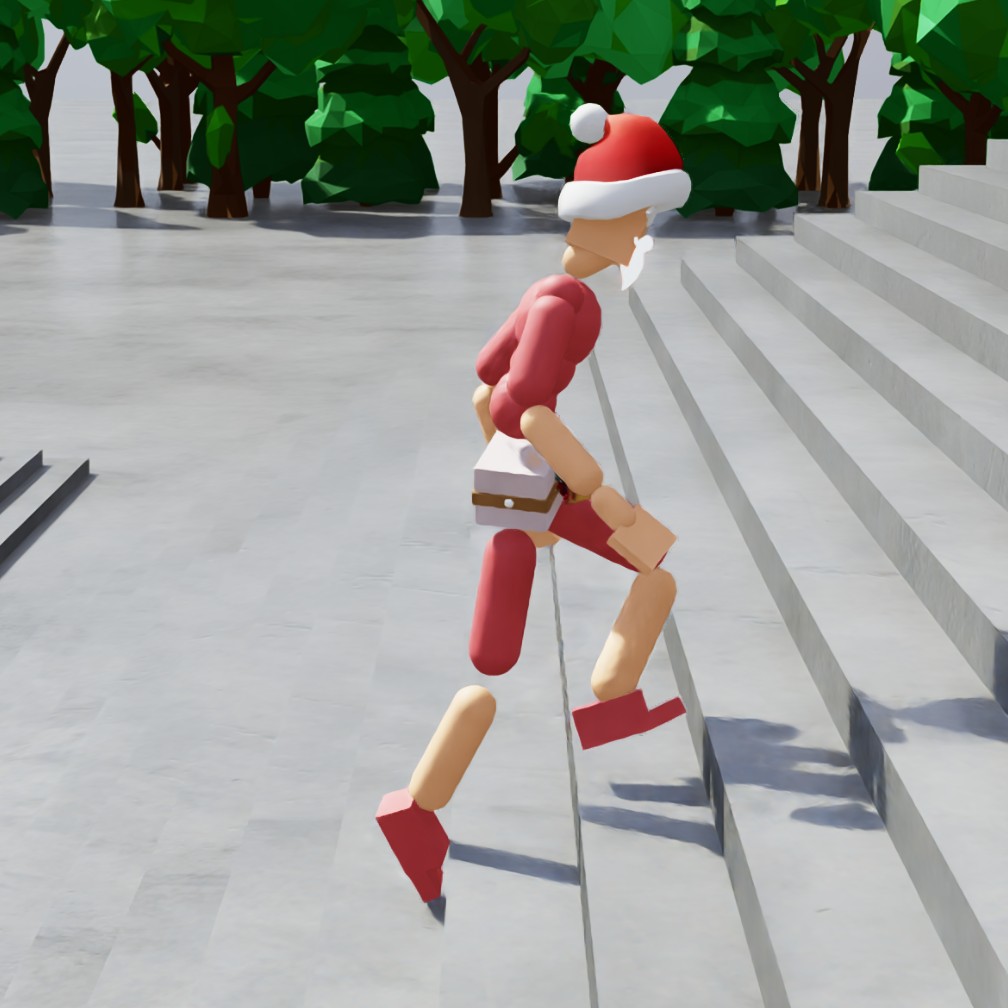}
        \caption{A person runs forward quickly.}
        \label{fig:t2m_run}
    \end{subfigure}\hfill
    \begin{subfigure}[t]{0.33333\textwidth}
        \centering
        \includegraphics[width=0.2425\linewidth]{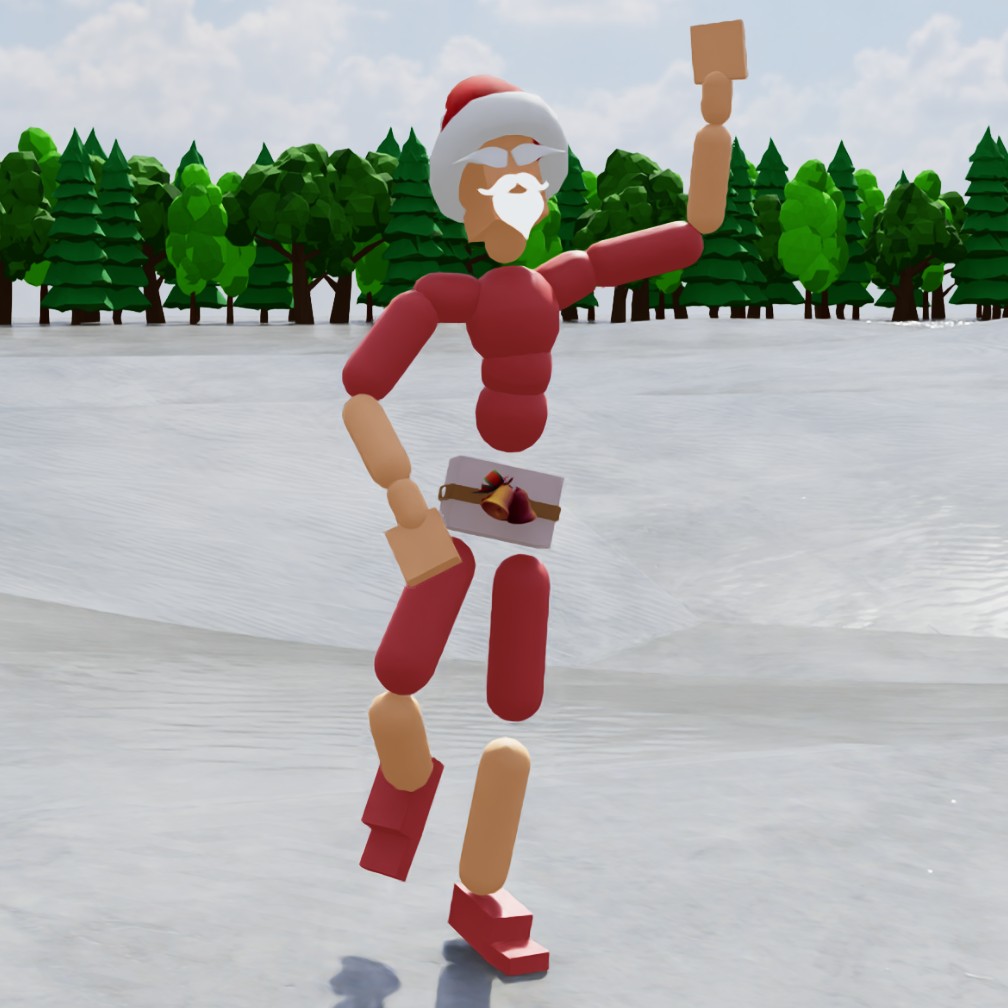}\hspace{0.5pt}%
        \includegraphics[width=0.2425\linewidth]{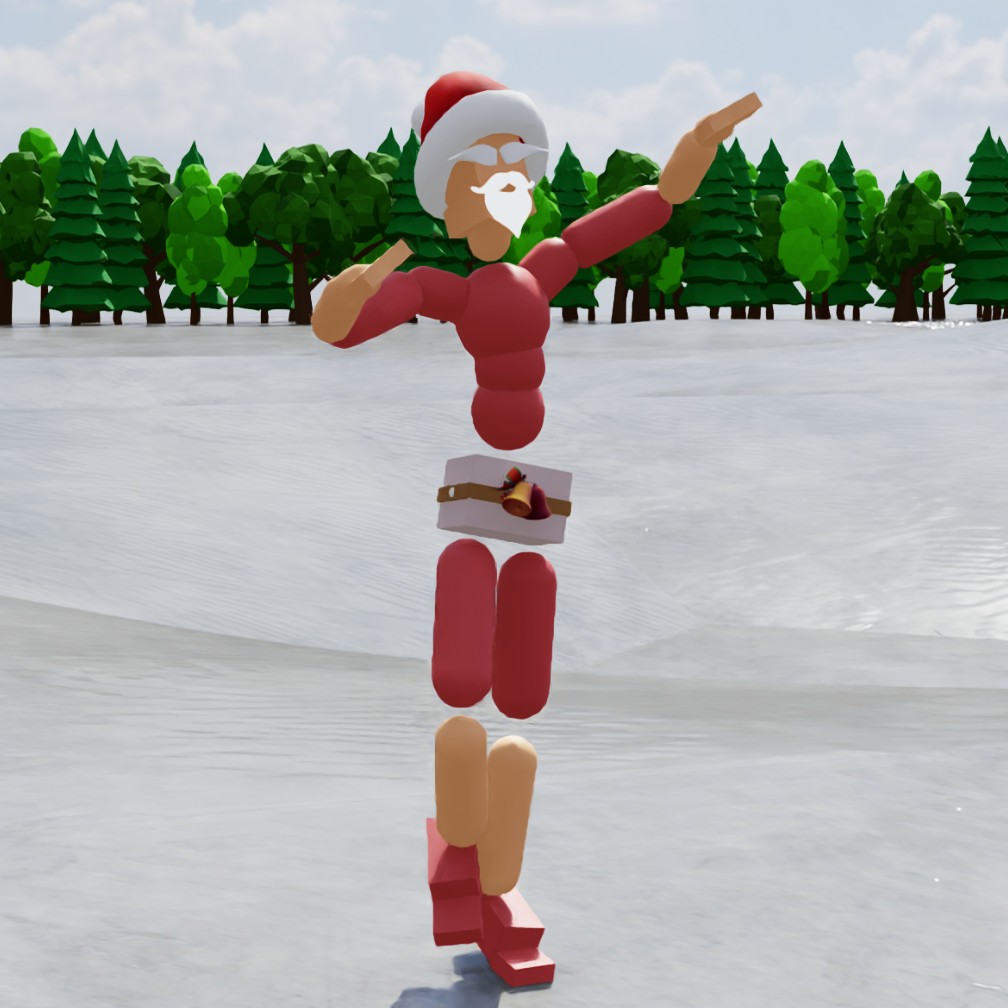}\hspace{0.5pt}%
        \includegraphics[width=0.2425\linewidth]{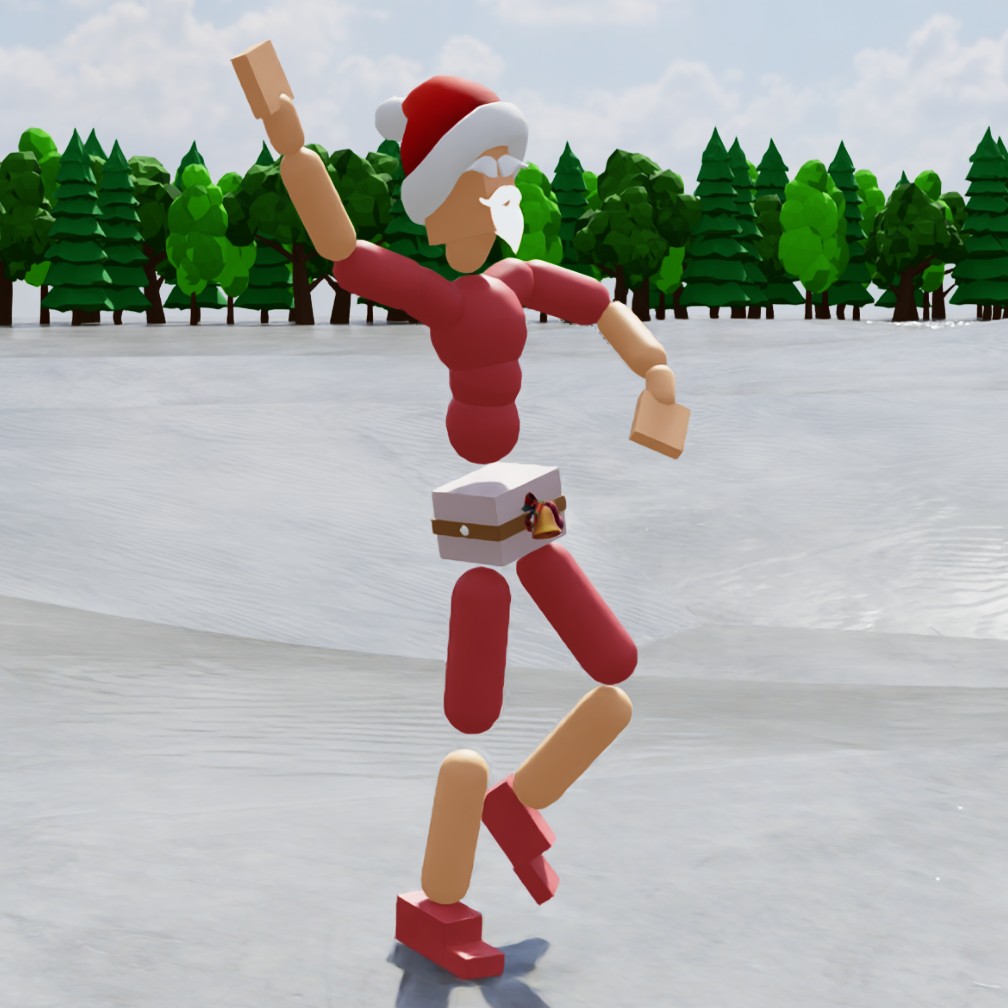}\hspace{0.5pt}%
        \includegraphics[width=0.2425\linewidth]{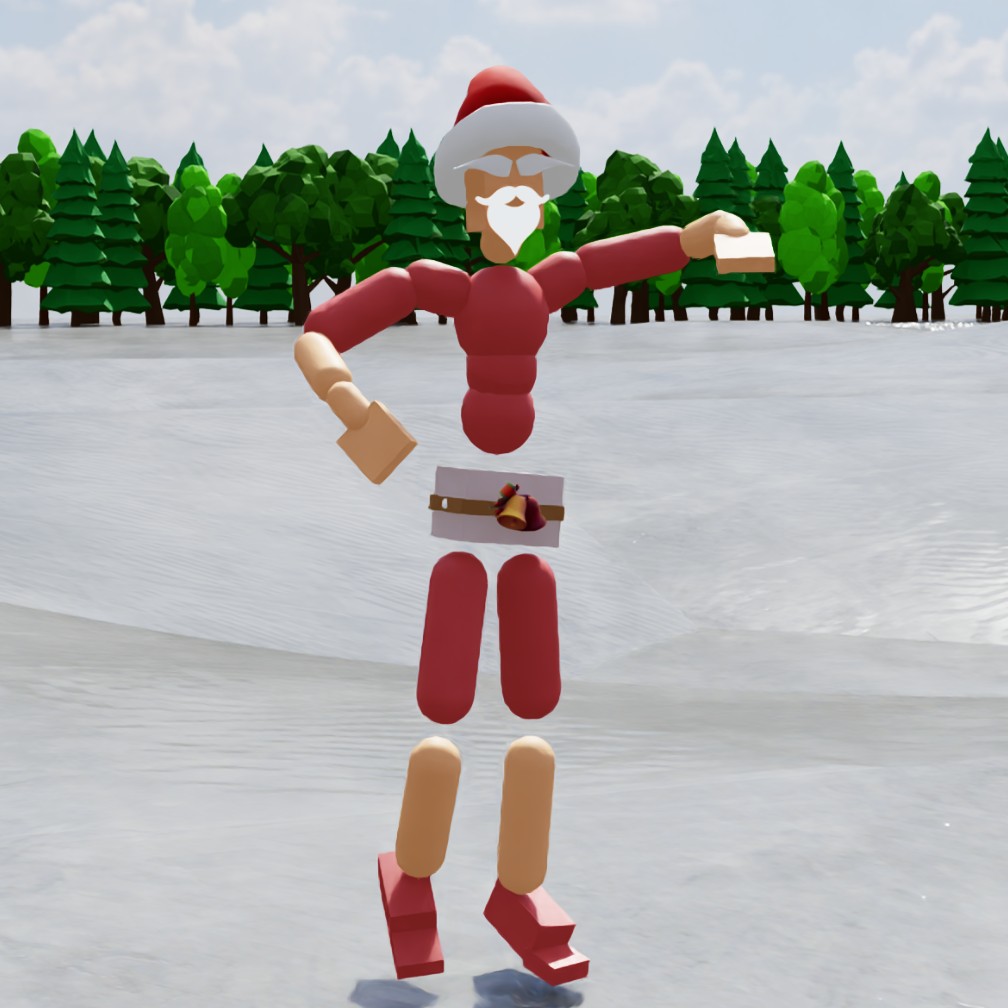}
        \caption{A person dances ballet.}
        \label{fig:t2m_ballet}
    \end{subfigure}\hfill
    \begin{subfigure}[t]{0.33333\textwidth}
        \centering
        \includegraphics[width=0.2425\linewidth]{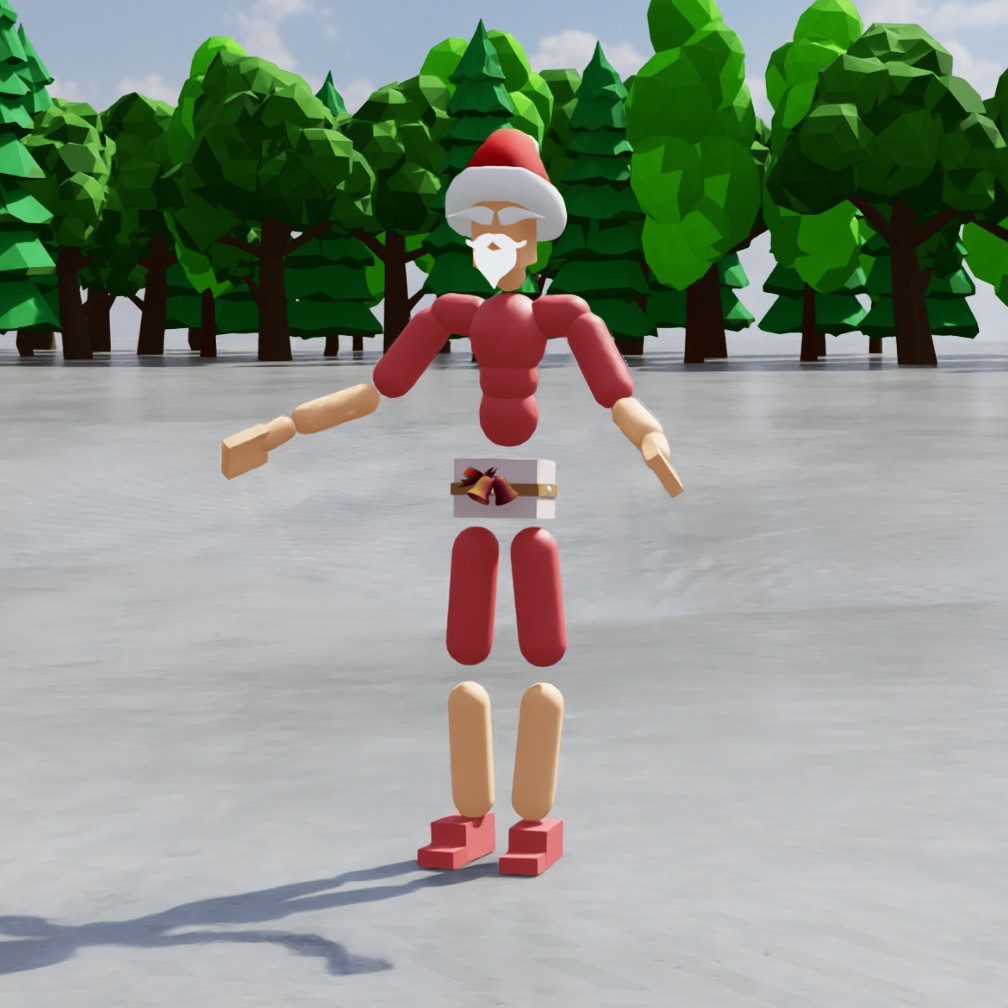}\hspace{0.5pt}%
        \includegraphics[width=0.2425\linewidth]{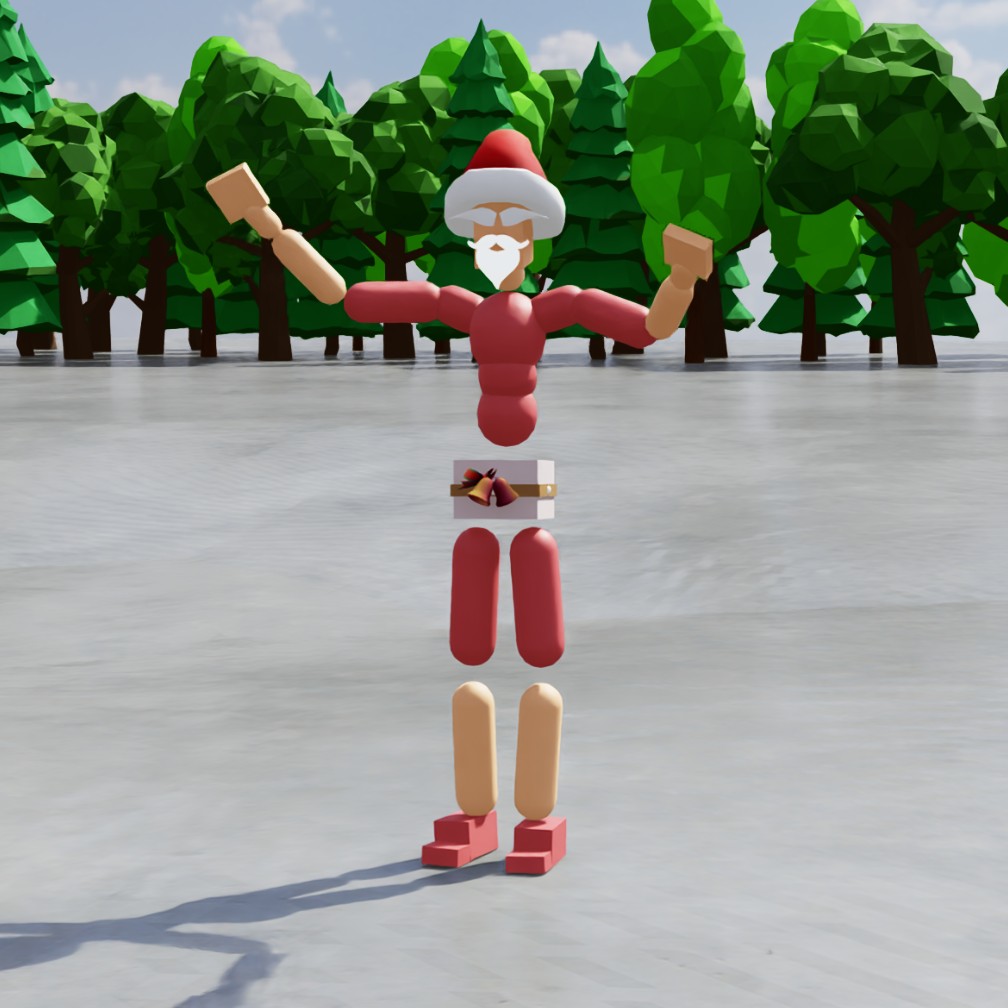}\hspace{0.5pt}%
        \includegraphics[width=0.2425\linewidth]{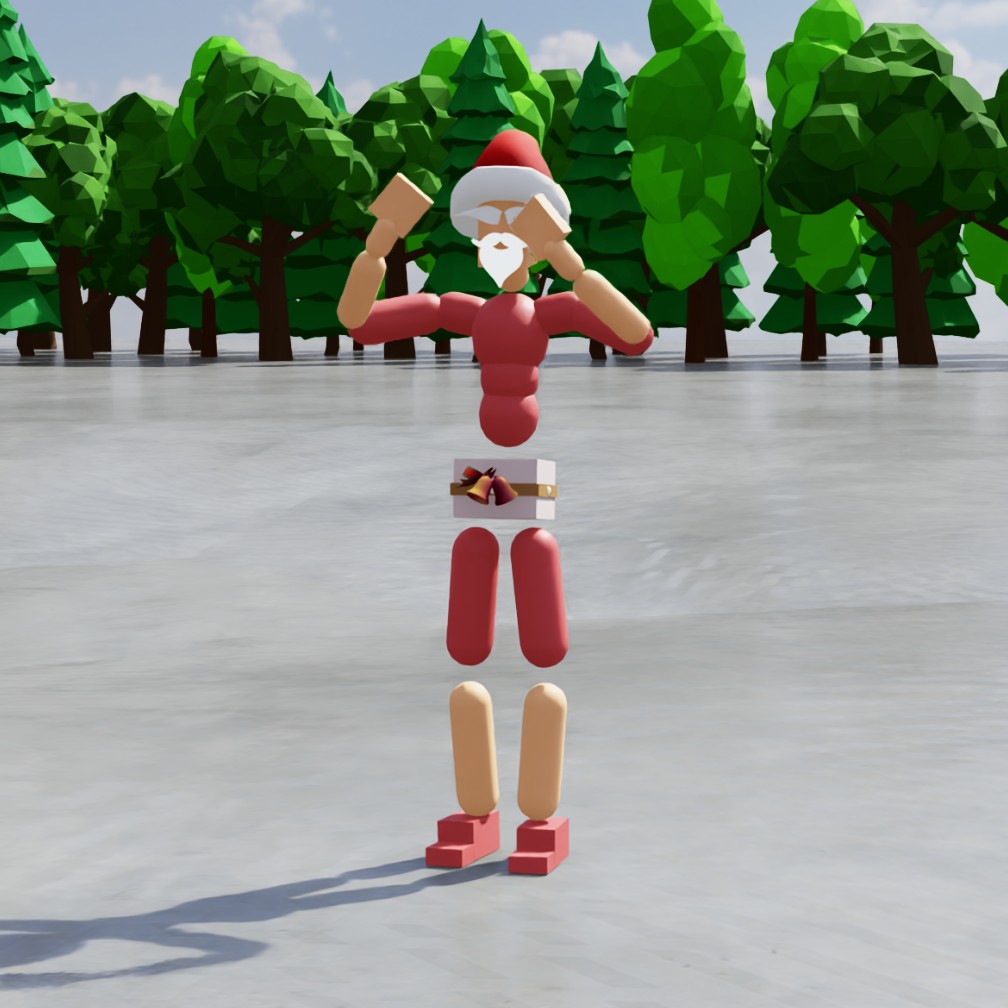}\hspace{0.5pt}%
        \includegraphics[width=0.2425\linewidth]{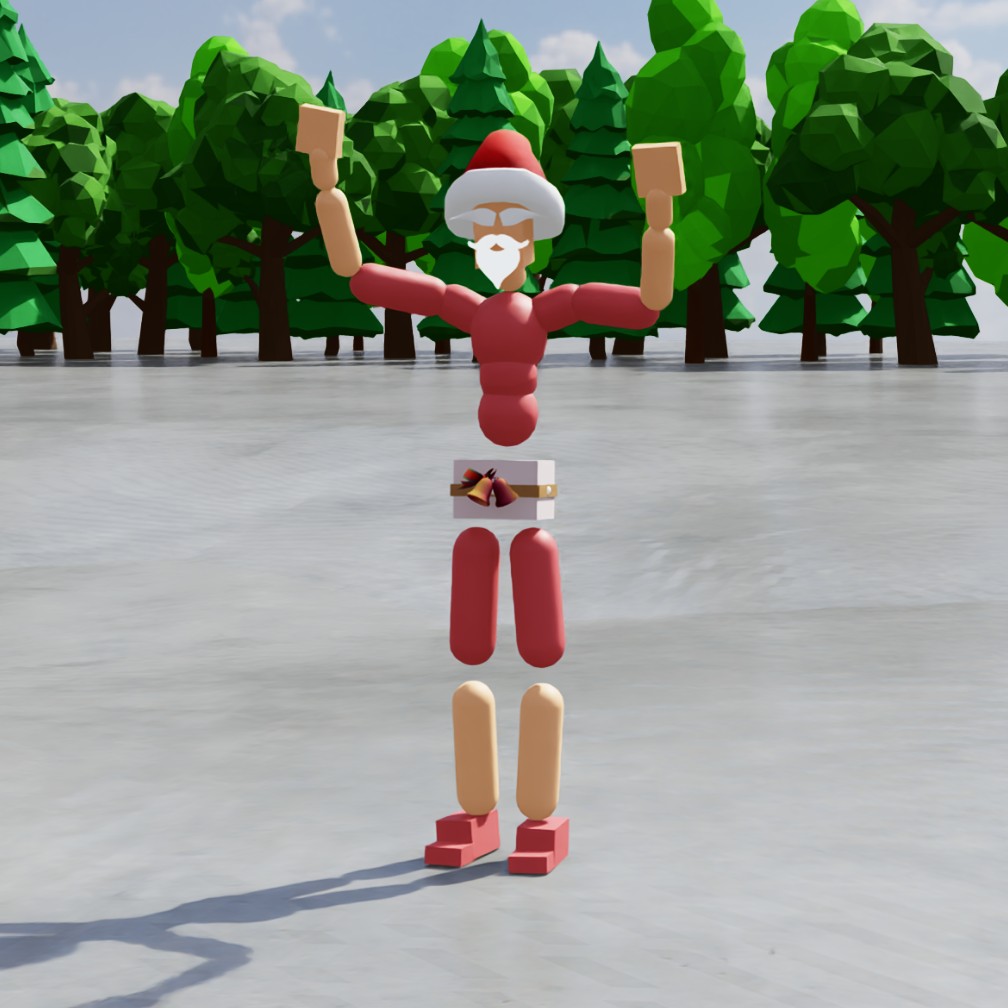}
        \caption{A person waves both hands.}
        \label{fig:t2m_wave}
    \end{subfigure}
    \caption{Qualitative results of text-to-motion generation. Given only natural-language descriptions, $\mathscr{F}^{\mathrm{t2m}}$ generates diverse full-body motions from a neutral-pose initialization, including acrobatic movements, locomotion, dance, crouching, lying down, running, and hand waving. These results show that the text-to-motion skill can produce semantically aligned and physically plausible motions even when the initial state does not match the target motion.}
    \label{fig:t2m_qualitative}
\end{figure*}

We evaluate the text-to-motion skill $\mathscr{F}^{t2m}$ from two complementary aspects. First, we evaluate pose-level robustness under two initialization protocols: starting from the ground-truth first frame of each target clip and starting from a neutral pose. A rollout is considered a failure if the mean per-joint position error (MPJPE) exceeds a predefined threshold at any frame~\citep{luo2021dynamics,tessler2024maskedmimic}, and we report success rates at thresholds of $0.3$ m and $0.5$ m in~\cref{tab:init_success}. Second, we evaluate semantic alignment following the HumanML3D retrieval protocol~\citep{guo2022generating}, where retrieval scores are computed using a pretrained TMR model~\citep{petrovich2023tmr}. Specifically, each generated motion is used as a query to retrieve its corresponding text description from a set of candidate texts in the shared TMR embedding space. We report R-Precision (R@N) and median rank (MedR), where R@N measures whether the correct match is ranked within the top N retrieved results, and MedR denotes the median rank of the correct match. Higher R@N and lower MedR indicate better semantic alignment between the generated motion and the input language.

\begin{table}[t]
\centering
\caption{Text-to-motion success rates on HumanML3D under two initialization protocols and two MPJPE failure thresholds. The upper block reports the main results, where rollouts start from either the ground-truth first pose or a neutral pose. The lower block ablates MID components, highlighting the importance of Randomized Memory Initialization and showing that the residual shortcut weakens language grounding.}
\label{tab:init_success}
\setlength{\tabcolsep}{6pt}
\renewcommand{\arraystretch}{1.1}
\begin{tabular}{l|cc|cc}
\hline\hline
& \multicolumn{2}{c|}{threshold: 0.3 m} & \multicolumn{2}{c}{threshold: 0.5 m} \\
\cline{2-5}
Method & First frame & Neutral & First frame & Neutral \\
\hline
MaskedMimic & 76.1\% & 19.8\% & 83.5\% & 34.0\% \\
CLoSD       & 30.9\% & 27.1\% & 44.2\% & 38.5\% \\
HetSkills   & 92.2\% & 60.2\% & 96.9\% & 81.7\% \\
\hline\hline
w/o RMI   & 92.6\% & 48.5\% & 97.7\% & 70.0\% \\
w/ residual   & 0\% & 0\% & 0\% & 0\% \\
\hline\hline
\end{tabular}
\end{table}

HetSkills outperforms both MaskedMimic~\citep{tessler2024maskedmimic} and CLoSD~\citep{tevet2024closd} across both initialization protocols in \cref{tab:init_success}. At the $0.5$ m threshold, HetSkills achieves $96.9\%$ success with the first-frame initialization and $81.7\%$ with the neutral-pose initialization, significantly surpassing all baselines. This advantage is especially noticeable under neutral-pose initialization, where other methods suffer from a sharp decline. This robustness is a direct result of MID, which removes the reliance on matched initial states and future context, forcing the model to ground motion generation solely in language semantics and current observations. The ablation results in~\cref{tab:init_success} further support this design: removing RMI causes a clear performance drop under neutral-pose initialization, while adding a residual branch alongside the text encoder leads to failure under both protocols due to the shortcut issue described in~\cref{sec:mid}. These results indicate that RMI is critical for preventing overfitting to matched motion histories and for maintaining effective language conditioning. In contrast to diffusion-based methods, the next-token prediction approach of HetSkills allows real-time closed-loop corrections, preventing positional errors from accumulating across motion segments. Qualitative results across diverse action categories are shown in~\cref{fig:t2m_qualitative}, with animated demonstrations provided in the supplementary video.

The retrieval results in~\cref{tab:t2m_retrieval} further show that HetSkills preserves strong text-motion alignment under the neutral-pose initialization protocol. Compared with the baselines, HetSkills achieves higher R@N scores and a lower MedR, indicating that the generated motions are more consistently retrieved as their corresponding language descriptions in the TMR embedding space. This result complements the MPJPE-based success rates: while MPJPE measures whether the generated motion follows the target pose sequence, R@N and MedR evaluate whether the motion remains semantically recognizable as the intended textual action. Together, these results suggest that HetSkills improves not only initialization robustness, but also language-motion consistency.

\begin{table}[t]
\centering
\caption{Text-to-motion retrieval performance under the HumanML3D evaluation protocol \cite{guo2022generating} with neutral-pose initialization. Retrieval metrics are computed using the pretrained TMR model. R@N measures the fraction of correct text-motion matches ranked within the top N retrieved results, while MedR denotes the median rank of the correct match. Higher R@N and lower MedR indicate better text-motion alignment.}
\label{tab:t2m_retrieval}
\setlength{\tabcolsep}{6pt}
\renewcommand{\arraystretch}{1.1}
\begin{tabular}{l|cccc|c}
\hline\hline
Method & R@1$\uparrow$ & R@2$\uparrow$ & R@3$\uparrow$ & R@5$\uparrow$ & MedR$\downarrow$ \\
\hline
Ground Truth & 71\% & 86\% & 91\% & 96\% & 1.00 \\
HetSkills    & 65\% & 82\% & 88\% & 93\% & 1.03 \\
CLoSD        & 41\% & 58\% & 67\% & 77\% & 2.08 \\
MaskedMimic  & 38\% & 54\% & 62\% & 72\% & 2.41 \\
\hline\hline
\end{tabular}
\end{table}

\begin{figure*}
    \centering
    \includegraphics[width=1\linewidth]{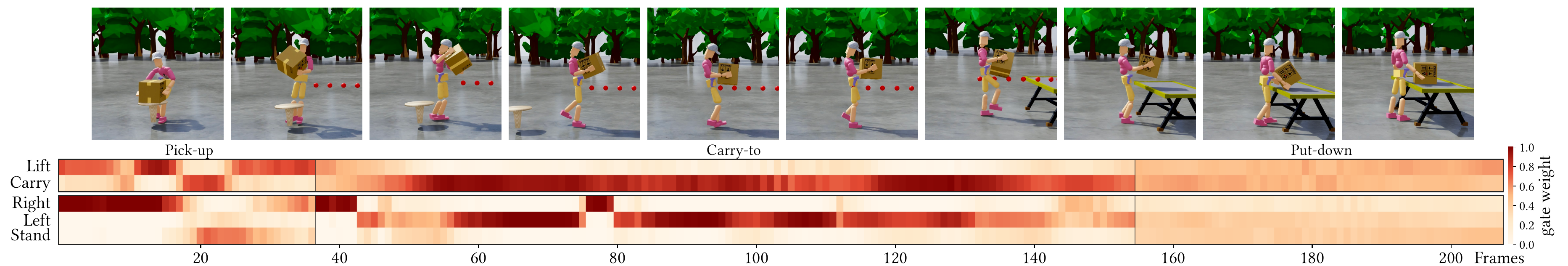}
    \caption{MoE gating dynamics during the pick-and-place task. The rendered sequence at the top shows rollout frames sampled every $20$ frames across the pick-up, carry-to, and put-down stages. The heatmap below shows how the Arm and Body branches assign time-varying weights to different language-conditioned priors. The changing weights indicate that the task-guidance module dynamically selects and blends different semantic priors as the task stage evolves; the full language prompts are listed in~\cref{tab:instructions}.}
    \label{fig:MoE_weight}
\end{figure*}

\begin{figure}[t]
    \centering
    \includegraphics[width=\linewidth,trim=0 0 0 0,clip]{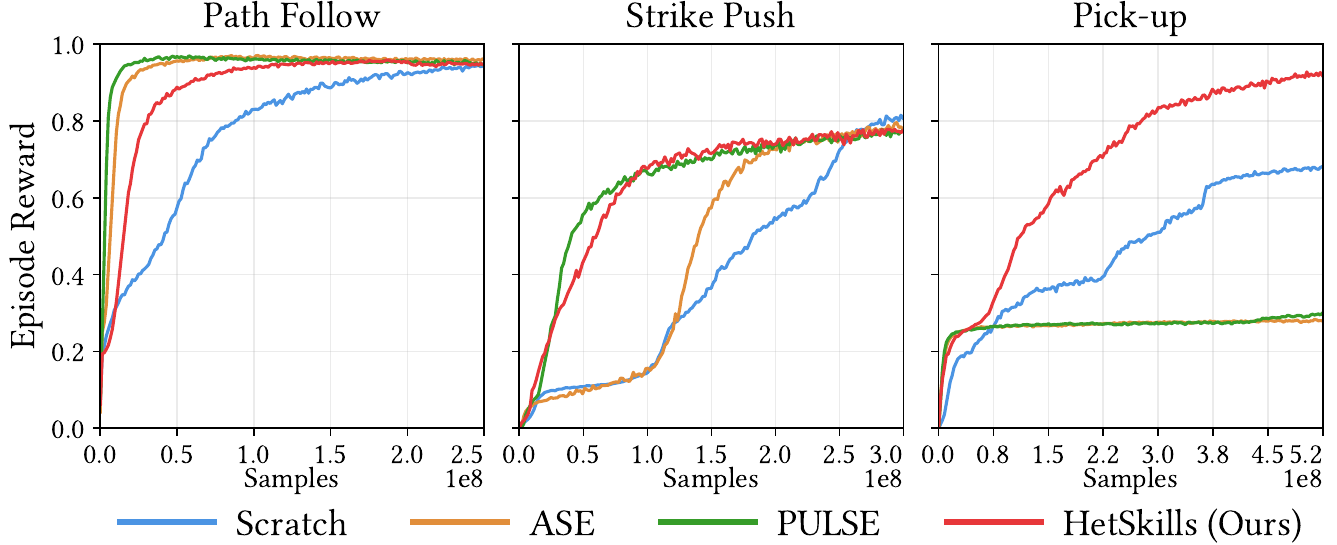}
    \caption{Downstream training curves on path follow, strike push, and pick-up. Episode rewards are plotted against environment samples and averaged over three random seeds. Compared with Scratch, ASE, and PULSE, HetSkills converges stably to high task rewards across all tasks, while preserving natural motion through the frozen language-conditioned prior.}
    \label{fig:RL-curve}
\end{figure}

\subsection{Downstream Task Adaptation}

HetSkills adopts the language-conditioned prior to restrict downstream exploration to a semantically meaningful distribution of natural human motions. This design improves motion naturalness and makes task-relevant behaviors easier to discover, especially when the desired behavior occupies only a small region of the latent space. We evaluate these properties through task performance across three downstream tasks. In particular, the pick-and-place task serves as a representative compositional setting, where multiple language-conditioned skill priors are routed and combined to accomplish a single object-interaction objective; we further analyze the MoE gating behavior to show how the model dynamically selects and composes these priors during execution.

\paragraph{Comparison Across Tasks.}
The training curves in~\cref{fig:RL-curve} show that HetSkills converges stably across all three downstream tasks. On path follow, all methods achieve comparable task rewards, but ASE and PULSE occasionally generate backward-walking gaits because their priors do not impose semantic constraints on locomotion direction. On strike, all methods can knock over the target, while the baselines often produce awkward postures due to insufficient constraints on the action distribution. In contrast, HetSkills generates more natural kicking and pushing motions. The largest difference appears on pick-up, where only HetSkills converges reliably. This task requires coordinated bimanual lifting, which occupies a small and sparse region of the latent space. Without semantic guidance, the policy struggles to find this region within the limited sample budget and can fall into local optima.

These results highlight two main advantages of HetSkills for downstream adaptation. First, the semantic constraint from the language-conditioned prior helps the policy preserve natural human-like motion while optimizing task rewards. More importantly, it narrows exploration toward task-relevant regions of the latent space, making sparse and coordinated behaviors easier to discover. Second, adapting to a new task only requires changing the language instructions, without task-specific motion data or prior retraining. This provides a simple and flexible interface for reusing the learned skill space across different downstream objectives. Additional rollout examples are included in the supplementary video.

\paragraph{Compositional Task Guidance.}
\cref{fig:MoE_weight} visualizes the gating weights assigned to different language instructions during the pick-and-place task. The visualization shows how the MoE module changes the active semantic prior across the pick-up, carry-to, and put-down stages. We observe three common patterns.

\textbf{Dominance.} The gating network assigns most of the weight to one instruction that best matches the current task stage. This indicates that the model can identify the most relevant semantic prior and use it as the main behavior source.

\textbf{Periodicity.} The weight of the dominant instruction changes periodically over time. This pattern reflects the intrinsic rhythm of the corresponding motion prior, such as the gait cycle in walking or the preparation, execution, and recovery phases in grasping.

\textbf{Complementarity.} When the dominant instruction enters a low-activity phase, the gating network shifts part of the weight to a semantically related instruction. This compensation helps fill behavioral gaps and maintain smooth transitions over long-horizon execution.

Overall, the MoE module provides an adaptive mechanism for semantic prior selection rather than assigning a fixed instruction to each task stage. It selects the most relevant prior according to the current task context, aligns the generated behavior with the corresponding temporal structure, and integrates complementary instructions when necessary. This mechanism enables a compact set of language priors to support coherent and coordinated long-horizon downstream behaviors.

\subsection{Long-Horizon Skill Composition}
\label{sec:long_horizon_composition}

\begin{figure*}[t]
    \centering
    \includegraphics[width=\textwidth]{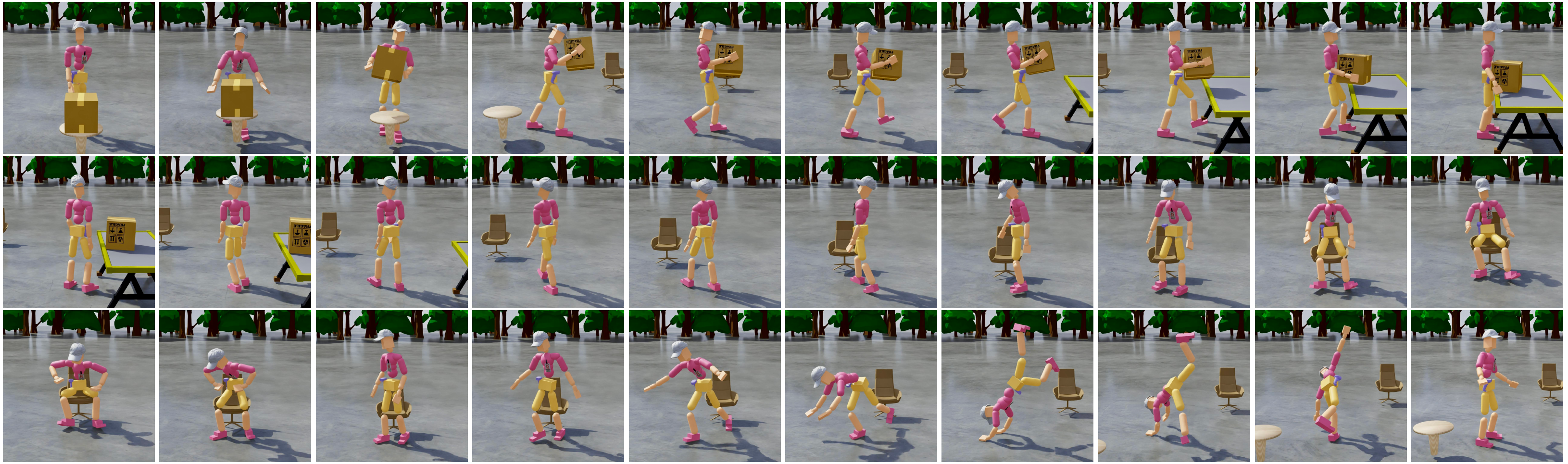}
    \caption{Long-horizon composition of heterogeneous skills. A high-level task is decomposed into a sequence of task units, including pick-up, carry-to, put-down, path follow, human-scene interaction, and text-to-motion, with each unit executed by its corresponding skill module within the same unified latent space. This demonstrates that skills learned from different data sources, supervision forms, and task objectives can be composed seamlessly without modifying the shared controller or manually designing low-level transition rules.}
    \label{fig:long_horizon_demo}
\end{figure*}

\begin{figure}[t]
\centering
\small
\begin{minipage}{0.98\linewidth}
\hrule
\vspace{0.4em}
\textbf{Prompt: Long-Horizon Skill Composition}
\vspace{0.4em}
\hrule
\vspace{0.6em}
\setlength{\parindent}{0pt}
\setlength{\parskip}{0.5em}
\sloppy
Please follow the Standardized Task Description to generate an executable long-horizon skill program for a SMPL character. The task depicts a humanoid that retrieves a box from a source table, transports it to a nearby target platform, moves toward a chair, interacts with the scene, and finally performs an open-ended text-conditioned motion. The detailed skill sequence is specified below.

\textbf{Pick-up Skill.}
Initialization: Spawn the humanoid next to the source table with the box.
Condition: Provide the box position.
Terminate: End when the humanoid stably lifts the box for several consecutive frames.

\textbf{Carry-to Skill.}
Initialization: Use the final pick-up state as the initial state.
Condition: Generate and follow a path from the current position to the front of the target platform.
Terminate: End when the humanoid reaches the neighborhood of the target platform.

\textbf{Put-down Skill.}
Initialization: Use the final carry-to state as the initial state.
Condition: Provide the target platform position to the put-down policy.
Terminate: End when the box is successfully placed for several consecutive frames.

\textbf{Path Follow Skill.}
Initialization: Start from the final state of the put-down Skill.
Condition: Generate and follow a walking path toward a target location in front of the chair.
Terminate: End when the humanoid reaches the target location in front of the chair.

\textbf{Human-Scene Interaction Skill.}
Initialization: Start from the final state of the path follow skill.
Condition: Specify a scheduled interaction sequence, including turning toward the chair, sitting down, and standing up, and assign a duration to each target pose.
Terminate: End when the interaction sequence is completed.

\textbf{Text-to-Motion Skill.}
Initialization: Start from the final state of the Human-scene Interaction Skill.
Condition: Set both prompts to ``a person doing cartwheel''.
Terminate: Continue until the episode is externally stopped.
\vspace{0.3em}
\hrule
\end{minipage}
\caption{Example LLM prompt for long-horizon skill composition.}
\label{fig:llm_prompt}
\end{figure}

To further demonstrate the compositional capability of HetSkills, we construct a long-horizon demonstration that sequentially combines multiple heterogeneous skills into one continuous behavior. The demonstration includes object manipulation, goal-directed locomotion, human-scene interaction, and text-conditioned motion generation, where the character picks up a box, places it on a target platform, walks toward a chair, sits down, and then rises from the chair to perform an open-ended text-conditioned motion.

The long-horizon behavior is generated through the standardized task interface introduced in~\cref{sec:task_description}. As shown in the prompt example in \cref{fig:llm_prompt}, the user only needs to describe the task at the skill level, where each stage is represented by its initialization, condition, and termination criterion. The large language model (LLM) \cite{chang2024survey,singh2025openai} then organizes the high-level instruction into a structured sequence of executable task units, making it easier to compose different skills without manually designing low-level transition logic. Importantly, the LLM is not used to generate motion trajectories directly. Instead, it serves as a convenient planning interface \cite{wang2025sims,yao2024moconvq,wu2025human} that translates a natural task description into task units executed by the corresponding skill modules.

This design makes long-horizon composition more flexible and easier to specify. Since all skills share the same latent control space and standardized task format, skills learned from different supervision forms and task settings can be combined within a single framework. New behaviors can be expressed by changing the task description, target conditions, or skill ordering, without manually designing detailed controller-switching logic.

As shown in the long-horizon demonstration in ~\cref{fig:long_horizon_demo}, the generated behavior remains coherent across different stages, even though the underlying skills involve different objectives and conditioning modalities. This suggests that HetSkills provides a practical interface for organizing heterogeneous skills into reusable and extensible long-horizon behavior programs. To further illustrate the compositional flexibility of HetSkills, we provide additional examples with different skill combinations and task sequences in the appendix \ref{appendix:skill}.

\section{Limitations and Future Work}
Our experiments reveal two main limitations. First, although HetSkills demonstrates progressive integration across several representative skill categories, the current experiments do not yet fully cover larger-scale skill accumulation, more complex long-horizon composition, or more open-ended human-object and human-scene interactions. Second, text-driven motion generation still struggles with high-difficulty actions and ambiguous language descriptions. MID improves robustness to different initial states and motion histories. However, the text-to-motion skill remains affected by the quality of language-motion annotations. In particular, semantically similar descriptions may correspond to substantially different motions, which makes fine-grained language-motion alignment more difficult.

Future work includes improving HetSkills along three directions. First, incorporating more diverse contact-rich and scene-aware motion data during the tracking stage could help the unified latent space better cover interaction behaviors. Second, contact-aware modules and higher-quality text-motion annotations could improve interaction stability and text-driven generation fidelity. Finally, extending HetSkills to a broader range of skills and real-world robot control is a promising direction. Another interesting direction is to further explore the role of large language models in high-level planning. In this work, an LLM is used to help organize high-level instructions into task sequences, illustrating the potential of combining standardized task interfaces with language-based planning. Future work could further investigate more robust task decomposition, automatic skill selection, and failure recovery for more open-ended long-horizon tasks.

\section{Conclusion}

We presented HetSkills, a physics-based character control framework that progressively learns heterogeneous skills within a unified latent space. The key idea is to treat the latent space as a shared executable interface, allowing skills from different data sources, supervision forms, and training stages to extend the same control substrate rather than requiring dedicated controllers. HetSkills integrates motion tracking, text-to-motion generation, motion completion, and downstream task adaptation, while supporting natural language as a flexible interface for both motion generation and task guidance. With part-wise latent control, motion intuition distillation, and language-guided latent composition, the framework preserves natural motion quality, enables robust skill reuse, and adapts to new tasks without task-specific demonstrations, motion-prior retraining, or shared-controller modification. Experiments show that HetSkills can effectively accommodate diverse skills and task objectives, suggesting that a unified latent space is a practical foundation for scalable, reusable, and progressively extensible character control.

\bibliographystyle{ACM-Reference-Format}
\bibliography{acmart}

\clearpage

\onecolumn

\appendix

\section{Implementation Details}
Unless otherwise specified, all trainable policy and prior modules are optimized with Adam~\cite{kingma2014adam}. We use a learning rate of $2\times10^{-5}$ for policy and prior networks, and a learning rate of $10^{-4}$ for critic networks. The arm and body latent dimensions are both set to $64$. Transformer-based prior modules embed all input tokens into $256$-dimensional features. The default latent regularization coefficients are $\lambda_{\mathrm{mr}}=1\times10^{-3}$, $\lambda_{\mathrm{ar}}=5\times10^{-3}$, and $\phi=0.99$.

\paragraph{Tracking skill $\mathscr{F}^{trc}$.} We train the part-wise tracking policy with PPO. The shared state encoder outputs a $512$-dimensional feature. Exploration is applied only in the final action space using a fixed diagonal Gaussian distribution with log standard deviation $-2.9$. We use a tracking-error termination threshold of $0.5\,\mathrm{m}$. Reference State Initialization (RSI) is applied with probability $0.8$; otherwise, the episode starts from the first frame of the reference motion. The tracking reward includes global joint position, global joint rotation, joint velocity, joint angular velocity, and root-height terms, with weights $w_{\mathrm{gp}}=0.5$, $w_{\mathrm{gr}}=0.3$, $w_{\mathrm{jv}}=0.1$, $w_{\mathrm{jav}}=0.1$, and $w_{\mathrm{rh}}=0.2$, and exponential coefficients $k_{\mathrm{gp}}=100$, $k_{\mathrm{gr}}=5$, $k_{\mathrm{jv}}=0.5$, $k_{\mathrm{jav}}=0.1$, and $k_{\mathrm{rh}}=100$. We further use a contact mismatch penalty, an action smoothness penalty, and an energy penalty with weights $w_{\mathrm{ct}}=-0.1$, $w_{\mathrm{sm}}=-0.02$, and $w_{\mathrm{eg}}=-10^{-5}$, respectively. The energy penalty is clipped from below at $-0.5$, and the reference contact signal is smoothed with a $7$-frame temporal window.

\paragraph{Text-to-motion skill $\mathscr{F}^{t2m}$.} The text-to-motion prior uses two part-wise Transformer branches, one for the arm latent and one for the body latent. The input tokens consist of one current-state token, one language token, and $6$ history tokens uniformly sampled from a $60$-step history window. The language token is obtained from the frozen text encoder used for language-motion alignment. During training, we apply RSI with probability $p_{\mathrm{rsi}}=0.7$ and Randomized Memory Initialization (RMI) with probability $p_{\mathrm{rmi}}=0.2$. This initialization scheme encourages the policy to rely on language semantics and current observations rather than matched initial states or memorized motion histories.

\paragraph{Motion completion skill $\mathscr{F}^{moc}$.} We instantiate the motion completion skill in three sparse-observation settings: (1) \textit{VR tracking}. The sparse target contains one future step of head and both hands states. The prior is a four-layer MLP with $1024$ hidden units, and outputs a $128$-dimensional latent vector split into $z_a,z_b\in\mathbb{R}^{64}$; (2) \textit{Motion in-betweening.} The prior uses two part-wise Transformer branches. The input consists of the current state, a future target keyframe pose, its time offset, and $3$ historical poses uniformly sampled from a $30$-step history window. The target offset is uniformly sampled from $5$ to $30$ future frames; (3) \textit{Human-scene interaction.} We train the human-scene interaction prior with PPO. The input consists of the current state, a future target keyframe pose, its time offset, and $3$ historical poses uniformly sampled from a $30$-step history window. The target offset is uniformly sampled from $5$ to $30$ future frames. We use $6$ PPO mini-epochs per update. The reward uses the same global joint position, global joint rotation, joint velocity, joint angular velocity, action smoothness, and energy terms as the tracking skill, but removes the root-height and contact mismatch terms. 

\paragraph{Downstream task adaptation $\mathscr{F}^{tsk^*}$.}
For downstream adaptation, the pretrained motion prior and low-level decoder are kept frozen, and only the task-guidance modules are optimized. The arm and body residual latent dimensions are both set to $64$. We use rollouts of length $32$ and $4$ PPO mini-epochs per update. The latent policy log standard deviation is annealed from $-2.5$ to $-3.0$ between epochs $500$ and $1000$. Adapted latents outside the interval $[-0.1,0.1]$ are penalized with coefficient $10.0$, and the latent smoothness penalty coefficient is set to $0.1$.

\section{Training Time and Compute}
The progressive design of HetSkills allows different skills to be trained independently on top of the shared latent space, instead of jointly optimizing all skills in a single monolithic model. This design keeps each training stage relatively lightweight. After the tracking skill learns the shared motion decoder, later stages only need to learn skill-specific latent mappings or task-guidance modules while reusing the frozen low-level controller.

In our implementation, the tracking skill is trained on two RTX 5090 GPUs for approximately $2$ days, and the text-to-motion skill is trained on two RTX 5090 GPUs for approximately $5$ days. For motion completion, the three instantiations are trained independently on a single RTX 5090 GPU. Human-scene interaction takes approximately $2$ days, VR tracking takes approximately $1$ day, and motion in-betweening takes approximately $2$ days.

For downstream task adaptation, each body part may be conditioned on multiple language instructions. These instruction branches are evaluated in parallel by batching them together, so increasing the number of instructions does not lead to a proportional increase in wall-clock training time. In our experiments, each downstream task can be trained within $5$ hours on a single RTX 5090 GPU.

\section{Downstream Task Rewards and Settings}\label{app:reward}

This section describes the reward functions and environment settings used for downstream task adaptation. All rewards are designed to encourage task completion while regularizing the motion with energy penalties, so that the adapted behaviors remain physically plausible and compatible with the pretrained motion prior.

\paragraph{Path Follow.}
The path follow reward encourages the character root to track an online-generated target path in the horizontal plane:
\begin{equation}
    r = \exp\!\left(-\|p_{\text{target}}^{xy} - p_{\text{root}}^{xy}\|^2\right) - \lambda \cdot \mathcal{P}_{\text{leg}}
\end{equation}
where $p_{\text{target}}^{xy}$ and $p_{\text{root}}^{xy}$ denote the horizontal positions of the current path target and the character root, respectively. The leg energy term is defined as
\begin{equation}
    \mathcal{P}_{\text{leg}} = \sum_i |\tau_i \dot{q}_i|,
\end{equation}
where the summation is taken over the $24$ leg DoFs, including the bilateral hip, knee, ankle, and toe joints. We set $\lambda = 10^{-5}$ and clip the resulting energy penalty to $[-0.1,0]$. The path is generated online with a maximum speed of $5\,\mathrm{m/s}$ and a maximum acceleration of $2\,\mathrm{m/s^2}$. An episode terminates if the horizontal distance between the root and the current path target exceeds $1.5\,\mathrm{m}$.

\paragraph{Strike.}
The strike reward encourages the character to approach the target, face it, move toward it, and apply a motion that tilts the target:
\begin{equation}
    r = 0.6\, r_{\text{rot}} + 0.2\, r_{\text{vel}} + 0.1\, r_{\text{prog}} + 0.1\, r_{\text{toward}} - \lambda \mathcal{P}.
\end{equation}
The reward components are defined as
\begin{equation}
\begin{alignedat}{2}
    r_{\text{rot}} &=
    \max(1 - \mathbf{u}^{\top} R_{\text{target}}\mathbf{u},\ 0),
    \qquad&
    r_{\text{vel}} &=
    \exp\!\left(-4(2.5 - v_{\text{along}})^2\right),
    \\
    r_{\text{prog}} &=
    \operatorname{clamp}\!\left(\frac{D_0 - D_t}{D_0 + \epsilon},\ 0,\ 1\right),
    \qquad&
    r_{\text{toward}} &=
    \operatorname{clamp}\!\left(\mathbf{d}_{\text{face}} \cdot \mathbf{d}_{\text{target}},\ 0,\ 1\right).
\end{alignedat}
\label{eq:reward_components}
\end{equation}
Here, $\mathbf{u}=[0,0,1]^{\top}$ is the world up vector, $R_{\text{target}}$ is the target orientation, and $v_{\text{along}}$ is the root velocity projected onto the horizontal direction toward the target. $D_0$ and $D_t$ denote the initial and current horizontal distances to the target, respectively. $\mathbf{d}_{\text{face}}$ is the character heading direction, and $\mathbf{d}_{\text{target}}$ is the direction from the character to the target in the horizontal plane. Once the target tilt exceeds approximately $78^\circ$, the strike reward is set to $1$. The energy term is defined as $\mathcal{P}=\sum_i|\tau_i\dot{q}_i|$ over all joints. We set $\lambda = 10^{-5}$ and clip the resulting energy penalty to $[-0.1,0]$. The target is initialized at a random horizontal distance between $0.5\,\mathrm{m}$ and $10.0\,\mathrm{m}$ from the character.

\paragraph{Pick-and-Place.}
The pick-and-place task is decomposed into three stages, including pick-up, carry-to, and put-down. Each stage uses a stage-specific reward while sharing the same energy regularization form. The pick-up reward encourages the character to face the box, grasp it with both hands, apply sufficient contact force, and lift it from the source platform:
\begin{equation}
    r = 0.4\, r_{\text{lift}} + 0.3\, r_{\text{grasp}} + 0.2\, r_{\text{force}} + 0.1\, r_{\text{face}} - \lambda \cdot \mathcal{P}
\end{equation}
The reward components are
\begin{equation}
\begin{alignedat}{2}
    r_{\text{lift}} &=
    \operatorname{clamp}\!\left(\frac{z_{\min} - h_{\text{src}}}{h_{\text{target}} - h_{\text{src}}},\ 0,\ 1\right),
    \quad&
    r_{\text{grasp}} &=
    \frac{1}{2}\left(e^{-3 d_R} + e^{-3 d_L}\right),
    \\
    r_{\text{force}} &=
    \operatorname{clamp}\!\left(\frac{\min(\|F_R\|, \|F_L\|)}{10},\ 0,\ 1\right)
    \cdot \mathbf{1}[\text{both hands close}],
    \quad&
    r_{\text{face}} &=
    \operatorname{clamp}\!\left(\mathbf{d}_{\text{face}} \cdot \mathbf{d}_{\text{box}},\ 0,\ 1\right).
\end{alignedat}
\label{eq:pickup_reward_components}
\end{equation}
Here, $z_{\min}$ is the lowest corner height of the box, $h_{\text{src}}$ is the source platform surface height, and $h_{\text{target}}=0.8\,\mathrm{m}$ is the target lifting height. $d_R$ and $d_L$ are the distances from the right and left hands to their nearest grasp points on opposite sides of the box. $F_R$ and $F_L$ are the right and left hand contact forces. The indicator $\mathbf{1}[\text{both hands close}]$ equals $1$ only when both hands are within $0.1\,\mathrm{m}$ of their corresponding grasp points. The energy term is computed over all joints with $\lambda=10^{-5}$, and the resulting energy penalty is clipped to $[-0.1,0]$. The source platform is initialized at a random horizontal distance between $0.5\,\mathrm{m}$ and $1.5\,\mathrm{m}$, with a surface height between $0.1\,\mathrm{m}$ and $0.5\,\mathrm{m}$.

The carry-to reward maintains the grasp while guiding the character toward the target platform:
\begin{equation}
r = 0.4\, r_{\text{path}} + 0.1\, r_{\text{face}} + 0.2\, r_{\text{grasp}} + 0.15\, r_{\text{force}} + 0.15\, r_{\text{lift}}' - \lambda \cdot \mathcal{P}
\end{equation}
The path follow and lift-maintenance terms are defined as
\begin{equation}
    r_{\text{path}} = \exp\!\left(-2\|p_{\text{target}}^{xy} - p_{\text{root}}^{xy}\|^2\right),
    \qquad
    r_{\text{lift}}' = \operatorname{clamp}\!\left(\frac{z_{\min}}{1.0},\ 0,\ 1\right).
\end{equation}
The grasp and force terms are reused from the pick-up reward. We set $\lambda=2\times10^{-5}$ and clip the resulting energy penalty to $[-0.2,0]$.

The put-down reward guides the character to place the box on the target platform and release it after stable placement:
\begin{equation}
    r = 0.3\, r_{\text{prog}} + 0.2\, r_{\text{place}} + 0.2\, r_{\text{succ}} + 0.1\, r_{\text{grasp}} + 0.1\, r_{\text{force}} + 0.1\, r_{\text{lift}}' - \lambda \mathcal{P}.
\end{equation}
The progress and placement terms are defined as
\begin{equation}
    r_{\text{prog}} =
    \operatorname{clamp}\!\left(\frac{D_0 - D_t}{D_0 + \epsilon},\ 0,\ 1\right),
    \qquad
    r_{\text{place}} =
    \exp\!\left(-5|z_{\text{box}} - z_{\text{target}}|\right).
\end{equation}
Here, $r_{\text{prog}}$ rewards horizontal progress toward the target platform, and $r_{\text{place}}$ is activated within $0.5\,\mathrm{m}$ of the target platform to encourage vertical alignment. The success term $r_{\text{succ}}$ is a binary reward triggered when the box is stably placed on the target platform and both hands have released it. Upon success, all sub-rewards are frozen at their maximum values. We set $\lambda=2\times10^{-5}$ and clip the resulting energy penalty to $[-0.2,0]$.

\section{Language Instruction Sets for Downstream Tasks}
\label{app:instructions}

For downstream task adaptation, we provide fixed language instruction sets for the arm and body branches separately. Each instruction specifies a candidate semantic motion prior in the text-to-motion latent space, such as lifting, carrying, standing, or moving in a particular direction. Given the current task context, the gating network $\mathcal{G}_p$ predicts part-wise mixture weights over these candidate instructions and dynamically blends the corresponding motion priors. This design allows the downstream policy to compose task-relevant behaviors from reusable language-conditioned priors, while keeping the pretrained motion decoder fixed. The instruction sets used for each downstream task are listed in~\cref{tab:instructions}.

\begin{table}[h]
\centering
\caption{Language-prior instruction sets for downstream adaptation. The table lists the natural-language prompts used by $\mathscr{F}^{t2m}$ to construct candidate motion priors for each downstream task and body branch. The arm and body prompt sets provide part-specific semantic priors for tasks such as pick-and place, strike, and path follow, while the MoE gating network dynamically blends these priors according to the current task observation.}
\label{tab:instructions}
\begin{tabular}{llp{9cm}}
\toprule
\textbf{Task} & \textbf{Branch} & \textbf{Instructions} \\
\midrule
Pick-and-Place & Arm & ``a man lifts something on his left and places it down on his right'' (Lift)\\
(all stages)   &     & ``the person is moving or carrying something'' (Carry)\\
\cmidrule{2-3}
               & Body & ``a person walks to his right'' (Right)\\
               &      & ``the person was walking forward then turn left'' (Left)\\
               &      & ``a man stands still'' (Stand)\\
\midrule
Strike (kick) & Arm \& Body & ``a person does a high kick with their right leg'' \\
              &             & ``this person kicked with their right leg'' \\
              &             & ``a man runs forward fast'' \\
\midrule
Strike (push) & Arm \& Body & ``a person pushes something forward with both hands'' \\
              &             & ``a person walks up to an object and shoves it'' \\
              &             & ``a man runs forward fast'' \\
\midrule
Path Follow & Arm \& Body & ``a person walks to his right'' \\
               &             & ``the person was walking forward then turn left'' \\
               &             & ``a person runs to the right then runs to the left then back to the middle'' \\
\bottomrule
\end{tabular}
\end{table} 

\section{Long-Horizon Skill Composition Examples}
\label{appendix:skill}

More long-horizon skill composition examples are provided to show different combinations and task sequences. Each example follows the same standardized task interface, where the overall behavior is decomposed into a sequence of executable skill units with specified initialization, condition, and termination criterion. The examples involve different combinations and orderings of heterogeneous skills, further illustrating that HetSkills can flexibly organize skills learned from different supervision forms and task settings into coherent long-horizon behaviors.

\begin{table}[htbp]
\centering
\small
\begin{minipage}{0.98\linewidth}
\hrule
\vspace{0.4em}
\textbf{Prompt A: Two-Box Sequential Placement}
\vspace{0.4em}
\hrule
\vspace{0.6em}
\setlength{\parindent}{0pt}
\setlength{\parskip}{0.5em}
\sloppy
Please follow the Standardized Task Description to generate an executable long-horizon skill program for a SMPL character. The task depicts a humanoid that retrieves a box from the first source table, transports it to a target desk, walks to a second source table, retrieves the second box, and transports it to the same target desk. The detailed skill sequence is specified below.

\textbf{Pick-up Skill (Box 1).}
Initialization: Spawn the humanoid next to the first source table with the first box.
Condition: Provide the box position of the first source table.
Terminate: End when the humanoid stably lifts the first box for several consecutive frames.

\textbf{Carry-to Skill (Box 1).}
Initialization: Use the final pick-up state as the initial state.
Condition: Generate and follow a path from the current position toward the target desk while carrying the first box.
Terminate: End when the humanoid reaches the neighborhood of the target desk.

\textbf{Put-down Skill (Box 1).}
Initialization: Use the final carry-to state as the initial state.
Condition: Provide the first target placement position on the desk.
Terminate: End when the first box is successfully placed for several consecutive frames.

\textbf{Path Follow Skill.}
Initialization: Start from the final state of the first put-down skill.
Condition: Generate and follow a walking path toward the second source table.
Terminate: End when the humanoid reaches the vicinity of the second source table.

\textbf{Pick-and-Place Skill (Box 2).}
Initialization: Start from the final state of the path follow skill, with the second box on the second source table.
Condition: Reuse the same pick-up, carry-to, and put-down procedure for Box 2 by providing the second box position, the path from the second source table to the desk, and the second target placement position.
Terminate: End when the second box is successfully placed on the desk.
\vspace{0.3em}
\hrule
\end{minipage}
\end{table}
\begin{figure*}[t]
    \centering
    \includegraphics[width=\textwidth]{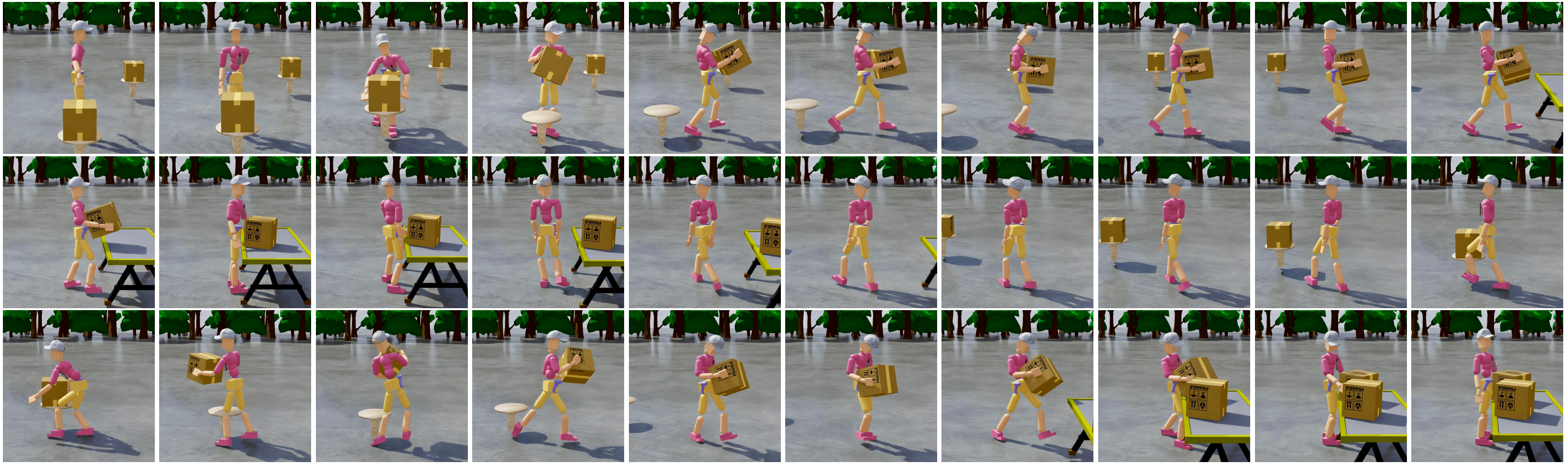}
    \caption{Long-horizon skill composition results for the two-box sequential placement task. The humanoid first picks up the first box from the source table and places it on the target desk, then walks to the second source table, picks up the second box, carries it back to the desk, and places it at the target position.}
    \vspace{2em}
\end{figure*}

\begin{table}[htbp]
\centering
\small
\begin{minipage}{0.98\linewidth}
\hrule
\vspace{0.4em}
\textbf{Prompt B: From Sitting to Striking and Celebration}
\vspace{0.4em}
\hrule
\vspace{0.6em}
\setlength{\parindent}{0pt}
\setlength{\parskip}{0.5em}
\sloppy
Please follow the Standardized Task Description to generate an executable long-horizon skill program for a SMPL character. The task depicts a humanoid that walks to a chair, sits down and stands up, moves toward a target object and strikes it down, and then celebrates with an open-ended text-conditioned jumping motion. The detailed skill sequence is specified below.

\textbf{Path Follow Skill.}
Initialization: Spawn the humanoid at the scene origin.
Condition: Generate and follow a straight walking path toward a target location in front of the chair.
Terminate: End when the humanoid reaches the target location in front of the chair.

\textbf{Human-Scene Interaction Skill.}
Initialization: Start from the final state of the path follow skill.
Condition: Provide a scheduled inpainting target sequence, including sitting down on the chair, holding the seated pose, and standing back up, with a specified duration for each target frame.
Terminate: End when the full interaction schedule is completed.

\textbf{Strike Skill.}
Initialization: Start from the final state of the human-scene interaction skill.
Condition: Provide the target object position, and generate a full-body striking motion that moves the humanoid toward the target object and knocks it down.
Terminate: End when the target object remains in a knocked-down orientation.

\textbf{Text-to-Motion Skill.}
Initialization: Start from the final state of the strike skill.
Condition: Set the text prompt to ``a person jumping up and down, with their hands above their head.''
Terminate: Continue until the episode is externally stopped.
\vspace{0.3em}
\hrule
\end{minipage}
\end{table}
\begin{figure*}[t]
    \centering
    \includegraphics[width=\textwidth]{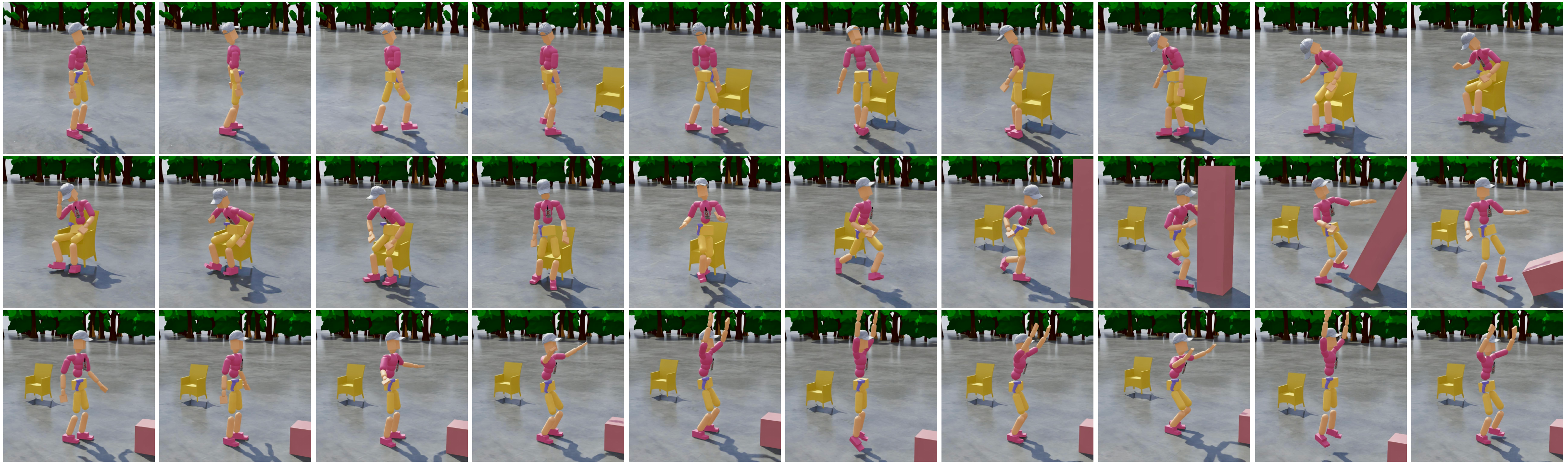}
    \caption{Long-horizon skill composition results for the sit-strike-celebrate task. The humanoid walks to the chair, sits down and stands up, approaches the target object, strikes it down, and then performs a jumping celebration motion.}
\end{figure*}

\begin{table}[htbp]
\centering
\small
\begin{minipage}{0.98\linewidth}
\hrule
\vspace{0.4em}
\textbf{Prompt: Text-Conditioned Motion Sequence}
\vspace{0.4em}
\hrule
\vspace{0.6em}
\setlength{\parindent}{0pt}
\setlength{\parskip}{0.5em}
\sloppy
Please follow the Standardized Task Description to generate an executable long-horizon skill program for a SMPL character. The task depicts a humanoid that sequentially performs three text-conditioned motions, including punching, kicking, and cartwheeling, each for a specified duration. The detailed skill sequence is specified below.

\textbf{Text-to-Motion: Punch.}
Condition: Set the language embedding to ``person was fighting with a left punch''.
Terminate: End after 5 seconds.

\textbf{Text-to-Motion: Forward Kick.}
Condition: Switch the language embedding to ``a person kicks forward.''
Terminate: End after 3 seconds.

\textbf{Text-to-Motion: Cartwheel.}
Condition: Switch the language embedding to ``a person does a cartwheel.''
Terminate: End after 5 seconds.
\vspace{0.3em}
\hrule
\end{minipage}
\end{table}

\begin{figure*}[t]
    \centering
    \includegraphics[width=\textwidth]{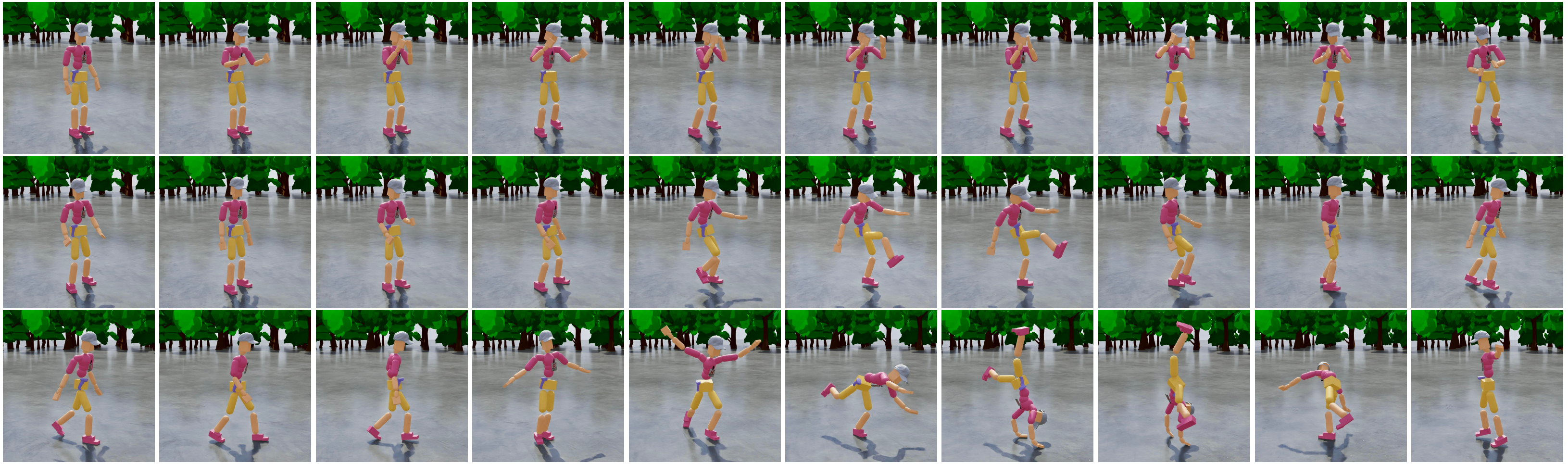}
    \caption{Long-horizon rollout of sequential text-conditioned motion generation. The humanoid switches across three language conditions and performs a left punch, a forward kick, and a cartwheel in sequence.}
\end{figure*}

\end{document}